%% file: main.tex
\documentclass[11pt,oneside]{book}

\usepackage[T1]{fontenc}
\usepackage[utf8]{inputenc}
\usepackage{lmodern}
\usepackage[margin=1in]{geometry}
\usepackage{graphicx}
\usepackage{amsmath,amssymb}
\usepackage{xcolor}
\usepackage{listings}
\usepackage{hyperref}
\usepackage{caption}
\usepackage{tcolorbox}
\usepackage{parskip}
\usepackage[strings]{underscore}
\usepackage{tocloft}
\tcbuselibrary{skins,breakable}

\cftsetpnumwidth{1.25em}
\hypersetup{
  colorlinks=true,
  linkcolor=blue!45!black,
  citecolor=blue!45!black,
  urlcolor=blue!45!black,
}

\definecolor{codebg}{rgb}{0.96,0.96,0.96}
\definecolor{kwcolor}{rgb}{0.0,0.3,0.6}
\definecolor{strcolor}{rgb}{0.0,0.45,0.35}
\definecolor{cmtcolor}{rgb}{0.45,0.45,0.45}

\lstdefinestyle{pycode}{
  language=Python,
  basicstyle=\ttfamily\small,
  keywordstyle=\color{kwcolor}\bfseries,
  commentstyle=\color{cmtcolor}\itshape,
  stringstyle=\color{strcolor},
  showstringspaces=false,
  breaklines=true,
  breakatwhitespace=true,
  backgroundcolor=\color{codebg},
  frame=single,
  rulecolor=\color{gray!40},
  framesep=4pt,
  numbers=none,
  columns=flexible,
  xleftmargin=0pt,
  aboveskip=8pt,
  belowskip=8pt,
}
\newcommand{\code}[1]{\texttt{\small #1}}

\newtcolorbox{warningbox}{
  colback=red!4, colframe=red!55!black, boxrule=0.6pt,
  title={Warning}, fonttitle=\bfseries, breakable
}
\newtcolorbox{notebox}{
  colback=blue!4, colframe=blue!35!black, boxrule=0.6pt,
  title={Note}, fonttitle=\bfseries, breakable
}
\newtcolorbox{repobox}{
  colback=teal!6, colframe=teal!45!black, boxrule=0.6pt, breakable,
  fontupper=\small\itshape, left=8pt
}

\newcommand{\reponotebookurl}{https://github.com/sbirchfield/sbirchfield.github.io/blob/main/cvintro/notebooks/}

\newcommand{\fullcode}[2]{%
  \begin{repobox}
  \textbf{\upshape Full runnable code} \textit{(including plotting and setup):} \href{\reponotebookurl#2.ipynb}{\texttt{#1.ipynb}} in the course repository.
  \end{repobox}
}

\newcommand{\citelink}[2]{\hyperlink{bib:#1}{#2}}
\newcommand{\bibentry}[3]{%
  \hypertarget{bib:#1}{}%
  \noindent\textbf{#2}#3\par\smallskip
}
\newcommand{\landmark}{\textcolor{orange!80!black}{$\bigstar$}\ }

\newcommand{\attrib}[1]{{\footnotesize\itshape #1}}

\begin{document}

\frontmatter

\begin{titlepage}
\centering
\vspace*{3.2cm}
{\Huge\bfseries Introduction to Computer Vision}\par
\vspace{0.7cm}
{\Large\itshape\color{gray!55!black} Lecture notes with working Python code}\par
\vspace{2.4cm}
{\Large Stan Birchfield}\par
\vspace{0.9cm}
{\normalsize\color{gray!45!black} NVIDIA\\University of Washington}\par
\vspace{2.6cm}
{\small\itshape\color{gray!45!black} Created with AI assistance}\par
\vspace{0.5cm}
\href{https://sbirchfield.github.io/cvintro}{\small\ttfamily sbirchfield.github.io/cvintro}\par
\vfill
{\normalsize 2026}\par
\vspace{8pt}
{\small\color{gray!45!black} This book is published under the Creative Commons Attribution 4.0\\
International License (CC-BY 4.0).}\par
\vspace*{2.2cm}
\end{titlepage}

\tableofcontents

\input{chapters/preface}

\mainmatter
\part{Classical 2D Image Processing (1960--1990)}

\input{chapters/lesson01}
\input{chapters/lesson02}
\input{chapters/lesson03}
\input{chapters/lesson04}
\input{chapters/lesson05}
\input{chapters/lesson06}
\input{chapters/lesson07}
\input{chapters/lesson08}
\input{chapters/lesson09}
\input{chapters/lesson10}
\input{chapters/lesson11}
\input{chapters/lesson12}
\input{chapters/lesson13}
\input{chapters/lesson14}
\input{chapters/lesson15}
\input{chapters/lesson16}
\input{chapters/lesson17}
\input{chapters/lesson18}
\input{chapters/lesson19}

\part{Classical 3D Computer Vision (1980--2010)}

\input{chapters/lesson20}
\input{chapters/lesson21}
\input{chapters/lesson22}
\input{chapters/lesson23}
\input{chapters/lesson24}
\input{chapters/lesson25}
\input{chapters/lesson26}
\input{chapters/lesson27}
\input{chapters/lesson28}
\input{chapters/lesson29}

\part{Deep Learning for Computer Vision (2012--2022)}

\input{chapters/lesson30}
\input{chapters/lesson31}
\input{chapters/lesson32}
\input{chapters/lesson33}
\input{chapters/lesson34}
\input{chapters/lesson35}
\input{chapters/lesson36}
\input{chapters/lesson37}
\input{chapters/lesson38}
\input{chapters/lesson39}
\input{chapters/lesson40}
\input{chapters/lesson41}
\input{chapters/lesson42}
\input{chapters/lesson43}
\input{chapters/lesson44}

\backmatter
\input{chapters/references}

\end{document}

%% file: chapters/preface.tex
\chapter*{Preface}
\addcontentsline{toc}{chapter}{Preface}

When I published my first book on computer vision (\textit{Image Processing and Analysis}, Cengage 2018), I hoped that it would last.  I had taken great pains to ensure that the material was thoroughly covered, and that it focused on the most important concepts in the field. Alas, since it took me 12 years to write the book, it was inevitably somewhat obsolete before the ink dried. The deep learning revolution eclipsed nearly all of the old methods.

In this book, I have attempted to update and simplify the material, as well as to combine it with working Python code. Thanks to the power of AI, this book was written in just one month.

This book is a community effort.  It was AI-distilled from the online course notes, which were themselves AI-written.  Without all the amazing online material that was digested by the AI model, this would not have been possible.  The AI model used was Claude Sonnet 5.  Claude was able to structure the course in ways that I would not have imagined, and it was able to blend the prose and code seamlessly in ways that I could never have done myself.  I take little credit in the writing of the book, other than to say that I did have to prompt Claude quite a bit to get the output I wanted.

Thanks to all the amazing folks at the University of Washington for allowing me to teach a computer vision course for the past 7 years, and for all the equally amazing students from whom I learn so much. Also thanks to NVIDIA for giving me the freedom.

I hope you will learn to love computer vision as much as I do.  Enjoy!

\bigskip
\begin{flushright}
Stan Birchfield\\
Redmond, Washington\\
2026
\end{flushright}

%% file: chapters/lesson01.tex
\chapter{Images as Arrays}
\label{ch:lesson01}

A digital image is a 2D grid of numbers. This chapter builds a small image from scratch with NumPy, examines how grayscale and color images are represented, and displays them.

\section{A grayscale image is a 2D array}

\begin{figure}[h]
\centering
\includegraphics[width=0.35\textwidth]{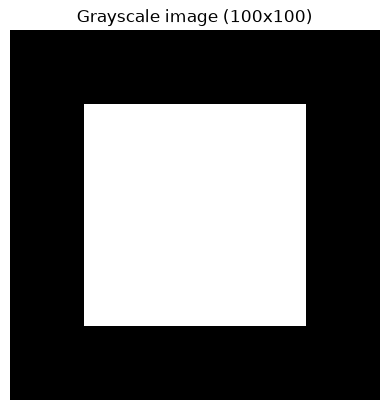}
\caption{A $100\times100$ grayscale array with a white square.}
\label{fig:l01-1}
\end{figure}

Each entry is a pixel intensity. For an 8-bit image, pixel values range from 0 (black) to 255 (white), as in Figure~\ref{fig:l01-1}.

\begin{lstlisting}
im = np.zeros((100, 100), dtype=np.uint8)
im[20:80, 20:80] = 255   # a white square on a black background
\end{lstlisting}

\subsection*{Simple pixel-level operations}

\begin{figure}[h]
\centering
\includegraphics[width=0.6\textwidth]{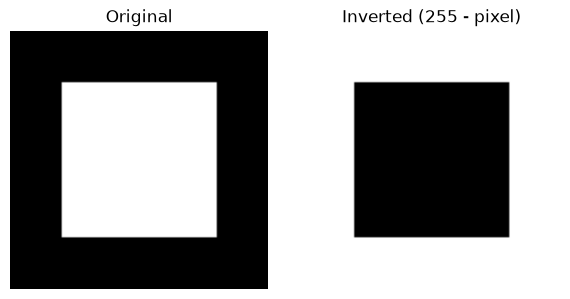}
\caption{Inversion is a single elementwise array operation.}
\label{fig:l01-2}
\end{figure}

Because an image is just an array of numbers, standard NumPy operations apply directly, e.g., inverting intensities with \code{im\_inv = 255 - im} (Figure~\ref{fig:l01-2}).

\section{A color image is a 3D array}

An image's dimensions are \code{(height, width, channels)} --- the output of \code{array.shape}. The \textbf{first} axis selects the row (moving down the image, i.e., $y$), the \textbf{second} axis selects the column (moving across, i.e., $x$); the \textbf{third} axis selects the color channel (red, green, or blue). By convention, the origin is the top-left corner.

\begin{warningbox}
Even though \code{array.shape} and NumPy indexing use \code{(row, column)} = \code{(y, x)} ordering, many OpenCV functions that take points or sizes --- \code{cv2.circle}, \code{cv2.rectangle}, \code{cv2.resize}, \code{cv2.warpAffine}'s \code{dsize}, etc. --- expect the \emph{opposite} order, \code{(x, y)} or \code{(width, height)}. This reversal is a common source of bugs; always check a function's argument order rather than assuming it matches array shape.
\end{warningbox}

\begin{lstlisting}
imc = np.zeros((50, 100, 3), dtype=np.uint8)
imc[:, :, 0] = 255          # red channel on for the whole image
imc[10:40, 20:80, 1] = 255  # add green in the middle -> looks yellow
\end{lstlisting}

\begin{figure}[h]
\centering
\includegraphics[width=0.55\textwidth]{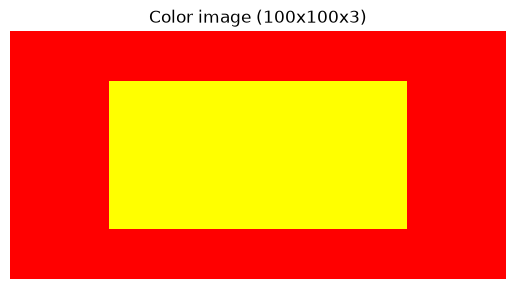}
\caption{Red + green renders as yellow.}
\label{fig:l01-3}
\end{figure}

The result is shown in Figure~\ref{fig:l01-3}.

\subsection*{Colors are typically in RGB order}

\begin{figure}[h]
\centering
\includegraphics[width=0.5\textwidth]{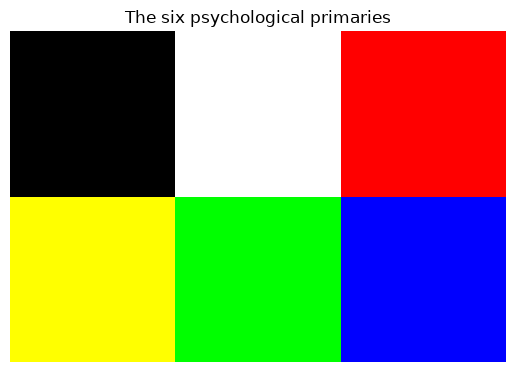}
\caption{The six psychological primaries as a $2\times3$ RGB image.}
\label{fig:l01-4}
\end{figure}

To illustrate channel order, build a tiny image out of the six \emph{psychological primaries} --- the six hues (black, white, red, yellow, green, blue) the human visual system treats as elementary (Figure~\ref{fig:l01-4}).

\begin{figure}[h]
\centering
\includegraphics[width=0.75\textwidth]{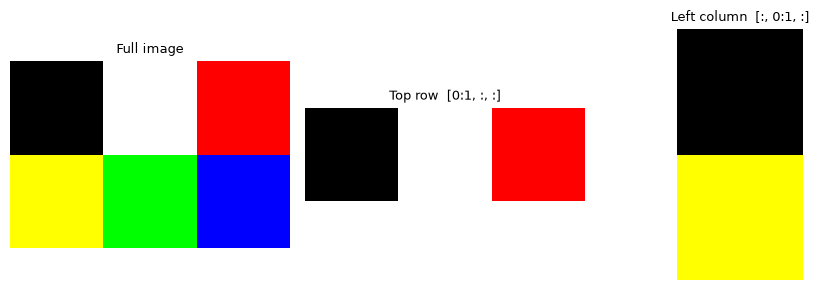}
\caption{Slicing the first axis selects a row; the second axis, a column.}
\label{fig:l01-5}
\end{figure}

Slicing the first axis selects whole rows (a horizontal band); slicing the second axis selects whole columns (a vertical band), as Figure~\ref{fig:l01-5} shows directly.

\begin{notebox}
A bare integer index (\code{im[0, :, :]}) \emph{drops} that axis entirely, turning shape \code{(1, 3, 3)} into \code{(3, 3)} --- \code{imshow} would then misread it as a $3\times3$ grayscale image rather than a 1-row RGB strip. Slicing with \code{0:1} keeps the axis alive as size 1.
\end{notebox}

\section{Loading a real photo: BGR vs. RGB}

Every color image so far was constructed manually with channels in RGB order --- matching what Matplotlib's \code{plt.imshow} expects. Real photos, however, are loaded with OpenCV's \code{cv2.imread}, which reads (and writes) color images with channels in \textbf{BGR} order --- blue first, red last --- the opposite of the RGB order every other Python imaging library assumes. Handing a BGR array straight to \code{imshow} misinterprets the channels, producing a color-shifted image.

There are two equivalent ways to fix this:
\begin{enumerate}
  \item \textbf{Reverse the channel axis with NumPy indexing:} \code{img[:, :, ::-1]} turns \code{[B, G, R]} into \code{[R, G, B]} --- a plain array operation, no OpenCV function needed.
  \item \textbf{\code{cv2.cvtColor}:} OpenCV's general-purpose color-conversion function; \code{cv2.COLOR\_BGR2RGB} performs exactly the same channel swap, but is more explicit about \emph{why}, and is the idiomatic choice in OpenCV code (the same function also handles conversions that are not a simple reversal, e.g.\ to grayscale or HSV, in later lessons).
\end{enumerate}

\begin{lstlisting}
im_rgb1 = im_bgr[:, :, ::-1]
im_rgb2 = cv2.cvtColor(im_bgr, cv2.COLOR_BGR2RGB)
# np.array_equal(im_rgb1, im_rgb2) -> True: both fixes agree exactly
\end{lstlisting}

\begin{figure}[h]
\centering
\includegraphics[width=0.85\textwidth]{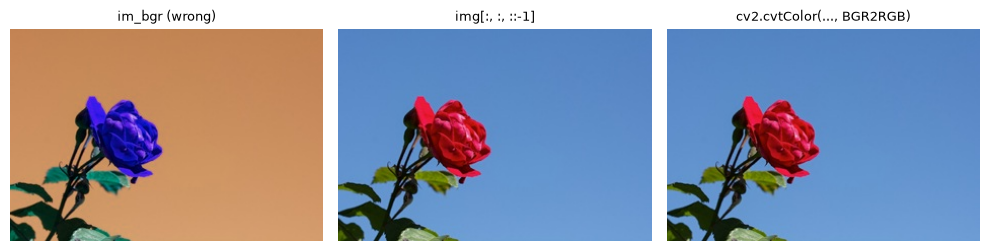}
\caption{Left: BGR array shown directly (wrong colors). Center and right: the two equivalent fixes, in agreement. \attrib{Photo: \href{https://picryl.com}{Picryl}.}}
\label{fig:l01-6}
\end{figure}

Figure~\ref{fig:l01-6} compares the raw BGR array against both fixes.

\subsection*{Exercises}
\begin{enumerate}
  \item Modify the color image \code{imc} so the rectangle is cyan instead of yellow. Which channel(s) need to change?
  \item In the primaries image, slice out the \emph{bottom-right} block (blue) using row and column ranges, and confirm with \code{np.array\_equal} that it matches a fresh block filled with \code{(0, 0, 255)}.
  \item Load \code{rose.jpg} and print \code{im\_bgr[0, 0, :]} alongside \code{im\_rgb2[0, 0, :]}. Confirm by hand that the three numbers are the same three numbers, just reordered.
\end{enumerate}

\vfill
\fullcode{lesson01}{lesson01_images_as_arrays}

%% file: chapters/lesson02.tex
\chapter{Image Arithmetic}
\label{ch:lesson02}

Since an image is a 2D array of numbers (Chapter~\ref{ch:lesson01}), basic arithmetic can edit it: add a constant to brighten, subtract to darken, blend two images by weighted averaging, and so on. This chapter highlights an important subtlety: when images are stored as 8-bit unsigned integers, \emph{how} you do the arithmetic matters.

\section{Brightening: add a constant}

A pixel's value represents the amount of light, so adding a positive constant to every pixel should brighten the whole image.

\begin{lstlisting}
img = np.zeros((150, 150), dtype=np.uint8)
cv2.circle(img, (75, 75), 50, 180, -1)
brightened_naive = img + np.uint8(80)
\end{lstlisting}

\begin{figure}[h]
\centering
\includegraphics[width=0.55\textwidth]{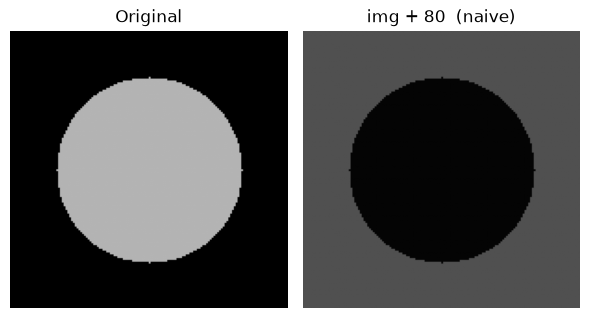}
\caption{Look closely at the circle: it became \emph{darker}, not brighter.}
\label{fig:l02-1}
\end{figure}

The result (Figure~\ref{fig:l02-1}) is not what was intended.

\subsection*{The problem: \code{uint8} overflow}

An 8-bit unsigned integer can only represent 0 to 255, so $220+80=300$ does not fit. NumPy's \code{uint8} arithmetic \textbf{wraps around} (like a car odometer rolling over), silently computing $300 \bmod 256 = 44$ instead of \textbf{clamping} at 255 --- the same fixed-width overflow any low-level integer type has, applied silently and with no warning:
\begin{equation}
\texttt{np.array([200, 250, 220], dtype=np.uint8) + np.uint8(80)} \;\longrightarrow\; \texttt{[24, 74, 44]}
\end{equation}

\subsection*{The fix: saturating arithmetic}

OpenCV's arithmetic functions (\code{cv2.add}, \code{cv2.subtract}, \ldots) use \textbf{saturating} arithmetic: results are clamped to the valid range (0--255 for \code{uint8}) rather than wrapping --- almost always what is actually wanted.

\begin{lstlisting}
brightened_saturated = cv2.add(img, 80)
\end{lstlisting}

\begin{figure}[h]
\centering
\includegraphics[width=0.9\textwidth]{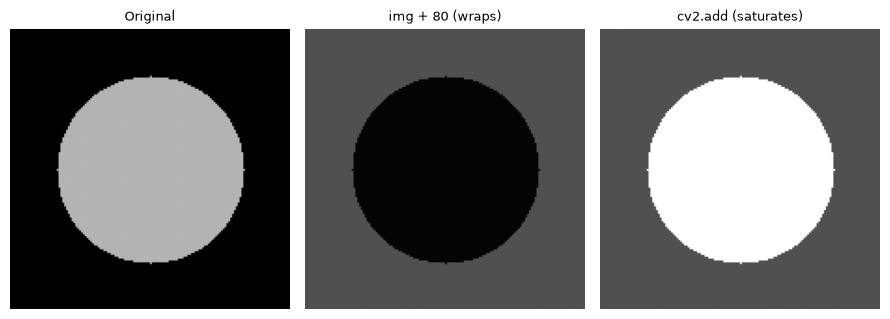}
\caption{Naive NumPy addition wraps (darkens); \code{cv2.add} saturates (clamps at 255) --- the same idea applies at the bottom end for \code{cv2.subtract}, clamped at 0 instead of wrapping to a large positive value.}
\label{fig:l02-2}
\end{figure}

Figure~\ref{fig:l02-2} compares all three results side by side.

\section{Brightening only part of an image, with a mask}

Often one wants to adjust brightness selectively --- e.g., lighten a foreground subject without touching the background. Given a binary mask (from thresholding or segmentation, Chapters~\ref{ch:lesson03}--\ref{ch:lesson04}), only copy the brightened result where the mask is set.

\begin{lstlisting}
fully_brightened = cv2.add(photo, np.full_like(photo, 60))
result = photo.copy()
result[foreground_mask > 0] = fully_brightened[foreground_mask > 0]
\end{lstlisting}

\begin{figure}[h]
\centering
\includegraphics[width=0.85\textwidth]{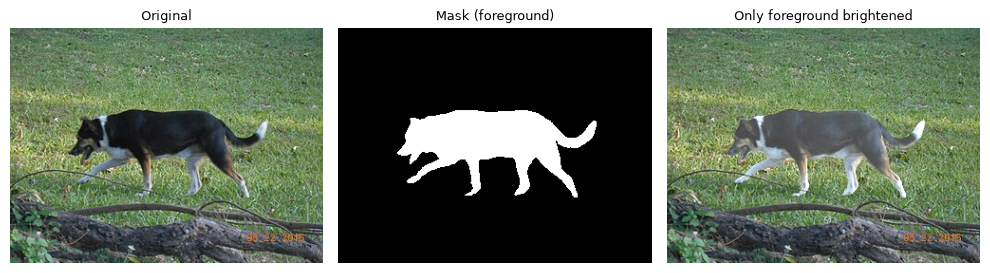}
\caption{Photo, foreground mask, and the selectively brightened result. \attrib{Photo: \href{https://commons.wikimedia.org/wiki/File:Dog_walking_sideview_grass.JPG}{Wikimedia Commons}.}}
\label{fig:l02-3}
\end{figure}

Figure~\ref{fig:l02-3} shows the photo, the mask, and the result.

\begin{notebox}
\code{cv2.add} (and several other OpenCV functions) also accept a \code{mask} argument directly, which performs exactly this select-and-copy in one call: \code{cv2.add(photo, 60, mask=foreground\_mask)} --- writing only into the masked region and leaving the rest of the destination untouched.
\end{notebox}

\section{Blending two images}

\begin{figure}[h]
\centering
\includegraphics[width=0.95\textwidth]{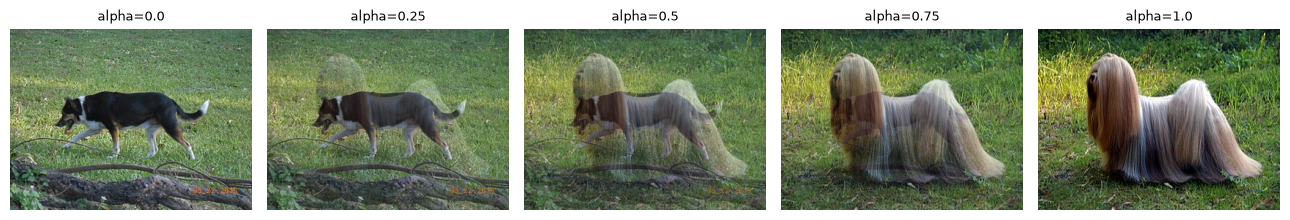}
\caption{\code{cv2.addWeighted} at five values of $\alpha$, cross-dissolving between two dog photos. \attrib{Photos: \href{https://commons.wikimedia.org/wiki/File:Dog_walking_sideview_grass.JPG}{Wikimedia Commons}, \href{https://commons.wikimedia.org/wiki/File:Lhasaapso.jpg}{Wikimedia Commons}.}}
\label{fig:l02-4}
\end{figure}

A weighted sum of two images --- a \textbf{cross-dissolve} --- is the same saturating arithmetic, generalized: \code{cv2.addWeighted(a, alpha, b, beta, gamma)} computes $a\cdot\alpha + b\cdot\beta + \gamma$, saturated, as swept across five values of $\alpha$ in Figure~\ref{fig:l02-4}.

\section{Difference imaging}

Subtracting two images (with an absolute value, so both directions of change matter equally) highlights exactly what differs between them --- the basis of simple motion/change detection, reused when comparing reconstructions against ground truth in later lessons.

\begin{lstlisting}
difference = cv2.absdiff(before, after)
\end{lstlisting}

\begin{figure}[h]
\centering
\includegraphics[width=0.85\textwidth]{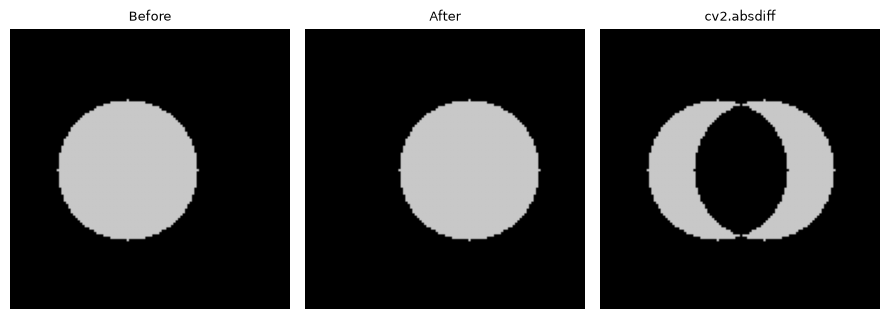}
\caption{The difference image is zero everywhere nothing changed, bright wherever the circle used to be or now is, and dark in the region where it is present in both frames.}
\label{fig:l02-5}
\end{figure}

Figure~\ref{fig:l02-5} shows the result.

\subsection*{Exercises}
\begin{enumerate}
  \item Predict, then check, what \code{cv2.subtract(img, 150)} does to a pixel whose value is 100 --- does it wrap around like the naive NumPy addition did, or saturate?
  \item Use \code{cv2.add} with its \code{mask} argument directly (instead of manually copying with boolean indexing) to reproduce the masked-brightening result, and confirm the two approaches give identical output.
  \item \code{cv2.addWeighted} does not require \code{alpha + beta} to sum to 1 (the \code{gamma} term is added on top). Try values that do not sum to 1 and describe the effect.
\end{enumerate}

\vfill
\fullcode{lesson02}{lesson02_image_arithmetic}

%% file: chapters/lesson03.tex
\chapter{Thresholding, Erosion, and Dilation}
\label{ch:lesson03}

Thresholding converts a grayscale image into a binary (black/white) image by comparing each pixel to a cutoff value. The result is often noisy, so it is cleaned up with two basic morphological operations: \textbf{erosion} (shrinks white regions, removes small specks) and \textbf{dilation} (grows white regions, fills small holes).

\section{Thresholding}

\begin{figure}[h]
\centering
\includegraphics[width=0.35\textwidth]{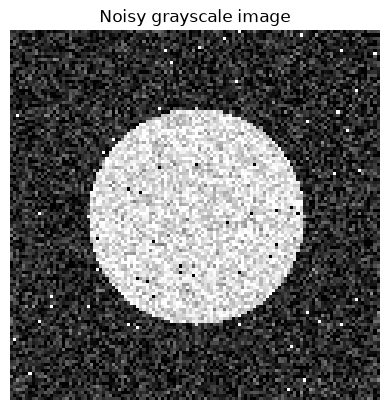}
\caption{A synthetic bright circle on a dark background, with added noise and salt-and-pepper specks.}
\label{fig:l03-1}
\end{figure}

\begin{figure}[h]
\centering
\includegraphics[width=0.35\textwidth]{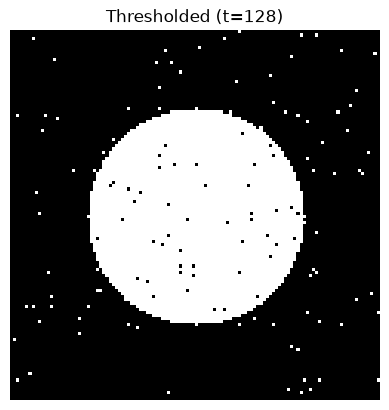}
\caption{Simple thresholding at $t=128$.}
\label{fig:l03-2}
\end{figure}

Thresholding is a per-pixel comparison: every pixel above the cutoff becomes white (255), every pixel at or below it becomes black (0). Figure~\ref{fig:l03-1} shows the noisy test image used throughout this chapter, and Figure~\ref{fig:l03-2} the result of thresholding it.

\begin{lstlisting}
binary = (noisy > threshold_value) * np.uint8(255)
\end{lstlisting}

OpenCV bundles the same operation into \code{cv2.threshold}, which for plain binary thresholding computes exactly the same result, and additionally supports variants such as inverted output, capping instead of zeroing, and automatic cutoff selection through the same function.

\subsection*{Otsu's method: choosing the threshold automatically}

The threshold above was picked by hand. \textbf{Otsu's method} (\citelink{otsu1979}{Otsu, 1979}) picks it automatically: it treats the image's histogram as a mixture of two classes (foreground and background) and searches over every possible cutoff for the one that minimizes the \emph{within-class} variance (equivalently, maximizes the variance \emph{between} the two classes).

\begin{lstlisting}
thresh_otsu, binary_otsu = cv2.threshold(
    noisy, 0, 255, cv2.THRESH_BINARY + cv2.THRESH_OTSU)
\end{lstlisting}

\begin{figure}[h]
\centering
\includegraphics[width=0.95\textwidth]{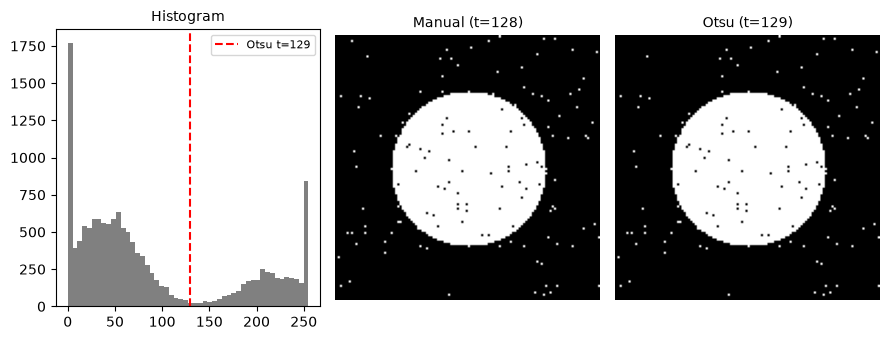}
\caption{Histogram with the Otsu-selected cutoff, compared against the hand-picked threshold; the two results are visually near-identical here.}
\label{fig:l03-3}
\end{figure}

Figure~\ref{fig:l03-3} shows the resulting histogram and cutoff.

Otsu is a good default choice whenever a threshold is needed, but it will not work well with badly-separated or multi-modal histograms.

\section{Erosion and dilation}

There are two basic \textbf{morphological operations} for cleaning up the salt-and-pepper specks and ragged edges above.

\textbf{Erosion} slides a small \textbf{structuring element} (kernel) over the image; a pixel stays white only if the \emph{entire} kernel fits inside the white region, shrinking white regions and removing small specks. \textbf{Dilation} does the opposite: a pixel becomes white if the kernel overlaps \emph{any} part of the white region, growing white regions and filling small black holes.

\begin{lstlisting}
kernel = np.ones((3, 3), np.uint8)
eroded  = cv2.erode(binary, kernel, iterations=1)
dilated = cv2.dilate(binary, kernel, iterations=1)
\end{lstlisting}

\begin{figure}[h]
\centering
\includegraphics[width=0.95\textwidth]{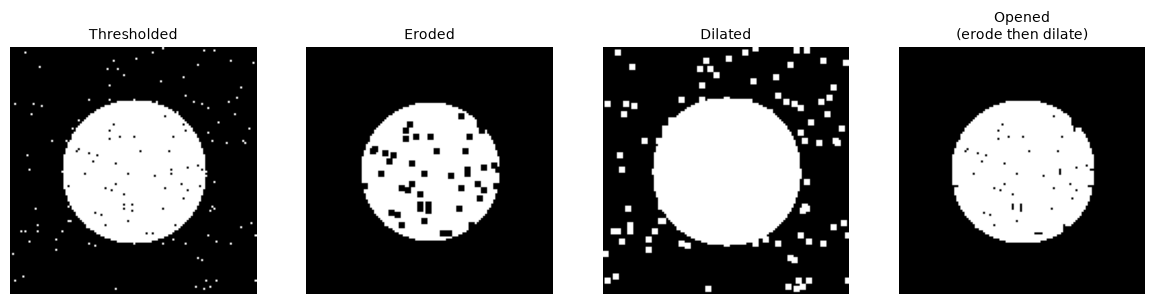}
\caption{Thresholded, eroded, dilated, and opened (erode then dilate) versions of the same binary image.}
\label{fig:l03-4}
\end{figure}

Applying erosion then dilation --- an \textbf{opening} --- removes small white specks while restoring the size of the main region, a common way to denoise a thresholded image: \code{cv2.morphologyEx(binary, cv2.MORPH\_OPEN, kernel)}. Figure~\ref{fig:l03-4} compares all four results.

\subsection*{Exercises}
\begin{enumerate}
  \item Try \code{cv2.MORPH\_CLOSE} (dilation followed by erosion) instead of \code{MORPH\_OPEN}. How does it differ, and which black-pixel noise does it fix that opening does not?
  \item Increase the kernel size to $(5,5)$. How does that change the result compared to $(3,3)$?
\end{enumerate}

\vfill
\fullcode{lesson03}{lesson03_thresholding_morphology}

%% file: chapters/lesson04.tex
\chapter{Flood Fill and Connected Components}
\label{ch:lesson04}

\begin{figure}[h]
\centering
\includegraphics[width=0.95\textwidth]{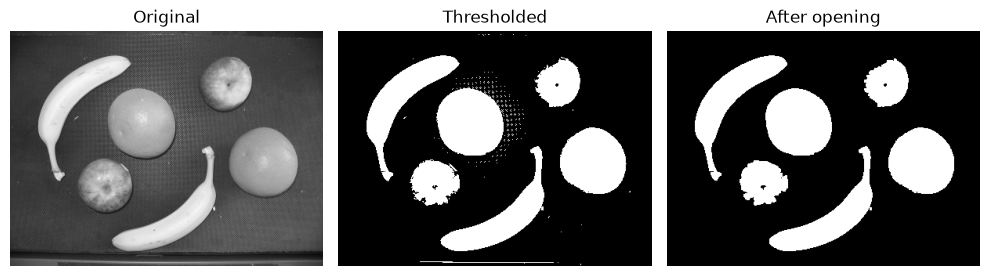}
\caption{A photo of fruit, thresholded with Otsu's method (Chapter~\ref{ch:lesson03}), then cleaned up with a morphological opening. \attrib{Photo: Stan Birchfield.}}
\label{fig:l04-1}
\end{figure}

Once we have a binary image (e.g., from thresholding, Chapter~\ref{ch:lesson03}), a natural next question is: \emph{how many separate blobs are there, and where are they?} Two tools answer this:
\begin{itemize}
  \item \textbf{Flood fill} grows a region from a single seed pixel, spreading to all connected neighbors that share a similar value.
  \item \textbf{Connected component labeling} finds \emph{all} such regions in a binary image at once, labeling each with a unique ID.
\end{itemize}
Figure~\ref{fig:l04-1} shows the test image used throughout this chapter.

\section{Flood fill from a seed point}

Flood fill starts at a seed pixel and spreads outward to all connected pixels within a tolerance of the seed value, painting them a new color. The classic algorithm keeps a \emph{frontier} of pixels still to visit:

\begin{lstlisting}
def flood_fill_stack(binary, seed):
    filled = np.zeros_like(binary, dtype=bool)
    h, w = binary.shape
    stack = [seed]
    while stack:
        x, y = stack.pop()
        if x < 0 or x >= w or y < 0 or y >= h:
            continue
        if filled[y, x] or binary[y, x] == 0:
            continue
        filled[y, x] = True
        stack.extend([(x+1, y), (x-1, y), (x, y+1), (x, y-1)])
    return filled
\end{lstlisting}

\begin{figure}[h]
\centering
\includegraphics[width=0.46\textwidth]{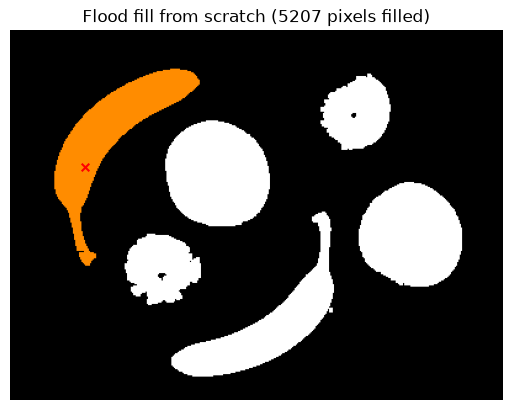}\hspace{0.02\textwidth}\includegraphics[width=0.46\textwidth]{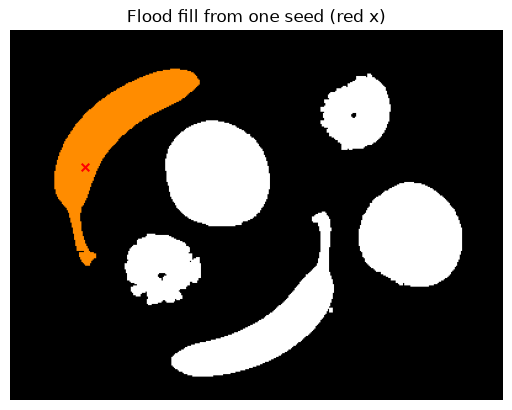}
\caption{Flood fill from a single seed (red $\times$): the hand-written stack version (left) and \code{cv2.floodFill} (right) fill an identical region.}
\label{fig:l04-2}
\end{figure}

\code{cv2.floodFill} does exactly the same thing, considerably faster because it is compiled. Figure~\ref{fig:l04-2} compares the two.

\section{Connected components: label every blob at once}

\begin{figure}[h]
\centering
\includegraphics[width=0.55\textwidth]{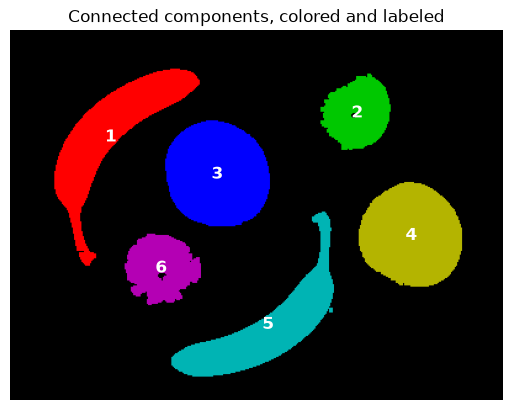}
\caption{Each connected blob labeled with a distinct color and its integer ID.}
\label{fig:l04-3}
\end{figure}

Instead of picking seeds by hand, connected-component labeling scans the whole image and assigns every blob its own integer label:

\begin{lstlisting}
num_labels, labels, stats, centroids = \
    cv2.connectedComponentsWithStats(im_bin, connectivity=8)
\end{lstlisting}

The classic algorithm --- omitted here for brevity --- is known as \emph{union-find}: two passes through the image, plus a traversal of an equivalence table. It also returns handy per-blob statistics (bounding box, area, centroid) essentially for free, since they require almost no extra computation (Chapter~\ref{ch:lesson05}). Figure~\ref{fig:l04-3} shows the labeled result.

\section{Filtering blobs by size}

\begin{figure}[h]
\centering
\includegraphics[width=0.75\textwidth]{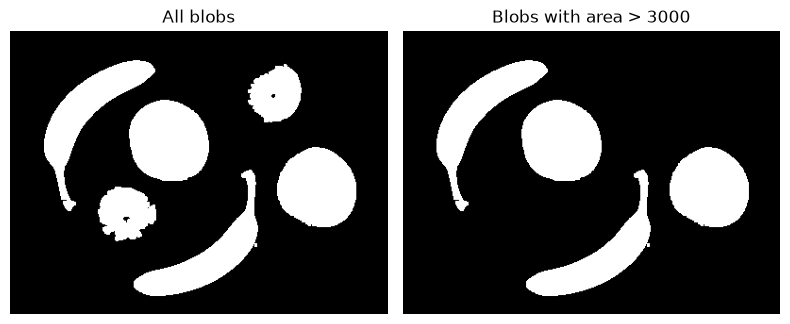}
\caption{All blobs (left) versus only blobs with area above a chosen cutoff (right), which discards the two smallest apples.}
\label{fig:l04-4}
\end{figure}

A common use of connected components is to discard small, noise-like blobs and keep only significant ones, by thresholding on \code{stats[label, cv2.CC\_STAT\_AREA]}, as in Figure~\ref{fig:l04-4}.

\subsection*{Exercises}
\begin{enumerate}
  \item Change \code{connectivity=8} to \code{connectivity=4}. Construct a binary image (e.g., a diagonal staircase of single pixels) where 4-connectivity and 8-connectivity give a \emph{different} number of components.
  \item Use \code{cv2.floodFill} with a nonzero \code{loDiff}/\code{upDiff} tolerance on a grayscale (not binary) image, and describe what changes.
\end{enumerate}

\vfill
\fullcode{lesson04}{lesson04_floodfill_connected_components}

%% file: chapters/lesson05.tex
\chapter{Image Moments}
\label{ch:lesson05}

Once a blob can be isolated (Chapter~\ref{ch:lesson04}), \textbf{moments} let us summarize its shape with a handful of numbers: its area, centroid, orientation, and even a description that stays the same under translation, scale, and rotation. Moments offer a classic, lightweight alternative to learned features (covered in later chapters) for simple shape matching.

\section{What is a moment?}

For a binary image, the raw moment $m_{pq}$ is defined as
\begin{equation}
m_{pq} = \sum_{x,y} x^p\, y^q\, I(x,y),
\end{equation}
where $I(x,y)$ is 1 inside the shape and 0 elsewhere. A few special cases are already familiar quantities:
\begin{itemize}
  \item $m_{00}$ = area (pixel count)
  \item $(\bar x, \bar y) = (m_{10}/m_{00},\, m_{01}/m_{00})$ yields the centroid
\end{itemize}
The \textbf{order} of a moment is $p+q$: $m_{00}$ is zeroth-order, $m_{10}$ and $m_{01}$ are first-order, and $m_{20}$, $m_{11}$, $m_{02}$ are second-order.

The formula is just a weighted sum over pixel coordinates, so for a binary image it reduces to finding every foreground pixel's $(x, y)$ and summing $x^py^q$ over them:

\begin{lstlisting}
ys, xs = np.nonzero(im_tiny)   # (row, col) = (y, x) of every foreground pixel
m00 = len(xs)                  # area
m10 = xs.sum()                 # sum of x^1 y^0
m01 = ys.sum()                 # sum of x^0 y^1
# cv2.moments(im_tiny, binaryImage=True) agrees exactly with this hand computation
\end{lstlisting}

\section{An elongated, rotated blob}

To make orientation easy to eyeball and check against, draw a rotated ellipse.

\begin{notebox}
OpenCV's \code{cv2.ellipse} produces minor rendering artifacts around the border; the figures below instead use a hand-written scan-conversion (an explicit inside/outside test against the rotated-ellipse equation) for an exact, artifact-free shape. See the repository for the drawing routine.
\end{notebox}

\section{Area and centroid from moments}

\begin{lstlisting}
m = cv2.moments(im_binary, binaryImage=True)
area = m['m00']
cx, cy = m['m10'] / m['m00'], m['m01'] / m['m00']
\end{lstlisting}

\begin{figure}[h]
\centering
\includegraphics[width=0.4\textwidth]{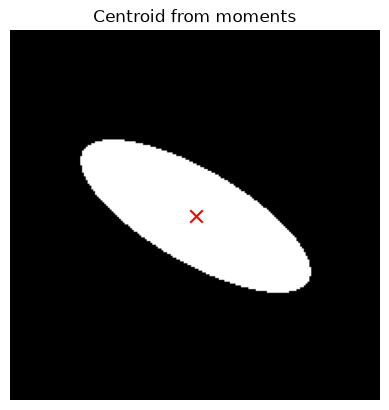}
\caption{The centroid recovered from moments, marked on the rotated ellipse.}
\label{fig:l05-1}
\end{figure}

Figure~\ref{fig:l05-1} marks the result on the ellipse.

\section{Central moments}

Raw moments $m_{pq}$ depend on where the shape sits in the image --- move the shape and every $m_{ij}$ except $m_{00}$ changes. \textbf{Central moments} fix this by measuring around the shape's own centroid instead of the image origin:
\begin{equation}
\mu_{pq} = \sum_{x,y} (x-\bar x)^p\,(y-\bar y)^q\, I(x,y).
\end{equation}
Recomputing this sum from scratch would revisit every pixel again; instead, a shift-of-origin identity --- the image-moment analogue of the parallel-axis theorem --- gives, e.g., $\mu_{20} = m_{20} - \bar x\, m_{10}$, computed directly from the raw moments already in hand.

\section{Orientation from central moments}

\begin{figure}[h]
\centering
\includegraphics[width=0.4\textwidth]{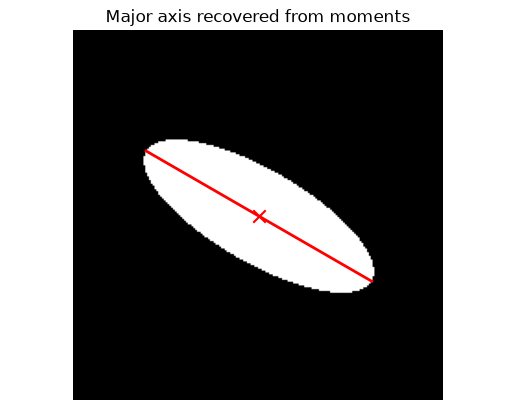}
\caption{The major axis recovered purely from second-order central moments matches the ellipse's true drawn orientation.}
\label{fig:l05-2}
\end{figure}

The central moments $\mu_{20}$, $\mu_{02}$, $\mu_{11}$ describe the spread of the shape around its centroid --- essentially its covariance matrix (revisited in later chapters). The angle of the major axis (direction of greatest spread) is
\begin{equation}
\theta = \tfrac{1}{2}\,\operatorname{atan2}\bigl(2\mu_{11},\; \mu_{20}-\mu_{02}\bigr).
\end{equation}
The eigenvalues of the (normalized) covariance matrix built from $\mu_{20}, \mu_{11}, \mu_{02}$ give the axis lengths of the equivalent ellipse, plotted in Figure~\ref{fig:l05-2}.

\section{Hu moments: a shape descriptor invariant to pose}

\begin{figure}[h]
\centering
\includegraphics[width=0.95\textwidth]{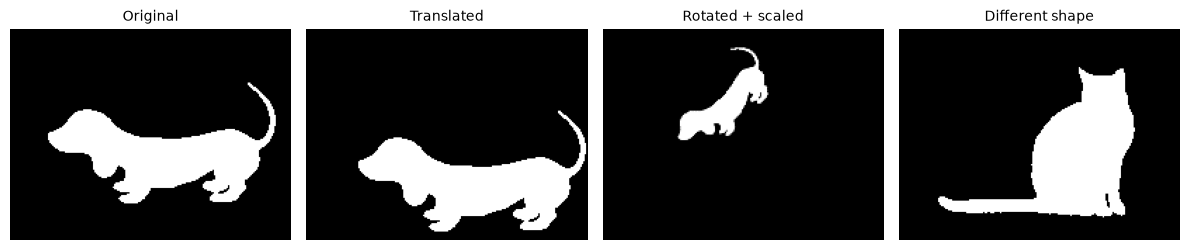}
\caption{Original, translated, rotated $+$ scaled, and a different shape entirely. \attrib{Images: \href{https://www.publicdomainpictures.net}{Public Domain Pictures}.}}
\label{fig:l05-3}
\end{figure}

\textbf{Hu moments} (\code{cv2.HuMoments}) are seven values calculated from the central moments that stay (nearly) the same regardless of the shape's position, size, and rotation --- useful for comparing two shapes without first aligning them. To test this, three versions of a dog silhouette (original, translated, and rotated $+$ scaled) are compared against each other and against a different shape (a cat silhouette), all shown in Figure~\ref{fig:l05-3}.

\begin{lstlisting}
def hu_log(im):
    m = cv2.moments(im, binaryImage=True)
    hu = cv2.HuMoments(m).flatten()
    return -np.sign(hu) * np.log10(np.abs(hu) + 1e-30)  # log-scale: raw values span many orders of magnitude
\end{lstlisting}

The translated and rotated+scaled versions yield similar (log-scaled) Hu moments to the original, whereas the different shape yields markedly different values --- exactly the invariance property that makes Hu moments useful for shape matching.

\subsection*{Exercises}
\begin{enumerate}
  \item Use \code{cv2.findContours} to get the outline of a blob from Chapter~\ref{ch:lesson04}'s binary image, then call \code{cv2.moments} on the \emph{contour} instead of the full binary mask. Compare the centroid to the one computed here.
  \item Draw a shape that is mirror-flipped rather than rotated. Are its Hu moments still close to the original? (Hint: think about what determinant/parity information moments do or do not capture.)
\end{enumerate}

\vfill
\fullcode{lesson05}{lesson05_moments}

%% file: chapters/lesson06.tex
\chapter{Eigenvectors and Eigenvalues of the Moment Covariance Matrix}
\label{ch:lesson06}

In Chapter~\ref{ch:lesson05} we used central moments to find a blob's orientation and, with a bit of extra algebra, its semi-axis lengths. This chapter makes that connection explicit and general: the central moments of \emph{any} blob (not just an ellipse) define a $2\times2$ \textbf{covariance matrix}, and the eigenvectors/eigenvalues of that matrix directly give the orientation and size of the \emph{equivalent ellipse} --- the ellipse with the same area and same second-moment spread as the blob.

\section{Refresher: what are eigenvectors and eigenvalues?}

For a square matrix $A$, a nonzero vector $v$ is an \textbf{eigenvector} with \textbf{eigenvalue} $\lambda$ if
\begin{equation}
Av = \lambda v.
\end{equation}
In words: applying $A$ to $v$ doesn't rotate $v$ off its own line --- it only \emph{scales} it, by a factor of $\lambda$. Most vectors get both rotated and scaled by $A$; eigenvectors are the special directions that only get scaled.

For a $2\times2$ \textbf{symmetric} matrix (like the covariance matrices built below), something even nicer is guaranteed: there are always two eigenvectors, they are perpendicular to each other, and their eigenvalues are real numbers. Geometrically, $A$ takes a circle of unit vectors and stretches it into an ellipse whose axes point along the eigenvectors, with each semi-axis length equal to the corresponding eigenvalue.

\begin{lstlisting}
A = np.array([[3.0, 1.0], [1.0, 1.5]])
eigvals, eigvecs = np.linalg.eigh(A)  # ascending order
\end{lstlisting}

\begin{figure}[h]
\centering
\includegraphics[width=0.42\textwidth]{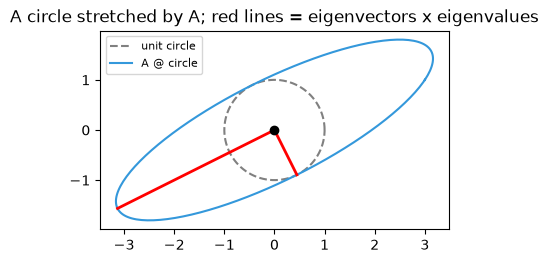}
\caption{A unit circle stretched by $A$; the red lines are the eigenvectors scaled by their eigenvalues.}
\label{fig:l06-1}
\end{figure}

Figure~\ref{fig:l06-1} shows this geometrically.

\subsection*{From one eigenvector to both at once: diagonalization}

$Av=\lambda v$ holds for a single eigenvector. Stack \emph{both} eigenvectors as the columns of $P=\begin{bmatrix}v_1 & v_2\end{bmatrix}$, and both eigenvalues on the diagonal of $\Lambda$. Then $Av_1=\lambda_1 v_1$ and $Av_2=\lambda_2 v_2$, side by side, become one matrix equation:
\begin{equation}
AP = P\Lambda.
\end{equation}
Because $A$ is symmetric, its eigenvectors are orthonormal, which makes $P$ an \textbf{orthogonal} matrix ($P^{-1}=P^\top$). That lets us solve for $A$ itself:
\begin{equation}
A = P\Lambda P^\top,
\end{equation}
the \textbf{diagonalization} of $A$. Geometrically this is three steps: $P^\top$ \emph{rotates} coordinates into the eigenvector frame, $\Lambda$ \emph{scales} along those (now axis-aligned) directions independently, and $P$ \emph{rotates back}. When $A$ is a covariance matrix, the off-diagonal entry that mixes $x$ and $y$ is exactly $\mu_{11}$: diagonalizing the covariance matrix and finding the ellipse's own natural axes are the same operation.

\section{From central moments to a covariance matrix}

Treat a blob's pixels as samples from a 2D distribution. Its covariance matrix, in terms of the central moments $\mu_{ij}$ and area $m_{00}$, is
\begin{equation}
A = \frac{1}{m_{00}}\begin{bmatrix}\mu_{20} & \mu_{11} \\ \mu_{11} & \mu_{02}\end{bmatrix}.
\end{equation}
This is exactly the covariance-matrix formula for a scatter of points $(x,y)$, just weighted by pixel membership instead of sample index. Diagonalizing it gives eigenvectors along the blob's own major and minor axes and eigenvalues measuring the spread along each. This same operation, applied to a covariance matrix of general (not necessarily 2D pixel) data, is called \textbf{principal component analysis (PCA)}: the eigenvectors ranked by eigenvalue are the \emph{principal components}, often used to reduce a high-dimensional dataset down to just its most informative few directions.

\begin{lstlisting}
def covariance_from_moments(binary):
    m = cv2.moments(binary, binaryImage=True)
    cx, cy = m['m10'] / m['m00'], m['m01'] / m['m00']
    cov = np.array([[m['mu20'], m['mu11']],
                     [m['mu11'], m['mu02']]]) / m['m00']
    return cov, (cx, cy)
\end{lstlisting}

\section{Eigen decomposition gives orientation and size}

For a filled ellipse with semi-axes $a \ge b$, the eigenvalues of its covariance matrix work out to $\lambda_{\max}=a^2/4$ and $\lambda_{\min}=b^2/4$, so
\begin{equation}
a = 2\sqrt{\lambda_{\max}}, \qquad b = 2\sqrt{\lambda_{\min}},
\end{equation}
with the corresponding eigenvectors pointing along the major and minor axes. The angle of the major axis matches the angle computed from central moments directly in Chapter~\ref{ch:lesson05}.

\begin{lstlisting}
def principal_axes(binary):
    cov, center = covariance_from_moments(binary)
    eigvals, eigvecs = np.linalg.eigh(cov)
    semi_axes = 2 * np.sqrt(np.clip(eigvals, 0, None))
    order = [1, 0]  # largest eigenvalue first
    return center, semi_axes[order], eigvecs[:, order].T
\end{lstlisting}

\begin{figure}[h]
\centering
\includegraphics[width=0.4\textwidth]{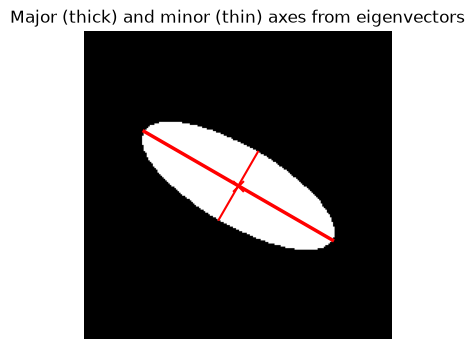}
\caption{Major (thick) and minor (thin) axes recovered from a known ellipse's eigenvectors: $a=70.6$, $b=25.5$, matching the drawn semi-axes (70, 25).}
\label{fig:l06-2}
\end{figure}

Figure~\ref{fig:l06-2} confirms the recovered axes against the known ellipse.

\section{It works on arbitrary shapes, too}

\begin{figure}[h]
\centering
\includegraphics[width=0.42\textwidth]{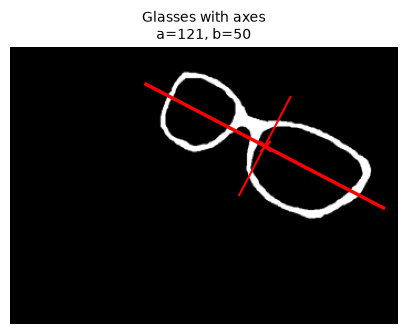}
\caption{Equivalent-ellipse axes for a pair of eyeglasses. The ellipse summarizes the mass distribution rather than tracing the outline, so it cuts across the bridge between lenses. \attrib{Photo: Stan Birchfield.}}
\label{fig:l06-3}
\end{figure}

The real power of this approach is that it doesn't require the blob to be an ellipse at all. Every binary shape has \emph{some} equivalent ellipse --- the one that matches its area, centroid, and second-moment spread, giving a compact 5-number summary (center, two semi-axis lengths, orientation) of an arbitrarily shaped blob, as in Figure~\ref{fig:l06-3}.

\section{Eccentricity: how elongated is a shape?}

The ratio of the eigenvalues (or semi-axes) gives a single number describing elongation, independent of orientation and overall size:
\begin{equation}
\text{eccentricity} = \sqrt{1 - \frac{\lambda_{\min}}{\lambda_{\max}}}.
\end{equation}
This ranges from 0 (a circle, both axes equal) to nearly 1 (a very thin, elongated shape). The glasses shape above ($a=120.9$, $b=49.9$) has eccentricity $0.911$ --- clearly elongated, as expected of a pair of glasses lying flat.

\subsection*{Exercises}
\begin{enumerate}
  \item Draw a perfect circle and confirm its eccentricity is (close to) 0.
  \item Overlay the \emph{actual} fitted ellipse outline (not just the axis lines) using \code{cv2.ellipse} with the center, \code{(2a, 2b)} as the full axes lengths, and the orientation angle from the eigenvectors. Compare it to \code{cv2.fitEllipse} applied to the shape's contour --- do they agree?
  \item Run \code{principal\_axes} on the filled blob of the glasses (after flood fill). How much do $a$, $b$, and the eccentricity change? Does that match your intuition for how moments depend on \emph{where} mass sits, not just the overall silhouette?
\end{enumerate}

\vfill
\fullcode{lesson06}{lesson06_eigen_covariance}

%% file: chapters/lesson07.tex
\chapter{Distance Measures}
\label{ch:lesson07}

``Distance'' on a pixel grid is not as simple as it sounds. This chapter looks at three different flavors:
\begin{enumerate}
  \item \textbf{Point-to-point distance metrics} (Manhattan, chessboard, and Euclidean) --- which measure how far apart two pixels are.
  \item \textbf{Curve-length estimators} (Freeman and a refined weighted estimator) --- which measure the length of a \emph{digitized curve} (e.g.\ the boundary of a blob) from its chain code.
  \item \textbf{Chamfer distance transform} --- a fast, image-wide approximation that gives every pixel its distance to the nearest member of a given set (e.g.\ the foreground), computed with two raster-scan passes.
\end{enumerate}

\section{Point-to-point distance metrics}

Given two pixels $p=(x_1,y_1)$ and $q=(x_2,y_2)$, let $d_x=x_2-x_1$ and $d_y=y_2-y_1$. Three common metrics:
\begin{itemize}
  \item \textbf{Manhattan / city-block ($L_1$)}: distance if you can move only along the grid axes.
    \begin{equation}
    D_4(p,q) = |d_x| + |d_y|
    \end{equation}
  \item \textbf{Chessboard ($L_\infty$)}: number of king moves on a chessboard (diagonal steps are free).
    \begin{equation}
    D_8(p,q) = \max(|d_x|,\,|d_y|)
    \end{equation}
  \item \textbf{Euclidean ($L_2$)}: ordinary straight-line distance.
    \begin{equation}
    D_E(p,q) = \sqrt{d_x^2+d_y^2}
    \end{equation}
\end{itemize}
The names $D_4$ and $D_8$ come from each being the shortest path length when only 4-connected (axis) or 8-connected (axis $+$ diagonal) moves are allowed, one unit per move.

\begin{lstlisting}
def manhattan(dx, dy):  return np.abs(dx) + np.abs(dy)
def chessboard(dx, dy): return np.maximum(np.abs(dx), np.abs(dy))
def euclidean(dx, dy):  return np.sqrt(dx**2 + dy**2)
\end{lstlisting}

\begin{figure}[h]
\centering
\includegraphics[width=0.95\textwidth]{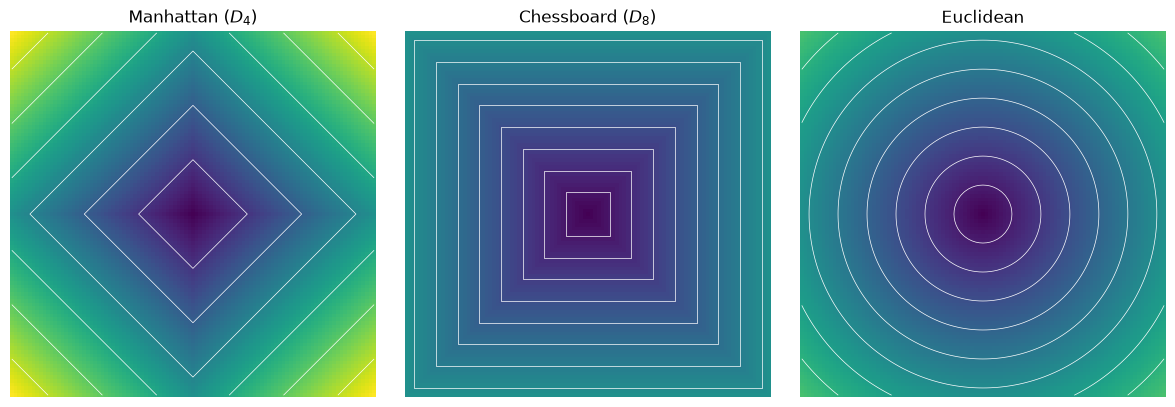}
\caption{Distance fields for the three metrics, contoured. Manhattan distance forms diamonds, chessboard distance forms squares, and Euclidean distance forms circles --- the shape most people intuitively think of as ``equal distance away.''}
\label{fig:l07-1}
\end{figure}

For every pixel in a grid, computing its distance to the center pixel under each metric and displaying it as an image (Figure~\ref{fig:l07-1}) shows the metrics agree only along the axes --- everywhere else they diverge.

\section{Estimating curve length from a chain code}

Now suppose we want to estimate the length of a \emph{curve}, not just the distance between two points. A test circle of known radius $60$ has known circumference $120\pi \approx 377$; its number of boundary pixels ($336$) is already a poor estimate, since $336 \ne 377$.

A \textbf{Freeman chain code} describes a digital curve as a sequence of unit steps in 8 possible directions ($0$--$7$, $45^\circ$ apart). Four directions are \emph{even} (horizontal/vertical, true length 1) and four are \emph{odd} (diagonal, true length $\sqrt2$). If a curve's chain code has $N_e$ even steps and $N_o$ odd steps, \textbf{Freeman's estimate} of its length is
\begin{equation}
L_{\text{Freeman}} = N_e + \sqrt{2}\,N_o.
\end{equation}
This systematically \emph{overestimates} smooth curves: a digitized circle's boundary alternates through many short zig-zag ``staircase'' steps that add up to more than the true arc length. Replacing the exact weights $(1,\sqrt2)$ with weights $(a,b)$ tuned to minimize average error over many digitized curves --- an idea sometimes associated with Kimura's chain-code length correction, with commonly cited values $a\approx0.948$, $b\approx1.343$ --- gives
\begin{equation}
L_{\text{corrected}} = a\,N_e + b\,N_o.
\end{equation}

\begin{lstlisting}
steps = np.diff(np.vstack([pts, pts[:1]]), axis=0)
is_diagonal = (steps[:, 0] != 0) & (steps[:, 1] != 0)
N_o, N_e = int(is_diagonal.sum()), int((~is_diagonal).sum())

L_freeman   = N_e + np.sqrt(2) * N_o
L_corrected = 0.948 * N_e + 1.343 * N_o
\end{lstlisting}

For radius 60 (true circumference $377.0$, $192$ even steps and $144$ odd steps): the Freeman estimate comes to $395.6$ (an error of $+4.9\%$), while the corrected estimate comes to $375.4$ (an error of just $-0.4\%$) --- much closer to the truth, without needing anything more than a count of even and odd steps.

\section{Chamfer distance transform}

Unlike the point-to-point metrics in Part 1, which measure the distance between two specific pixels, a \textbf{distance transform} labels \emph{every} pixel of a binary image with its distance to the nearest member of a given set of pixels --- typically the foreground, or the background --- useful for skeletonization, shape matching, and path planning. Computing it \emph{exactly} with Euclidean distance requires comparing every pixel against every pixel in that set (slow), or a more careful algorithm.

The \textbf{chamfer distance transform} is a fast approximation: instead of a single global search, it propagates local distances across the image in just two raster-scan passes (top-left to bottom-right, then bottom-right to top-left), adding a small fixed cost for each step to a neighbor. Using integer weights $a=3$ for axis-aligned neighbors and $b=4$ for diagonal neighbors (then dividing by 3 at the end) gives the classic \textbf{3-4 chamfer distance} --- the same idea as the Freeman/corrected chain-code weights above, but applied to every pixel instead of just a boundary. \code{cv2.distanceTransform} with \code{DIST\_L2} computes this; \code{maskSize} trades accuracy for speed.

\begin{lstlisting}
im_dist3 = cv2.distanceTransform(img, cv2.DIST_L2, maskSize=3)  # 3-4 chamfer
im_dist5 = cv2.distanceTransform(img, cv2.DIST_L2, maskSize=5)  # more accurate
\end{lstlisting}

\begin{figure}[h]
\centering
\includegraphics[width=0.95\textwidth]{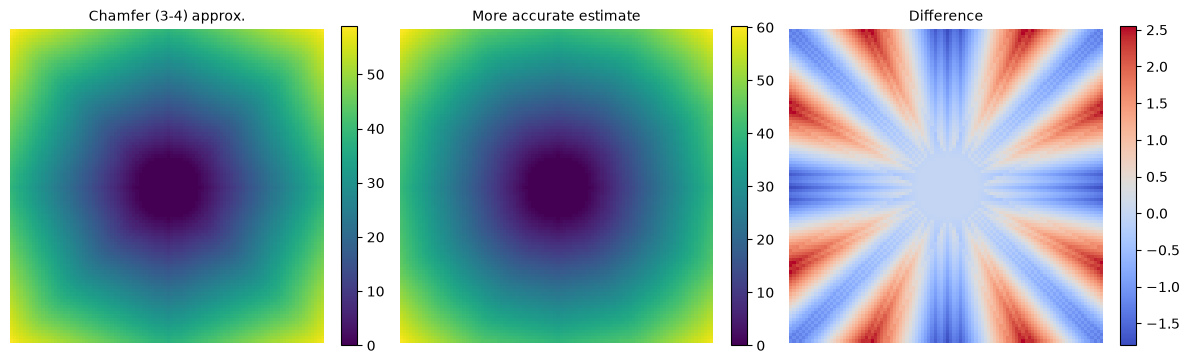}
\caption{Chamfer (mask size 3) vs.\ a more accurate distance transform (mask size 5) and their difference; the two agree closely (max error under half a pixel).}
\label{fig:l07-2}
\end{figure}

Figure~\ref{fig:l07-2} compares the two mask sizes.

\subsection*{Exercises}
\begin{enumerate}
  \item Repeat the circle experiment for a few different radii (e.g.\ 10, 20, 40, 80). Does the Freeman estimate's percent error stay roughly constant, or does it shrink as the circle gets bigger? Why might that be?
  \item The chessboard distance $D_8$ is the exact shortest path length when diagonal moves are allowed and cost 1. Can you construct a modified chessboard-like metric, using a diagonal cost of $\sqrt2$ instead of 1, that behaves like a coarse local version of the chain-code length estimators above? Compare it to true Euclidean distance for a few points.
\end{enumerate}

\vfill
\fullcode{lesson07}{lesson07_distance_measures}

%% file: chapters/lesson08.tex
\chapter{Geometric Transformations}
\label{ch:lesson08}

This chapter covers how to move, resize, and reshape images: flipping, cropping, rotating, and scaling, and then the more general hierarchy of 2D transformations --- \textbf{Euclidean}, \textbf{similarity}, and \textbf{affine} --- that those operations are all special cases of.

\section{A test image with no symmetry}

\begin{figure}[h]
\centering
\includegraphics[width=0.3\textwidth]{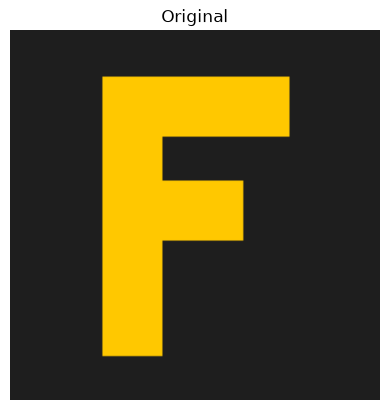}
\caption{The test image: an asymmetric letter F.}
\label{fig:l08-1}
\end{figure}

A letter ``F'' (Figure~\ref{fig:l08-1}) has no rotational or mirror symmetry --- every flip and rotation produces a visibly different result, which makes it easy to tell the transformations apart.

\section{Basic operations}

\begin{figure}[h]
\centering
\includegraphics[width=0.9\textwidth]{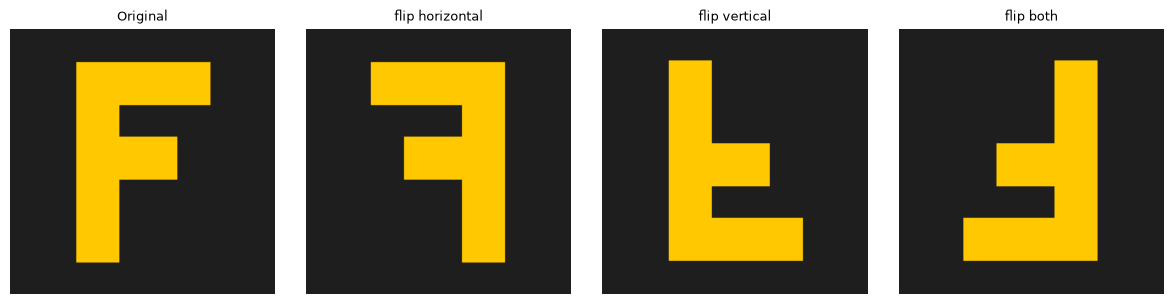}
\caption{Horizontal, vertical, and both-axis flips.}
\label{fig:l08-2}
\end{figure}

Some basic operations:

\textbf{Flipping.} \code{cv2.flip} mirrors an image: code \code{1} flips horizontally, \code{0} flips vertically, \code{-1} flips both (Figure~\ref{fig:l08-2}).

\begin{figure}[h]
\centering
\includegraphics[width=0.3\textwidth]{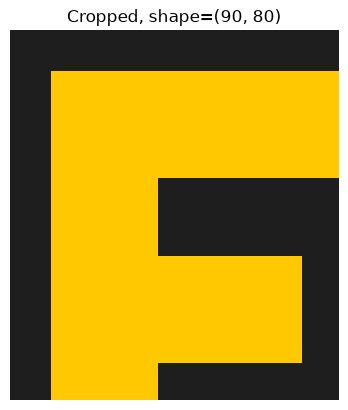}
\caption{A crop, obtained by plain NumPy slicing.}
\label{fig:l08-3}
\end{figure}

\textbf{Cropping} is just array slicing --- no OpenCV function needed: \code{image[y0:y1, x0:x1]} keeps rows \code{y0..y1} and columns \code{x0..x1} (Figure~\ref{fig:l08-3}).

\textbf{Scaling (resizing).} \code{cv2.resize} changes an image's dimensions, optionally with different factors for width and height. The ratio of width to height is the \textbf{aspect ratio}. Uniform scaling preserves shape; non-uniform scaling stretches or squashes it.

\begin{lstlisting}
uniform    = cv2.resize(img, None, fx=0.5, fy=0.5)
stretched  = cv2.resize(img, None, fx=1.5, fy=0.6)
\end{lstlisting}

\begin{figure}[h]
\centering
\includegraphics[width=0.85\textwidth]{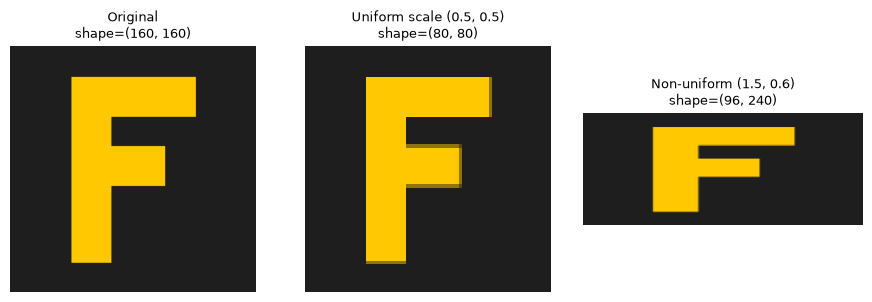}
\caption{Uniform scaling preserves aspect ratio; non-uniform scaling does not.}
\label{fig:l08-4}
\end{figure}

Figure~\ref{fig:l08-4} shows both.

\textbf{Rotating.} \code{cv2.getRotationMatrix2D} builds a $2\times3$ matrix for rotating by an arbitrary angle around a chosen center, which \code{cv2.warpAffine} then applies to the image.

\begin{lstlisting}
M = cv2.getRotationMatrix2D(center, angle=25, scale=1.0)
rotated = cv2.warpAffine(img, M, (w, h))
\end{lstlisting}

\begin{figure}[h]
\centering
\includegraphics[width=0.3\textwidth]{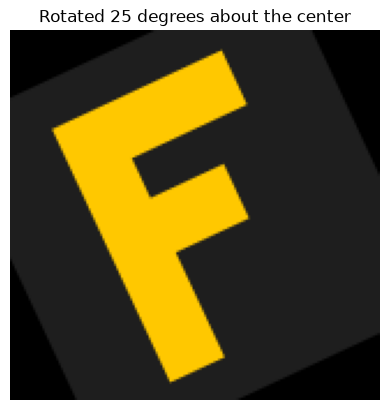}
\caption{Rotation about the image center.}
\label{fig:l08-5}
\end{figure}

Figure~\ref{fig:l08-5} shows the result.

\section{The transformation hierarchy}

Flipping, rotating, scaling, and translating are all instances of a general $2\times3$ matrix transform
\begin{equation}
\begin{bmatrix}x'\\y'\end{bmatrix} = \begin{bmatrix}a & b\\c & d\end{bmatrix}\begin{bmatrix}x\\y\end{bmatrix} + \begin{bmatrix}t_x\\t_y\end{bmatrix}
\end{equation}
applied with \code{cv2.warpAffine}. Restricting the $2\times2$ part in different ways gives a nested family of transformations, from most to least restrictive:

\begin{center}
\begin{tabular}{lll}
\hline
\textbf{Transform} & \textbf{Free parameters} & \textbf{Preserves} \\
\hline
Euclidean (rigid) & rotation $\theta$, translation & lengths, angles \\
Similarity & \quad + uniform scale $s$ & angles, ratios of lengths \\
Affine & any invertible $2{\times}2$ matrix & parallelism, ratios along a line \\
\hline
\end{tabular}
\end{center}

\begin{figure}[h]
\centering
\includegraphics[width=0.85\textwidth]{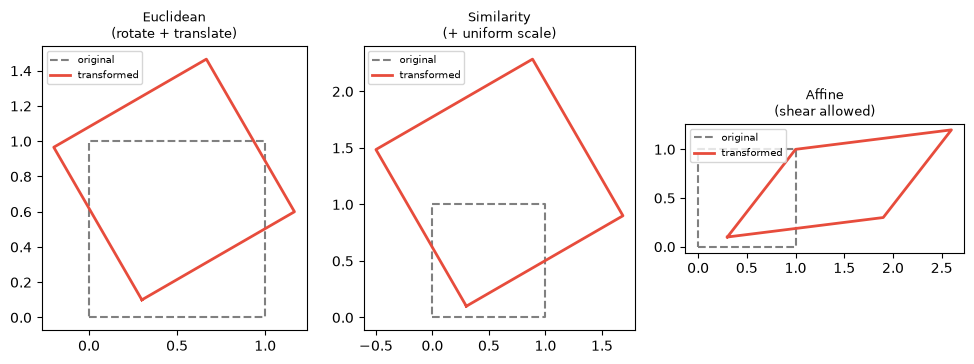}
\caption{A unit square under each transform. Only the affine square stops looking like a (possibly resized) square --- its corners are no longer $90^\circ$, because affine transforms allow shear. All three keep opposite sides parallel. (The more general projective transforms, e.g.\ camera perspective, are covered in a later chapter.)}
\label{fig:l08-6}
\end{figure}

Two things change going down the table, in opposite directions. The \emph{set of allowed transforms} grows: every Euclidean transform is a special case of a similarity transform (just $s=1$), and every similarity transform is a special case of an affine transform (just a rotation-and-scale matrix instead of an arbitrary one). In exchange for that extra flexibility, the \emph{properties guaranteed to survive} shrink: Euclidean preserves both lengths and angles; similarity gives up preserving lengths themselves (only their ratios survive); affine gives up angles too. Affine transforms are the most general of the three: they can shear a square into a parallelogram, something neither Euclidean nor similarity transforms can do, as Figure~\ref{fig:l08-6} shows.

\subsection*{Fitting an affine transform from point correspondences}

In practice we often don't know the transform matrix directly --- e.g., a flat object was photographed \emph{at an angle}, and we want to undo the distortion. This requires identifying a few landmark points in both the crooked photo and where they \emph{should} be in a canonical, front-on view. \code{cv2.getAffineTransform} takes 3 such point correspondences (the minimum needed to determine all 6 affine parameters) and returns the matrix that maps one set onto the other --- solve for the warped-to-canonical direction, and applying it \emph{rectifies} the photo.

\begin{lstlisting}
M_recovered = cv2.getAffineTransform(warped_pts, canonical_pts)
rectified = cv2.warpAffine(warped, M_recovered, (w, h))
\end{lstlisting}

\begin{figure}[h]
\centering
\includegraphics[width=0.95\textwidth]{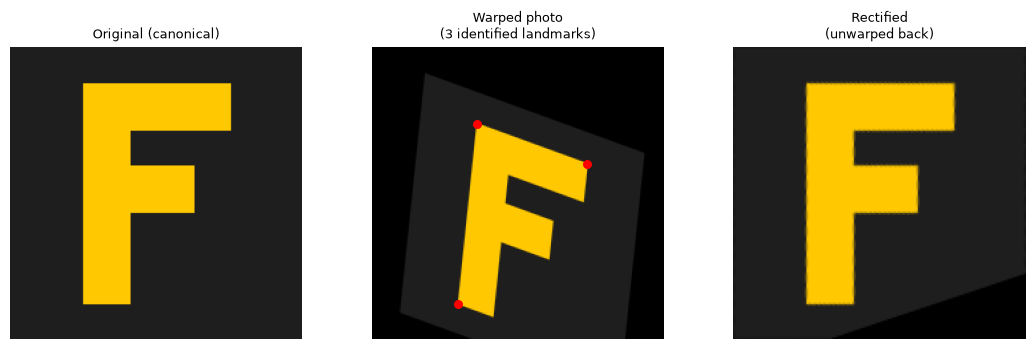}
\caption{Original, a simulated skewed photo with 3 identified landmarks, and the rectified result --- mean absolute pixel error against the true original is just $3.14$ (out of 255).}
\label{fig:l08-7}
\end{figure}

Figure~\ref{fig:l08-7} shows the original, the simulated photo, and the rectified result.

\subsection*{Exercises}
\begin{enumerate}
  \item \code{cv2.flip} is not part of the affine family written in matrix form above (rotation matrices $R$ always have determinant $+1$). What determinant does a flip's $2\times2$ matrix have? Construct the $2\times2$ matrix for \code{flip(1)} and check.
  \item Build a similarity transform matrix with $s=1$ and confirm it matches a pure Euclidean transform --- i.e.\ similarity is a strict generalization of Euclidean, not a different family.
  \item Modify the affine matrix in the unit-square example so that opposite sides are \emph{not} parallel. What has to break for that to happen? (Hint: this is exactly the boundary that separates affine from projective transforms.)
\end{enumerate}

\vfill
\fullcode{lesson08}{lesson08_geometric_transformations}

%% file: chapters/lesson09.tex
\chapter{Image Warping, Inverse Mapping, and Bilinear Interpolation}
\label{ch:lesson09}

\begin{figure}[h]
\centering
\includegraphics[width=0.3\textwidth]{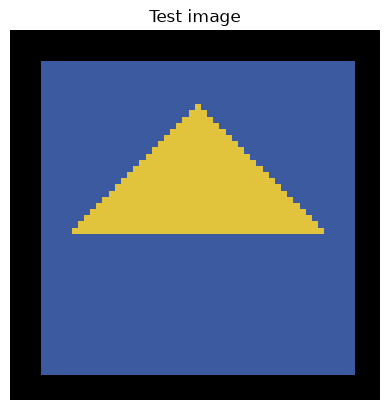}
\caption{The test image used throughout this chapter.}
\label{fig:l09-1}
\end{figure}

Chapter~\ref{ch:lesson08} built transformation matrices; this chapter is about actually \emph{applying} one to resample an image. The naive way to do this --- push every source pixel to its transformed location --- turns out to be broken. We see why, fix it with \textbf{inverse mapping}, and then deal with the fact that inverse mapping lands on fractional pixel coordinates that need to be \emph{interpolated}. Figure~\ref{fig:l09-1} shows the test image used throughout.

\section{The problem with forward mapping}

The obvious way to warp an image: for every source pixel $(x,y)$, compute its destination $(x',y')=M(x,y)$ and copy the pixel value there. This is called \textbf{forward mapping}. It has a fundamental flaw: whereas the source pixels form a \emph{complete} grid, their transformed locations generally do not. When a transform enlarges the image (or rotates it, which locally stretches the grid along the diagonal), the destination locations spread out and leave gaps --- no source pixel happens to land exactly on some destination pixels.

\begin{lstlisting}
dst_pts = M @ src_pts
dxi, dyi = np.round(dst_pts[0]).astype(int), np.round(dst_pts[1]).astype(int)
valid = (dxi >= 0) & (dxi < w) & (dyi >= 0) & (dyi < h)
forward[dyi[valid], dxi[valid]] = img[ys.ravel()[valid], xs.ravel()[valid]]
\end{lstlisting}

\begin{figure}[h]
\centering
\includegraphics[width=0.3\textwidth]{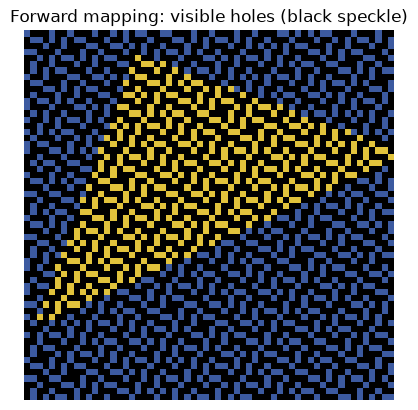}
\caption{Forward mapping: visible holes (black speckles) where no source pixel happened to land.}
\label{fig:l09-2}
\end{figure}

For a $25^\circ$ rotation combined with a $1.6\times$ scale, only $1408$ of $3600$ destination pixels ($39\%$) receive a value, leaving visible holes (Figure~\ref{fig:l09-2}).

\section{The fix: inverse mapping}

Instead of asking ``where does each source pixel go?'', ask the opposite question for every \emph{destination} pixel: ``where in the source image did this come from?'' That means applying the \textbf{inverse} transform $M^{-1}$ to each destination coordinate. Since we now iterate over a complete destination grid, every output pixel is guaranteed to get a value --- no holes, by construction. The catch: $M^{-1}(x',y')$ almost never lands exactly on an integer source coordinate, so a value must be \emph{interpolated} from the surrounding source pixels.

\begin{lstlisting}
M_inv = cv2.invertAffineTransform(M)
src_coords = M_inv @ dst_grid   # fractional source coordinates for every destination pixel
\end{lstlisting}

\subsection*{Nearest-neighbor interpolation}

\begin{figure}[h]
\centering
\includegraphics[width=0.3\textwidth]{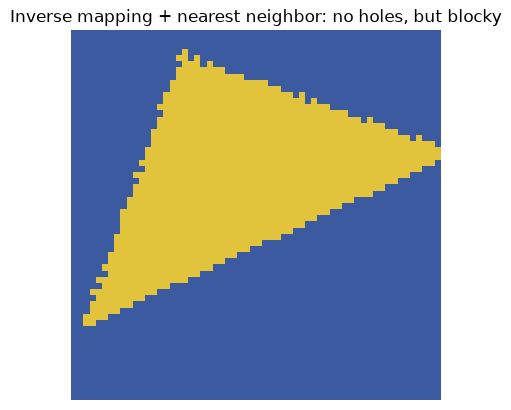}
\caption{Inverse mapping with nearest-neighbor sampling: no holes, but blocky.}
\label{fig:l09-3}
\end{figure}

The simplest option: round to the nearest integer source pixel. This has no holes, but is blocky --- many destination pixels round to the \emph{same} source pixel, so the image looks pixelated wherever the transform enlarges it (Figure~\ref{fig:l09-3}).

\subsection*{Bilinear interpolation}

Instead of snapping to the single nearest pixel, \textbf{bilinear interpolation} blends the 4 neighboring pixels, weighted by how close it is to each one. Let $(x_0,y_0)$ be the integer pixel just below-left of $(x,y)$, and $d_x=x-x_0$, $d_y=y-y_0$, with $0\le d_x,d_y<1$. Then
\begin{equation}
I(x,y) \approx (1{-}d_x)(1{-}d_y)\,I(x_0,y_0) + d_x(1{-}d_y)\,I(x_1,y_0) + (1{-}d_x)\,d_y\,I(x_0,y_1) + d_x\,d_y\,I(x_1,y_1),
\end{equation}
where $x_1=x_0+1$, $y_1=y_0+1$. This is just two 1D linear interpolations (along $x$, then along $y$) composed --- hence \emph{bi}linear.

\begin{lstlisting}
Ia, Ib, Ic, Id = image[y0,x0], image[y0,x1], image[y1,x0], image[y1,x1]
out = (1-dx)*(1-dy)*Ia + dx*(1-dy)*Ib + (1-dx)*dy*Ic + dx*dy*Id
\end{lstlisting}

\begin{figure}[h]
\centering
\includegraphics[width=0.55\textwidth]{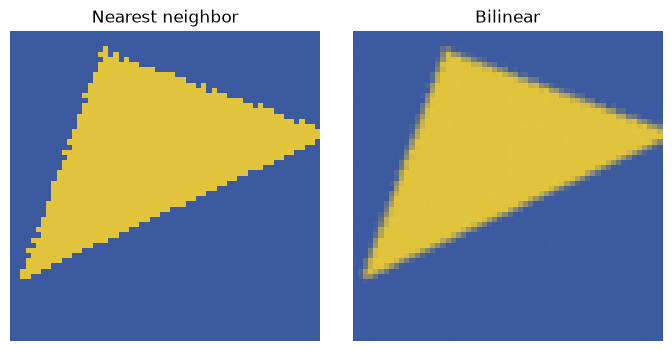}
\caption{Nearest-neighbor (left) vs.\ bilinear (right). Bilinear has smooth edges around the triangle and rectangle instead of jagged steps --- the classic trade-off is a softer, slightly blurrier image in exchange for removing aliasing artifacts.}
\label{fig:l09-4}
\end{figure}

Figure~\ref{fig:l09-4} compares the two interpolation methods.

\subsection*{Sanity check against OpenCV}

\begin{figure}[h]
\centering
\includegraphics[width=0.85\textwidth]{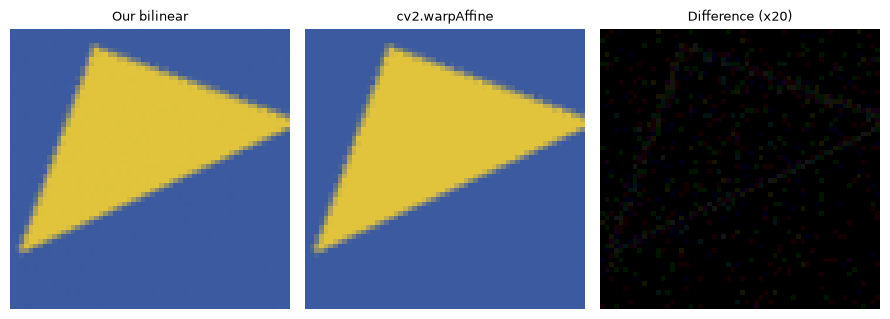}
\caption{Our bilinear implementation vs.\ \code{cv2.warpAffine(..., flags=cv2.INTER\_LINEAR)}: max pixel difference is $1$ (out of 255), mean difference $0.092$. The tiny remaining gap comes from OpenCV using fixed-point rounding internally for speed rather than full floating-point arithmetic.}
\label{fig:l09-5}
\end{figure}

\code{cv2.warpAffine} does exactly this --- inverse mapping plus interpolation --- internally, agreeing closely with the from-scratch version (Figure~\ref{fig:l09-5}).

\subsection*{Exercises}
\begin{enumerate}
  \item Why doesn't the forward-mapping holes problem happen when the transform only \emph{shrinks} the image? Try a scale of \code{0.5} instead of \code{1.6} in the forward-mapping example and see what fraction of destination pixels get filled.
  \item Extend \code{bilinear\_sample} to bicubic interpolation conceptually: instead of a $2\times2$ neighborhood, what neighborhood size would you need, and what property would you want the weights to satisfy at the sample points?
  \item Time \code{bilinear\_sample} against \code{cv2.warpAffine} on a larger image (e.g.\ $800\times800$) using \code{\%timeit}. By how much faster is the OpenCV version, and why?
\end{enumerate}

\vfill
\fullcode{lesson09}{lesson09_warping_interpolation}

%% file: chapters/lesson10.tex
\chapter{Convolution}
\label{ch:lesson10}

Convolution is the core operation behind blurring, sharpening, and edge detection --- and it's also the operation at the heart of a convolutional neural network's convolution layers (a later chapter). This chapter builds it from scratch, clarifies a common point of confusion (convolution vs.\ correlation), and shows a few classic filters.

\section{1D convolution}

Before tackling 2D images, start with the simpler problem of convolving two 1D arrays. For a signal $I$ and a kernel $K$, the discrete convolution is
\begin{equation}
(I * K)(x) = \sum_{i} K(i)\, I(x - i).
\end{equation}
Concretely: \textbf{flip} the kernel end-to-end, \textbf{slide} it along the signal one position at a time, and at each position take the dot product between the flipped kernel and the overlapping chunk of signal.

\begin{lstlisting}
def convolve1d(signal, kernel):
    ksize, pad = len(kernel), len(kernel) // 2
    padded = np.pad(signal, pad, mode='constant')
    flipped = kernel[::-1]
    out = np.zeros_like(signal, dtype=np.float64)
    for n in range(len(signal)):
        out[n] = np.dot(padded[n:n + ksize], flipped)
    return out
\end{lstlisting}

\begin{figure}[h]
\centering
\includegraphics[width=0.7\textwidth]{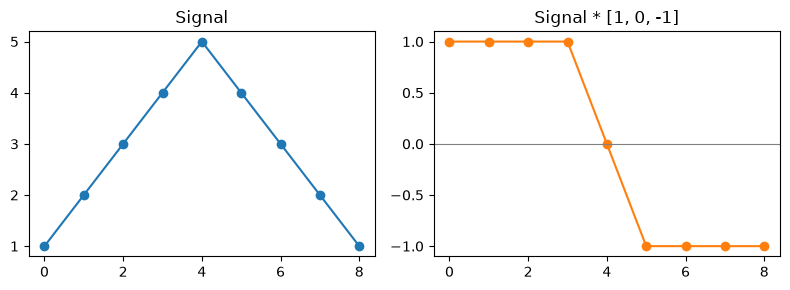}
\caption{A ramp-up-then-down signal and its convolution with a derivative kernel: positive while rising, zero at the peak, negative while falling.}
\label{fig:l10-1}
\end{figure}

With signal $[1,2,3,4,5,4,3,2,1]$ and a 3-tap derivative kernel $[-1,0,1]$ (scaled by $0.5$), the from-scratch result matches \code{np.convolve} exactly (Figure~\ref{fig:l10-1}).

\section{2D convolution}

The 2D version (what actually gets used on images) is the same flip-and-slide idea, over two axes:
\begin{equation}
(I * K)(x, y) = \sum_{i}\sum_{j} K(i, j)\, I(x - i,\, y - j).
\end{equation}
Just as in 1D, the key detail is the \emph{minus} signs: the kernel is flipped ($180^\circ$ rotated) before it's slid across the image, which matters whenever the kernel is not symmetric. \textbf{Correlation} is the same idea without the flip:
\begin{equation}
(I \star K)(x, y) = \sum_{i}\sum_{j} K(i, j)\, I(x + i,\, y + j).
\end{equation}
In practice, most image-processing libraries --- including OpenCV's \code{cv2.filter2D} --- actually implement \emph{correlation}, not convolution, even though people casually call it ``convolving with a kernel.'' For symmetric kernels (box blur, Gaussian) the two are identical, so the distinction rarely matters in practice. But it is worth knowing the difference exists.

\subsection*{Convolution from scratch}

Flip the kernel both horizontally and vertically, pad the image so the kernel never runs off the edge, then at every position multiply the flipped kernel elementwise by the pixels underneath it and sum. For speed, the implementation below loops over the kernel's (few) entries rather than the image's (many) pixels, shifting and accumulating the whole padded image at once for each kernel weight.

\begin{lstlisting}
def convolve2d(image, kernel, border=cv2.BORDER_REFLECT101):
    kh, kw = kernel.shape
    ph, pw = kh // 2, kw // 2
    flipped = kernel[::-1, ::-1]
    padded = cv2.copyMakeBorder(image, ph, ph, pw, pw, border)
    out = np.zeros(image.shape, dtype=np.float64)
    for i in range(kh):
        for j in range(kw):
            out += flipped[i, j] * padded[i:i+image.shape[0], j:j+image.shape[1]]
    return out
\end{lstlisting}

\begin{figure}[h]
\centering
\includegraphics[width=0.85\textwidth]{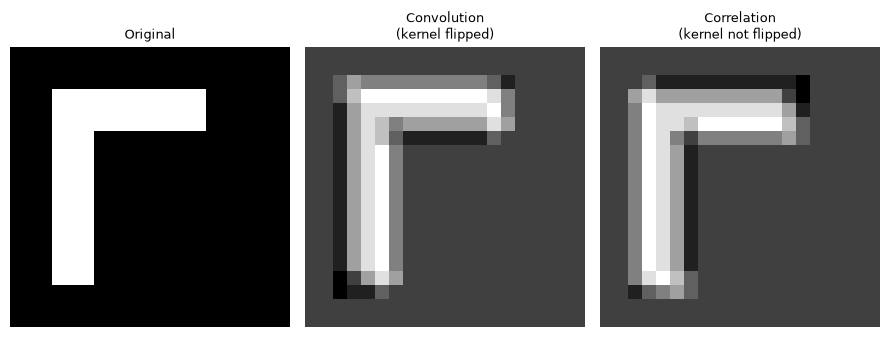}
\caption{An asymmetric test shape, its convolution (kernel flipped), and its correlation (kernel not flipped) with the same asymmetric kernel --- visibly different results.}
\label{fig:l10-2}
\end{figure}

Checked against \code{cv2.filter2D} fed the pre-flipped kernel (since \code{filter2D} computes correlation): max absolute difference is exactly $0.0$. Figure~\ref{fig:l10-2} also shows what correlation (no flip) gives instead.

\section{Boundary handling}

Near the edges, the kernel hangs off the image. \code{convolve2d} uses \code{cv2.copyMakeBorder} to pad first; the padding mode changes the result at the border. Common choices: constant (zero) padding, replicate (extend the edge pixel), and reflect (mirror across the edge; \code{REFLECT101} avoids duplicating the edge pixel itself).

\section{Classic filters as kernels}

\begin{figure}[h]
\centering
\includegraphics[width=0.95\textwidth]{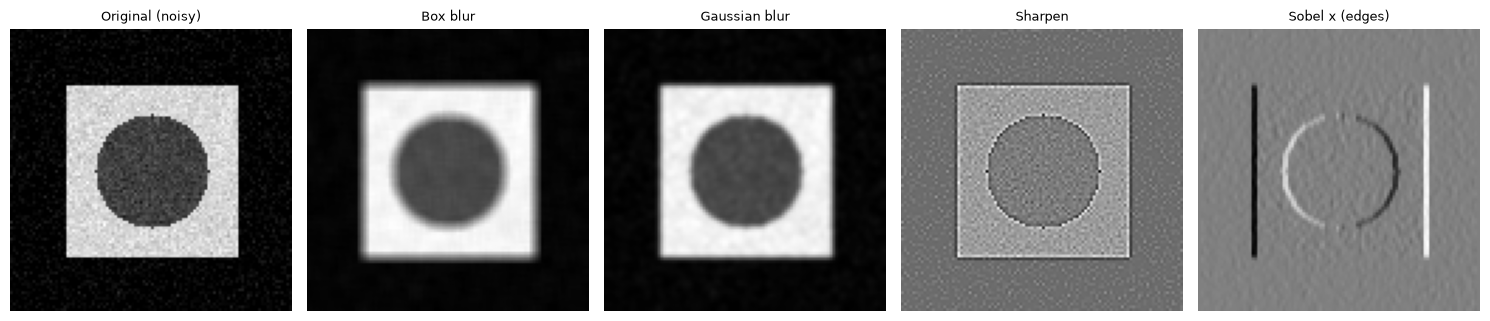}
\caption{A noisy synthetic photo, then box blur, Gaussian blur, sharpen, and Sobel-$x$ edge detection, all implemented with the same \code{convolve2d} function above.}
\label{fig:l10-3}
\end{figure}

Once convolution works, most filters are just ``pick a kernel'':
\begin{itemize}
  \item \textbf{Box blur}: every neighbor weighted equally --- averages out noise but blurs edges.
  \item \textbf{Gaussian blur}: neighbors weighted by a bell curve --- smoother falloff, less ``boxy'' artifacting than a box blur.
  \item \textbf{Sharpen}: boosts the center pixel relative to its neighbors.
  \item \textbf{Sobel (edge detection)}: an asymmetric kernel that responds strongly to intensity changes in one direction --- a case where the convolution-vs-correlation flip actually matters, since the kernel is asymmetric (results in a sign flip).
\end{itemize}
Figure~\ref{fig:l10-3} shows all four filters applied to the same noisy photo.

\subsection*{Separable kernels: a speed trick}

The Gaussian kernel above is \textbf{separable}: it can be written as the outer product of two 1D kernels, $K=k_x\,k_y^\top$. That means convolving with the full $n\times n$ kernel is equivalent to convolving with a $1\times n$ kernel, then an $n\times1$ kernel --- turning $O(n^2)$ work per pixel into $O(2n)$.

\begin{lstlisting}
separable_result = convolve2d(convolve2d(photo, k1d.reshape(1, -1)), k1d.reshape(-1, 1))
full_result = convolve2d(photo, gaussian)
# separable_result == full_result, to floating-point precision
\end{lstlisting}

\subsection*{Exercises}
\begin{enumerate}
  \item Verify that the box kernel is also separable, by writing it as an outer product of two 1D uniform kernels.
  \item \code{sobel\_x} above is not symmetric. Compute both the convolution and correlation of the noisy photo with it, and describe how the two outputs differ (hint: think about which direction of intensity change each responds to).
  \item Time \code{convolve2d} against \code{cv2.filter2D} on a $300\times300$ image with a $15\times15$ Gaussian kernel using \code{\%timeit}. Then time the separable two-pass version against the full 2D version. How much faster is each speedup?
\end{enumerate}

\vfill
\fullcode{lesson10}{lesson10_convolution}

%% file: chapters/lesson11.tex
\chapter{Smoothing and Gaussian Pyramids}
\label{ch:lesson11}

Chapter~\ref{ch:lesson10} introduced Gaussian blur as one convolution kernel among several. Here we look at \emph{why} blurring matters beyond just ``softening'' an image: it's the key ingredient that makes downsampling safe. That leads directly to the \textbf{Gaussian pyramid} --- a stack of progressively smaller, blurrier versions of an image, used throughout computer vision for multi-scale analysis.

\section{Why downsampling needs blurring first}

Naively shrinking an image by keeping every $k$-th pixel (``nearest-neighbor downsampling'') can produce \textbf{aliasing}: fine periodic detail that oscillates faster than the new pixel spacing can represent folds into a completely different, fake low-frequency pattern. A photo of a brick wall demonstrates this well --- the repeating rows of bricks and mortar lines are exactly the kind of fine, regular detail that aliases badly.

\begin{lstlisting}
factor = 4
naive_downsample = brick_wall[::factor, ::factor]
safe_downsample = cv2.GaussianBlur(brick_wall, (0, 0), sigmaX=factor/2)[::factor, ::factor]
\end{lstlisting}

\begin{figure}[h]
\centering
\includegraphics[width=0.95\textwidth]{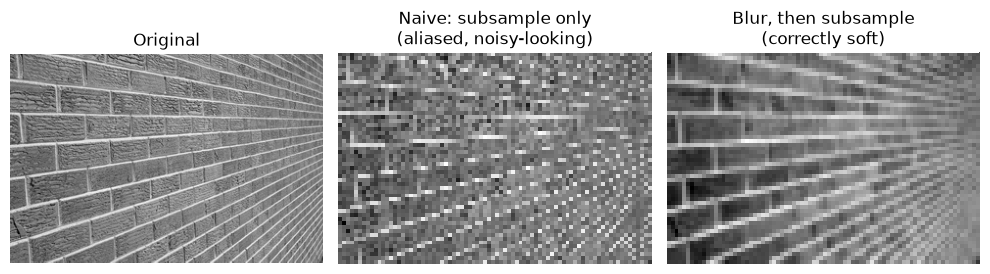}
\caption{Naive subsampling turns the regular brick pattern into fine, noise-like speckle --- a classic aliasing artifact (the same effect that makes a car's wheels look like they're spinning backwards on camera). Blurring first keeps the wall recognizable. \attrib{Photo: \href{https://www.publicdomainpictures.net/}{publicdomainpictures.net}.}}
\label{fig:l11-1}
\end{figure}

Figure~\ref{fig:l11-1} shows the difference directly.

\section{Choosing sigma and kernel size}

\begin{figure}[h]
\centering
\includegraphics[width=0.95\textwidth]{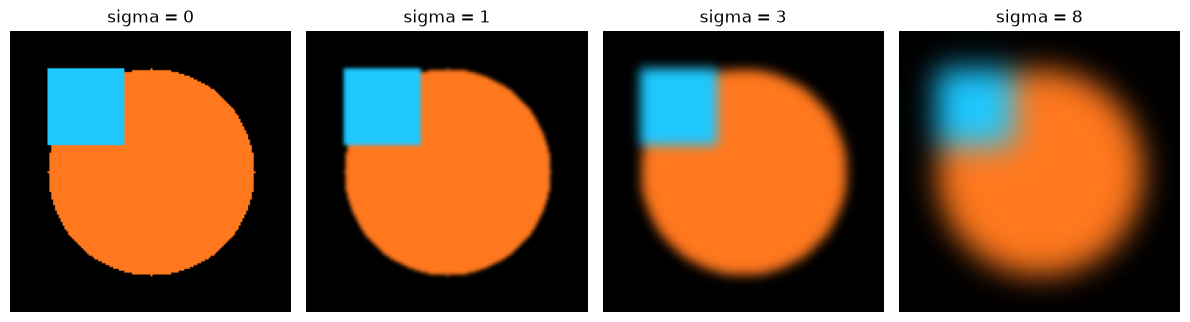}
\caption{Larger sigma averages over a wider neighborhood, removing finer detail, as long as the kernel is large enough to represent it.}
\label{fig:l11-2}
\end{figure}

Blurring correctly before downsampling needs two things: the right \code{sigma}, and a kernel \emph{large enough to represent that sigma}. Concretely, blurring a single pixel with a $3\times3$ Gaussian kernel is nothing more than a weighted average of its $3\times3$ neighborhood. For \code{sigma=1.0}, the manual weighted-sum computation matches \code{cv2.GaussianBlur}'s output for that pixel exactly (up to integer rounding: $139.40$ manually vs.\ $139$ from OpenCV). Figure~\ref{fig:l11-2} shows the effect of sigma at several values.

\subsection*{Kernel size has to keep up with sigma}

\begin{figure}[h]
\centering
\includegraphics[width=0.85\textwidth]{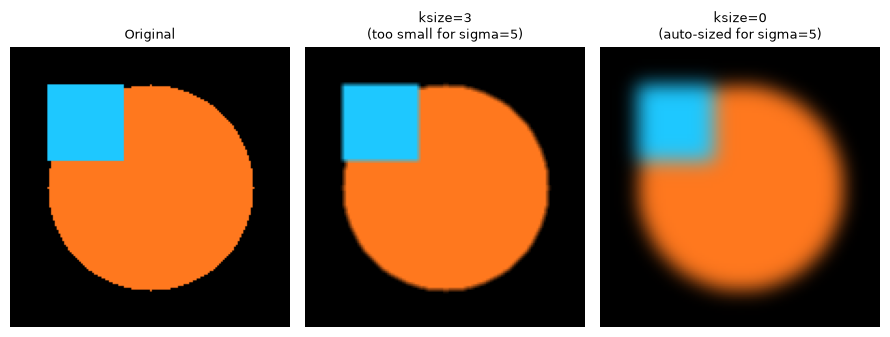}
\caption{A $3\times3$ kernel is far too small to represent \code{sigma=5}: mean absolute difference against the correctly auto-sized kernel is $9.31$ gray levels.}
\label{fig:l11-3}
\end{figure}

\code{cv2.getGaussianKernel} always renormalizes its weights to sum to 1, no matter how few taps it's given --- so a too-small kernel doesn't just clip the Gaussian's tails, it silently reshapes the whole kernel into something close to a plain box average. A 3-tap kernel built for \code{sigma=5.0} comes out as $[0.331,\ 0.338,\ 0.331]$ --- nearly identical to a plain box average $[0.33,0.33,0.33]$, the wide bell curve squashed away entirely (Figure~\ref{fig:l11-3}).

This is exactly why the pyramid code below (and the aliasing fix above) passes \code{ksize=(0, 0)}: letting OpenCV choose a kernel wide enough for the requested sigma, rather than risking a silently-too-small one.

\section{Building a Gaussian pyramid}

A \textbf{Gaussian pyramid} repeats ``blur, then downsample by 2'' over and over, producing a stack of images each half the width and height of the previous one. Each level is a properly anti-aliased, coarser view of the same scene --- not just a smaller crop.

\begin{lstlisting}
def gaussian_pyramid(image, num_levels, sigma=1.0):
    pyramid = [image]
    current = image
    for _ in range(num_levels - 1):
        blurred = cv2.GaussianBlur(current, (0, 0), sigmaX=sigma)
        current = blurred[::2, ::2]
        pyramid.append(current)
    return pyramid
\end{lstlisting}

\begin{figure}[h]
\centering
\includegraphics[width=0.95\textwidth]{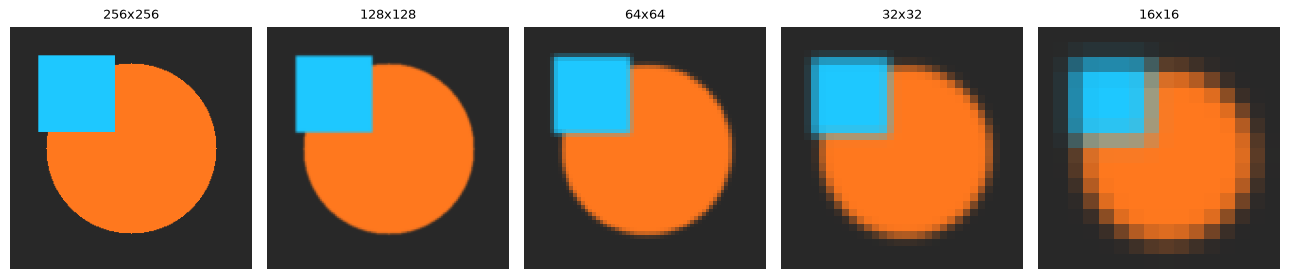}
\caption{A 5-level Gaussian pyramid: $256\times256$ down to $16\times16$.}
\label{fig:l11-4}
\end{figure}

Figure~\ref{fig:l11-4} shows the resulting pyramid.

\subsection*{Comparison to OpenCV's built-in \code{cv2.pyrDown}}

OpenCV provides \code{cv2.pyrDown}, which does the same blur-then-downsample idea using a fixed, carefully designed 5-tap binomial kernel (approximating a Gaussian) instead of an arbitrary sigma. The output sizes match ours exactly; pixel values are close but not identical, since the kernels differ slightly --- mean absolute difference grows from $0.00$ at the original size to just $0.19$ at $16\times16$.

\section{Pyramids lose information}

\begin{figure}[h]
\centering
\includegraphics[width=0.85\textwidth]{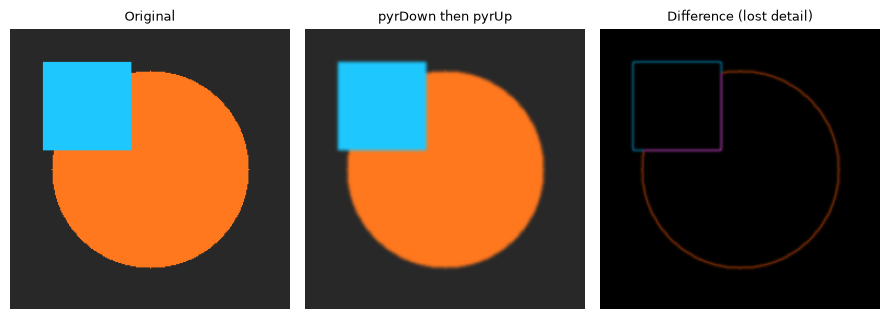}
\caption{\code{pyrDown} then \code{pyrUp} loses detail permanently: mean absolute reconstruction error is $1.59$ gray levels. The sharp edges of the circle and square come back soft and shifted-looking --- \code{pyrUp} (interpolation) can only guess, not restore, the discarded detail.}
\label{fig:l11-5}
\end{figure}

Downsampling is not reversible: upsampling a lower pyramid level back to the original size (\code{cv2.pyrUp}) cannot recover detail that blurring/subsampling discarded. This gap between an upsampled coarse level and the original is exactly what a \textbf{Laplacian pyramid} captures at each level (Chapter~\ref{ch:lesson13}), but the information loss is already visible directly, in Figure~\ref{fig:l11-5}.

\subsection*{Exercises}
\begin{enumerate}
  \item Repeat the brick-wall experiment with \code{factor = 2} instead of \code{4}. Does naive subsampling still show visible aliasing? Why might a smaller downsampling factor alias less?
  \item Gaussian pyramids are used for coarse-to-fine search (e.g.\ quickly finding an approximate object location at a small pyramid level, then refining at larger levels). Why would skipping the blur step and just subsampling directly break this strategy?
  \item Build a 6-level pyramid of a real photo-like image and measure how many levels it takes before the image becomes too small to recognize any shape at all. What real-world resolution would that correspond to for, say, a $1920\times1080$ photo?
\end{enumerate}

\vfill
\fullcode{lesson11}{lesson11_smoothing_pyramids}

%% file: chapters/lesson12.tex
\chapter{Differentiation and Edge Detection}
\label{ch:lesson12}

An edge is, informally, a place where intensity changes quickly. That's a statement about a \emph{derivative}. This chapter builds up discrete image derivatives --- Prewitt, Sobel, and Scharr operators --- as convolution kernels (Chapter~\ref{ch:lesson10}), uses them inside the classic \textbf{Canny} edge detector, and finishes by turning the resulting edge maps into actual shapes with the \textbf{Hough transform}.

\section{From derivatives to gradients}

Treat the image as a function $I(x,y)$. Its gradient $\nabla I=(I_x,I_y)$ points in the direction of steepest intensity increase, where $I_x=\partial I/\partial x$ and $I_y=\partial I/\partial y$ are the \textbf{partial derivatives} of image intensity --- how fast the intensity changes as you move one pixel right, or one pixel down. Two numbers summarize the gradient at each pixel:
\begin{itemize}
  \item \textbf{magnitude} $\|\nabla I\| = \sqrt{I_x^2+I_y^2}$ --- how sharp the change is (large at edges, near zero on flat regions)
  \item \textbf{direction} $\theta = \operatorname{atan2}(I_y,I_x)$ --- which way intensity is increasing fastest (perpendicular to the edge)
\end{itemize}
$I_x$ and $I_y$ are themselves computed by convolving the image with small derivative kernels.

\section{Prewitt, Sobel, and Scharr: derivative + smoothing}

The simplest $I_x$ estimate is a 1D central-difference kernel $[-1,\,0,\,1]$; it works, but with no smoothing in the perpendicular direction, it's maximally sensitive to noise. Real operators combine a difference in one direction with a \emph{smoothing} average in the perpendicular direction --- this is what makes them robust to noise. They differ only in how they weight that smoothing:

\begin{center}
\begin{tabular}{lll}
\hline
\textbf{Operator} & $x$\textbf{-kernel} & \textbf{Perpendicular weighting} \\
\hline
Prewitt & $\begin{bmatrix}-1&0&1\\-1&0&1\\-1&0&1\end{bmatrix}$ & uniform: 1, 1, 1 \\[1.2em]
Sobel & $\begin{bmatrix}-1&0&1\\-2&0&2\\-1&0&1\end{bmatrix}$ & binomial: 1, 2, 1 \\[1.2em]
Scharr & $\begin{bmatrix}-3&0&3\\-10&0&10\\-3&0&3\end{bmatrix}$ & 3, 10, 3 \\
\hline
\end{tabular}
\end{center}

\begin{figure}[h]
\centering
\includegraphics[width=0.95\textwidth]{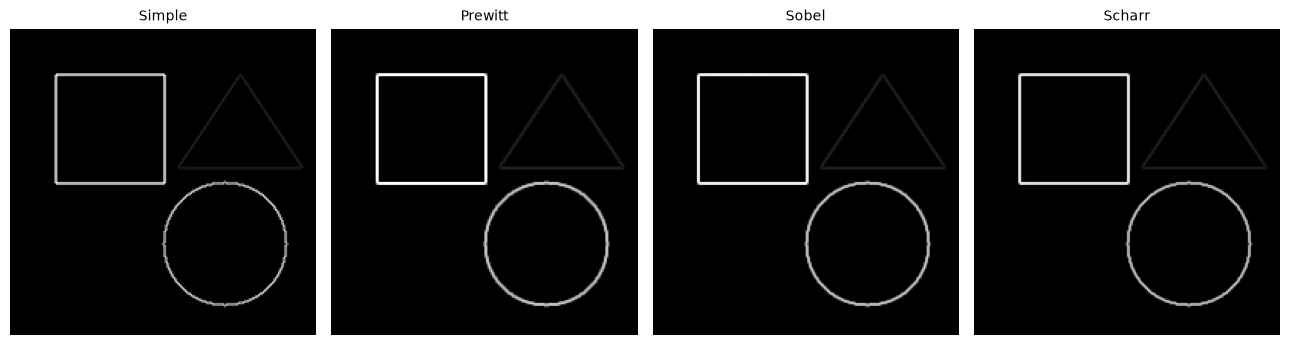}
\caption{Gradient magnitude from the simple difference kernel and the three smoothed operators, on a clean test image. On a low-noise image like this, the four results look nearly identical.}
\label{fig:l12-1}
\end{figure}

Each $y$-kernel is just the transpose of its $x$-kernel. Sobel's binomial weights approximate a small Gaussian, which is why it's the most commonly used default. Scharr's weights were numerically optimized specifically to make the \emph{estimated gradient direction} as rotationally accurate as possible --- which matters most in applications like optical flow that depend on precise gradient angles rather than just edge location. Figure~\ref{fig:l12-1} compares all four operators.

\subsection*{Noise robustness: why the perpendicular smoothing matters}

To make the benefit of perpendicular smoothing concrete, measure each kernel's response to pure noise on a flat (edge-free) region, after first rescaling each kernel to the \emph{same gain} on an ideal step edge (dividing by the sum of its positive weights) so a kernel with larger raw coefficients doesn't look noisier just from its overall scale.

\begin{center}
\renewcommand{\arraystretch}{1.4}
\begin{tabular}{lr}
\hline
\textbf{Kernel} & \textbf{Noise std (gain-normalized)} \\
\hline
Simple $[-1,0,1]$ & 20.93 \\
Prewitt & 12.29 \\
Sobel & 12.97 \\
Scharr & 14.32 \\
\hline
\end{tabular}
\end{center}

The plain difference kernel is noticeably noisier than any of the $3\times3$ operators --- averaging over 3 rows (or columns) while differencing cuts down the noise response substantially, which is exactly the point of building smoothing into the derivative kernel.

\section{From gradients to edges: the Canny detector}

\begin{figure}[h]
\centering
\includegraphics[width=0.9\textwidth]{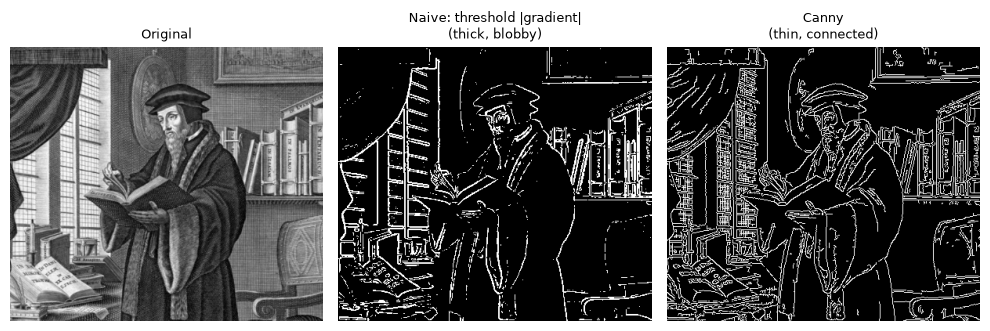}
\caption{Naively thresholding $\|\nabla I\|$ gives thick, blobby edges; \code{cv2.Canny} gives thin, connected ones. \attrib{Image source: \href{https://www.nga.gov/artworks/37890-john-calvin}{National Gallery of Art}.}}
\label{fig:l12-2}
\end{figure}

Simply thresholding gradient magnitude gives thick, noisy edge blobs. The \textbf{Canny} edge detector (\citelink{canny1986}{Canny, 1986}) refines this with a multi-stage pipeline:
\begin{enumerate}
  \item \textbf{Smooth} with a Gaussian, to suppress noise before differentiating.
  \item \textbf{Compute gradients} (Sobel, internally) to get magnitude and direction at every pixel.
  \item \textbf{Non-maximum suppression}: at each pixel, keep the gradient magnitude only if it's a local maximum \emph{along the gradient direction} --- this thins wide gradient ridges down to single-pixel-wide lines.
  \item \textbf{Double thresholding + hysteresis}: pixels above a high threshold are definite edges; pixels below a low threshold are discarded; pixels in between are kept only if they connect to a definite edge.
\end{enumerate}
Figure~\ref{fig:l12-2} compares the naive and Canny results.

\subsection*{Threshold sensitivity}

\begin{figure}[h]
\centering
\includegraphics[width=0.9\textwidth]{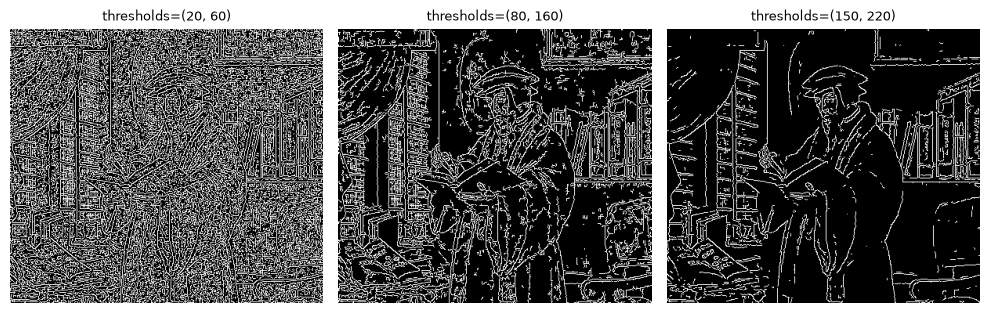}
\caption{With low thresholds on a noisy image, speckle noise survives as spurious edge fragments. At the highest thresholds, faint detail disappears completely, while the strongest outlines survive --- Canny's threshold trading off noise rejection against sensitivity to faint-but-genuine edges.}
\label{fig:l12-3}
\end{figure}

Canny's two thresholds trade off completeness against noise. Too low, and noise gets picked up as spurious edges; too high, and real (but low-contrast) edges are missed, as Figure~\ref{fig:l12-3} shows.

\section{From edges to lines: the Hough transform}

Edge detection finds \emph{where} intensity changes sharply, but it doesn't know that a scattered set of edge pixels forms a straight line. The \textbf{Hough transform} (\citelink{hough1962}{Hough, 1962}; \citelink{dudahart1972}{Duda \& Hart, 1972}) answers a different question: given a set of edge points, which ones are consistent with lying on a common line?

\begin{figure}[h]
\centering
\includegraphics[width=0.55\textwidth]{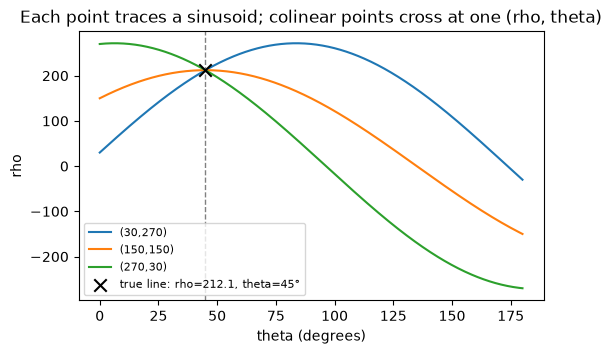}
\caption{Three points on the line $x+y=300$ each trace a sinusoid in $(\rho,\theta)$ space; all three cross at the line's true parameters, $\rho=212.1$, $\theta=45^\circ$.}
\label{fig:l12-4}
\end{figure}

Parameterize a line not as $y=mx+b$ (which blows up for vertical lines) but by its distance from the origin and the angle of its normal:
\begin{equation}
\rho = x\cos\theta + y\sin\theta.
\end{equation}
For a \emph{fixed} edge point $(x,y)$, this traces a sinusoidal curve in $(\rho,\theta)$ space as $\theta$ varies --- every $(\rho,\theta)$ pair on that curve describes a line through $(x,y)$. The key idea: if several edge points are \textbf{collinear}, their sinusoids all cross at the \emph{same} $(\rho,\theta)$ --- the parameters of the line they share, as demonstrated in Figure~\ref{fig:l12-4}.

\subsection*{The accumulator: voting for lines}

\begin{figure}[h]
\centering
\includegraphics[width=0.95\textwidth]{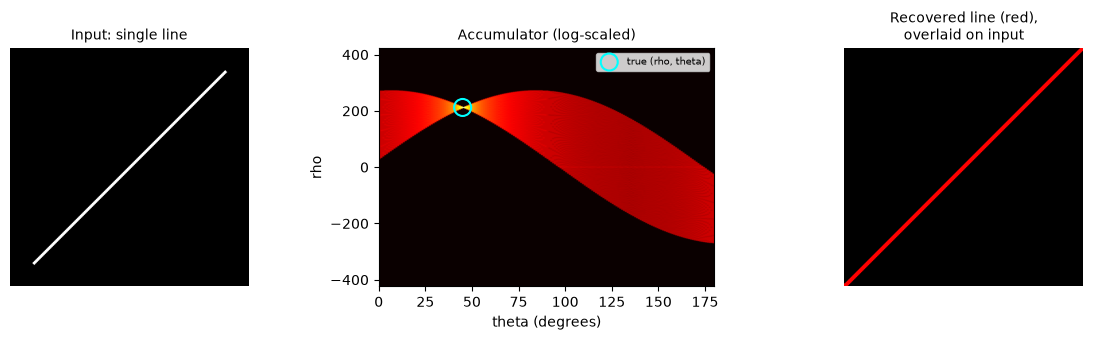}
\caption{Input line, the log-scaled $(\rho,\theta)$ accumulator (true parameters marked in cyan), and the recovered line overlaid in red. The basic Hough transform returns an infinite line, not a line segment.}
\label{fig:l12-5}
\end{figure}

In practice, every edge pixel votes for its entire sinusoid of possible $(\rho,\theta)$ lines, into a discretized 2D \textbf{accumulator} array; the bin with the most votes is the line most consistent with the data. Built from scratch for a single synthetic line and checked against \code{cv2.HoughLines}: the manual accumulator peak lands at $\rho=212$, $\theta=45.0^\circ$, and \code{cv2.HoughLines} reports $\rho=211$, $\theta=45.0^\circ$ --- both within a pixel of the true $\rho=212.1$. Figure~\ref{fig:l12-5} shows the accumulator and recovered line.

\subsection*{\code{cv2.HoughLinesP}: the practical version}

\begin{figure}[h]
\centering
\includegraphics[width=0.85\textwidth]{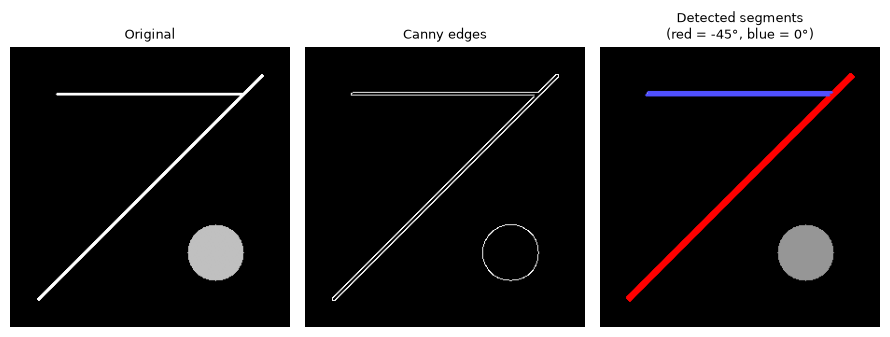}
\caption{Two lines (true angles $-45^\circ$ and $0^\circ$) plus a distractor circle with no straight edges. Every detected segment lands almost exactly on one of the two true angles; the circle correctly produces no line detections. Each line typically yields two or three near-duplicate detections, since Canny finds an edge on \emph{each} side of a stroke with finite width.}
\label{fig:l12-6}
\end{figure}

The probabilistic Hough transform additionally returns line \emph{segment} endpoints, and is what's normally used in practice, run on top of an actual Canny edge map, as in Figure~\ref{fig:l12-6}.

\subsection*{Beyond lines: Hough circles}

\begin{figure}[h]
\centering
\includegraphics[width=0.35\textwidth]{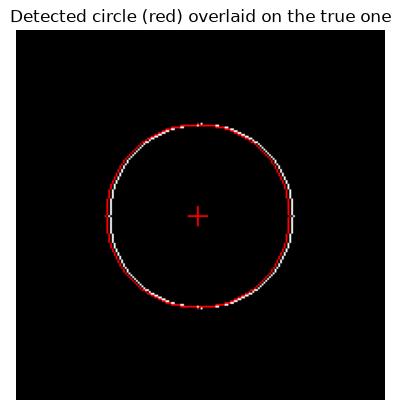}
\caption{Detected circle (red) overlaid on the true one (white).}
\label{fig:l12-7}
\end{figure}

The same voting idea generalizes to any shape describable by a small number of parameters. A circle needs 3 (center $x$, center $y$, radius $r$), so \code{cv2.HoughCircles} accumulates votes in a 3D parameter space instead of 2D. For a true circle at center $(100,100)$, radius $50$: \code{cv2.HoughCircles} reports center $(98.5, 100.5)$, radius $50.0$, shown in Figure~\ref{fig:l12-7}.

\begin{figure}[h]
\centering
\includegraphics[width=0.85\textwidth]{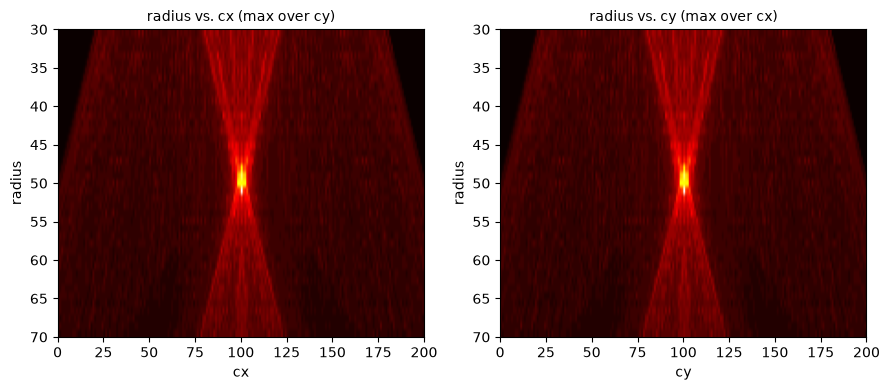}
\caption{2D slices of the 3D circle accumulator: radius vs.\ $c_x$ (max over $c_y$) and radius vs.\ $c_y$ (max over $c_x$). Each slice shows a bright X where every edge point's votes cross at the true center and radius.}
\label{fig:l12-8}
\end{figure}

\code{cv2.HoughCircles} doesn't expose its internal accumulator, but building one from scratch follows the same voting idea, with one more dimension: each edge pixel votes for every $(c_x,c_y,r)$ triple it could be consistent with, for every candidate radius. A from-scratch accumulator peak lands at center $(100,100)$, radius $51$ --- matching the true circle almost exactly, visualized in Figure~\ref{fig:l12-8}.

\subsection*{Exercises}
\begin{enumerate}
  \item Increase the noise standard deviation in \code{flat\_noisy} and re-run the noise-robustness comparison. Does the relative ordering of the four kernels change?
  \item \code{cv2.Canny} accepts an \code{apertureSize} parameter for its internal Sobel step (default 3). Try \code{apertureSize=5} and compare the result to the default on the noisy image.
  \item Canny's hysteresis step needs a \emph{connected} path of above-low-threshold pixels between a weak edge and a strong one. Construct a small binary example (by hand, as a NumPy array) where a real edge is broken into two segments with a 1-pixel gap, and confirm that hysteresis fails to link them even though a human viewer would clearly see one edge.
  \item Lower \code{cv2.HoughLinesP}'s \code{threshold} argument well below 60 on the two-line scene. Do you start getting spurious short segments from noise in the circle's Canny boundary, even though it has no straight edges? What does raising \code{minLineLength} do to fix this?
  \item Add a third line at a shallow angle (e.g.\ from $(30,200)$ to $(270,150)$), and predict its angle before checking \code{HoughLinesP}'s output against your prediction.
  \item \code{cv2.HoughCircles}'s \code{param2} argument is roughly an accumulator vote threshold --- lower values report more (and more spurious) circles. Sweep \code{param2} from 30 down to 10 on a noisy version of the circle image and observe when false-positive circles start appearing.
\end{enumerate}

\vfill
\fullcode{lesson12}{lesson12_differentiation_edges}

%% file: chapters/lesson13.tex
\chapter{Second Derivatives and the Laplacian}
\label{ch:lesson13}

Chapter~\ref{ch:lesson12} found edges as \emph{peaks} in the first derivative (gradient magnitude). This chapter uses the \textbf{second} derivative instead, where edges show up as \textbf{zero crossings} --- the basis of the classic Marr-Hildreth edge detector. We then reuse the same blur-and-difference idea to build a \textbf{Laplacian pyramid}, which fixes the information-loss problem from Chapter~\ref{ch:lesson11}'s Gaussian pyramid.

\section{Why zero crossings? A 1D intuition}

\begin{figure}[h]
\centering
\includegraphics[width=0.5\textwidth]{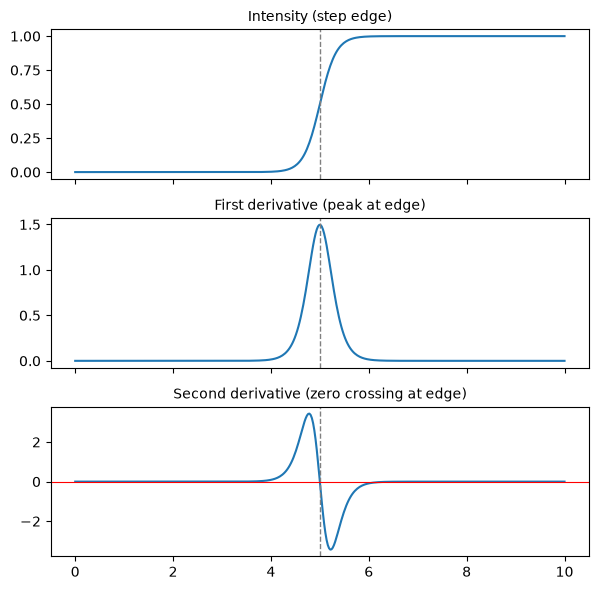}
\caption{A smoothed step edge, its first derivative (a peak), and its second derivative (a zero crossing), all at the same location.}
\label{fig:l13-1}
\end{figure}

Consider a single step edge along one row. The first derivative is a spike at the edge; the second derivative swings from positive to negative (or vice versa) and crosses exactly zero right at the edge location, as shown in Figure~\ref{fig:l13-1}.

\section{The Laplacian: a 2D second derivative}

\begin{figure}[h]
\centering
\includegraphics[width=0.6\textwidth]{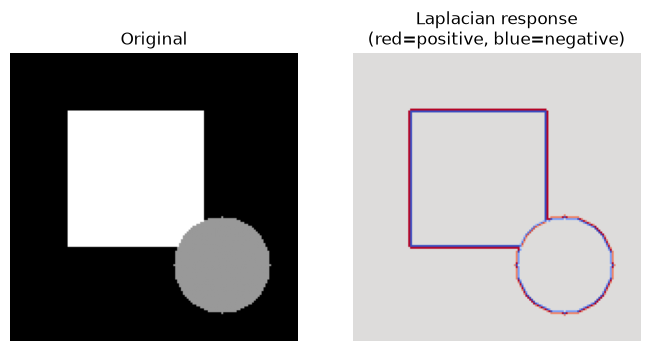}
\caption{Every edge produces a positive lobe on one side and a negative lobe on the other --- the edge itself sits at the zero crossing between them. The response is shown on a red-blue diverging color scale specifically because the Laplacian is signed --- red marks positive values, blue marks negative, making the sign flip at each zero crossing easy to see at a glance.}
\label{fig:laplacian-response}
\end{figure}

The Laplacian sums the unmixed second partial derivatives of a 2D image:
\begin{equation}
\nabla^2 I = \frac{\partial^2 I}{\partial x^2} + \frac{\partial^2 I}{\partial y^2}.
\end{equation}
Unlike the gradient, it's a single scalar at each pixel (no direction) --- it just measures how much a pixel differs from the average of its neighbors. Two common discrete kernels:
\begin{equation}
K_4 = \begin{bmatrix}0&1&0\\1&-4&1\\0&1&0\end{bmatrix} \qquad K_8 = \begin{bmatrix}1&1&1\\1&-8&1\\1&1&1\end{bmatrix}.
\end{equation}
$K_4$ uses only the 4-connected neighbors; $K_8$ also includes the diagonals for a stronger, more isotropic response. Convolving with $K_4$ matches \code{cv2.Laplacian(img, cv2.CV\_64F, ksize=1)} exactly, shown in Figure~\ref{fig:laplacian-response}.

\section{The Marr-Hildreth detector: Laplacian of Gaussian (LoG)}

\begin{figure}[h]
\centering
\includegraphics[width=0.9\textwidth]{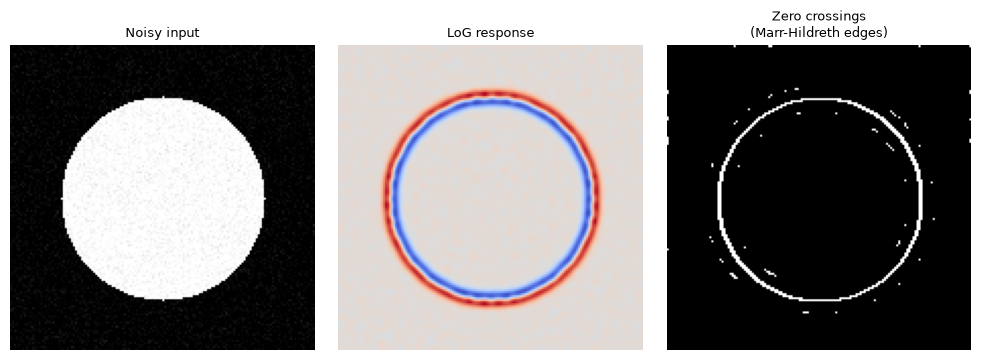}
\caption{A noisy disk, its LoG response, and the resulting zero-crossing edge map (Marr-Hildreth edges). The LoG response uses the same red/blue diverging scale as Figure~\ref{fig:laplacian-response}, for the same reason.}
\label{fig:l13-3}
\end{figure}

The Laplacian, like any derivative, amplifies high-frequency noise --- and a \emph{second} derivative amplifies it even more than a first derivative does. Marr and Hildreth's fix (\citelink{marrhildreth1980}{Marr \& Hildreth, 1980}): \textbf{smooth first}, then take the Laplacian. Since convolution is associative, this is equivalent to convolving directly with a single combined kernel, the \textbf{Laplacian of Gaussian (LoG)}:
\begin{equation}
\text{LoG}_\sigma(I) = \nabla^2(G_\sigma * I) = (\nabla^2 G_\sigma) * I.
\end{equation}
Edges are then found as \textbf{zero crossings} of the LoG response, rather than by thresholding its magnitude directly, as in Figure~\ref{fig:l13-3}.

\subsection*{The cheaper approximation: Difference of Gaussians (DoG)}

\begin{figure}[h]
\centering
\includegraphics[width=0.6\textwidth]{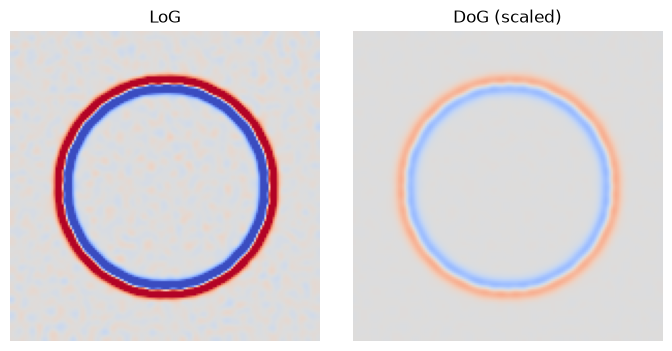}
\caption{LoG response vs.\ the scaled DoG approximation --- visually almost indistinguishable.}
\label{fig:l13-4}
\end{figure}

Computing a true LoG kernel is more expensive than blurring. A widely used shortcut: the \emph{difference} of two Gaussian blurs at slightly different scales approximates a scaled LoG:
\begin{equation}
\text{DoG} = G_{\sigma} - G_{k\sigma} \;\approx\; -(k-1)\sigma^2 \cdot \text{LoG}_\sigma.
\end{equation}
This is the same building block later reused by SIFT for keypoint detection at multiple scales. On a test image, the correlation between the true LoG and the scaled DoG comes to $0.967$ --- highly correlated but not identical, since DoG is an approximation, not an exact substitute; a much cheaper one, though (two blurs and a subtraction vs.\ an explicit second-derivative kernel). Figure~\ref{fig:l13-4} shows the two side by side.

\section{Laplacian pyramids: recovering what Gaussian pyramids throw away}

Chapter~\ref{ch:lesson11} showed that a Gaussian pyramid loses information: blurring and downsampling repeatedly, then trying to upsample back, doesn't reconstruct the original. A \textbf{Laplacian pyramid} fixes this by storing, at each level, exactly the detail that would otherwise be lost --- the \emph{difference} between a Gaussian level and the coarser level upsampled back to match it. That difference is a discrete approximation of the Laplacian, which is why it shares the name.

\begin{lstlisting}
def laplacian_pyramid(gauss_pyramid):
    lap_pyramid = []
    for i in range(len(gauss_pyramid) - 1):
        size = (gauss_pyramid[i].shape[1], gauss_pyramid[i].shape[0])
        upsampled = cv2.pyrUp(gauss_pyramid[i + 1], dstsize=size)
        detail = gauss_pyramid[i].astype(np.int16) - upsampled.astype(np.int16)
        lap_pyramid.append(detail)
    lap_pyramid.append(gauss_pyramid[-1].astype(np.int16))  # smallest level: no finer detail to subtract
    return lap_pyramid
\end{lstlisting}

\begin{figure}[h]
\centering
\includegraphics[width=0.95\textwidth]{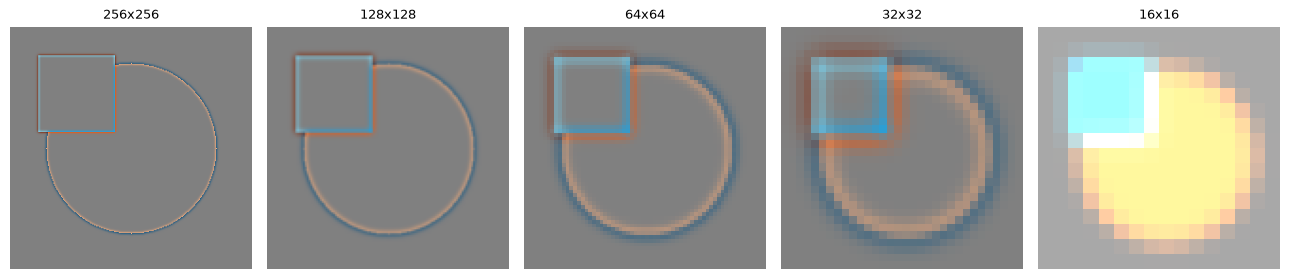}
\caption{A 5-level Laplacian pyramid, each level shifted by $+128$ gray levels so negative values are visible. Each level is mostly flat gray except right at edges --- exactly where detail is lost by blurring and downsampling. The final, smallest level stores actual image content rather than a difference, since there's nothing coarser left to compare it to.}
\label{fig:l13-5}
\end{figure}

Figure~\ref{fig:l13-5} shows the resulting pyramid.

\subsection*{Exact reconstruction}

\begin{figure}[h]
\centering
\includegraphics[width=0.55\textwidth]{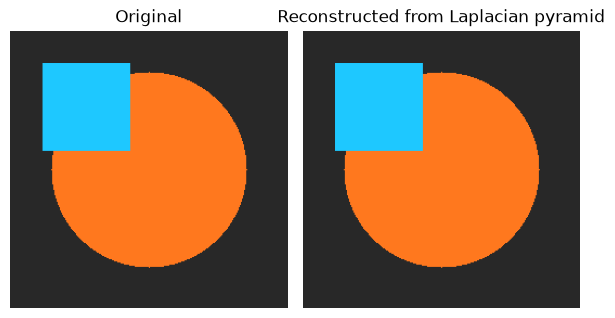}
\caption{Original vs.\ reconstructed-from-Laplacian-pyramid --- pixel-perfect.}
\label{fig:l13-6}
\end{figure}

Because each level stores exactly what its Gaussian-pyramid counterpart discarded, summing back up the pyramid reconstructs the original image exactly (up to integer rounding): max reconstruction error is $0$ gray levels, mean error $0.0000$ (Figure~\ref{fig:l13-6}).

Compare this to Chapter~\ref{ch:lesson11}'s \code{pyrDown} + \code{pyrUp} result, which had a visibly nonzero reconstruction error: the Laplacian pyramid stores just enough extra information at each level to make the process perfectly reversible, at the cost of needing to keep all the levels around (not just the smallest one).

\subsection*{Exercises}
\begin{enumerate}
  \item Try \code{threshold=1.0} and \code{threshold=15.0} in \code{zero\_crossings}. How does the number of detected edge pixels change, and why does a \emph{higher} threshold on a \emph{second}-derivative sign-change criterion behave differently than a magnitude threshold on a first derivative?
  \item Increase the noise level in the noisy disk and see how large \code{sigma} needs to be before the zero-crossing edge map stops being dominated by spurious noise loops.
  \item Laplacian pyramids are the classic tool behind seamless image blending (e.g.\ blending two photos along a mask, level by level). Sketch out --- in words or code --- how you'd blend two Laplacian pyramids before reconstructing, and why blending in this representation avoids sharp seams that blending the original images directly would produce.
\end{enumerate}

\vfill
\fullcode{lesson13}{lesson13_laplacian_pyramids}

%% file: chapters/lesson14.tex
\chapter{Nonlinear Filters}
\label{ch:lesson14}

Every filter so far (box, Gaussian, Sobel, Laplacian, \ldots) has been a convolution: a fixed, linear combination of neighboring pixel values. But linear filters are limited in what they can do --- they can't reason about which neighboring pixels are ``outliers'' or which are ``on the other side of an edge.'' This chapter looks at filters that break that linearity: \textbf{median filtering}, grayscale \textbf{erosion/dilation as min/max filters}, and the edge-preserving \textbf{bilateral filter}.

\section{Median filtering: robust to outliers}

A median filter replaces each pixel with the \textbf{median} (not the mean) of its neighborhood. Whereas a single wildly wrong pixel value (outlier) can drastically affect the mean, it barely moves a median, if at all --- so median filtering is especially effective against \textbf{salt-and-pepper noise} (isolated pixels randomly slammed to black or white), which linear smoothing handles poorly.

\begin{lstlisting}
median_result = cv2.medianBlur(salt_pepper, 5)
gaussian_result = cv2.GaussianBlur(salt_pepper, (5, 5), sigmaX=1.5)
\end{lstlisting}

\begin{figure}[h]
\centering
\includegraphics[width=0.9\textwidth]{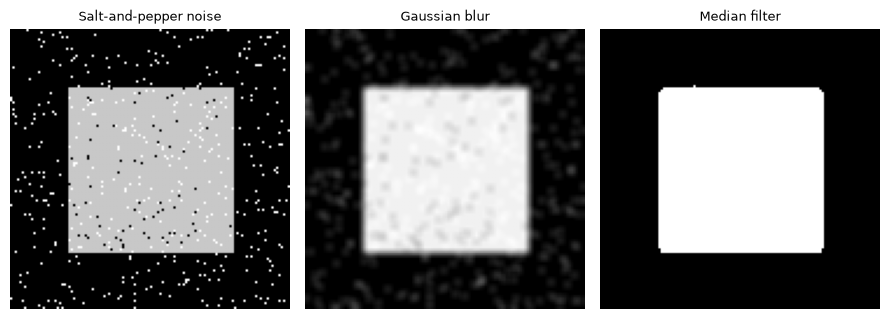}
\caption{Gaussian blur smears each corrupted pixel's extreme value into its neighbors, leaving faint speckle everywhere. The median filter simply outvotes an isolated bad pixel with its many good neighbors, removing nearly all of the noise while keeping edges sharp.}
\label{fig:l14-1}
\end{figure}

Measured against the clean, noise-free image: Gaussian blur leaves a mean absolute error of $9.56$ gray levels, while the median filter leaves just $0.15$ (Figure~\ref{fig:l14-1}).

\section{Min and max filters: grayscale erosion and dilation}

Chapter~\ref{ch:lesson03} introduced erosion and dilation on \emph{binary} images. On a \textbf{grayscale} image, they generalize naturally: erosion replaces each pixel with the \textbf{minimum} value in its neighborhood, and dilation with the \textbf{maximum}. Both are nonlinear (not a weighted sum), and both are edge-preserving in a different way than blurring --- they shift edges rather than blurring them.

\begin{lstlisting}
def min_filter(image, ksize):
    r = ksize // 2
    padded = cv2.copyMakeBorder(image, r, r, r, r, cv2.BORDER_REPLICATE)
    out = np.zeros_like(image)
    for i in range(image.shape[0]):
        for j in range(image.shape[1]):
            out[i, j] = padded[i:i+ksize, j:j+ksize].min()
    return out
\end{lstlisting}

A from-scratch min filter matches \code{cv2.erode} with a $3\times3$ all-ones kernel exactly, confirming that grayscale erosion really is just a sliding-window minimum.

\section{The problem linear blur can't solve: smoothing without crossing edges}

Gaussian blur treats every neighbor equally regardless of how different its value is --- that's exactly what makes it blur across edges. The \textbf{bilateral filter} fixes this by weighting each neighbor by \emph{two} factors: how close it is spatially (like a Gaussian blur) \textbf{and} how similar its intensity is to the center pixel:
\begin{equation}
I'(p) = \frac{1}{W_p}\sum_{q \in N(p)} G_{\sigma_s}(\|p - q\|) \cdot G_{\sigma_r}(|I(p) - I(q)|) \cdot I(q),
\end{equation}
where $W_p$ normalizes the weights to sum to 1. Neighbors that are spatially close \emph{and} have similar intensity get high weight (smoothed together); neighbors that are spatially close but have very different intensity (i.e.\ on the other side of an edge) get down-weighted almost to zero, so the edge survives.

\begin{lstlisting}
def bilateral_filter(image, radius, sigma_spatial, sigma_range):
    image_f = image.astype(np.float64)
    ys, xs = np.mgrid[-radius:radius+1, -radius:radius+1]
    spatial_weight = np.exp(-(xs**2 + ys**2) / (2 * sigma_spatial**2))
    padded = cv2.copyMakeBorder(image_f, radius, radius, radius, radius, cv2.BORDER_REFLECT101)
    out = np.zeros_like(image_f)
    for i in range(image.shape[0]):
        for j in range(image.shape[1]):
            patch = padded[i:i+2*radius+1, j:j+2*radius+1]
            range_weight = np.exp(-(patch - image_f[i, j])**2 / (2 * sigma_range**2))
            weight = spatial_weight * range_weight
            out[i, j] = (weight * patch).sum() / weight.sum()
    return out
\end{lstlisting}

\code{cv2.bilateralFilter} uses the same idea with some internal approximations for speed, so a close but not pixel-exact match is expected: max absolute difference is $3.40$ gray levels, mean absolute difference $0.56$.

\subsection*{Edge preservation, side by side}

\begin{figure}[h]
\centering
\includegraphics[width=0.95\textwidth]{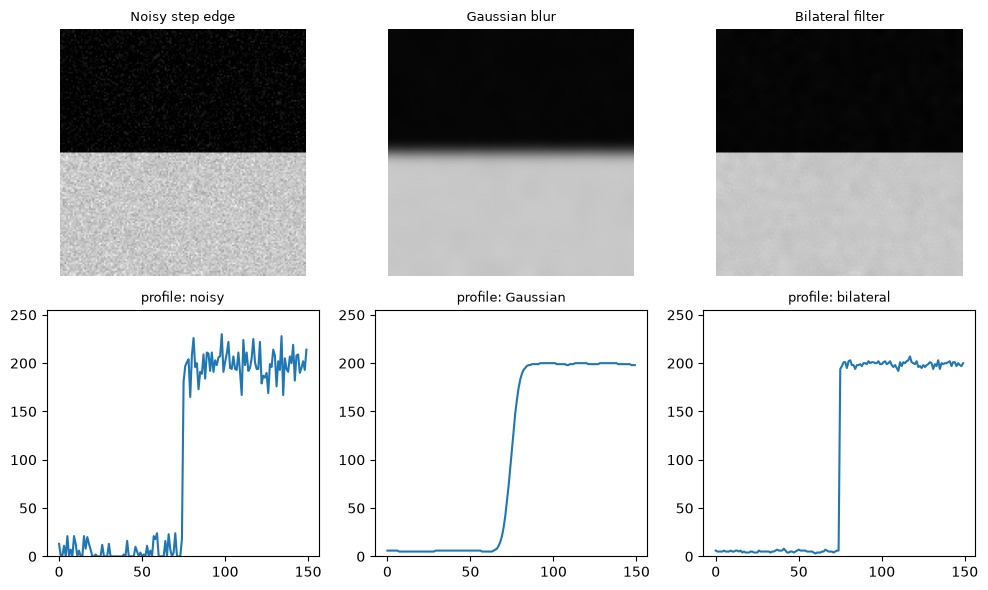}
\caption{Top: a noisy step edge, Gaussian-blurred, and bilateral-filtered. Bottom: intensity profiles along a column crossing the edge. The Gaussian-blurred profile ramps gradually across many rows --- the edge has been visibly softened. The bilateral-filtered profile stays flat on each side (noise removed) but still jumps sharply at the true edge --- it survived because pixels across the boundary were too different in intensity to be averaged together, no matter how spatially close they were.}
\label{fig:l14-2}
\end{figure}

The real comparison that matters: on a noisy step edge, does the filter smooth the noise while keeping the edge sharp, or does it blur the edge along with the noise? Figure~\ref{fig:l14-2} answers this directly.

\subsection*{Exercises}
\begin{enumerate}
  \item Apply a median filter (instead of Gaussian or bilateral) to the noisy step edge and add its profile to the comparison. How does it compare to the bilateral filter at preserving the edge?
  \item In \code{bilateral\_filter}, set \code{sigma\_range} very large (e.g.\ 1000). What should this reduce to, and does your result confirm it?
  \item Bilateral filtering is much slower than Gaussian blur because the weights must be recomputed at every pixel (they depend on local intensity, not just position). Time the from-scratch \code{bilateral\_filter} against \code{cv2.bilateralFilter} on a $100\times100$ image with \code{\%timeit}, and describe in words what OpenCV's implementation likely does differently to be faster.
\end{enumerate}

\vfill
\fullcode{lesson14}{lesson14_nonlinear_filters}

%% file: chapters/lesson15.tex
\chapter{The Fourier Transform and Frequency-Domain Filtering}
\label{ch:lesson15}

Every filter so far has worked directly on pixel neighborhoods. The Fourier transform offers a completely different view: any image can be rewritten as a sum of sinusoidal gratings at different frequencies and orientations. Filtering can then be done by directly boosting or suppressing specific frequencies --- and, thanks to the \textbf{convolution theorem}, this turns out to be mathematically the same operation as spatial convolution, just seen from a different angle.

\section{1D warm-up: a signal as a sum of sinusoids}

\begin{figure}[h]
\centering
\includegraphics[width=0.75\textwidth]{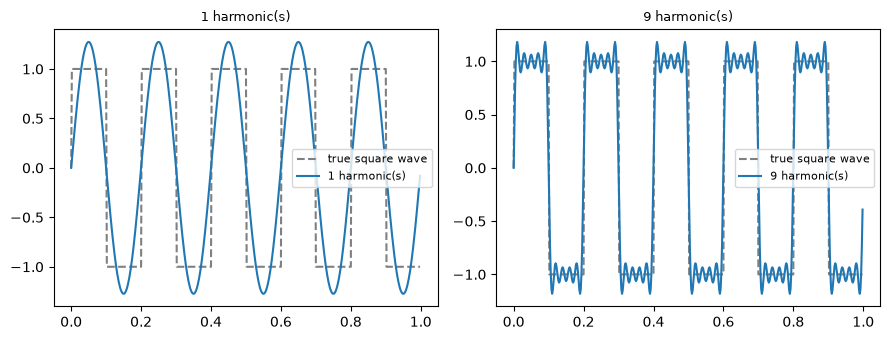}
\caption{A square wave approximated with 1 harmonic and with 9 harmonics --- more harmonics track the true square wave more closely, especially near the sharp transitions.}
\label{fig:l15-1}
\end{figure}

Any signal can be approximated by adding up sine and cosine waves of different frequencies. A square wave, for example, is the sum of sine waves at increasing (odd) frequencies with decreasing amplitude. The Fourier transform is the tool that goes the other way: given a signal, it tells you exactly which frequencies (and how much of each) are present, as Figure~\ref{fig:l15-1} shows by approximating a square wave.

\section{The 2D Fourier transform of an image}

\begin{figure}[h]
\centering
\includegraphics[width=0.65\textwidth]{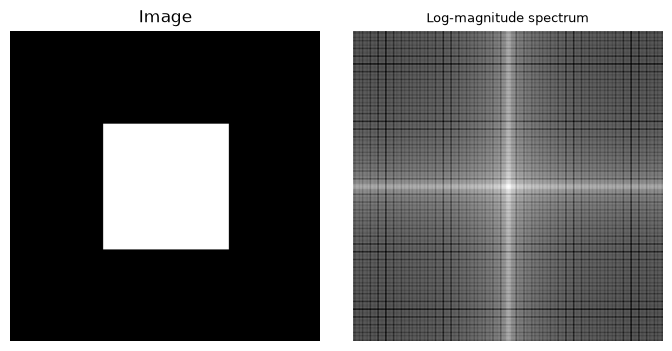}
\caption{A filled square and its log-magnitude spectrum. The bright cross through the center comes from the square's sharp horizontal and vertical edges --- a hard step edge contains energy at \emph{every} frequency along the direction perpendicular to it, which is why a sharp-edged shape has such a spread-out spectrum.}
\label{fig:l15-2}
\end{figure}

The 2D discrete Fourier transform (DFT) of an image decomposes it into 2D sinusoidal gratings at every combination of horizontal and vertical frequency, computed via the fast Fourier transform (\code{np.fft.fft2}), which produces a complex-valued 2D array. As with the image gradient, we usually view the Fourier transform's \textbf{magnitude} (how much of each frequency is present). Two common visualization tricks, shown in Figure~\ref{fig:l15-2}: show the magnitude on a log scale, and shift the zero-frequency (average brightness) term from the top-left corner to the center.

\subsection*{A grating reveals its frequency directly as a spectrum peak}

A pure sinusoidal grating is the simplest possible test: its spectrum should be (almost) a single pair of bright dots, at a distance from the center equal to its frequency, in the direction perpendicular to its stripes.

\begin{figure}[h]
\centering
\includegraphics[width=0.6\textwidth]{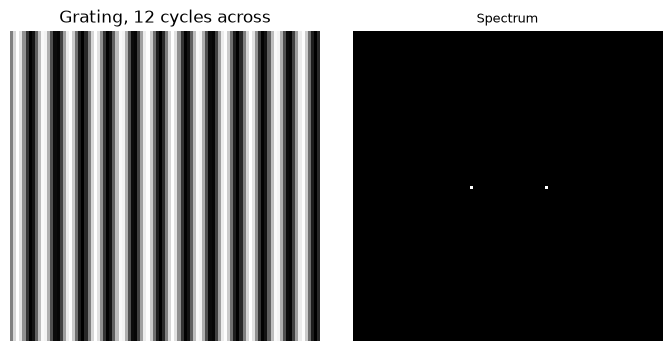}
\caption{A grating with 12 cycles across the image and its spectrum. The brightest spectrum point lands at offset $-12$ from center, matching the grating's true frequency of 12.}
\label{fig:l15-3}
\end{figure}

Why a \emph{pair}, not one dot? A real-valued sine wave is, by Euler's formula, a mix of two complex exponentials in equal amounts: $\sin(2\pi fx) = \tfrac{1}{2i}(e^{i2\pi fx} - e^{-i2\pi fx})$. The Fourier transform of the first exponential is a spike at frequency $f$, and of the other at $-f$. So a real sinusoid produces two spikes at $+f$ and $-f$, symmetric about the zero-frequency (DC) center --- in fact, for \emph{any} real-valued image, its Fourier magnitude is symmetric: $|F(-f)|=|F(f)|$, as confirmed by the grating in Figure~\ref{fig:l15-3}.

\section{Filtering by editing the spectrum}

An \textbf{ideal low-pass filter} keeps only frequencies inside a circle around the center (blurs); an \textbf{ideal high-pass filter} keeps only frequencies outside it (edge-enhances). Once an image is transformed to the frequency domain, filtering is masking frequencies directly, then transforming back with the inverse transform --- since the mask is the same size as the image, this is a plain elementwise multiply.

\begin{lstlisting}
def apply_frequency_filter(image, mask):
    F = np.fft.fftshift(np.fft.fft2(image))
    filtered = np.fft.ifft2(np.fft.ifftshift(F * mask))
    return np.real(filtered)
\end{lstlisting}

\begin{figure}[h]
\centering
\includegraphics[width=0.85\textwidth]{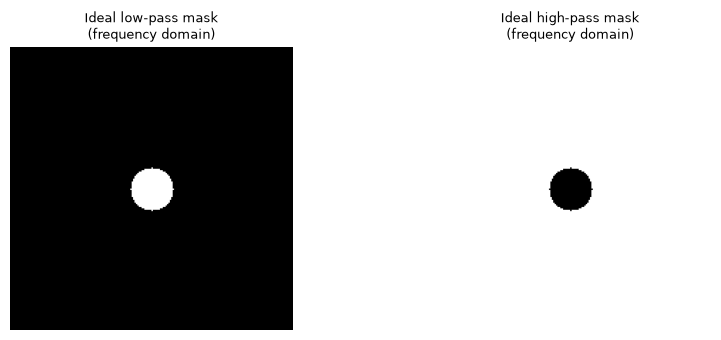}
\caption{Ideal low-pass and high-pass masks in the frequency domain.}
\label{fig:l15-4}
\end{figure}

\begin{figure}[h]
\centering
\includegraphics[width=0.9\textwidth]{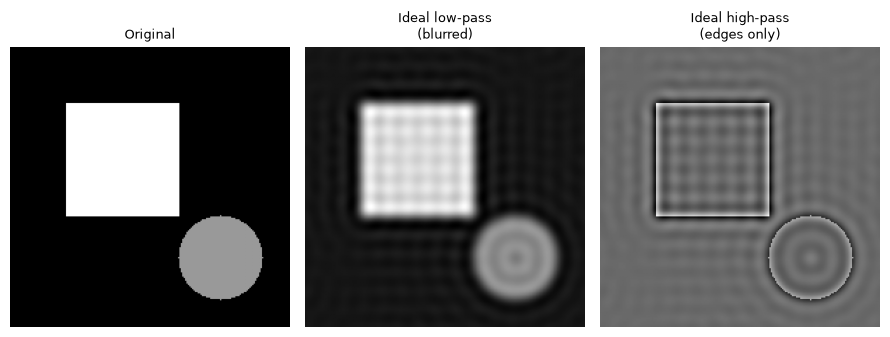}
\caption{Original image, ideal low-pass result (blurred), and ideal high-pass result (edges only).}
\label{fig:l15-5}
\end{figure}

Figure~\ref{fig:l15-4} shows the two masks, and Figure~\ref{fig:l15-5} the resulting filtered images.

\subsection*{The cost of a sharp cutoff: ringing}

\begin{figure}[h]
\centering
\includegraphics[width=0.65\textwidth]{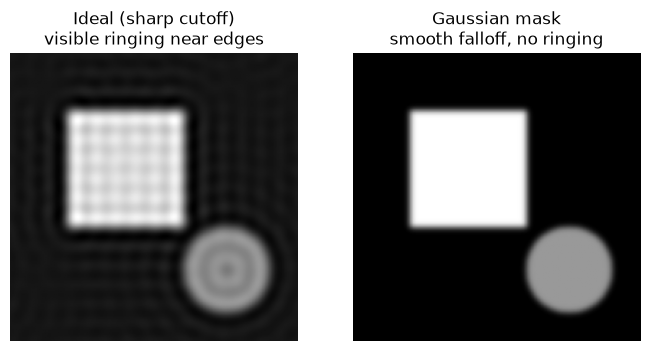}
\caption{Ideal low-pass (sharp cutoff, visible ringing near edges) vs.\ a Gaussian mask (smooth falloff, no ringing).}
\label{fig:l15-6}
\end{figure}

Faint ripples radiate from the shape edges in the ideal-filter results above. An \emph{ideal} filter has a perfectly sharp cutoff in frequency, which corresponds (by the same convolution theorem) to convolving with a spatial kernel that has infinite ripples of its own (a sinc function) --- the \textbf{Gibbs phenomenon}. A Gaussian mask, which falls off smoothly instead of cutting off sharply, avoids this, as Figure~\ref{fig:l15-6} shows.

\section{The convolution theorem}

Why does frequency-domain filtering work? The \textbf{convolution theorem}
\begin{equation}
\mathcal{F}\{f * g\} = \mathcal{F}\{f\} \cdot \mathcal{F}\{g\}
\end{equation}
says that convolving two signals in the spatial domain is \emph{exactly} the same as multiplying their Fourier transforms elementwise, then transforming back --- convolution becomes multiplication, and vice versa.

\subsection*{A catch: the DFT gives circular convolution}

The DFT implicitly treats the image as if it were infinitely repeating (seamless tiles extending forever in both directions) --- there is no ``outside the image,'' only ``the next copy of the same image.'' So multiplying two DFTs and transforming back doesn't give the ordinary (``linear'') convolution from Chapter~\ref{ch:lesson10}, where the kernel eventually slides off the edge into some explicit border; rather, it gives \textbf{circular convolution}, where the kernel that slides off the right edge wraps around and picks up pixels from the left edge instead. The two only agree if the border handling on the spatial side is \emph{also} wraparound (\code{cv2.BORDER\_WRAP}). Checked directly on a $64\times64$ random test image with a $15\times15$ Gaussian kernel: the max absolute difference between spatial (wraparound-border) convolution and frequency-domain multiplication is $1.14\times10^{-13}$ --- floating-point roundoff, not a real discrepancy.

In practice, though, frequency-domain filtering is rarely used in everyday computer vision. Most kernels --- box blur, Gaussian, Sobel, sharpen --- are small ($3\times3$ or $5\times5$), and direct spatial convolution is already fast at those sizes, so the fixed overhead of two forward FFTs and one inverse FFT usually isn't worth paying. The DFT earns its keep mainly once kernels get large.

\subsection*{Exercises}
\begin{enumerate}
  \item Rotate the grating in the spectrum-peak experiment (make the stripes diagonal instead of vertical) by building it as a function of $x\cos\theta + y\sin\theta$. Predict, then verify, where the spectrum peak moves to.
  \item Build a \textbf{band-pass} frequency mask (an annulus: keep frequencies between an inner and outer radius, block everything else) and apply it to a test photo. What kind of image content survives?
  \item Try \code{apply\_frequency\_filter} with a \emph{very} small Gaussian mask sigma (e.g.\ 3) versus a large one (e.g.\ 60) on a photo-like test image, and describe the trend as sigma increases toward the image size.
  \item Get \emph{linear} (not circular) convolution out of an FFT: zero-pad both the image and kernel to at least (image size $+$ kernel size $- 1$) along each axis before transforming, multiply the spectra, inverse-transform, then crop back down. Compare the result to \code{cv2.filter2D(..., borderType=cv2.BORDER\_CONSTANT)} instead of \code{BORDER\_WRAP}.
  \item Using \code{\%timeit}, measure direct convolution (\code{cv2.filter2D}) against FFT-based convolution for Gaussian kernel sizes 5, 15, 31, 63, and 127 on a $256\times256$ image. At roughly what kernel size does the FFT approach start winning?
\end{enumerate}

\vfill
\fullcode{lesson15}{lesson15_fourier_frequency_filtering}

%% file: chapters/lesson16.tex
\chapter{Wavelets and Gabor Filters}
\label{ch:lesson16}

The Fourier transform (Chapter~\ref{ch:lesson15}) tells you \emph{which} frequencies are present in a signal, but not \emph{where} --- a sine wave basis function extends across the entire image, so a localized feature (an edge, a texture patch) gets smeared across the whole spectrum. \textbf{Wavelets} and \textbf{Gabor filters} are two different fixes for this: both are built from small, spatially localized oscillations instead of infinite sinusoids, giving a joint sense of \emph{where} and \emph{what frequency}.

\section{The Haar wavelet: the simplest possible wavelet}

The Haar wavelet transform splits a signal into an \textbf{approximation} (local averages) and a \textbf{detail} (local differences), computed on non-overlapping pairs of samples:
\begin{equation}
a_k = \frac{x_{2k} + x_{2k+1}}{\sqrt{2}}, \qquad d_k = \frac{x_{2k} - x_{2k+1}}{\sqrt{2}}.
\end{equation}
This should look familiar: it's essentially the same ``blur $+$ keep the residual'' idea as the Laplacian pyramid (Chapter~\ref{ch:lesson13}), just computed pairwise instead of with a Gaussian kernel, and without overlap between neighborhoods. A manual NumPy computation matches \code{pywt.dwt(signal, 'haar')} exactly.

\section{From Haar to Daubechies: smoother wavelets}

\begin{figure}[h]
\centering
\includegraphics[width=0.75\textwidth]{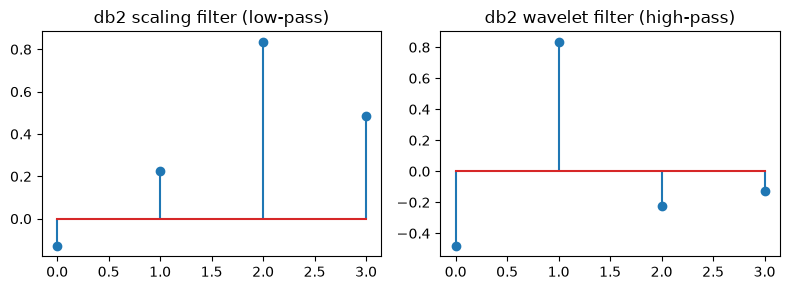}
\caption{\code{db2}'s scaling (low-pass) and wavelet (high-pass) filters, four taps each.}
\label{fig:l16-1}
\end{figure}

The Haar wavelet is discontinuous (a hard step), which gives it poor frequency localization --- its own frequency content is spread out, the opposite of what we want. \textbf{Daubechies wavelets} (\citelink{daubechies1988}{Daubechies, 1988}) use longer, smoother filters with more \emph{vanishing moments}, trading a wider spatial support for much better frequency behavior. \code{db2} (sometimes called ``D4'' for its 4 filter taps) is the next step up in smoothness from Haar, shown in Figure~\ref{fig:l16-1}.

The Haar transform above is just a convolve-and-downsample-by-2 operation with the 2-tap filters $[\tfrac{1}{\sqrt2},\tfrac{1}{\sqrt2}]$ and $[\tfrac{1}{\sqrt2},-\tfrac{1}{\sqrt2}]$. Daubechies wavelets follow the exact same recipe with longer filters. A from-scratch periodic-convolution implementation of \code{db2}'s decomposition matches \code{pywt} exactly, and inverting it (\code{pywt.idwt}) reconstructs the original signal perfectly.

\section{2D wavelet decomposition of an image}

\begin{figure}[h]
\centering
\includegraphics[width=0.95\textwidth]{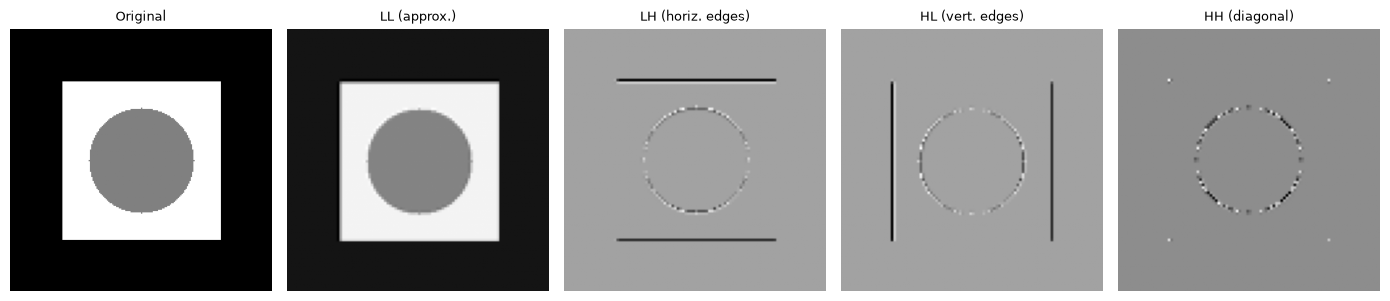}
\caption{A test image and its four \code{db2} subbands.}
\label{fig:l16-2}
\end{figure}

Like the separable Gaussian filter in Chapter~\ref{ch:lesson10}, a 2D wavelet transform is applied as two 1D passes: rows, then columns. One level of decomposition splits an image into \textbf{four subbands}:
\begin{itemize}
  \item \textbf{LL}: low-pass both directions --- a coarser, half-resolution copy of the image (like one Gaussian pyramid level)
  \item \textbf{LH}: low-pass rows, high-pass columns --- responds to horizontal edges
  \item \textbf{HL}: high-pass rows, low-pass columns --- responds to vertical edges
  \item \textbf{HH}: high-pass both directions --- responds to diagonal detail and corners
\end{itemize}
Figure~\ref{fig:l16-2} shows all four subbands for a test image.

Just like the Laplacian pyramid, this decomposition is exactly invertible: \code{pywt.idwt2} reconstructs the original from the four subbands with essentially no loss (max error $1.4\times10^{-13}$, floating-point roundoff).

\section{Gabor filters: localized oscillations tuned to orientation}

A \textbf{Gabor filter} is a sinusoidal grating multiplied by a Gaussian envelope --- a wave that's localized in space, tuned to a specific frequency \emph{and} orientation. Unlike Daubechies wavelets (built for compact, orthogonal, invertible multi-resolution decomposition), Gabor filters are used more for \emph{feature extraction}: detecting oriented texture and edges at a chosen scale.

\begin{figure}[h]
\centering
\includegraphics[width=0.6\textwidth]{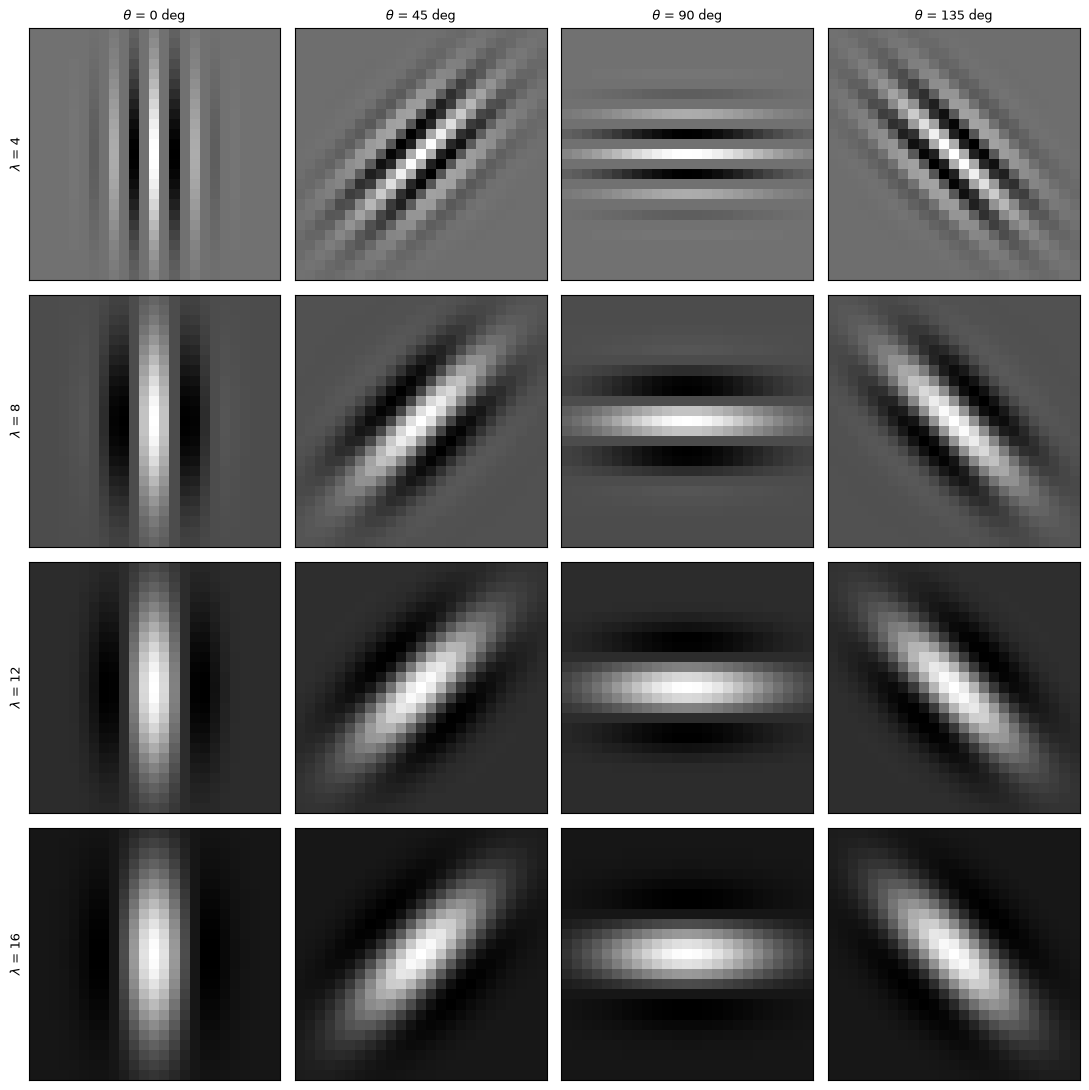}
\caption{Gabor kernels across four orientations ($\theta$) and four wavelengths ($\lambda$).}
\label{fig:l16-3}
\end{figure}

Gabor filters also have a striking biological connection. Recording neurons in a cat's visual cortex, Hubel and Wiesel (\citelink{hubelwiesel1959}{Hubel \& Wiesel, 1959}) found ``simple cells'' that fire selectively for a bar or edge at one specific orientation and position, and barely at all for other orientations --- a foundational discovery in visual neuroscience. It was later shown (\citelink{marcelja1980}{Marcelja, 1980}; \citelink{daugman1985}{Daugman, 1985}) that a 2D Gabor function is a remarkably good mathematical model of these simple-cell receptive fields, part of why Gabor filters became a standard, biologically-motivated tool for orientation-selective feature extraction in computer vision. Figure~\ref{fig:l16-3} shows a bank of Gabor kernels across several orientations and wavelengths.

\subsection*{Orientation selectivity, demonstrated quantitatively}

\begin{figure}[h]
\centering
\includegraphics[width=0.35\textwidth]{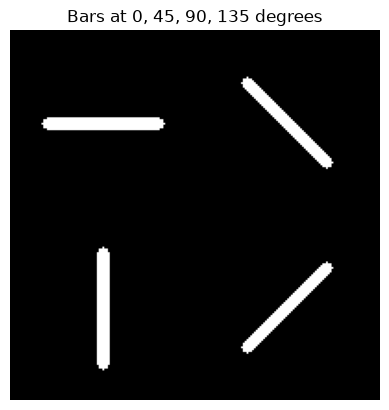}
\caption{Test bars at $0^\circ$, $45^\circ$, $90^\circ$, $135^\circ$.}
\label{fig:l16-4}
\end{figure}

Four bars at four different orientations (Figure~\ref{fig:l16-4}), each filtered with a Gabor kernel tuned to each orientation; a filter should respond most strongly to the bar matching its tuning and weakly to the others.

\begin{center}
\renewcommand{\arraystretch}{1.3}
\begin{tabular}{lrrrr}
\hline
\textbf{Filter tuned for} & \multicolumn{4}{c}{\textbf{Response to bar at}} \\
 & $0^\circ$ & $45^\circ$ & $90^\circ$ & $135^\circ$ \\
\hline
$0^\circ$   & \textbf{2580} & 612 & 343 & 612 \\
$45^\circ$  & 495 & \textbf{3277} & 495 & 428 \\
$90^\circ$  & 343 & 612 & \textbf{2580} & 612 \\
$135^\circ$ & 495 & 428 & 495 & \textbf{3277} \\
\hline
\end{tabular}
\end{center}

The response matrix is strongly diagonal-dominant: each filter's largest response (bold) lands squarely on the bar it was tuned for, exactly the orientation selectivity Hubel and Wiesel observed biologically.

\subsection*{Exercises}
\begin{enumerate}
  \item Try \code{db4} (\code{pywt.Wavelet('db4')}) instead of \code{db2} for the 2D image decomposition. How does the LH/HL/HH subband appearance change, given \code{db4}'s longer, smoother filters?
  \item Apply \code{pywt.dwt2} a second time to the LL subband from the image decomposition above, to get a second, coarser level --- this is a \textbf{wavelet pyramid}, directly analogous to the Gaussian/Laplacian pyramids of Chapters~\ref{ch:lesson11} and \ref{ch:lesson13}.
  \item Increase \code{lambd} (the sinusoid's wavelength) in the Gabor kernel while keeping \code{sigma} fixed. What happens to the number of visible stripes inside the Gaussian envelope, and how would you expect that to change which real-image texture scale the filter responds to?
\end{enumerate}

\vfill
\fullcode{lesson16}{lesson16_wavelets_gabor}

%% file: chapters/lesson17.tex
\chapter{Image Compression --- Lossless and Lossy}
\label{ch:lesson17}

Storing an image as raw pixels costs 1 byte per channel per pixel, no matter what the image contains. Compression exploits various types of redundancy in images --- i.e., the fact that real images are not random. This chapter covers both \textbf{lossless} compression (the decoded image is bit-for-bit identical to the original --- e.g., PNG) and \textbf{lossy} compression (the decoded image is only an approximation --- e.g., JPEG).

\section{Lossless compression}

\subsection*{Run-length encoding (RLE)}

The simplest lossless scheme: instead of storing every pixel, store \code{(value, run length)} pairs for consecutive runs of identical pixels. This works great on images with large flat regions (synthetic graphics, scanned text, masks), but it does nothing useful on real photographs.

\begin{lstlisting}
def rle_encode(array):
    flat = array.ravel()
    change_points = np.where(np.diff(flat) != 0)[0] + 1
    starts = np.concatenate([[0], change_points])
    ends = np.concatenate([change_points, [len(flat)]])
    return [(flat[s], e - s) for s, e in zip(starts, ends)]
\end{lstlisting}

On a $80\times80$ image with a single flat rectangle (raw size $6400$ bytes), RLE reconstructs the image exactly and compresses it to roughly $369$ bytes --- a $17.3\times$ reduction.

\subsection*{Huffman coding: fewer bits for common values}

A more sophisticated lossless method is \textbf{Huffman coding}, which takes advantage of the fact that some pixel values (e.g., a common background gray) occur far more often than others. It builds a variable-length code where frequent values get short codes and rare values get longer ones --- which is provably optimal among prefix codes.

Shannon's \textbf{entropy} gives the theoretical floor on average bits/pixel for \emph{any} code based only on the value distribution (ignoring spatial structure):
\begin{equation}
H = -\sum_i p_i \log_2 p_i,
\end{equation}
where $p_i$ is the probability that a certain pixel value occurs in the image. On a synthetic noisy test image: naive fixed-width encoding costs $8.00$ bits/pixel, the Shannon entropy floor is $5.17$, and the from-scratch Huffman coder achieves $5.20$ --- very close to the theoretical bound (it can only match it exactly when every probability happens to be a power of 2).

Real lossless formats like PNG combine an idea like this (Huffman/arithmetic coding) with a \textbf{predictive filter} first --- predicting each pixel from its neighbors and encoding only the (usually small) prediction error, which has much lower entropy than the raw pixel values.

\section{Lossy compression: the DCT}

JPEG's core idea is to take advantage of the human visual system's tendency to overlook fine details. JPEG divides the image into small $8\times8$ blocks, then transforms each block into the frequency domain so that it can discard the high-frequency components that are barely noticeable to the human eye. JPEG uses the \textbf{discrete cosine transform (DCT)}, a close relative of the Fourier transform from Chapter~\ref{ch:lesson15}. Like the DFT, the DCT re-expresses a block as a sum of frequency components --- but it uses only cosines (no imaginary part) to avoid boundary artifacts.

On an $8\times8$ block straddling a sharp edge: $83.1\%$ of the block's energy lands in just the top-left $2\times2$ DCT coefficients, and $92.4\%$ in the top-left $4\times4$ --- the DCT concentrates most of a block's information into a handful of low-frequency coefficients.

\subsection*{Quantization: throwing away the coefficients that matter least}

\begin{figure}[h]
\centering
\includegraphics[width=0.85\textwidth]{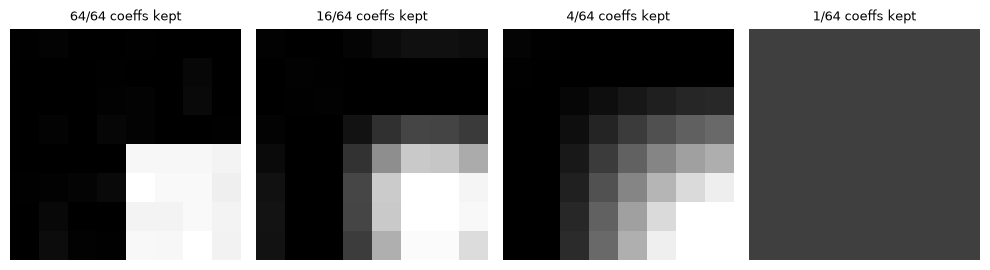}
\caption{Reconstructing an $8\times8$ block from only the top-left $n\times n$ DCT coefficients, for $n=8,4,2,1$. Even keeping just the top-left $2\times2$ (4 of 64 coefficients --- a $16\times$ reduction) preserves most of the block's overall structure; only the finest detail is lost.}
\label{fig:l17-1}
\end{figure}

Compression happens by \textbf{quantizing} the DCT coefficients --- dividing by some value (from a quantization table designed around human visual sensitivity) and rounding. Many coefficients, especially high-frequency ones carrying the least energy, become exactly zero and take almost no space to store, as Figure~\ref{fig:l17-1} illustrates.

\section{Real JPEG: the rate-distortion tradeoff}

\begin{figure}[h]
\centering
\includegraphics[width=0.85\textwidth]{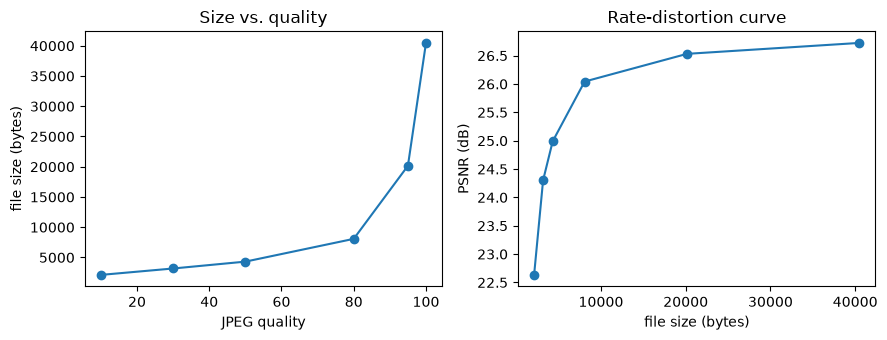}
\caption{File size vs.\ JPEG quality (left), and the rate-distortion curve: PSNR vs.\ file size (right).}
\label{fig:l17-2}
\end{figure}

Real JPEG encoding follows this same DCT-then-quantize recipe per block (with a carefully designed, frequency-dependent quantization table, plus run-length and Huffman coding of the resulting sparse coefficients --- combining the lossy and lossless ideas from this chapter). Sweeping OpenCV's JPEG quality setting on a synthetic color image and measuring both file size and PSNR (peak signal-to-noise ratio), plotted in Figure~\ref{fig:l17-2}:

\begin{center}
\renewcommand{\arraystretch}{1.3}
\begin{tabular}{rrrr}
\hline
\textbf{Quality} & \textbf{File size (bytes)} & \textbf{Compression ratio} & \textbf{PSNR (dB)} \\
\hline
10  & 2{,}113  & 56.8$\times$ & 22.63 \\
30  & 3{,}173  & 37.8$\times$ & 24.30 \\
50  & 4{,}307  & 27.9$\times$ & 24.99 \\
80  & 8{,}074  & 14.9$\times$ & 26.04 \\
95  & 20{,}156 & 6.0$\times$  & 26.53 \\
100 & 40{,}473 & 3.0$\times$  & 26.72 \\
\hline
\end{tabular}
\end{center}

This is the fundamental tradeoff of lossy compression: every extra byte you're willing to spend buys diminishing returns in quality --- the rate-distortion curve flattens out at high quality, and there is no free lunch, only a choice of where on the curve to sit. For a typical photograph, JPEG quality of 75 will not be noticeable unless you zoom in, and it will reduce the file size by 6--8$\times$.

\section{JPEG or PNG? Choosing the right format}

The trade-offs above translate into a practical rule:
\begin{itemize}
  \item Use \textbf{JPEG for photographs}. Photographs are dominated by smooth gradients --- exactly what the DCT concentrates into a few low-frequency coefficients, so discarding the rest is barely visible and JPEG buys a large size reduction for a small perceptual loss.
  \item Use \textbf{PNG for graphics} (screenshots, diagrams, text, icons, anything with flat regions or sharp edges). Graphics are the opposite case: a sharp edge spreads energy across \emph{all} DCT frequencies, so the same quantization that photos tolerate shows up as visible ringing and blocking around edges and text. Lossless compression handles flat regions and sharp edges almost for free (that's exactly what RLE and the predictive filter above are good at), so PNG is often both artifact-free \emph{and} smaller than JPEG for this kind of content.
\end{itemize}

\begin{figure}[h]
\centering
\includegraphics[width=0.75\textwidth]{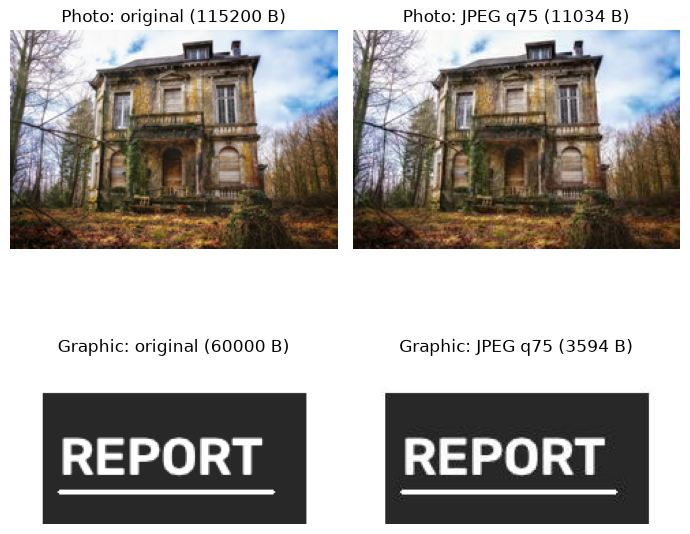}
\caption{Photograph vs.\ graphic, original and JPEG-compressed at quality 75. \attrib{Photo by \href{https://unsplash.com/@tama66}{Peter Herrmann} on \href{https://unsplash.com/photos/brown-and-white-concrete-building-eZaEWy2rAIc}{Unsplash}.}}
\label{fig:l17-3}
\end{figure}

On a real photograph ($115{,}200$ bytes raw): PNG compresses to $84{,}639$ bytes, JPEG (quality 75) to $11{,}034$ bytes --- JPEG wins by $7.7\times$. On a synthetic graphic with flat regions and text ($60{,}000$ bytes raw): PNG compresses to just $1{,}920$ bytes, JPEG (quality 75) to $3{,}594$ bytes --- here PNG wins, at roughly $1.9\times$ smaller than JPEG. Figure~\ref{fig:l17-3} shows both cases.

\subsection*{A closer look at JPEG artifacts}

\begin{figure}[h]
\centering
\includegraphics[width=0.75\textwidth]{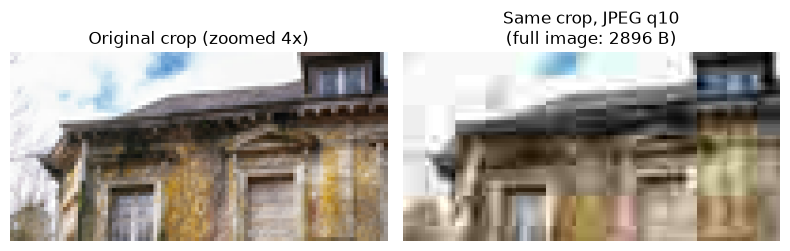}
\caption{A $4\times$ zoomed crop, original vs.\ JPEG quality 10 --- visible blocking and ringing at the roofline.}
\label{fig:l17-4}
\end{figure}

Zooming into a high-contrast region (a roofline against the sky) at a much lower quality setting makes the cost of aggressive quantization visible directly (Figure~\ref{fig:l17-4}): blocky $8\times8$ squares and ringing around the sharp edge.

\subsection*{Exercises}
\begin{enumerate}
  \item Run the RLE experiment on a noisy image instead of the flat-region one. How many runs does it produce, and what does that say about RLE's suitability for noisy natural images?
  \item \code{keep\_top\_left} is a deliberately crude stand-in for real quantization (which shrinks \emph{all} coefficients gradually rather than hard-zeroing a block of them). Replace it with \code{np.round(dct\_block / step) * step} for a few different \code{step} values, and compare the visual artifacts to the top-left-corner version.
  \item At very low JPEG quality, you may notice a visible grid pattern aligned with the $8\times8$ block boundaries (``blocking artifacts''). Given what block boundaries have to do with the DCT, explain why quantizing each block \emph{independently} would produce exactly this kind of artifact.
\end{enumerate}

\vfill
\fullcode{lesson17}{lesson17_compression}

%% file: chapters/lesson18.tex
\chapter{Color Spaces}
\label{ch:lesson18}

Previous chapters converted color images to grayscale and moved on. This chapter looks at what that conversion actually does, and at the other ways to represent color besides RGB, each suited to a different task: \textbf{HSV} (separating color from brightness), \textbf{YCbCr} (separating luminance from chrominance, the basis of video compression from Chapter~\ref{ch:lesson17}), and \textbf{L*a*b*} (designed so distances match human perception).

\section{Grayscale is already a color-space choice}

The simplest way to convert color to grayscale is to average the three channels, $Y=(R+G+B)/3$. The human visual system is actually more sensitive to green than to the other colors, so it's better to weight green more; a formula based on actual experiments on human subjects gives the weights used in \code{cv2.cvtColor(img, cv2.COLOR\_BGR2GRAY)}:
\begin{equation}
Y = 0.299R + 0.587G + 0.114B.
\end{equation}
These weights come from human luminance sensitivity: the eye is far more sensitive to green than to blue, so equal-intensity green and blue should \emph{not} map to the same gray level, even though a naive channel average would treat them identically. On pure green vs.\ pure blue patches: the naive average gives $85$ for both (indistinguishable), while the perceptually weighted version correctly reports green as much brighter ($150$) than blue ($29$) at the same channel intensity.

\subsection*{Why this formula is outdated (even though widely used, and still useful)}

Two things are hiding in that formula worth knowing about:
\begin{enumerate}
  \item \textbf{Old primaries.} The weights $0.299, 0.587, 0.114$ come from the 1953 NTSC standard (Rec.\ 601), derived from the specific red/green/blue phosphors used in early color CRT televisions. Modern LED displays use more saturated primaries, so the colorimetrically correct weights for a modern screen are actually $Y=0.2126R+0.7152G+0.0722B$ (Rec.\ 709/sRGB) --- noticeably \emph{more} green-weighted and \emph{less} red-weighted. \code{cv2.cvtColor} still uses the old Rec.\ 601 weights, mainly for backward compatibility with decades of existing code.
  \item \textbf{Assumes linear light.} The weighted-sum formula is only true luminance when $R,G,B$ are \emph{linear} light values. But stored pixel values are gamma-\emph{encoded}, not linear. \code{cv2.cvtColor} applies the weights directly to the raw encoded 8-bit values without decoding gamma first --- fast, and close enough for most computer-vision purposes, but not the physically correct luminance a display-calibration or photometry application would need. (To see the problem: a pure blue image, RGB $=(0,0,255)$, converts to a gray value of $29$ everywhere --- which will incorrectly appear almost completely dark, even though a fully saturated blue is a fairly bright color to the eye.)
\end{enumerate}

\section{RGB channels are highly correlated}

\begin{figure}[h]
\centering
\includegraphics[width=0.85\textwidth]{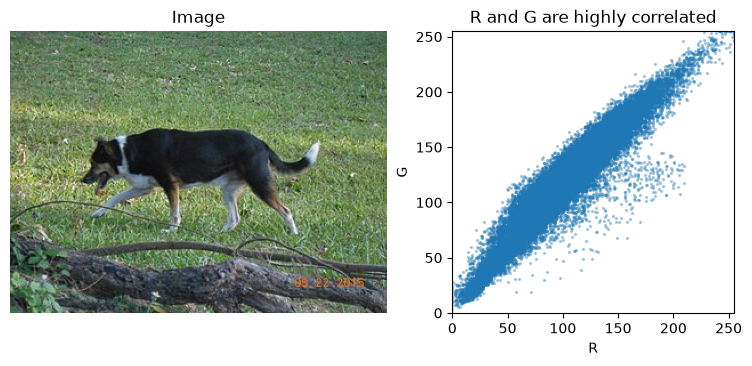}
\caption{A real photo and a scatter plot of its R and G channel values --- tightly correlated (correlation coefficient $0.954$ on this image), because shading scales all three channels together.}
\label{fig:l18-1}
\end{figure}

In most real images, lighting/shading variation dominates: a surface gets brighter or darker as a whole, scaling all three channels together. That means R, G, and B end up strongly correlated with each other --- not a very efficient or convenient way to separate ``what color is this'' from ``how brightly lit is this,'' as the scatter plot in Figure~\ref{fig:l18-1} shows.

\section{HSV: separating color from brightness}

HSV (Hue, Saturation, Value) reparametrizes color so that Hue captures \emph{which} color (independent of how bright or washed-out it is), Saturation captures how vivid/pure it is, and Value captures brightness alone. This makes color-based segmentation dramatically more robust to lighting than thresholding directly in RGB.

\begin{figure}[h]
\centering
\includegraphics[width=0.9\textwidth]{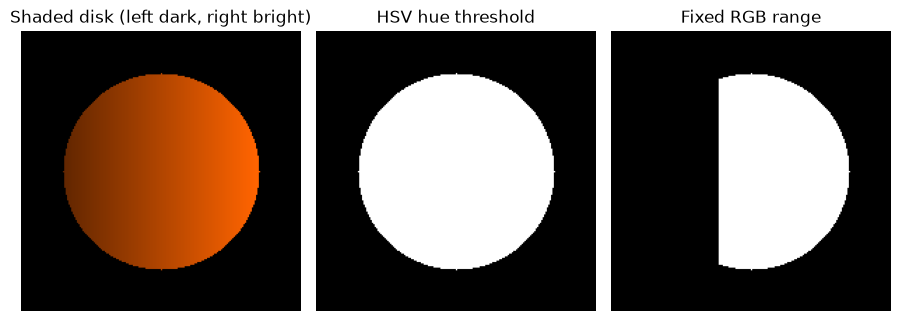}
\caption{A shaded disk (dark on the left, bright on the right), an HSV hue threshold, and a fixed RGB range. Hue thresholding recovers the disk regardless of the lighting gradient; the RGB range misses the darkened side entirely.}
\label{fig:l18-2}
\end{figure}

On a synthetic orange disk with a strong left-to-right lighting gradient: thresholding on HSV hue recovers the disk perfectly (IoU $1.000$ against the true disk), while a fixed RGB color range misses the darkened side entirely (IoU $0.710$), as Figure~\ref{fig:l18-2} shows.

\section{YCbCr: luma and chroma, and why video compresses color more aggressively}

YCbCr (used internally by JPEG and almost all video codecs) splits an image into \textbf{luma} ($Y$, roughly brightness) and two \textbf{chroma} channels ($C_b, C_r$, roughly ``how blue'' and ``how red''). The human visual system resolves fine spatial detail in luma far better than in chroma --- the biological basis for \textbf{chroma subsampling} (storing chroma at lower resolution than luma, e.g.\ video's common ``4:2:0'' format), one of the free compression wins used throughout Chapter~\ref{ch:lesson17}'s JPEG pipeline.

\begin{figure}[h]
\centering
\includegraphics[width=0.9\textwidth]{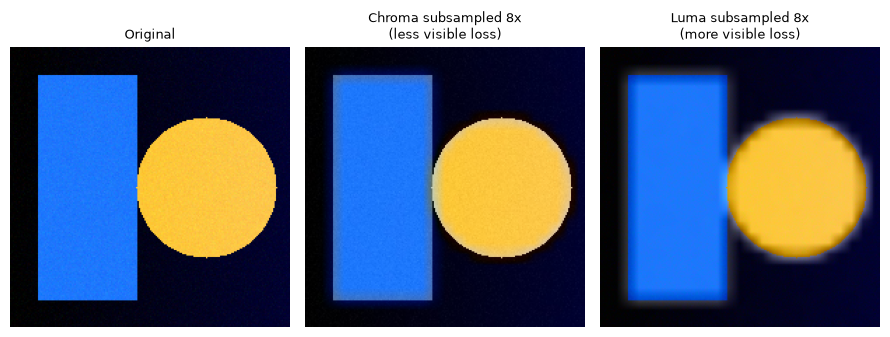}
\caption{Original, chroma subsampled $8\times$ (barely visible loss), and luma subsampled $8\times$ (clearly visible loss) --- the same detail loss is simply less objectionable when it happens to color rather than brightness.}
\label{fig:l18-3}
\end{figure}

Subsampling either channel $8\times$ and measuring reconstruction error against the original: degrading chroma costs a mean absolute error of $6.44$ gray levels, while degrading luma by the same factor costs $10.15$ --- discarding the same amount of resolution costs noticeably less error in chroma than in luma, as Figure~\ref{fig:l18-3} shows.

\section{L*a*b*: a perceptually uniform space}

RGB's Euclidean distance is a poor stand-in for how different two colors actually \emph{look}: equal RGB distances can correspond to wildly different perceived differences, depending on where in color space they fall. CIELAB was explicitly designed so that Euclidean distance between two L*a*b* colors approximates perceptual difference much more consistently --- useful for color-based quality metrics, clustering, and matching.

\begin{figure}[h]
\centering
\includegraphics[width=0.55\textwidth]{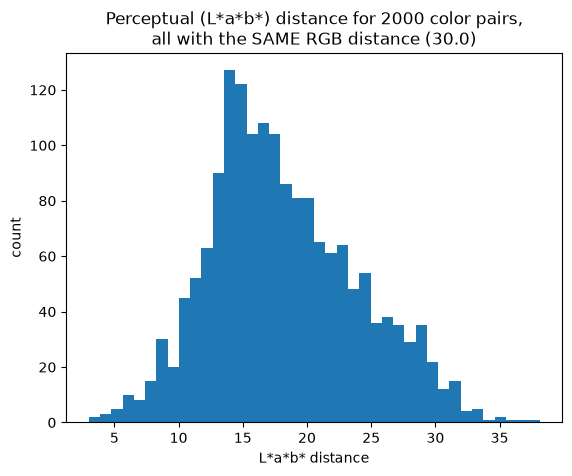}
\caption{Histogram of L*a*b* distances for 2000 color pairs all held at the same RGB distance. Any algorithm using raw RGB distance as a proxy for perceptual similarity (nearest-neighbor color matching, k-means color clustering, background-subtraction thresholds) inherits this same distortion; switching to Lab distance is a simple, standard fix.}
\label{fig:l18-4}
\end{figure}

Generating 2000 random color pairs, each held at \emph{exactly} the same RGB distance ($30.0$): the resulting L*a*b* distances range from a minimum of $3.0$ to a maximum of $38.1$ (median $17.3$) --- more than a $12\times$ spread, depending purely on \emph{where} in color space the pair sits, as the histogram in Figure~\ref{fig:l18-4} shows.

\subsection*{Exercises}
\begin{enumerate}
  \item Repeat the HSV vs.\ RGB segmentation demo with a shading gradient that also shifts hue slightly (e.g.\ blend toward a different color at one edge, not just changing brightness). Does hue-only thresholding still work as well?
  \item Increase the chroma/luma subsampling factor from 8 to 20. Does the qualitative story (chroma loss less visible than luma loss) still hold, or does it eventually break down?
  \item Find, by inspecting the color pairs in the Lab experiment, one example pair with a very \emph{small} Lab distance and one with a very \emph{large} L*a*b* distance (despite both having the same RGB distance). Display the four colors as swatches and describe, in your own words, why the low-Lab-distance pair looks more similar.
\end{enumerate}

\vfill
\fullcode{lesson18}{lesson18_color_spaces}

%% file: chapters/lesson19.tex
\chapter{Clustering --- k-Means, GMMs, and DBSCAN}
\label{ch:lesson19}

Every prior chapter in this part assumed some \emph{target} --- a threshold, a template, a known transform. \textbf{Clustering} asks a different kind of question, with no target at all: given a pile of points (or pixels), group them into a small number of sensible categories, using only how similar they are to each other. This is \textbf{unsupervised learning}'s classical starting point --- no labels, just a distance function and an assumption about what ``a cluster'' should look like. This chapter builds three clustering algorithms from scratch, each built on a different assumption about cluster shape, and shows concretely where each one wins and fails: \textbf{k-means} (round, similarly-sized clusters), \textbf{Gaussian mixture models} (elliptical clusters, soft assignment), and \textbf{DBSCAN} (arbitrary shape, defined by density rather than a center).

\section{k-means on real color data}

Chapter~\ref{ch:lesson18} built a perceptually meaningful color distance (L*a*b* Euclidean distance). Clustering is the natural next step: given a pile of pixel colors and no labels at all, group them into regions automatically.

K-means, also known as Lloyd's algorithm, is here unchanged from its textbook form: pick $k$ random points as initial centers, assign every point to its nearest center, recompute each center as the mean of its assigned points, repeat until nothing moves.

\begin{figure}[h]
\centering
\includegraphics[width=0.9\textwidth]{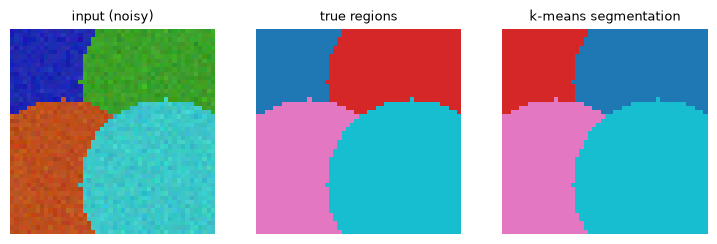}
\caption{A synthetic noisy 4-region color image, the true regions, and the k-means segmentation on L*a*b* pixel colors --- purity $1.000$ against the true regions.}
\label{fig:l19-1}
\end{figure}

With compact, roughly equal-sized, roughly round clusters in L*a*b* space (exactly what this synthetic image has), k-means recovers the true regions almost perfectly, as Figure~\ref{fig:l19-1} shows. Two real caveats worth flagging even in this success case: $k$ (the number of clusters) has to be chosen in advance --- nothing in the algorithm discovers it --- and different random initializations can converge to different, sometimes noticeably worse, local optima; \textbf{k-means++} (\citelink{arthur2007}{Arthur \& Vassilvitskii, 2007}) fixes this by seeding centers to be spread apart rather than purely at random.

\section{Where k-means breaks: non-convex clusters}

\begin{figure}[h]
\centering
\includegraphics[width=0.7\textwidth]{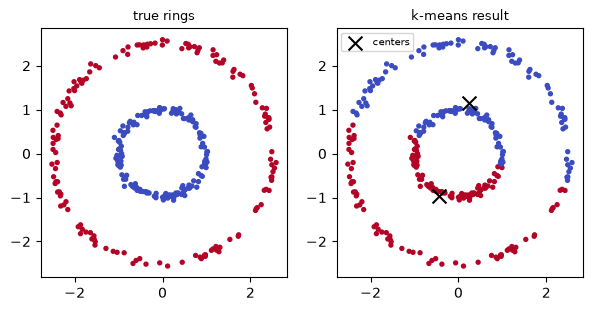}
\caption{True concentric rings vs.\ the k-means result (centers marked with $\times$). The clusters aren't ambiguous to a human eye at all; the \emph{algorithm's assumption} (round, center-based clusters) is simply the wrong tool for this shape.}
\label{fig:l19-2}
\end{figure}

K-means' update rule --- assign to the nearest \emph{center} --- can only carve space into convex regions. A dataset that isn't convex at all: two concentric rings. No placement of two centers can separate them correctly, no matter how many times the algorithm iterates. K-means lands right at chance ($54.0\%$, vs.\ $50\%$ for random guessing on this 2-class problem) --- it splits the two rings radially, half of each ring going to each cluster, because that's the only kind of boundary (a straight line, equidistant between two centers) the algorithm is capable of drawing, as Figure~\ref{fig:l19-2} shows.

\section{Gaussian mixture models: soft, elliptical clusters}

A \textbf{Gaussian mixture model (GMM)} generalizes k-means two ways: each cluster gets its own full covariance (so clusters can be elongated ellipses, not just circles), and assignment is \emph{soft} --- every point gets a probability of belonging to each cluster, not a hard label. Fit by \textbf{expectation-maximization (EM)} (\citelink{dempster1977}{Dempster, Laird, \& Rubin, 1977}): the E-step computes each point's responsibility (probability) under the current Gaussians; the M-step re-fits each Gaussian's mean, covariance, and weight to its responsibility-weighted points.

On the concentric rings, GMM does no better than k-means: $55.7\%$ accuracy, also near chance. That's an important, easy-to-miss point: GMM fixes k-means' \emph{round-only} assumption, but it's still a \textbf{unimodal-per-cluster} model --- each cluster is described by one Gaussian bump. A ring isn't elliptical any more than it's circular; no single Gaussian, of any shape, fits an annulus well. GMM and k-means fail this dataset for the same underlying reason.

\section{DBSCAN: clusters defined by density, not shape}

\textbf{DBSCAN} (density-based spatial clustering, \citelink{ester1996}{Ester et al., 1996}) never fits a parametric shape at all. A point is a \textbf{core point} if at least \code{min\_samples} other points lie within distance \code{eps} of it. Clusters are formed by chaining together core points that are within \code{eps} of each other (and their neighbors), so a cluster can be any shape --- including a ring --- as long as it's a \emph{connected, sufficiently dense} region of space. Points that end up in no core point's neighborhood are labeled \textbf{noise}, not forced into the nearest cluster.

\begin{figure}[h]
\centering
\includegraphics[width=0.95\textwidth]{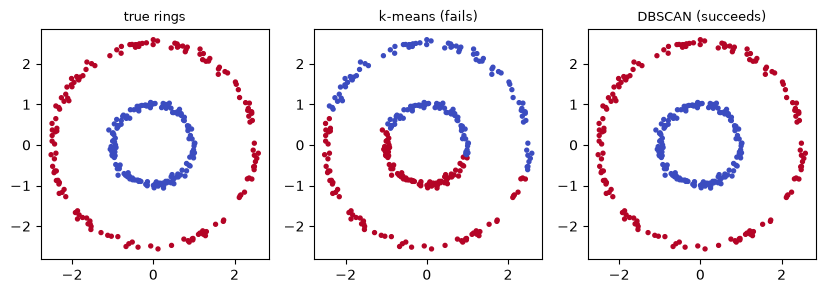}
\caption{True rings, the k-means result (fails), and the DBSCAN result (succeeds). DBSCAN never assumed a center or a shape, only that points \emph{within} a ring are densely connected to their neighbors, while the gap between rings is not.}
\label{fig:l19-3}
\end{figure}

On the same rings: DBSCAN finds exactly 2 clusters, 0 noise points, and $100.0\%$ accuracy (Figure~\ref{fig:l19-3}).

This comes at a real cost: \code{eps} and \code{min\_samples} have to be chosen by hand (a poor choice fragments each ring into dozens of tiny arcs), and DBSCAN struggles when different true clusters have very different densities, since one global \code{eps} can't be simultaneously right for a sparse cluster and a dense one.

\section{Noise robustness: a real advantage of ``no forced assignment''}

Add clutter --- points scattered uniformly across the whole region, belonging to neither ring --- and compare how each algorithm handles them. k-means and GMM have no concept of ``doesn't belong to any cluster'': every point gets assigned somewhere, dragging cluster centers toward the clutter. DBSCAN can simply call clutter what it is: out of 60 clutter points added, k-means correctly flags $0$ as anything other than ``definitely part of a ring'' (it has no vocabulary for that at all), while DBSCAN correctly isolates $8$ of them as noise. The rest happen to land within \code{eps} of a real ring point (or of each other, forming a small cluster of their own) --- a fair outcome, not a bug, since density-based clustering only calls a point noise if it's genuinely isolated. The qualitative point stands regardless of the exact count: DBSCAN has a real, usable category for ``this doesn't fit anywhere,'' whereas k-means structurally cannot.

\section{Where this goes next}

This is the end of Part 1. Many of the algorithms encountered so far operate directly on \textbf{raw or hand-picked features} --- pixel color, $(x,y)$ coordinates, texture --- and rely on human-chosen distance functions, such as those used here. Later chapters revisit many of these problems using deep learning to automatically discover the features and the rules for combining them: Chapter 43's instance segmentation uses a similar clustering idea, and Chapter 48's autoencoder learns the feature space itself, purely from reconstruction error, before any grouping is asked for at all. Where this chapter's methods assume a shape and fit it, the deep-learning methods later in this course learn a representation and let structure emerge from it.

\subsection*{Exercises}
\begin{enumerate}
  \item Rerun the color-segmentation k-means with $k=3$ and $k=6$ instead of the true $4$. Since nothing in the algorithm can ``know'' the right number of clusters, how does the segmentation degrade in each direction, and can you tell purely from the output which $k$ is likely wrong?
  \item The GMM above was evaluated only on the ring dataset, where it fails for the same reason k-means does. Construct a two-cluster dataset where the clusters are genuinely elliptical (e.g.\ using a multivariate normal with a non-diagonal covariance) and compare GMM against k-means there. Does GMM's extra flexibility (fitting the shape, not just the center) show a real accuracy advantage this time?
  \item In the noisy-clutter DBSCAN demo, shrink \code{eps} from $0.6$ to $0.3$. Does DBSCAN start mislabeling real ring points as noise, real clutter points as belonging to a ring, or both --- and does that match the earlier warning about \code{eps} being a hand-tuned, dataset-specific choice rather than something the algorithm discovers?
\end{enumerate}

\vfill
\fullcode{lesson19}{lesson19_clustering}

%% file: chapters/lesson20.tex
\chapter{Feature Detection and Matching}
\label{ch:lesson20}

This begins Part 2: classical 3D computer vision, where the algorithms increasingly work across \emph{multiple} images of a scene --- video frames, or several cameras --- rather than a single one. The first problem that creates: two images of the same scene rarely put the same content at the same pixel location, since rotation, scale, and viewpoint all shift things around. This chapter builds toward feature-based matching, which survives exactly those changes: first \textbf{corner detectors} (Harris, Shi-Tomasi) that find distinctive, repeatable points, then \textbf{SIFT}, which adds scale-invariance and a descriptor that can be matched between two different-looking images of the same scene.

\section{The structure tensor: what makes a good point to match?}

\begin{figure}[h]
\centering
\includegraphics[width=0.3\textwidth]{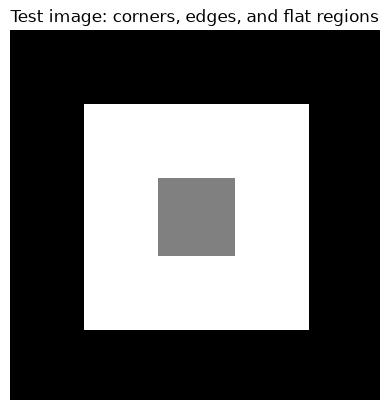}
\caption{A test image containing corners, edges, and flat regions.}
\label{fig:l20-1}
\end{figure}

At each pixel, build the $2\times2$ \textbf{structure tensor} from the local window of gradients (Chapter~\ref{ch:lesson12}):
\begin{equation}
M = \sum_{\text{window}} \begin{bmatrix}I_x^2 & I_xI_y \\ I_xI_y & I_y^2\end{bmatrix}.
\end{equation}
This should look familiar: it's exactly the moment covariance matrix from Chapter~\ref{ch:lesson06}, but built from gradients instead of pixel coordinates. Its eigenvalues $\lambda_1\ge\lambda_2$ describe the local intensity structure: both small means a \textbf{flat} region (gradients weak in every direction), one large and one small means an \textbf{edge} (gradient strong perpendicular to it, weak along it), and both large means a \textbf{corner} (gradient strong in every direction --- exactly the kind of point that can be precisely relocalized in another image), as illustrated on a test image in Figure~\ref{fig:l20-1}.

\section{Harris corner detector}

\begin{figure}[h]
\centering
\includegraphics[width=0.7\textwidth]{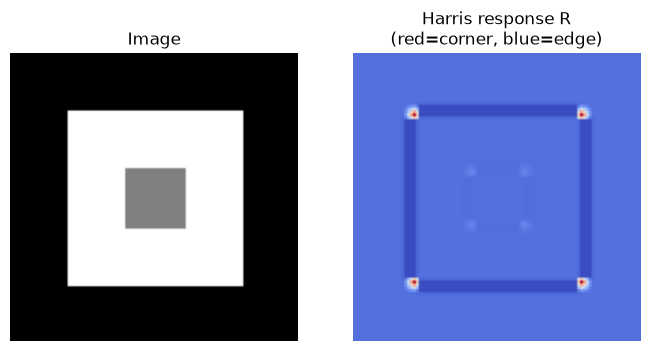}
\caption{The test image and its Harris response $R$ (red = corner, blue = edge).}
\label{fig:l20-2}
\end{figure}

Directly computing eigenvalues everywhere is a bit expensive, so Harris and Stephens (\citelink{harrisstephens1988}{1988}) proposed a cheaper proxy using only $\det(M)$ and $\operatorname{trace}(M)$ (computable without ever finding eigenvalues):
\begin{equation}
R = \det(M) - k \cdot \operatorname{trace}(M)^2, \qquad k \approx 0.04.
\end{equation}
$R$ is large and positive at corners, negative at edges, and near zero on flat regions. A from-scratch implementation correlates perfectly ($1.0$) with \code{cv2.cornerHarris}, shown in Figure~\ref{fig:l20-2}.

\section{Shi-Tomasi: ``Good Features to Track''}

\begin{figure}[h]
\centering
\includegraphics[width=0.3\textwidth]{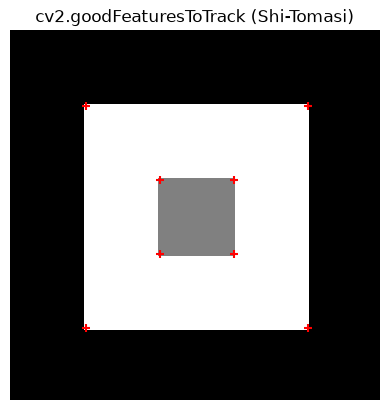}
\caption{\code{cv2.goodFeaturesToTrack} (Shi-Tomasi) corners marked in red.}
\label{fig:l20-3}
\end{figure}

Shi and Tomasi (\citelink{shitomasi1994}{1994}) argued for using the \textbf{minimum eigenvalue} of $M$ directly instead of Harris's determinant/trace proxy:
\begin{equation}
R_{\text{ST}} = \min(\lambda_1, \lambda_2).
\end{equation}
A point is only a strong corner if \emph{both} eigenvalues are large --- the minimum being large is exactly that condition, and (unlike Harris's $R$) it has a direct, easily interpretable meaning: it's proportional to the worst-case tracking precision in any direction. A from-scratch computation correlates perfectly ($1.0$) with \code{cv2.cornerMinEigenVal}, as Figure~\ref{fig:l20-3} shows.

\section{The scale problem}

Harris and Shi-Tomasi both operate at a single, fixed window size. Zoom the same corner in or out, and it may stop looking like a corner at that fixed scale (a sharp corner becomes a gentle curve when zoomed out enough) --- these detectors are not scale-invariant. \textbf{SIFT} (Scale-Invariant Feature Transform) (\citelink{lowe2004}{Lowe, 1999/2004}) fixes this by searching for keypoints across an entire scale-space, not just one window size.

\section{SIFT: scale-space extrema and descriptors}

\begin{figure}[h]
\centering
\includegraphics[width=0.9\textwidth]{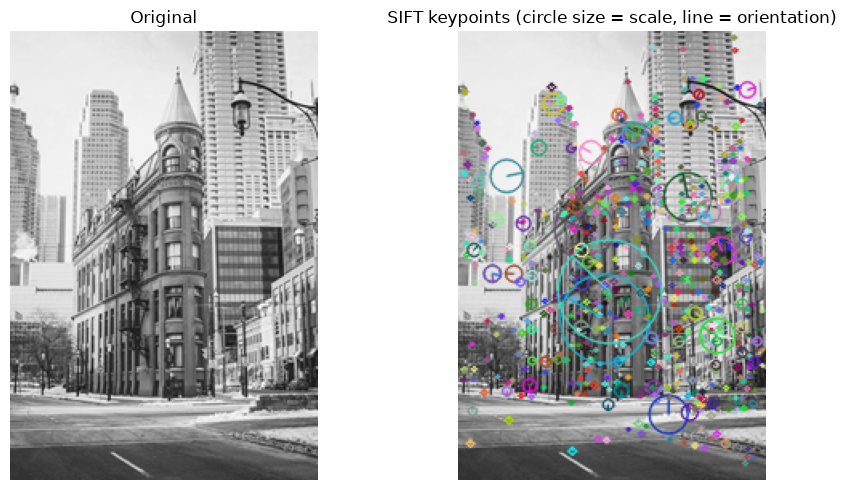}
\caption{SIFT keypoints overlaid on the original (circle size = scale, line = orientation). \attrib{Photo by \href{https://unsplash.com/@dawsonlovell}{Dawson Lovell} on \href{https://unsplash.com/photos/flatiron-building-in-toronto-W_MUqtuHwyY}{Unsplash}.}}
\label{fig:sift-keypoints}
\end{figure}

SIFT's pipeline, at a glance:
\begin{enumerate}
  \item Build a \textbf{Difference-of-Gaussians (DoG) scale-space} --- exactly the DoG approximation to the Laplacian from Chapter~\ref{ch:lesson13}, computed at many blur levels.
  \item Find keypoints as local extrema of DoG response across both space \emph{and} scale --- a point that's a maximum compared to its 26 neighbors (8 in the current scale, 9 in each scale level up and down).
  \item Assign each keypoint a dominant orientation from the local gradient histogram, so descriptors can be made rotation-invariant.
  \item Build a 128-dimensional descriptor from histograms of gradient orientations in a $4\times4$ grid of subregions around the keypoint.
\end{enumerate}
The result: a keypoint with a position, a scale, an orientation, and a descriptor vector designed to be very similar between two images even under moderate rotation, scaling, and illumination change. On a real photo, \code{cv2.SIFT\_create()} finds $723$ keypoints, each with a 128-dim descriptor, shown in Figure~\ref{fig:sift-keypoints}.

\section{Matching under rotation and scale}

Rotating and shrinking the image with a \emph{known} transform ($35^\circ$ rotation, $0.7\times$ scale), detecting SIFT features independently in both, and matching descriptors with a nearest-neighbor search: $723$ keypoints in the original, $365$ in the rotated-and-scaled version. \textbf{Lowe's ratio test} keeps a match only if the best candidate is meaningfully closer than the \emph{second}-best candidate --- a simple, effective way to reject ambiguous matches; here it keeps $191$ of $723$ candidate pairs.

\subsection*{How many matches are actually correct?}

\begin{figure}[h]
\centering
\includegraphics[width=0.95\textwidth]{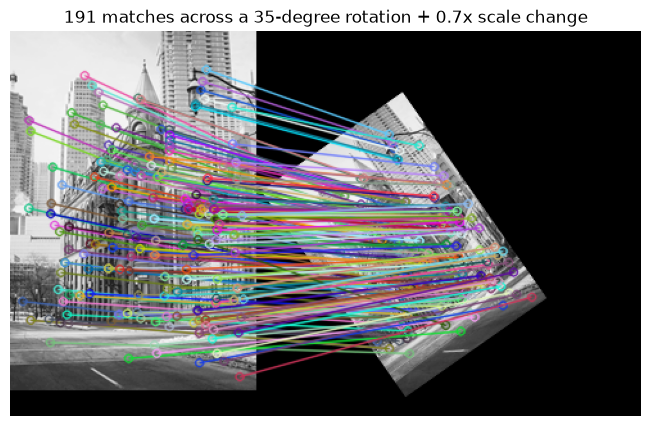}
\caption{$191$ matches across a $35^\circ$ rotation $+$ $0.7\times$ scale change. Nearly all the ratio-test survivors are genuinely correct correspondences, despite the rotation and scale change --- something a detector fixed to one scale and orientation simply couldn't recover.}
\label{fig:l20-5}
\end{figure}

Because the exact transform used to create the second image is known, each matched keypoint can be mapped through it and checked directly against where it was actually matched: $175$ of the $191$ ratio-test survivors ($92\%$) are geometrically correct (Figure~\ref{fig:l20-5}).

This is exactly what makes SIFT-style features the classic approach for tasks like image stitching, object recognition, and visual localization.

\subsection*{Exercises}
\begin{enumerate}
  \item Increase the rotation angle to $90^\circ$ and the scale to $0.3$. Does the number of good matches and the geometric-correctness percentage hold up, or degrade? At what point does SIFT start to struggle?
  \item Try \code{cv2.ORB\_create()} (a much faster, free binary-descriptor alternative to SIFT) with \code{cv2.NORM\_HAMMING} in the \code{BFMatcher} instead of the default L2 norm. Compare the number of good matches and qualitatively compare speed using \code{\%timeit} on \code{detectAndCompute}.
  \item Lower the ratio-test threshold from $0.75$ to $0.6$. How do the number of good matches and the fraction that are geometrically correct both change? What does this tell you about the threshold's role in a precision/recall tradeoff?
\end{enumerate}

\vfill
\fullcode{lesson20}{lesson20_feature_detection_matching}

%% file: chapters/lesson21.tex
\chapter{Optical Flow and Motion Estimation}
\label{ch:lesson21}

Optical flow estimates apparent motion between two frames of a video: for each pixel (or a chosen set of points), a 2D vector $(u,v)$ describing where it moved to. This chapter derives the classic \textbf{Lucas-Kanade} method from the same structure tensor used for corner detection in Chapter~\ref{ch:lesson20}, confronts the fundamental \textbf{aperture problem}, contrasts it with the globally-optimized \textbf{Horn-Schunck} method, and finishes with dense flow via the \textbf{Farneb\"ack} method.

\section{The brightness constancy assumption}

Optical flow assumes a point's intensity doesn't change as it moves: $I(x,y,t)=I(x+u,y+v,t+1)$. A first-order Taylor expansion of the right side gives the \textbf{optical flow constraint equation}:
\begin{equation}
I_x u + I_y v + I_t = 0,
\end{equation}
where $I_x,I_y$ are the spatial gradients (Chapter~\ref{ch:lesson12}) and $I_t$ is the frame-to-frame intensity difference. This is \textbf{one equation with two unknowns} ($u$ and $v$) at every single pixel --- not enough information on its own to solve for the flow.

\section{The aperture problem}

\begin{figure}[h]
\centering
\includegraphics[width=0.7\textwidth]{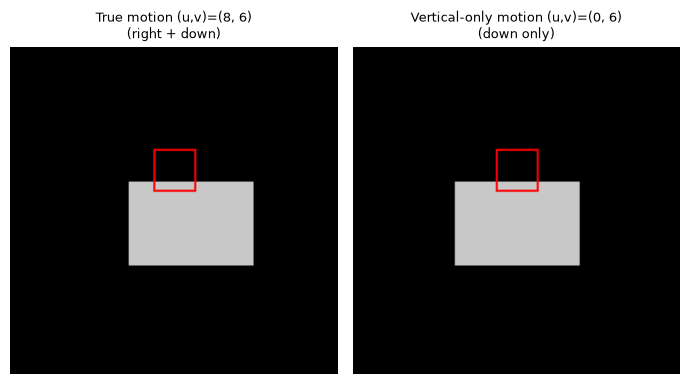}
\caption{Two different true motions (right$+$down vs.\ down-only). Inside a small window straddling the rectangle's top (horizontal) edge, the two motions look pixel-for-pixel identical --- verified directly: a local measurement genuinely cannot tell them apart.}
\label{fig:l21-1}
\end{figure}

Looking through a small window at a moving edge, only the motion \textbf{perpendicular to the edge} is visible; motion \textbf{along the edge} produces no visible change at all, as Figure~\ref{fig:l21-1} demonstrates. This is exactly why the flow constraint equation is underdetermined: it only ever constrains the component of $(u,v)$ along the gradient direction $(I_x,I_y)$, leaving the perpendicular component completely unconstrained.

\section{Lucas-Kanade: solving the aperture problem with a window}

Lucas and Kanade's fix (\citelink{lucaskanade1981}{1981}): assume the flow $(u,v)$ is \textbf{constant over a small window}, then combine the flow constraint equation from every pixel in that window into an overdetermined least-squares system:
\begin{equation}
\underbrace{\begin{bmatrix}\sum I_x^2 & \sum I_xI_y \\ \sum I_xI_y & \sum I_y^2\end{bmatrix}}_{M}\begin{bmatrix}u\\v\end{bmatrix} = -\begin{bmatrix}\sum I_xI_t\\ \sum I_yI_t\end{bmatrix}.
\end{equation}
$M$ is exactly the structure tensor from Chapter~\ref{ch:lesson20}! This is not a coincidence: solving this system requires $M$ to be invertible, i.e.\ to have two large eigenvalues --- precisely the Shi-Tomasi ``good feature to track'' condition. A flat region ($M$ near zero) or an edge (one small eigenvalue --- the aperture problem again) gives an ill-conditioned or singular system; a corner gives a well-conditioned one. \textbf{Corners are trackable for exactly the same reason they're good corners.}

\begin{figure}[h]
\centering
\includegraphics[width=0.5\textwidth]{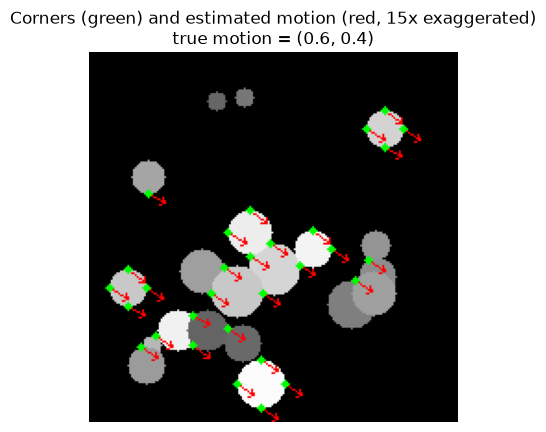}
\caption{Corners (green) and estimated motion (red arrows, exaggerated $15\times$ since the true motion is sub-pixel).}
\label{fig:l21-2}
\end{figure}

For a sub-pixel motion of $(0.6, 0.4)$ on a synthetic textured scene, a from-scratch Lucas-Kanade solve at 30 detected corners averages to $(0.612, 0.393)$ --- close to the true motion (Figure~\ref{fig:l21-2}).

\subsection*{Larger motions need iteration}

\begin{figure}[h]
\centering
\includegraphics[width=0.9\textwidth]{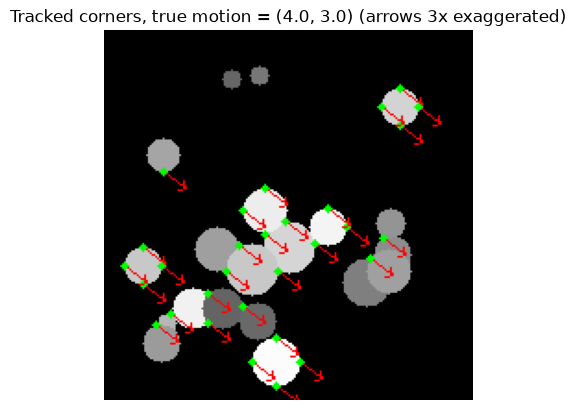}
\caption{Tracked corners with a larger true motion, using \code{cv2.calcOpticalFlowPyrLK}.}
\label{fig:l21-3}
\end{figure}

The single-shot linear solve relies on the Taylor approximation, which only holds for small motions. For a larger true shift of $(4.0, 3.0)$: the single-shot manual solve degrades badly, averaging to $(0.846, 0.690)$, while \code{cv2.calcOpticalFlowPyrLK} recovers it essentially exactly at $(4.000, 3.000)$, as shown in Figure~\ref{fig:l21-3}.

OpenCV's \code{calcOpticalFlowPyrLK} handles large motions by running Lucas-Kanade \textbf{iteratively} (re-warping and re-linearizing until convergence) on an \textbf{image pyramid} (Chapter~\ref{ch:lesson11}) --- estimate coarsely on a small, blurry version of the image first, then refine level by level. This combination lets it recover large motions accurately even though the underlying linear approximation is only valid locally, one small step at a time.

\section{Horn-Schunck: a global alternative to windowed flow}

\begin{figure}[h]
\centering
\includegraphics[width=0.9\textwidth]{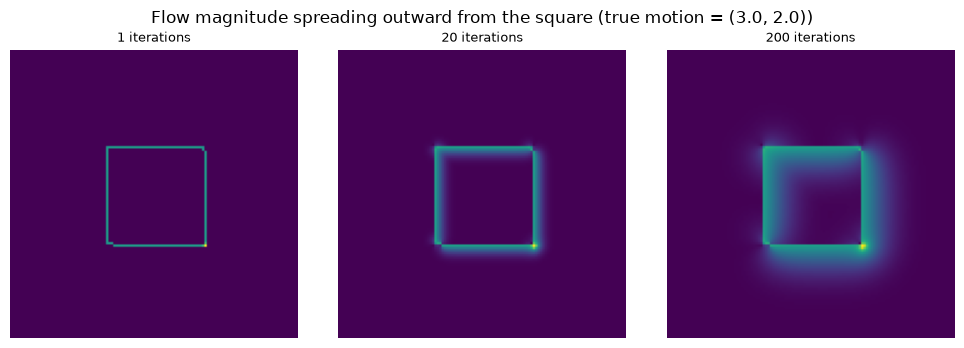}
\caption{Flow magnitude spreading outward from a single textured square, after 1, 20, and 200 Gauss-Seidel iterations. After just 1 iteration, flow is only nonzero right at the square's edges; by 200 iterations it has spread well beyond, though still far from every pixel --- a real solver would run more iterations (or use a pyramid) to converge everywhere.}
\label{fig:l21-4}
\end{figure}

Lucas-Kanade's window is a \emph{local} smoothness assumption: flow is constant over a small neighborhood, estimated independently at each point. The same year, Horn and Schunck (\citelink{hornschunck1981}{1981}) proposed a \emph{global} alternative: instead of many independent per-window solves, minimize a single energy over the whole image at once, trading off how well the flow satisfies the optical flow constraint equation against how smoothly it varies between neighboring pixels:
\begin{equation}
E(u,v) = \sum_{x,y} \underbrace{(I_x u + I_y v + I_t)^2}_{\text{data term}} \;+\; \alpha^2 \underbrace{\left(\|\nabla u\|^2 + \|\nabla v\|^2\right)}_{\text{smoothness term}}.
\end{equation}
The smoothness term is what makes this global: it couples every pixel's flow to its neighbors', so minimizing $E$ (e.g.\ by iterative Gauss-Seidel updates) lets reliable flow estimates near edges \emph{propagate} into flat, textureless regions that have no local information of their own --- exactly the failure mode the aperture problem produces at its worst, as Figure~\ref{fig:l21-4} shows.

\section{Farneb\"ack's method for dense optical flow}

Horn-Schunck's global optimization is elegant but expensive and slow to converge; for dense (every-pixel) flow, Farneb\"ack's method (\citelink{farneback2003}{2003}) --- exposed as \code{cv2.calcOpticalFlowFarneback} --- is a more popular method actually used in practice. It stays local, like Lucas-Kanade, but drops the constant-flow-in-a-window assumption in favor of locally approximating each neighborhood with a polynomial and comparing the polynomial expansions between frames.

\begin{figure}[h]
\centering
\includegraphics[width=0.85\textwidth]{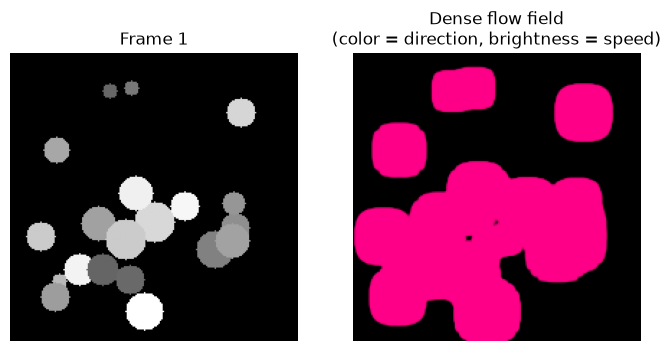}
\caption{A synthetic frame and its dense Farneb\"ack flow field, visualized with color = direction, brightness = speed. Every moving circle produces the same color here, since they all share the same true motion.}
\label{fig:l21-5}
\end{figure}

On a synthetic scene with true motion $(5.0,-3.0)$: averaging the dense flow over \emph{all} pixels gives a biased $(2.25,-1.35)$, but restricting to textured (high-gradient) pixels recovers $(5.00,-3.00)$ almost exactly. This is the aperture problem again, at its most extreme: over flat, textureless background, there's no local information at all to estimate motion from, dragging the whole-image average away from the true value. Figure~\ref{fig:l21-5} shows the resulting flow field.

\subsection*{A real video: the Army sequence}

\begin{figure}[h]
\centering
\includegraphics[width=0.85\textwidth]{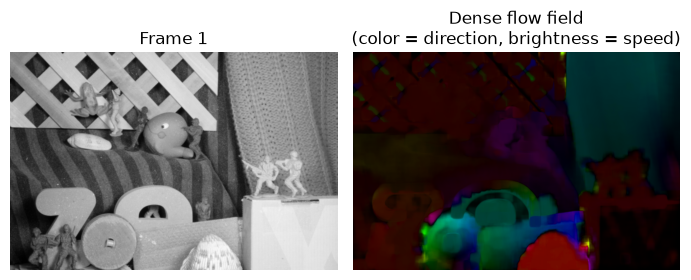}
\caption{A frame from the Army sequence and its dense flow field. \attrib{Image source: \href{https://vision.middlebury.edu/flow/data}{Middlebury Optical Flow}.}}
\label{fig:l21-6}
\end{figure}

On a real video with independently moving objects, this color-coding immediately separates different motions at a glance (Figure~\ref{fig:l21-6}) --- exactly why it's the standard way to visualize dense flow fields.

\subsection*{Exercises}
\begin{enumerate}
  \item Rerun the sparse Lucas-Kanade comparison with a true motion of $(10.0, 8.0)$. Does \code{cv2.calcOpticalFlowPyrLK} still recover it accurately? At what point (try increasingly large motions) does it start to fail, and why would you expect a pyramid to help push that limit further out?
  \item Modify the synthetic scene so the circles move with \emph{different} velocities (e.g.\ half moving one way, half another). Re-run the dense Farneb\"ack color visualization and confirm the two motions appear as two distinct colors.
  \item In the aperture-problem demo, construct a small textured (non-edge) patch instead of a straight edge, and show that \emph{both} components of an arbitrary motion vector are recoverable from it --- unlike the edge case, this should not have an invisible direction of motion.
\end{enumerate}

\vfill
\fullcode{lesson21}{lesson21_optical_flow}

%% file: chapters/lesson22.tex
\chapter{Stereo Matching}
\label{ch:lesson22}

Two cameras viewing the same scene from slightly different positions see the same 3D points shifted by different amounts depending on depth --- nearby points shift more, distant points shift less. \textbf{Stereo matching} finds these shifts (the \textbf{disparity}) at every pixel, which convert into depth via triangulation. The key problem is correspondence: finding, for every pixel in the left image, the matching pixel in the right. This chapter builds a \textbf{block matcher} that solves it by comparing small patches with sum-of-squared differences (SSD) --- a close cousin of Chapter~\ref{ch:lesson21}'s optical flow, but restricted to a 1D search along a single row instead of a full 2D neighborhood.

\section{Rectified stereo: why the search is 1D}

For a pair of cameras that are side-by-side, pointed the same direction, with parallel image planes (a \textbf{rectified} stereo pair --- real camera rigs are calibrated and warped to approximate this), a fundamental fact of epipolar geometry applies: the corresponding point for any pixel in the left image lies \textbf{on the same row} in the right image. This collapses the search from a 2D image search down to a 1D scan along one row, and the horizontal offset between the two matching positions is the \textbf{disparity} $d$.

Disparity relates to depth by
\begin{equation}
Z = \frac{f \cdot B}{d},
\end{equation}
where $f$ is the focal length and $B$ is the baseline (distance between the two camera centers). For a rectified pair, depth is \textbf{inversely proportional to disparity}: nearby objects have large disparity, distant objects have small disparity, and an object infinitely far away has zero disparity.

\section{A synthetic stereo pair with known ground truth}

\begin{figure}[h]
\centering
\includegraphics[width=0.9\textwidth]{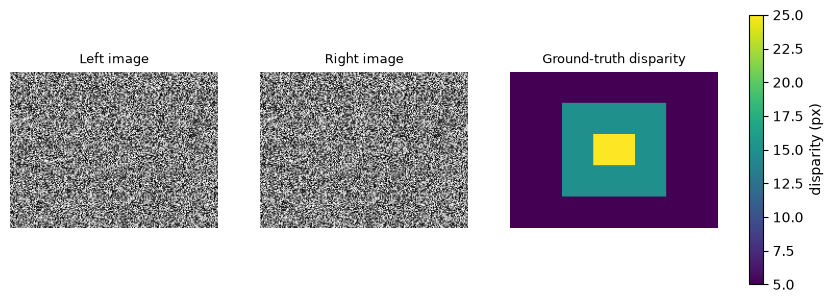}
\caption{Left image, right image, and ground-truth disparity (three depth planes).}
\label{fig:l22-1}
\end{figure}

A left image of pure random texture (so every patch is locally distinctive --- no aperture-problem ambiguity), with a right image constructed by shifting each pixel left by its true disparity, set to three different constant values for three depth ``planes'': a background and two nearer rectangles. Nearer surfaces uncover background pixels that have no corresponding pixel in the left image at all --- an \textbf{occlusion}, patched with fresh random noise. Figure~\ref{fig:l22-1} shows the resulting stereo pair and ground truth.

\section{Block matching along the scanline}

For each pixel in the left image, slide a small window along the \emph{same row} of the right image over a range of candidate disparities and keep the disparity with the lowest sum-of-squared-differences.

\begin{lstlisting}
for d in range(max_disp + 1):
    xr = x - d
    right_patch = right_f[y-half:y+half+1, xr-half:xr+half+1]
    cost = ((left_patch - right_patch)**2).sum()
    if cost < best_cost:
        best_cost, best_d = cost, d
\end{lstlisting}

\begin{figure}[h]
\centering
\includegraphics[width=0.9\textwidth]{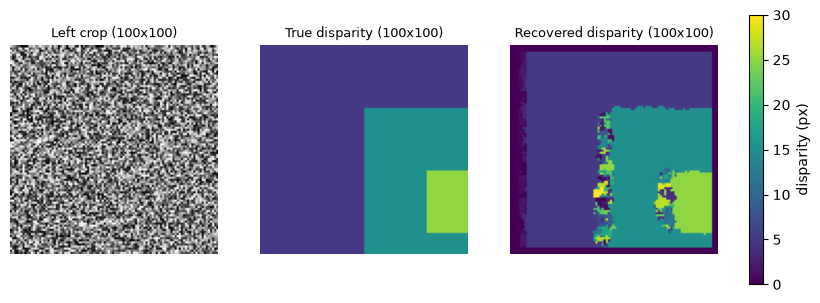}
\caption{A $100\times100$ crop, its true disparity, and the recovered disparity from a from-scratch block matcher --- the three depth planes are cleanly separated away from the border and depth-boundary regions.}
\label{fig:l22-2}
\end{figure}

As Figure~\ref{fig:l22-2} shows, within the interior of the rectangles, the recovered disparity matches the ground truth almost perfectly (mean disparity $5.03$, $15.03$, $25.04$ against true values $5$, $15$, $25$). The exception is the speckled band at each rectangle's edge: a window straddling a depth boundary mixes pixels from two different true disparities, so no single candidate disparity fits the whole window well --- a preview of the occlusion problem tackled next.

\section{Catching bad matches: the left-right consistency check}

\begin{figure}[h]
\centering
\includegraphics[width=0.9\textwidth]{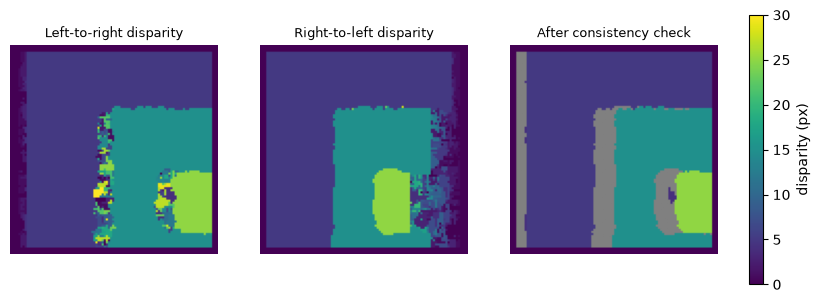}
\caption{Left-to-right disparity, right-to-left disparity, and the result after the consistency check (gray = rejected). On this crop, $1440$ of $10{,}000$ pixels are marked inconsistent, concentrated at occlusions and depth-boundary windows.}
\label{fig:l22-3}
\end{figure}

Occlusions and depth-boundary ambiguity produce wrong matches, like the speckle above. The fix: match both left-to-right and right-to-left, then keep only the pixels where the two results agree, discarding the rest as unreliable. Computing the right-to-left disparity needs no new code --- flip both images left-right, swap which one plays ``left,'' run the exact same matcher, then flip the result back. Figure~\ref{fig:l22-3} shows both directions and the consistency result.

\section{The same thing, at full resolution, with OpenCV}

\begin{figure}[h]
\centering
\includegraphics[width=0.7\textwidth]{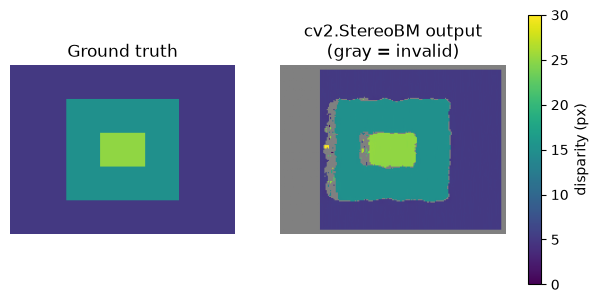}
\caption{Ground truth vs.\ \code{cv2.StereoBM} output at full resolution (gray = invalid).}
\label{fig:l22-4}
\end{figure}

\code{cv2.StereoBM} implements this same block-matching idea (with efficiency and post-processing refinements) fast enough to run on the whole image. \code{StereoBM} marks a pixel invalid wherever it isn't confident in a unique match, e.g.\ too close to the image border to search the full disparity range, as Figure~\ref{fig:l22-4} shows.

The left-right consistency check is also built into \code{cv2.StereoBM} --- set \code{disp12MaxDiff} to a small positive number (typically 1) to enable it. On the full synthetic image: $21{,}706$ valid pixels without the check, $21{,}442$ with it --- $264$ pixels newly flagged invalid.

\section{Window size: detail vs.\ noise tradeoff}

\begin{figure}[h]
\centering
\includegraphics[width=0.9\textwidth]{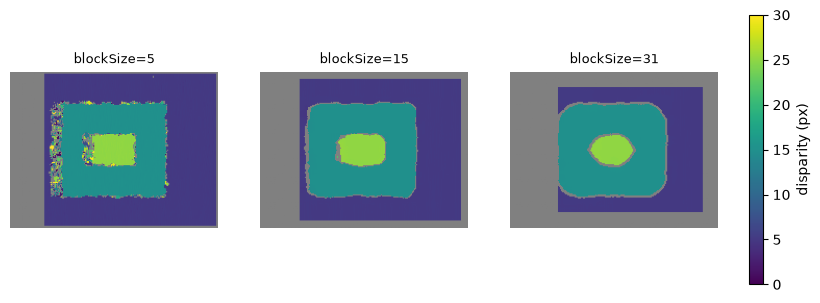}
\caption{\code{blockSize} 5, 15, and 31 on the same stereo pair --- small windows preserve sharp edges but are noisier; large windows are smoother but blur depth boundaries.}
\label{fig:l22-5}
\end{figure}

A larger matching window averages over more pixels, making the match more robust to noise but blurring across depth discontinuities. A smaller window preserves sharp depth boundaries but is more easily fooled by noise or repetitive texture, as Figure~\ref{fig:l22-5} shows.

\section{A real stereo pair: Tsukuba}

\begin{figure}[h]
\centering
\includegraphics[width=0.9\textwidth]{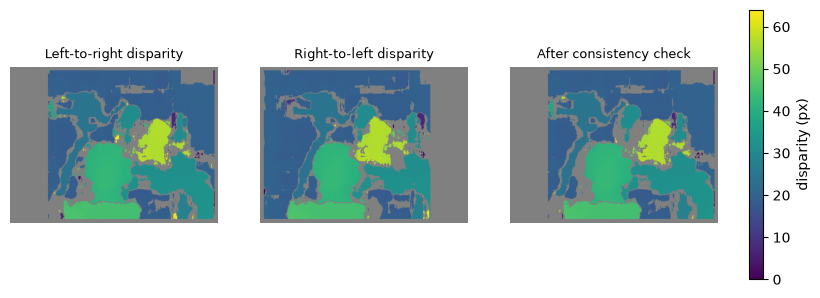}
\caption{Left-to-right, right-to-left, and consistency-checked disparity on the Tsukuba stereo pair. The lamp, the nearest object in the scene, stands out with the largest disparity; the bookshelves recede smoothly behind it. Only $1121$ of $110{,}592$ pixels ($1.0\%$) get flagged --- concentrated exactly along depth discontinuities: book spines, shelf edges, the head's silhouette. \attrib{Image source: \href{https://vision.middlebury.edu/stereo/data/scenes2001/}{Middlebury Stereo Datasets} (University of Tsukuba).}}
\label{fig:l22-6}
\end{figure}

Everything so far has used synthetic images, precisely so the ground truth is known. The same three passes (left-to-right, right-to-left, consistency check) apply exactly as before on an actual rectified stereo pair of photographs, shown in Figure~\ref{fig:l22-6}.

\section{From disparity to depth map}

\begin{figure}[h]
\centering
\includegraphics[width=0.45\textwidth]{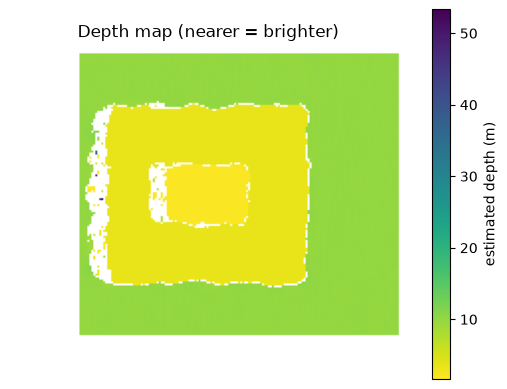}
\caption{The synthetic scene's disparity map converted to metric depth.}
\label{fig:l22-7}
\end{figure}

\begin{figure}[h]
\centering
\includegraphics[width=0.55\textwidth]{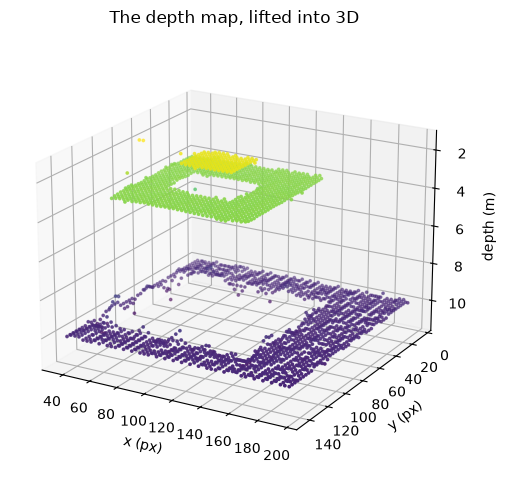}
\caption{The depth map lifted into 3D: three flat terraces at three different heights, with holes where \code{StereoBM} had no confident match cut cleanly out of each one.}
\label{fig:l22-8}
\end{figure}

Given a focal length and baseline (in whatever consistent units), $Z=fB/d$ converts a disparity map directly into a metric depth map (Figure~\ref{fig:l22-7}), which can then be lifted into a 3D point cloud (Figure~\ref{fig:l22-8}).

\subsection*{Exercises}
\begin{enumerate}
  \item Replace the random-texture left image with a flat gray region for the background (keeping the two rectangles textured). Run \code{cv2.StereoBM} again and describe what happens to the disparity estimate over the flat area --- this is the aperture problem from Chapter~\ref{ch:lesson21}, now in a stereo-matching context.
  \item Try \code{cv2.StereoSGBM\_create} (semi-global block matching, which enforces smoothness across neighboring disparities rather than matching each pixel completely independently) in place of \code{StereoBM}. Compare the amount of speckle noise in flat/background regions between the two methods.
  \item Increase the baseline in the depth conversion. How does the estimated depth map change, and why would a wider-baseline stereo rig give more precise depth estimates for distant objects (at the cost of a larger minimum-distance blind spot near the cameras)?
\end{enumerate}

\vfill
\fullcode{lesson22}{lesson22_stereo_matching}

%% file: chapters/lesson23.tex
\chapter{Model Fitting, Robust Estimation, and RANSAC}
\label{ch:lesson23}

Many computer vision problems boil down to fitting a model --- a line, a circle, a plane, a homography (Chapter~\ref{ch:lesson24}), a fundamental matrix --- to a set of corresponding points. In practice, that data almost always contains outliers: correspondences that are simply wrong. This chapter confronts that directly: least-squares fitting is catastrophically fragile to even a few bad points, and \textbf{RANSAC} is the standard fix.

\section{Ordinary vs.\ total least squares}

Fitting $y=mx+b$ by minimizing $\sum(y_i-(mx_i+b))^2$ --- \textbf{ordinary least squares (OLS)} --- measures error \emph{vertically}. That's the right choice when $x$ is known exactly and only $y$ is noisy, but not when both coordinates carry comparable noise, as they typically do for 2D image points: the fit gets biased toward whichever direction has less spread. \textbf{Total least squares (TLS)} instead minimizes the \emph{perpendicular} distance from each point to the line, treating both coordinates symmetrically.

For a 2D line, TLS needs no new machinery: the best-fit line passes through the centroid, and its direction is the eigenvector of the (centered) points' covariance matrix with the \emph{largest} eigenvalue --- exactly Chapter~\ref{ch:lesson06}'s covariance-and-eigenvectors idea, just picking out the axis of maximum spread instead of describing a blob's shape.

\begin{figure}[h]
\centering
\includegraphics[width=0.55\textwidth]{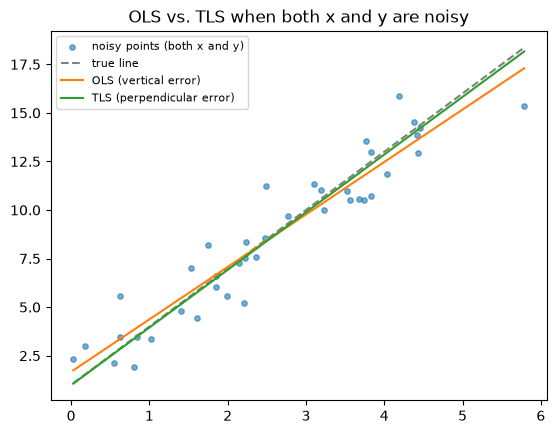}
\caption{OLS (vertical error) vs.\ TLS (perpendicular error) when both $x$ and $y$ are noisy.}
\label{fig:l23-1}
\end{figure}

On a steep true line $y=3.00x+1.00$ with comparable noise on both axes: OLS fits $y=2.70x+1.67$ (biased, since $x$ is noisy too), while TLS fits $y=2.97x+0.98$ --- noticeably closer to the true slope, as Figure~\ref{fig:l23-1} shows.

\section{Why least squares breaks under outliers}

\begin{figure}[h]
\centering
\includegraphics[width=0.55\textwidth]{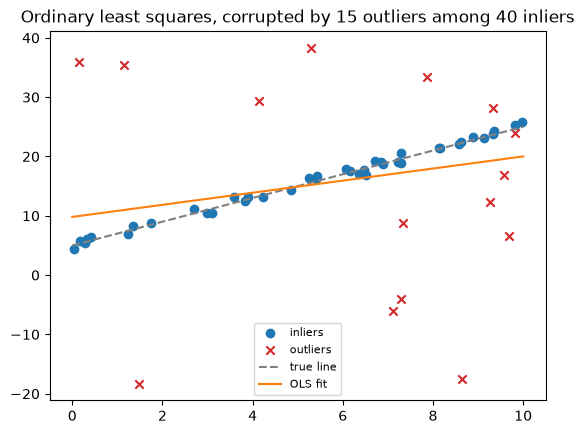}
\caption{Ordinary least squares, corrupted by 15 outliers among 40 inliers.}
\label{fig:l23-2}
\end{figure}

Least-squares fitting minimizes the \emph{sum of squared} residuals. But squaring has a drawback: every outlier (point far from the model) contributes an enormous amount to the total error, causing the whole fit to move away from all the other, correct points. On data with 40 true inliers plus 15 unrelated outliers: the true line $y=2.00x+5.00$ becomes an OLS fit of $y=1.02x+9.81$ --- badly dragged off course, as Figure~\ref{fig:l23-2} shows.

\section{RANSAC: fit from the inside out}

\textbf{RANSAC} (RANdom SAmple Consensus, \citelink{fischlerbolles1981}{Fischler \& Bolles, 1981}) flips the strategy: instead of using \emph{all} the data and hoping outliers don't matter, it repeatedly picks the \emph{smallest possible} random subset needed to define a candidate model, counts how many of the remaining points agree with it (the \textbf{consensus set}), and keeps whichever candidate has the most agreement. A final least-squares refit on the winning inlier set gives the polished result.

\begin{figure}[h]
\centering
\includegraphics[width=0.55\textwidth]{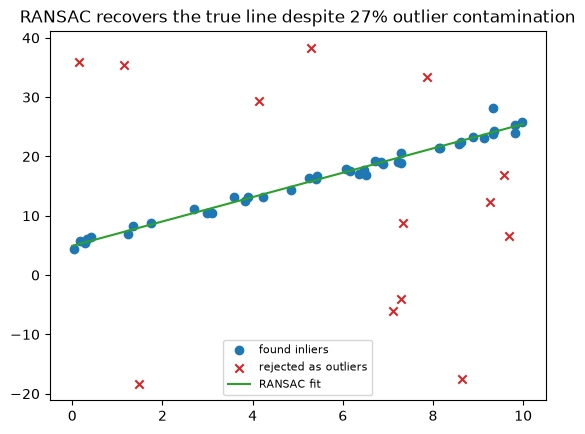}
\caption{RANSAC's found inliers (blue) vs.\ rejected outliers (red), and the resulting fit.}
\label{fig:l23-3}
\end{figure}

On the same contaminated data: RANSAC fits $y=2.05x+4.94$, finding $42$ of $55$ points as inliers (planted: $40$ true inliers) --- essentially recovering the true line despite $27\%$ outlier contamination, as Figure~\ref{fig:l23-3} shows.

\subsection*{How many iterations does RANSAC need?}

If a fraction $w$ of the data are inliers, and the model needs a minimal sample of $n$ points, the probability that any one random sample is entirely inliers is $w^n$. To be at least $p$ confident of drawing an all-inlier sample at least once across $N$ independent tries:
\begin{equation}
N = \frac{\log(1-p)}{\log(1-w^n)}.
\end{equation}

\begin{center}
\renewcommand{\arraystretch}{1.3}
\begin{tabular}{rrrr}
\hline
\textbf{Inlier \%} & \textbf{Line ($n{=}2$)} & \textbf{Homography ($n{=}4$)} & \textbf{Fund.\ matrix ($n{=}8$)} \\
\hline
90\% & 3  & 5   & 9 \\
70\% & 7  & 17  & 78 \\
50\% & 17 & 72  & 1{,}177 \\
30\% & 49 & 567 & 70{,}188 \\
\hline
\end{tabular}
\end{center}

Fitting a line only needs 2 points, so even fairly heavy contamination ($50\%$ outliers) needs just a few dozen iterations. Fitting a homography needs a minimal sample of 4 points, so the same inlier fraction needs an order of magnitude more iterations --- the price of a more complex model. A fundamental matrix (minimal sample 8) is far more sensitive still: at $30\%$ inliers, over $70{,}000$ iterations would be needed for the same confidence.

\section{A gentler alternative: robust loss functions}

RANSAC makes a hard inlier/outlier decision. An alternative family, \textbf{M-estimators}, instead reweights every point's contribution smoothly --- e.g.\ the \textbf{Huber loss} behaves like ordinary squared error for small residuals but switches to linear (much less aggressive) growth beyond a threshold, so a single very-wrong point can no longer dominate the total cost. RANSAC and M-estimators are complementary in practice: RANSAC is excellent at rejecting \emph{gross} outliers, while an M-estimator refinement afterward can down-weight smaller, more subtle deviations among the remaining inliers.

\begin{figure}[h]
\centering
\includegraphics[width=0.55\textwidth]{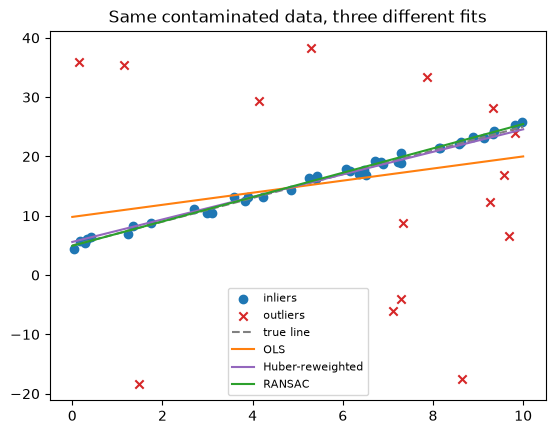}
\caption{Same contaminated data, three different fits: OLS, Huber-reweighted, and RANSAC.}
\label{fig:l23-4}
\end{figure}

On the same data: OLS gives $y=1.02x+9.81$, Huber-reweighting gives $y=1.90x+5.58$, and RANSAC gives $y=2.05x+4.94$ --- Huber lands between the two, much closer to the true line ($y=2.00x+5.00$) than OLS, though not quite as exact as RANSAC on data this heavily contaminated, since a handful of gross outliers still pull a little on every iteration rather than being cut out entirely. Figure~\ref{fig:l23-4} compares all three fits.

\subsection*{Exercises}
\begin{enumerate}
  \item Increase the outlier count until roughly $70\%$ of the points are outliers. Does 200 iterations remain enough to reliably recover the true line? Use the iteration-count formula to check whether 200 is even theoretically sufficient at that contamination level.
  \item Lower the RANSAC distance threshold from $2.0$ to $0.5$. Does the number of found inliers change, and why might too-strict a threshold actually start rejecting \emph{genuine} inliers?
  \item Try the Huber fit with a much smaller \code{delta} (e.g.\ $0.5$) and a much larger one (e.g.\ $5.0$). How does \code{delta} trade off between OLS-like behavior and RANSAC-like hard rejection?
\end{enumerate}

\vfill
\fullcode{lesson23}{lesson23_model_fitting_ransac}

%% file: chapters/lesson24.tex
\chapter{Projective Geometry}
\label{ch:lesson24}

Chapter~\ref{ch:lesson08} built a hierarchy of 2D transforms --- Euclidean, similarity, affine --- and noted that all three preserve parallel lines. This chapter covers the next, most general step: the \textbf{projective transform (homography)}, which models what a camera actually does when it looks at a flat surface from an angle, and which does \emph{not} preserve parallelism. That's not a bug --- it's exactly the phenomenon of a vanishing point, and it's the tool behind perspective correction and image stitching.

\section{Homogeneous coordinates}

Represent a 2D point $(x,y)$ as a 3-vector $(x,y,1)$. Any $3\times3$ matrix $H$ can then act on it by ordinary matrix multiplication; converting back to 2D means dividing by the third coordinate:
\begin{equation}
\begin{bmatrix}x'\\y'\\w'\end{bmatrix} = H\begin{bmatrix}x\\y\\1\end{bmatrix}, \qquad (x'_{\text{2D}}, y'_{\text{2D}}) = \left(\frac{x'}{w'}, \frac{y'}{w'}\right).
\end{equation}
Two big payoffs: first, a $3\times3$ matrix can represent translation too (impossible with a $2\times2$ matrix alone, which is why Chapter~\ref{ch:lesson08} needed a separate translation vector). Second, that division by $w'$ is what lets a homography model genuine \emph{perspective} effects --- not just the affine transforms of Chapter~\ref{ch:lesson08}, but photos where parallel lines converge toward a vanishing point, exactly what a real camera does when it looks at a flat surface from an angle. Note also that $H$ and $cH$ (any nonzero scalar multiple) represent the \emph{exact same} transform, since the division cancels the scale --- a homography has only 8 independent degrees of freedom, not 9.

\section{Parallel lines stop being parallel}

\begin{figure}[h]
\centering
\includegraphics[width=0.85\textwidth]{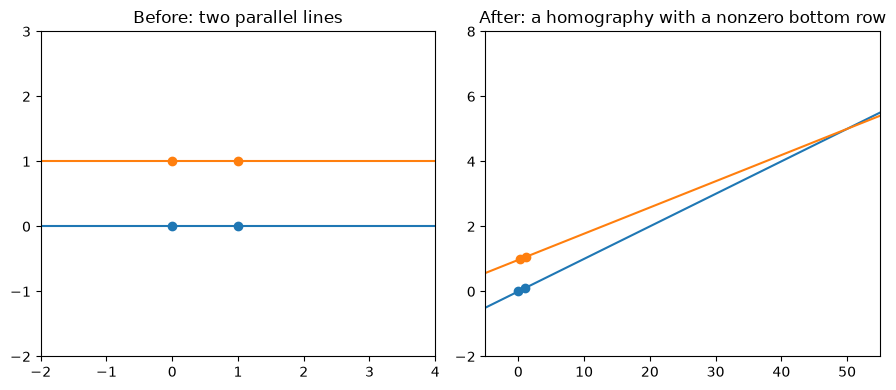}
\caption{Two parallel lines, before and after a homography with a nonzero bottom row --- they now converge toward a vanishing point.}
\label{fig:l24-1}
\end{figure}

Extending Chapter~\ref{ch:lesson08}'s hierarchy: an affine matrix has a fixed bottom row $(0,0,1)$; a projective matrix allows any $(g,h,1)$. That's the only difference, and it alone is enough to destroy parallelism, as Figure~\ref{fig:l24-1} shows.

\subsection*{Vanishing points, computed two ways}

Where do these two lines actually meet? Intersecting the transformed line segments directly gives $(50, 5)$. More elegantly: transforming the \emph{point at infinity} in the original lines' shared direction $(1,0)$, represented in homogeneous coordinates as $(1,0,0)$, gives the exact same answer, $(50, 5)$ --- confirming that a homography applied to a point at infinity lands exactly on the corresponding vanishing point.

Points aren't the only thing homogeneous coordinates represent elegantly --- a line $ax+by+c=0$ is just as naturally the 3-vector $l=(a,b,c)$, with a point $p=(x,y,1)$ lying on it exactly when $l^\top p=0$. Two useful consequences follow from the cross product: the line through two points is $l=p_1\times p_2$, and the intersection of two lines is $p=l_1\times l_2$ --- both give the same vanishing point, $(50,5)$, computed a third way.

\section{When does a homography actually apply?}

A homography exactly relates two photos in exactly two situations: photographing a flat surface from two different positions, or photographing \emph{any} 3D scene, flat or not, from the same camera position while only rotating the camera between shots. That second case is what makes Chapter~\ref{ch:lesson25}'s panorama stitching work on ordinary, non-planar scenes --- a translating camera or a genuinely 3D scene viewed from two different positions has no single homography that relates the two images exactly.

\section{Perspective correction: rectifying a photographed document}

\begin{figure}[h]
\centering
\includegraphics[width=0.85\textwidth]{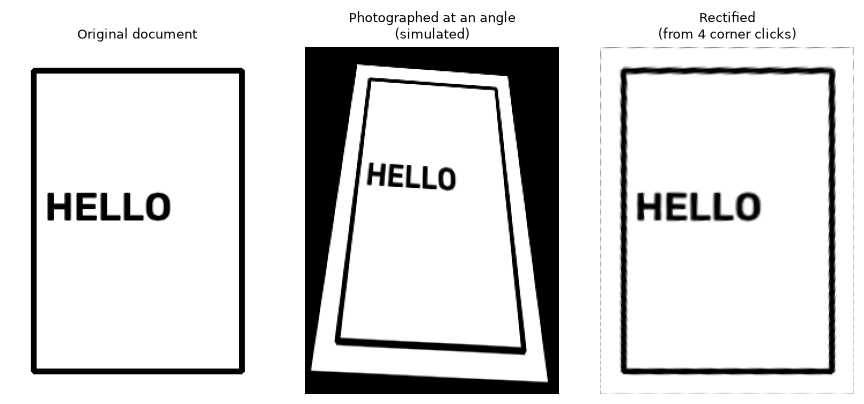}
\caption{A document, a simulated photograph at an angle (strong keystoning), and the rectified result from 4 corner correspondences. Mean absolute pixel difference after the distort-then-rectify round trip is $4.59$ --- small resampling loss only.}
\label{fig:l24-2}
\end{figure}

The most common practical use of a homography: given 4 point correspondences (e.g.\ the 4 corners of a document), \code{cv2.getPerspectiveTransform} solves for the unique homography mapping one set of 4 points to the other, and \code{cv2.warpPerspective} applies it. The trapezoid shape a photographed rectangle takes on --- edges that were parallel in real life visibly converging in the photo --- is called \textbf{keystoning}, and it's exactly the vanishing-point effect above, now applied to a document, as Figure~\ref{fig:l24-2} shows.

\section{Solving for a homography from many correspondences}

4 correspondences exactly determine a homography's 8 degrees of freedom, with no slack for error. With \emph{more} than 4 (typically from automatic feature matching, Chapter~\ref{ch:lesson20}), the problem becomes an overdetermined least-squares fit --- the Direct Linear Transform (DLT) algorithm that \code{cv2.findHomography} implements, optionally wrapped in RANSAC (Chapter~\ref{ch:lesson23}) to reject bad correspondences as outliers.

With 30 clean synthetic correspondences: \code{findHomography} recovers the true homography to 4 decimal places, with all 30 points marked as inliers.

\subsection*{RANSAC in practice: homography estimation with bad matches}

With clean correspondences, passing \code{cv2.RANSAC} to \code{findHomography} makes no visible difference --- there's nothing to reject. Injecting 15 deliberately garbage correspondences alongside the 30 genuine ones shows the real effect: RANSAC finds exactly the 30 genuine correspondences (inliers: $30/45$) and reconstructs the homography to a mean reprojection error of $1.49\times10^{-6}$ pixels on the true inliers. A plain least-squares fit on the same contaminated data reaches a mean reprojection error of $1.06\times10^{4}$ pixels --- off by ten orders of magnitude more, effectively useless for anything requiring pixel-level accuracy, because $15$ bad matches out of $45$ ($33\%$) was more than enough to corrupt an unweighted sum-of-squares fit.

\section{Looking ahead: projective geometry in 3D}

Everything in this chapter has been 2D-to-2D: a homography relates two planes (images or flat surfaces in the world). Later chapters extend the same projective machinery one dimension further, deriving the camera's projection matrix that maps a 3D scene onto a 2D image in the first place.

\subsection*{Exercises}
\begin{enumerate}
  \item Add zero-mean Gaussian noise (e.g.\ std 2 pixels) to the second point set before calling \code{cv2.findHomography}. How much does the estimated homography drift from the true one, and does increasing the number of correspondences (say, from 30 to 200) reduce that drift?
  \item Increase the number of injected bad matches from 15 until RANSAC starts to fail. Using the iteration-count formula from Chapter~\ref{ch:lesson23} (minimal sample size 4 for a homography), roughly what outlier fraction is that, and how many iterations would it theoretically require to stay $99\%$ confident of success?
  \item In the vanishing-point demo, change the two lines to be \emph{vertical} instead of horizontal (e.g.\ $x=0$ and $x=1$), and predict, then verify, where their vanishing point ends up using the point-at-infinity trick with direction $(0,1,0)$.
\end{enumerate}

\vfill
\fullcode{lesson24}{lesson24_projective_geometry}

%% file: chapters/lesson25.tex
\chapter{Image Stitching and Mosaicking}
\label{ch:lesson25}

This chapter brings together feature detection and matching (Chapter~\ref{ch:lesson20}), robust fitting (Chapter~\ref{ch:lesson23}), and homography estimation (Chapter~\ref{ch:lesson24}) into a complete pipeline that stitches two overlapping photos into a single seamless panorama.

\section{The pipeline, at a glance}

\begin{enumerate}
  \item \textbf{Detect and match features} between the two photos (Chapter~\ref{ch:lesson20}: SIFT $+$ ratio test).
  \item \textbf{Estimate a homography} relating one image's plane to the other's, robustly (Chapter~\ref{ch:lesson24}'s homography, Chapter~\ref{ch:lesson23}'s RANSAC).
  \item \textbf{Warp} one image into the other's coordinate frame (Chapter~\ref{ch:lesson09}: \code{cv2.warpPerspective}).
  \item \textbf{Composite and blend} the two images onto one canvas, feathering across the overlap so the seam is invisible.
\end{enumerate}

\section{An example}

\subsection*{Two overlapping photos}

\begin{figure}[h]
\centering
\includegraphics[width=0.9\textwidth]{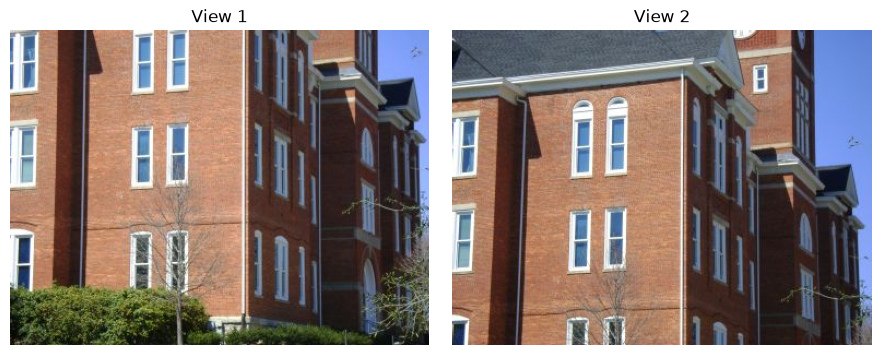}
\caption{The two input views. \attrib{Photo: Stan Birchfield.}}
\label{fig:l25-1}
\end{figure}

Two photos of the same brick building (Figure~\ref{fig:l25-1}), taken from slightly different positions with substantial overlap.

\subsection*{Step 1: feature matching}

\begin{figure}[h]
\centering
\includegraphics[width=0.9\textwidth]{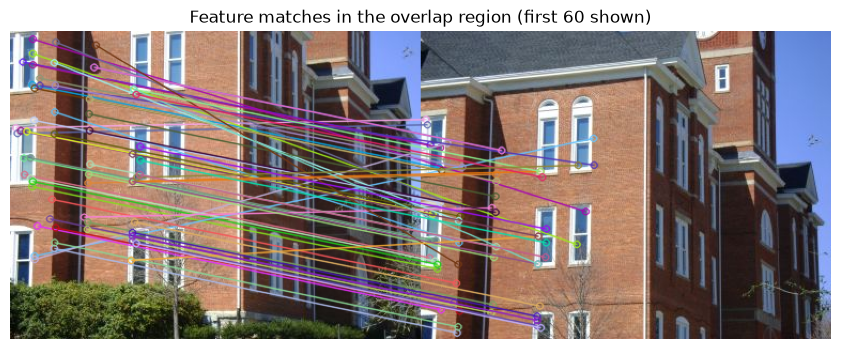}
\caption{Feature matches in the overlap region (first 60 shown).}
\label{fig:l25-2}
\end{figure}

SIFT finds $938$ keypoints in view 1 and $708$ in view 2; Lowe's ratio test keeps $315$ good matches, shown in Figure~\ref{fig:l25-2}.

\subsection*{Step 2: robust homography}

Solve for the homography that maps points in view 2 into view 1's coordinate frame, so warping view 2 through it lands it in the right place on a shared canvas: \code{cv2.findHomography} with RANSAC finds $263$ of $315$ matches as inliers.

\subsection*{Steps 3-4: warp, composite, and blend}

\begin{figure}[h]
\centering
\includegraphics[width=0.6\textwidth]{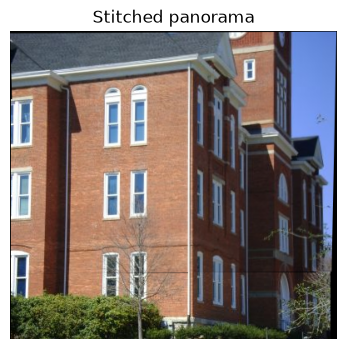}
\caption{The stitched panorama. The roof (top, fully visible only in view 2) and the bush (bottom, fully visible only in view 1) both appear in the mosaic, confirming that the composite genuinely draws from both source images.}
\label{fig:l25-3}
\end{figure}

First, figure out how big the output canvas needs to be, by mapping both images' corners through $H$ (Chapter~\ref{ch:lesson24}) and taking the bounding box --- unlike a synthetic case, these photos aren't simple axis-aligned crops of a shared frame, so the canvas size and the offset of view 1 within it both have to be computed rather than assumed. Warp view 2 onto that canvas, then blend the overlap region with a simple \textbf{feather}: a linear alpha ramp from ``fully view 1'' to ``fully view 2'' across the overlap, the same weighted-sum blending as \code{cv2.addWeighted} in Chapter~\ref{ch:lesson02}, just with a spatially-varying weight instead of a constant one. Figure~\ref{fig:l25-3} shows the resulting panorama.

\subsection*{How good is the fit?}

\begin{figure}[h]
\centering
\includegraphics[width=0.55\textwidth]{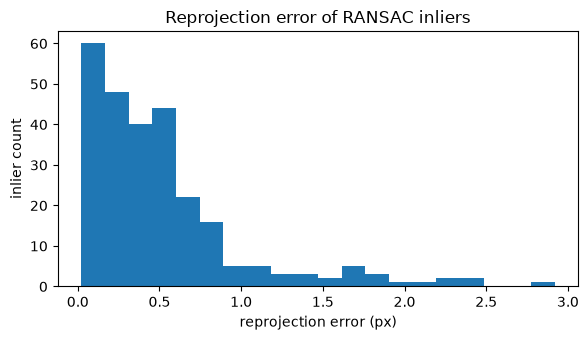}
\caption{Reprojection error of the RANSAC inliers, histogrammed.}
\label{fig:l25-4}
\end{figure}

There's no synthetic ground truth to compare against with a real photograph, but we don't need one: the RANSAC inliers themselves give a direct measure of geometric fit. For each inlier correspondence, project the point from view 2 through $H$ and measure the distance to its matched point in view 1 --- the \textbf{reprojection error}. Here: mean reprojection error $0.520$ px, max $2.919$ px --- sub-pixel-scale error, well under the RANSAC inlier threshold of 3 px, confirming $H$ is a good geometric fit, consistent with the clean seam visible in the stitched panorama. Figure~\ref{fig:l25-4} histograms the reprojection error across all inliers.

\section{In practice: \code{cv2.Stitcher}}

\begin{figure}[h]
\centering
\includegraphics[width=0.5\textwidth]{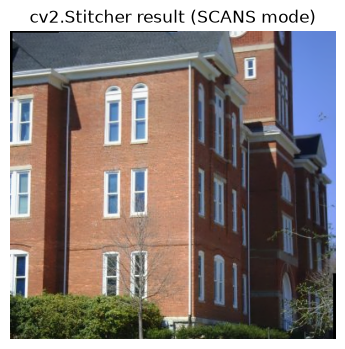}
\caption{\code{cv2.Stitcher} in \code{SCANS} mode: status \code{OK}.}
\label{fig:l25-5}
\end{figure}

OpenCV bundles this entire pipeline (plus more robust blending, exposure compensation, and support for many images at once) behind a single high-level call. Its default \code{PANORAMA} mode assumes the images come from a camera rotating about its optical center, and warps each image onto a sphere before compositing --- great for wide panoramas, but it bows straight lines when the scene is dominated by flat, rectilinear structure (like a building facade) and the camera translated rather than purely rotated. \code{SCANS} mode instead composites with the planar homographies directly, matching the approach used above, as confirmed by the successful \code{Stitcher} run in Figure~\ref{fig:l25-5}.

\subsection*{Exercises}
\begin{enumerate}
  \item Reduce the overlap between the two views (e.g.\ crop view 2 down to its rightmost quarter before matching). At what point does SIFT matching find too few good matches for \code{findHomography} to produce a reliable result?
  \item Replace the linear feather with a hard cutoff (no blending: just pick whichever image covers each pixel, splitting the overlap down the middle) and compare the visible seam quality to the feathered version.
  \item $H$ here is a mild general homography, not a pure translation, because the two photos were taken from slightly different positions rather than a pure sideways pan. Print the ratio of $H$'s last row to $[0,0,1]$ as a rough measure of how much perspective distortion it captures, and compare it to what you'd get by forcing an affine fit (\code{cv2.estimateAffinePartial2D}) instead --- how well does the affine approximation stitch the images compared to the full homography?
\end{enumerate}

\vfill
\fullcode{lesson25}{lesson25_stitching_mosaicking}

%% file: chapters/lesson26.tex
\chapter{Image Formation}
\label{ch:lesson26}

Up to now we have been processing images without thinking about where they come from. This chapter unpacks that mystery by considering: \textbf{geometry} (how the projection matrix is derived from a pinhole camera model), \textbf{optics} (what a real lens does that a pinhole camera fails to model --- focus and blur), and \textbf{sensing/color} (how a real sensor captures only one color per pixel and needs to reconstruct the rest, along with the two classic pitfalls --- gamma and white balance --- that trip up naive image processing).

\section{The pinhole projection matrix, derived}

\begin{figure}[h]
\centering
\includegraphics[width=0.4\textwidth]{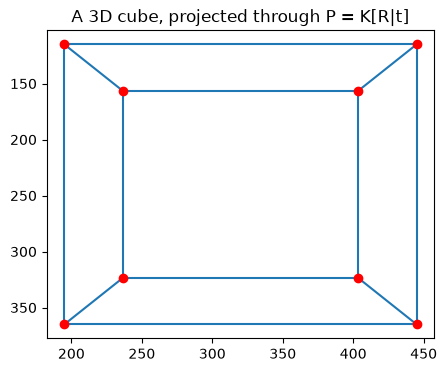}
\caption{A 3D cube, projected through $P=K[R|t]$ onto the image plane.}
\label{fig:l26-1}
\end{figure}

A pinhole camera only lets through rays of light passing through a particular point (the \emph{center of projection}). By similar triangles, a 3D point $(X,Y,Z)$ (camera-centered coordinates) lands on the image plane at
\begin{equation}
x = f\frac{X}{Z}, \qquad y = f\frac{Y}{Z},
\end{equation}
where $f$ is the distance from the pinhole to the image plane. Converting to pixel units requires the \textbf{intrinsics} $K$; incorporating the camera's position and orientation relative to the world requires the \textbf{extrinsics} $[R|t]$. Put together, in homogeneous coordinates:
\begin{equation}
\underbrace{\begin{bmatrix}u\\v\\1\end{bmatrix}}_{\text{pixel}} \;\propto\; \underbrace{\begin{bmatrix}f_x&0&c_x\\0&f_y&c_y\\0&0&1\end{bmatrix}}_{K}\underbrace{\begin{bmatrix}R & t\end{bmatrix}}_{\text{extrinsics}}\begin{bmatrix}X\\Y\\Z\\1\end{bmatrix}_{\text{world}}.
\end{equation}
This is exactly the $P=K[R|t]$ that Chapters~\ref{ch:lesson27}--\ref{ch:lesson29} (Camera Calibration, Epipolar Geometry, and Structure from Motion) will use. Every symbol traces back to a physical cause: $f_x,f_y$ to focal length and pixel size; $(c_x,c_y)$ to where the optical axis hits the sensor (the \emph{principal point}); $R,t$ to the camera's pose. Figure~\ref{fig:l26-1} shows a cube projected through $P$.

\section{Real lenses: focus and defocus blur}

\begin{figure}[h]
\centering
\includegraphics[width=0.85\textwidth]{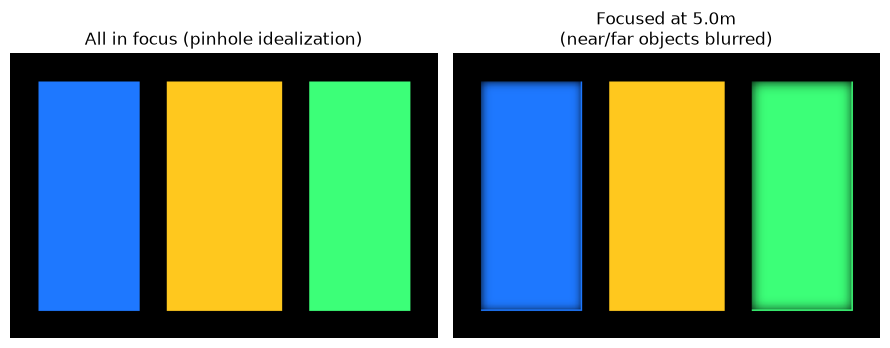}
\caption{All in focus (pinhole idealization) vs.\ focused at $5$m --- near and far objects blur, the mid-depth object stays sharp.}
\label{fig:l26-2}
\end{figure}

A pinhole camera lets in so little light that it needs absurdly long exposures. Real cameras use a lens to gather much more light, but lenses focus sharply only for objects at a particular distance (the \textbf{focal plane}). Points nearer or farther than that spread their light over a small disk on the sensor (the \textbf{circle of confusion}) instead of at a point --- exactly the blurring convolution kernels from Chapters~\ref{ch:lesson09}--\ref{ch:lesson10}, just applied with a kernel size that depends on depth. Focusing at $5$m: a near object (depth $2$m) gets kernel size $19$, the object at the focal plane gets kernel size $1$ (unblurred), and a far object (depth $9$m) gets kernel size $25$ (Figure~\ref{fig:l26-2}).

This explains why a wider aperture gathers more light (at the expense of a shallower depth of field), while a narrow aperture produces a sharp image (but lets in less light) --- the classic photographic aperture/depth-of-field tradeoff, and the reason portrait photos often blur the background while keeping the subject sharp.

\section{The Bayer color filter array: one color per pixel}

\begin{figure}[h]
\centering
\includegraphics[width=0.95\textwidth]{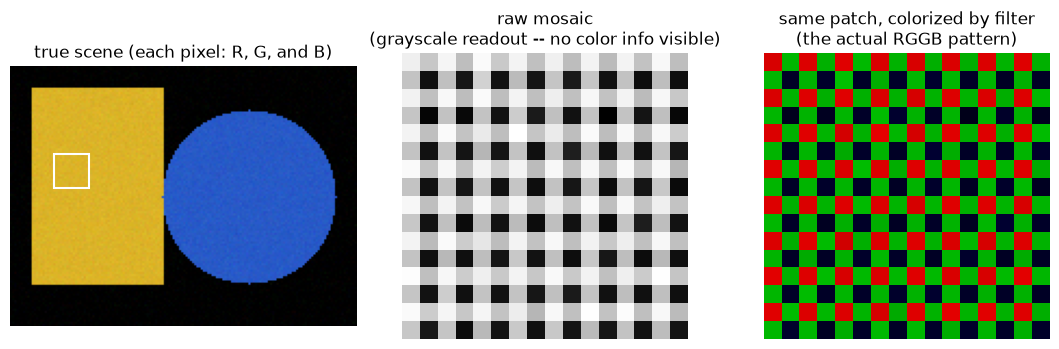}
\caption{A true-color scene, a $16\times16$ crop of the raw Bayer mosaic (grayscale readout, no color visible), and the same crop colorized by filter, revealing the RGGB pattern.}
\label{fig:l26-3}
\end{figure}

To keep costs down, most camera sensors don't measure red, green, and blue separately at every pixel. Instead, a single monochrome photosensor array sits under a mosaic of tiny color filters, so each individual pixel physically records only \emph{one} of the three colors. The most common arrangement (RGGB) --- the \textbf{Bayer filter} (\citelink{bayer1976}{Bayer, 1976}) --- repeats in a $2\times2$ tile pattern: one red, one blue, and \emph{two} green filters, since the human eye is most sensitive to green. Every pixel in the raw sensor output is therefore missing two-thirds of its color information by construction, as Figure~\ref{fig:l26-3} shows.

\subsection*{Demosaicking: reconstructing RGB from the mosaic}

\begin{figure}[h]
\centering
\includegraphics[width=0.9\textwidth]{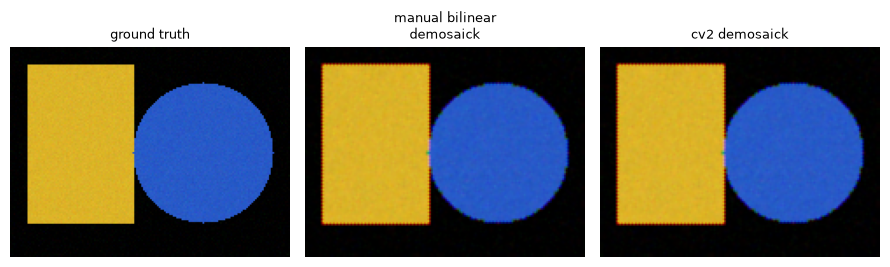}
\caption{Ground truth, manual bilinear demosaick, and \code{cv2} demosaick --- nearly indistinguishable.}
\label{fig:l26-4}
\end{figure}

\textbf{Demosaicking} fills in each pixel's two missing colors by interpolating from same-color neighbors --- the simplest version is bilinear: average the nearby known samples of a color, weighted by distance, exactly Chapter~\ref{ch:lesson09}'s bilinear interpolation, just interpolating across a sparse, patterned grid of known samples. A from-scratch bilinear demosaicker reaches a mean absolute error of $3.37$ against ground truth; \code{cv2.cvtColor(..., cv2.COLOR\_BayerBG2BGR)} reaches $3.43$ --- both land within a fraction of a pixel value of each other, the residual mostly just sensor noise already present in the mosaic (Figure~\ref{fig:l26-4}).

\subsection*{The pitfall: getting the mosaic pattern wrong}

\begin{figure}[h]
\centering
\includegraphics[width=0.6\textwidth]{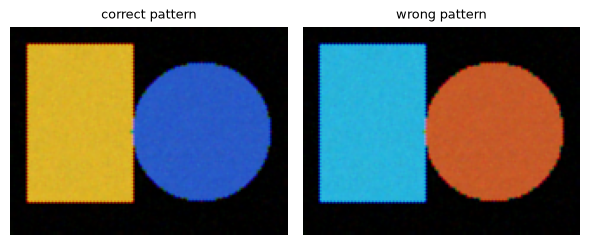}
\caption{Correct vs.\ wrong Bayer pattern assumption --- a strongly color-shifted image.}
\label{fig:l26-5}
\end{figure}

The R/G/B arrangement matters, and there's more than one convention (RGGB, BGGR, GRBG, GBRG, depending on the sensor). Demosaicking with the \emph{wrong} assumed pattern doesn't fail gracefully --- it silently produces a plausible-looking but badly wrong-colored image, since every pixel gets a value, just the wrong one: the correct pattern gives MAE $3.43$, the wrong pattern jumps to MAE $64.69$ --- nearly a $19\times$ increase, from every red sample being treated as blue and vice versa, as Figure~\ref{fig:l26-5} shows.

\section{Gamma correction: pixel values are not linear light}

Cameras and displays don't store brightness proportionally to physical light intensity (\textbf{linear radiance}). Instead, values are \textbf{gamma-encoded}: \code{encoded = linear ** (1/gamma)} (typically gamma $\approx 2.2$, close to the sRGB standard), which allocates more encoded levels to darker tones --- matching human vision's greater sensitivity to shadows than highlights. Decoding reverses it: \code{linear = encoded ** gamma}.

\subsection*{The pitfall: averaging encoded pixels is not averaging light}

\begin{figure}[h]
\centering
\includegraphics[width=0.55\textwidth]{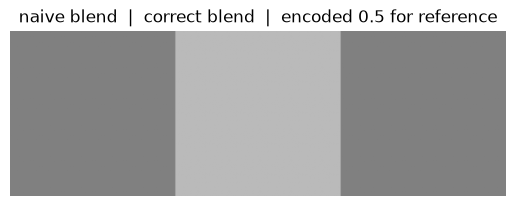}
\caption{Naive blend of encoded black/white (left) vs.\ the physically correct blend (middle) vs.\ encoded $0.5$ for reference (right) --- the naive blend is visibly, dramatically darker.}
\label{fig:l26-6}
\end{figure}

Blending, resizing, and blurring (Chapters~\ref{ch:lesson08}--\ref{ch:lesson10}) all implicitly assume a linear quantity is being averaged. Pixel values usually aren't linear --- they're gamma-encoded --- so naively averaging two encoded pixel values gives the \emph{wrong physical brightness}. Averaging encoded black and white directly gives $0.500$; the physically correct 50\%-linear-light gray, properly decoded/averaged/re-encoded, is $0.730$ --- the naive result represents only $21.8\%$ of the true linear brightness, not $50\%$, visibly so in Figure~\ref{fig:l26-6}.

This is a real and common bug when image-processing code (blurring, mipmap generation, alpha blending) operates directly on gamma-encoded pixel values instead of linearizing first.

\section{White balance: color depends on the light source}

A camera records the \emph{product} of a surface's reflectance and the illuminant's color, not the surface's ``true'' color alone --- a white sheet of paper looks orange under warm tungsten light and blue-ish under overcast sky, even though our visual system usually compensates for this automatically (color constancy). Cameras have to do the same correction deliberately: \textbf{white balancing}.

\subsection*{The gray-world assumption}

\begin{figure}[h]
\centering
\includegraphics[width=0.9\textwidth]{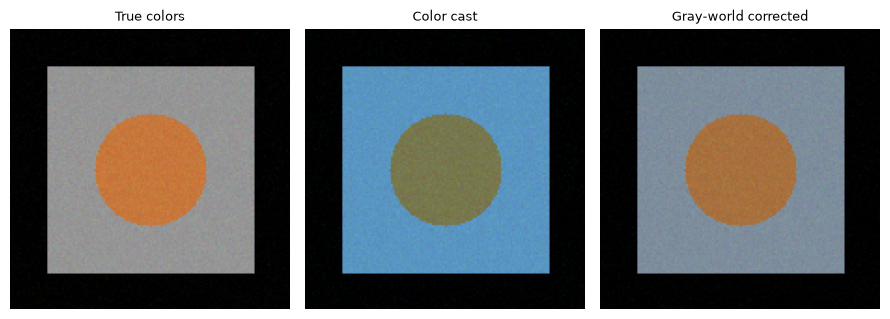}
\caption{True colors, under a warm tungsten cast, and gray-world corrected. Mean absolute pixel error drops from $18.98$ to $7.43$ after correction --- substantial, though not to zero, since this small scene (dominated by a colored circle) isn't perfectly neutral on average.}
\label{fig:l26-7}
\end{figure}

A simple, classic white-balancing algorithm assumes that, averaged over an entire real-world scene, colors roughly cancel out to neutral gray. Under that assumption, any systematic difference between a photo's per-channel averages reveals the illuminant's color cast, so rescaling each channel to equalize the averages should approximately undo it, as Figure~\ref{fig:l26-7} demonstrates.

That residual error is exactly why real cameras often combine gray-world with other cues (an actual detected gray/white reference patch, learned scene statistics, or user-specified presets) rather than relying on the assumption alone.

\subsection*{Exercises}
\begin{enumerate}
  \item Change the defocus \code{strength} parameter to make the depth-of-field effect much shallower (larger \code{strength}) or much deeper (smaller \code{strength}). At \code{strength=0}, what should happen, and does the code produce that?
  \item Build and decode a BGGR mosaic instead of RGGB (swap which corner is red vs.\ blue). Find the matching OpenCV code by trying each of \code{cv2.COLOR\_BayerBG2BGR}, \code{cv2.COLOR\_BayerGB2BGR}, \code{cv2.COLOR\_BayerGR2BGR}, \code{cv2.COLOR\_BayerRG2BGR} and checking which one drops the MAE back down to the noise floor --- OpenCV's naming convention for these codes is notoriously easy to get backwards, itself a small demonstration of this chapter's point about silent pattern-mismatch errors.
  \item Repeat the gamma-blending experiment, but blend $0.2$ and $0.8$ (instead of pure black and white) in both the naive and correct ways. Is the discrepancy between them larger or smaller than the black/white case, and why might that make sense given the shape of the gamma curve?
  \item Modify the white-balance demo so the scene is dominated by a large \emph{saturated red} region instead of a neutral gray card (i.e.\ make gray-world's core assumption clearly false). Does gray-world white balancing still improve the result, make no difference, or actively make it worse?
\end{enumerate}

\vfill
\fullcode{lesson26}{lesson26_image_formation}

%% file: chapters/lesson27.tex
\chapter{Camera Calibration}
\label{ch:lesson27}

Chapter~\ref{ch:lesson26} introduced the camera intrinsics matrix $K$ to model a pinhole camera. The classic approach (\citelink{zhang2000}{Zhang, 2000}) to \textbf{camera calibration} recovers $K$ by showing the camera a known flat pattern (a checkerboard) from several different angles: extracting a homography (Chapter~\ref{ch:lesson24}) between the checkerboard plane and each image, then combining constraints from several such homographies to estimate $K$.

\section{What calibration recovers}

Camera calibration recovers two separate things:
\begin{enumerate}
  \item \textbf{Intrinsics} $K=\begin{bmatrix}f_x&s&c_x\\0&f_y&c_y\\0&0&1\end{bmatrix}$: focal lengths, principal point, and skew --- the ideal pinhole part of the camera model used throughout the next two chapters. Skew can safely be assumed zero for real cameras.
  \item \textbf{Distortion coefficients} $(k_1,k_2,p_1,p_2,k_3,\ldots)$: real lenses aren't perfect pinholes. Radial distortion ($k_1,k_2,k_3$) bows straight lines into curves (barrel or pincushion, depending on sign); tangential distortion ($p_1,p_2$) accounts for the lens not being perfectly parallel to the sensor.
\end{enumerate}
Every equation in the next two chapters secretly assumes a pinhole camera model --- i.e.\ that distortion has already been removed, which requires calibration.

\section{A synthetic calibration rig}

Create synthetic images of a camera viewing a checkerboard: define a ground-truth $K$ and distortion, define a flat checkerboard's 3D corner positions, and generate several ``photos'' of it from different poses by projecting the corners with \code{cv2.projectPoints} (which applies distortion exactly as a real lens would). This gives the same kind of input \code{cv2.calibrateCamera} expects from real detected checkerboard corners, while also providing ground truth to check against. Of 40 attempted random poses, $11$ land fully inside the frame and are usable.

\subsection*{Detecting the target}

\begin{figure}[h]
\centering
\includegraphics[width=0.75\textwidth]{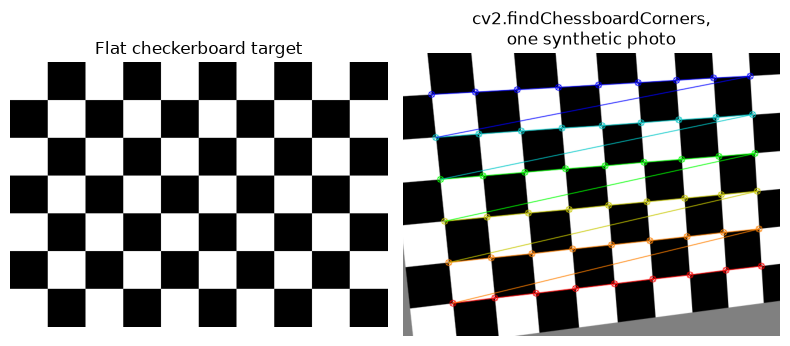}
\caption{The flat checkerboard target, and \code{cv2.findChessboardCorners}' detections on one synthetic photo.}
\label{fig:l27-1}
\end{figure}

A real calibration pipeline knows the checkerboard's own flat 2D layout, but not the board's pose in each shot, and not where its corners land in each photo --- so the first step is finding those 2D positions with \code{cv2.findChessboardCorners}, exactly as on a real photograph. Corners are detected in $8$ of the $11$ synthetic photos, with a mean per-view corner error of $1.83$ px (max $2.37$ px) against the ground truth --- realistic detector noise, not a bug. Figure~\ref{fig:l27-1} shows the target and one view's detections.

\section{Running calibration}

Zhang's calibration algorithm (\code{cv2.calibrateCamera}) takes the known corner positions on the checkerboard and their \emph{detected} 2D image positions across all views, and jointly solves for $K$, distortion, and each view's own pose. With only 8 noisy views: the RMS reprojection error comes to $0.264$ px, and $K$ is recovered reasonably close ($f_x=822.6$, $f_y=826.3$ vs.\ true $800, 800$) --- but the distortion coefficients drift much further from the true values ($k_1$ estimated at $0.0054$ vs.\ true $-0.3$). Using noisy corner detections, calibration does not recover the ground truth parameters exactly; this is why in practice you'd use as many views, from as wide a variety of angles, as practical --- more (and more diverse) constraints average out detector noise and better constrain the estimation, especially for distortion.

\section{Straight lines, bent by a lens}

\begin{figure}[h]
\centering
\includegraphics[width=0.9\textwidth]{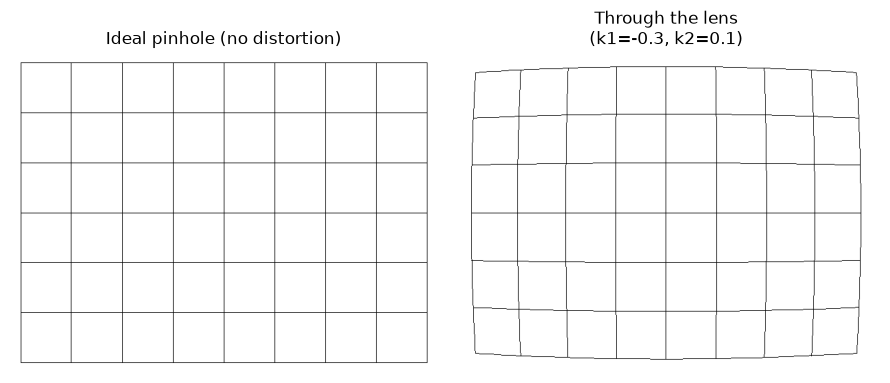}
\caption{An ideal pinhole grid (straight lines) vs.\ the same grid through a lens with $k_1=-0.3$, $k_2=0.1$ --- visibly bowed.}
\label{fig:l27-2}
\end{figure}

Lens distortion causes straight lines to become curved. In normalized (pre-$K$) camera coordinates:
\begin{align}
x_d &= x(1+k_1r^2+k_2r^4+k_3r^6) + 2p_1xy + p_2(r^2+2x^2), \\
y_d &= y(1+k_1r^2+k_2r^4+k_3r^6) + p_1(r^2+2y^2) + 2p_2xy,
\end{align}
where $r^2=x^2+y^2$ --- exactly the model \code{cv2.projectPoints} applied internally above. Figure~\ref{fig:l27-2} shows the visible effect on a grid of straight lines.

\section{Undoing distortion with the estimated parameters}

\begin{figure}[h]
\centering
\includegraphics[width=0.9\textwidth]{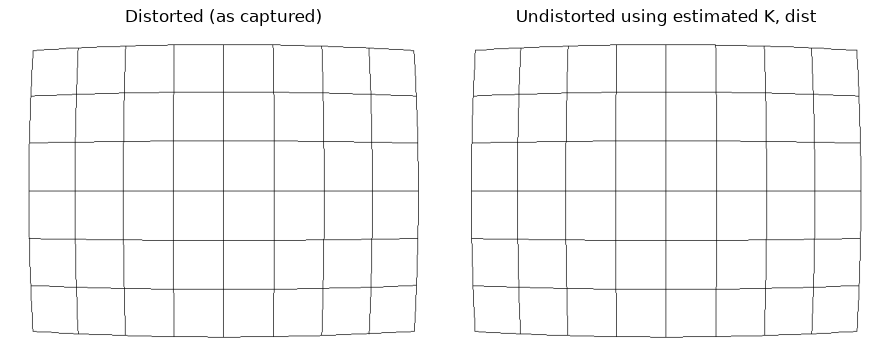}
\caption{Distorted grid (as captured) vs.\ undistorted using the estimated $K$ and distortion --- the lines straighten out, though a faint bow remains, since the estimated parameters are themselves only approximately correct.}
\label{fig:l27-3}
\end{figure}

\code{cv2.undistortPoints} inverts the distortion model, converting distorted pixel coordinates back to normalized undistorted coordinates. Applying it to the bent grid lines, using the \emph{estimated} (not the true) calibration, straightens them back out, with a max recovery error of $3.4\times10^{-2}$ in normalized coordinates (Figure~\ref{fig:l27-3}).

This process of calibrating a real camera should be performed, if possible, before handing real photos to any of the geometric machinery from the next two chapters, all of which implicitly assumes an ideal, distortion-free pinhole model. Keep in mind, however, that many modern pipelines perform auto-calibration without checkerboard images, using simplified camera models (typically just one focal length parameter --- ignoring aspect ratio, non-central principal point, and lens distortion).

\subsection*{Exercises}
\begin{enumerate}
  \item Add extra synthetic pixel noise (e.g.\ std $0.5$) on top of the real detector noise, to each view's detected corners before calibrating. How much does the RMS reprojection error grow, and how does the estimated $K$ drift further from the true one?
  \item This setup ends with only a handful of successfully-detected views. Try reducing that down to just 3 views before calibrating. Does the estimate noticeably degrade, especially for the distortion coefficients (which need corners spread widely across the frame, including near the edges, to be well constrained)?
  \item Set the true distortion to all zeros (a perfect pinhole) and rerun. Confirm the ``distorted'' and ``ideal'' grids become identical, and that calibration still recovers the true $K$ reasonably well even with zero distortion to estimate.
\end{enumerate}

\vfill
\fullcode{lesson27}{lesson27_camera_calibration}

%% file: chapters/lesson28.tex
\chapter{Epipolar Geometry}
\label{ch:lesson28}

Chapter~\ref{ch:lesson22}'s stereo matching assumed a \emph{rectified} pair, where corresponding points always fall on the same row. In the more general case, stereo cameras may not be rectified. This chapter covers the general two-view relationship between \textbf{any} pair of images of a static scene: a point's match in the other image isn't just \emph{somewhere} --- it's constrained to lie on a particular line. We cover both the calibrated and uncalibrated cases.

\section{The epipolar constraint}

For a point $x_1$ in image 1 and its true match $x_2$ in image 2 (both in homogeneous pixel coordinates), there's a $3\times3$ rank-deficient matrix $F$ --- the \textbf{fundamental matrix} --- such that
\begin{equation}
x_2^\top F x_1 = 0
\end{equation}
for \emph{every} corresponding pair, regardless of scene geometry. $F$ depends only on the two cameras' relative pose and (uncalibrated) intrinsics. Rearranged, $l_2=Fx_1$ is the \textbf{epipolar line} in image 2: the 1D line along which $x_1$'s match is guaranteed to lie. This is exactly the mechanism that made Chapter~\ref{ch:lesson22}'s row-restricted search valid --- rectification is just the special camera arrangement where every epipolar line happens to be horizontal, which collapses the 1D search to a single row.

\subsection*{Why a line? A geometric picture}

\begin{figure}[h]
\centering
\includegraphics[width=0.55\textwidth]{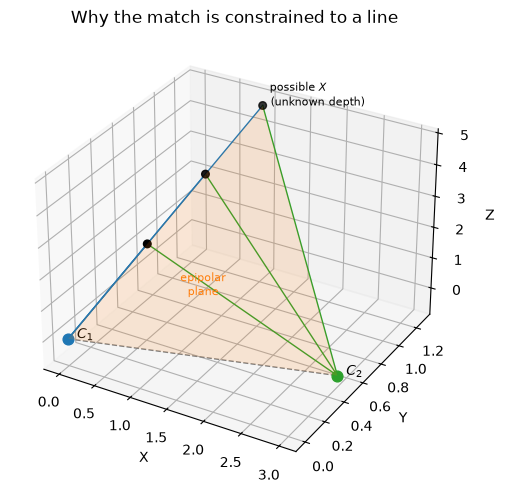}
\caption{Three candidate depths for the same pixel $x_1$, all lying in one epipolar plane; their projections into camera 2 all fall on the same epipolar line.}
\label{fig:l28-1}
\end{figure}

Camera 1 knows only the \emph{direction} to a point, not its depth: the true 3D point $X$ could be anywhere along the ray from $C_1$ through $x_1$. But every point on that ray, together with $C_1$ and $C_2$, lies in a single plane --- the \textbf{epipolar plane}. Camera 2 sees this plane edge-on, as a single line: wherever $X$ actually sits along the ray, its projection into image 2 always falls on that one line, the epipolar line, as Figure~\ref{fig:l28-1} illustrates.

\section{A synthetic two-camera scene}

\begin{figure}[h]
\centering
\includegraphics[width=0.9\textwidth]{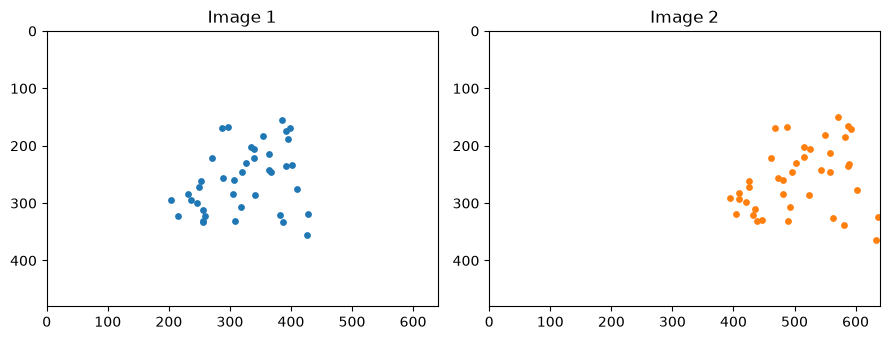}
\caption{40 3D points projected into both cameras.}
\label{fig:l28-2}
\end{figure}

Build two cameras with known intrinsics $K$ and a known relative rotation/translation, create random 3D points, project them into both cameras (Figure~\ref{fig:l28-2}), and use the resulting correspondences to recover $F$.

\section{SVD: singular value decomposition}

In Chapter~\ref{ch:lesson23}, we saw that a real symmetric positive-semidefinite matrix $A$ can be written as $A=P\Lambda P^\top$, with $P$ the eigenvectors and $\Lambda$ the (non-negative, diagonal) eigenvalues. The \textbf{singular value decomposition (SVD)} generalizes this: any matrix $A$ --- symmetric or not, square or not --- can be written as $A=U\Sigma V^\top$, with $U,V$ orthogonal (the \textbf{singular vectors}) and $\Sigma$ diagonal and non-negative (the \textbf{singular values}, sorted largest to smallest). Two facts matter here:

\textbf{Solving $Ax=0$ as closely as possible, subject to $\|x\|=1$.} The answer is $V$'s last column --- the right singular vector for the smallest singular value. This is Chapter~\ref{ch:lesson23}'s total-least-squares trick, generalized: $V$'s columns are the eigenvectors of $A^\top A$, sorted by eigenvalue.

\textbf{Enforcing a rank constraint.} Zeroing the smallest singular value(s) and reconstructing gives the closest lower-rank matrix (in a least-squares sense) to the original --- exactly what's needed to force the estimated $F$, which must be exactly rank 2, back onto that constraint after an unconstrained solve drifts off it slightly.

\section{The 8-point algorithm}

Each correspondence gives one linear equation in the 9 unknown entries of $F$ (expanding $x_2^\top F x_1=0$). $F$ is only defined up to scale (like the homography in Chapter~\ref{ch:lesson24}), so it really has just 8 independent unknowns, solvable as a homogeneous least-squares problem via SVD --- the same DLT machinery used for the homography fit. Since $F$ must be exactly rank 2, project the 9-parameter solution back onto the nearest rank-2 matrix using the SVD trick above.

$F$ actually has only 7 true degrees of freedom, since the rank-2 constraint removes one more; a 7-point algorithm exists but is nonlinear, which is why the linear 8-point algorithm (with its separate rank-2 cleanup step) remains popular. Raw pixel coordinates (in the hundreds) also make the linear system badly conditioned, so points are first rescaled to be centered at the origin with average distance $\sqrt2$, solved in normalized coordinates, then the rescaling is undone on the resulting $F$ --- the \textbf{normalized 8-point algorithm} (\citelink{hartley1997}{Hartley, 1997}), the version used here and worth reaching for in practice.

\begin{lstlisting}
_, _, Vt = np.linalg.svd(A)
F = Vt[-1].reshape(3, 3)
U, S, Vt2 = np.linalg.svd(F)   # enforce rank-2 by zeroing the smallest singular value
S[-1] = 0
F = U @ np.diag(S) @ Vt2
\end{lstlisting}

\begin{figure}[h]
\centering
\includegraphics[width=0.95\textwidth]{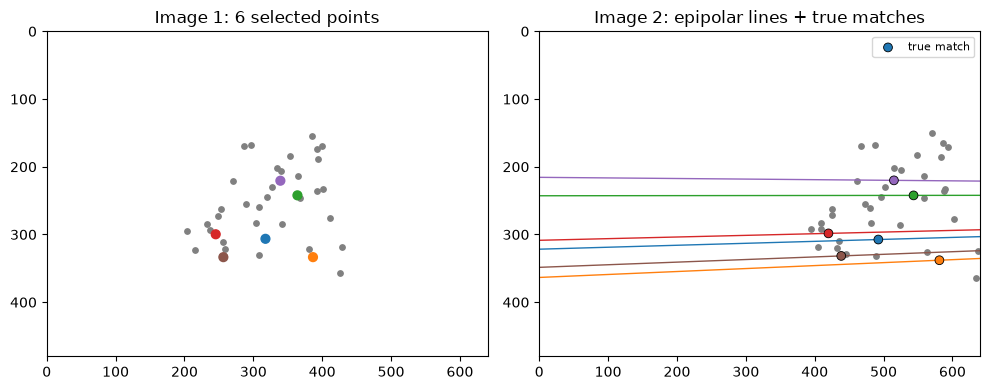}
\caption{6 selected points in image 1, and their predicted epipolar lines $l_2=Fx_1$ in image 2 --- every true match lands exactly on its predicted line, confirming a point's search in the second image really does collapse from 2D to 1D, even without rectifying the images first.}
\label{fig:l28-3}
\end{figure}

A from-scratch implementation matches \code{cv2.findFundamentalMat(x1, x2, cv2.FM\_8POINT)} to 5 decimal places, and both achieve a mean $|x_2^\top F x_1|$ residual on the order of $10^{-7}$--$10^{-15}$ --- effectively zero. Figure~\ref{fig:l28-3} plots several points' predicted epipolar lines directly, using this recovered $F$.

\section{From fundamental to essential: adding calibration}

The fundamental matrix works in raw pixel coordinates, assuming the camera intrinsics are unknown. If the intrinsics $K$ are known (from calibration, Chapter~\ref{ch:lesson27}), the intrinsics can be removed, giving the \textbf{essential matrix}:
\begin{equation}
E = K_2^\top F K_1 \qquad \text{(same camera: } K_1=K_2=K\text{)}.
\end{equation}
Unlike $F$'s 7 degrees of freedom, $E$ has only 5 --- it's built entirely from a relative rotation $R$ and translation direction $t$: $E=[t]_\times R$, where $[t]_\times$ is the skew-symmetric cross-product matrix of $t$. This means $E$ can be \emph{decomposed} back into $R$ and $t$, which is how camera motion is recovered from image correspondences alone. Estimating $E$ robustly with \code{cv2.findEssentialMat(..., method=cv2.RANSAC, ...)} --- the same outlier-rejecting fit built from scratch in Chapter~\ref{ch:lesson23} --- agrees with $E$ derived directly from $F$ and $K$ to 4 decimal places.

\subsection*{Recovering camera motion}

\code{cv2.recoverPose} decomposes $E$ into a rotation and a translation \emph{direction} --- not a full translation vector, since the magnitude is fundamentally unrecoverable from two views alone (a scene twice as large, viewed by cameras with twice the baseline, produces identical images --- the same scale ambiguity familiar from monocular vision). On the synthetic scene: the recovered rotation matches the true $15^\circ$ rotation to 3 decimal places, and the recovered translation direction matches the true direction up to the expected sign convention.

From feature correspondences alone (Chapter~\ref{ch:lesson20}), we've recovered the relative rotation and translation (up to scale) between the two cameras --- the starting point for structure from motion (Chapter~\ref{ch:lesson29}).

\subsection*{Exercises}
\begin{enumerate}
  \item Add pixel noise (e.g.\ std $0.5$) to the correspondences before running the 8-point algorithm. How much does the mean epipolar residual grow, and does normalizing coordinates actually matter here --- try skipping the normalization step and compare.
  \item The epipole in image 2 is the projection of camera 1's center, and satisfies $Fe_1=0$ for the epipole $e_1$ in image 1 (and $F^\top e_2=0$ for the epipole $e_2$ in image 2). Compute both epipoles as the null space of $F$ (via SVD) and check whether they fall inside or outside the visible image region for this camera configuration.
  \item Increase the rotation angle between the two cameras to $60^\circ$ and rerun pose recovery. Does \code{cv2.recoverPose} still find the correct rotation? At what point would you expect correspondence matching itself (Chapter~\ref{ch:lesson20}) to become the bottleneck rather than the geometry?
\end{enumerate}

\vfill
\fullcode{lesson28}{lesson28_epipolar_geometry}

%% file: chapters/lesson29.tex
\chapter{Structure from Motion}
\label{ch:lesson29}

This chapter assembles feature matching (Chapter~\ref{ch:lesson20}), robust estimation (Chapter~\ref{ch:lesson23}), calibration (Chapter~\ref{ch:lesson27}), and the essential matrix and pose recovery (Chapter~\ref{ch:lesson28}) into a complete pipeline that takes 2D image correspondences from two views and recovers \textbf{both} the cameras' relative motion \textbf{and} the 3D positions of the points that were being viewed --- \textbf{S}tructure \textbf{f}rom \textbf{M}otion. The one ingredient not yet developed in detail is \textbf{triangulation}: turning a matched 2D point pair, plus known camera poses, into a 3D point. (Chapter~\ref{ch:lesson22} showed how to convert disparity to depth for rectified images; this chapter generalizes the idea.)

\section{The classic pipeline, at a glance}

\begin{figure}[h]
\centering
\includegraphics[width=0.9\textwidth]{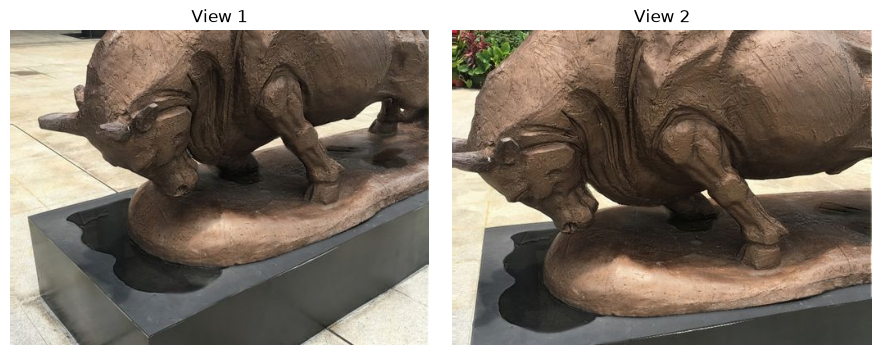}
\caption{The two input views. \attrib{Image source: \href{https://github.com/YoYo000/BlendedMVS}{BlendedMVS} (CC BY 4.0).}}
\label{fig:l29-1}
\end{figure}

\begin{enumerate}
  \item \textbf{Match features} between two images (Chapter~\ref{ch:lesson20}: SIFT $+$ ratio test).
  \item \textbf{Estimate the essential matrix} $E$ from those correspondences, robustly (Chapter~\ref{ch:lesson23}: RANSAC), using known intrinsics $K$ (Chapter~\ref{ch:lesson27}: calibration).
  \item \textbf{Recover relative pose} $(R,t)$ from $E$ (Chapter~\ref{ch:lesson28}: \code{cv2.recoverPose}) --- up to an unknown scale on $t$.
  \item \textbf{Triangulate}: for every matched pair, intersect the two corresponding rays in 3D to recover a 3D point.
\end{enumerate}
Two realistic synthetic photos of a bronze sculpture from the BlendedMVS dataset (\citelink{yao2020}{Yao et al., 2020}) are used --- unlike a photo pair shot casually, these images come with ground-truth camera poses, calibration, and geometry for checking the entire pipeline. Figure~\ref{fig:l29-1} shows the pair.

\section{An example}

\subsection*{Step 1: feature matching}

\begin{figure}[h]
\centering
\includegraphics[width=0.9\textwidth]{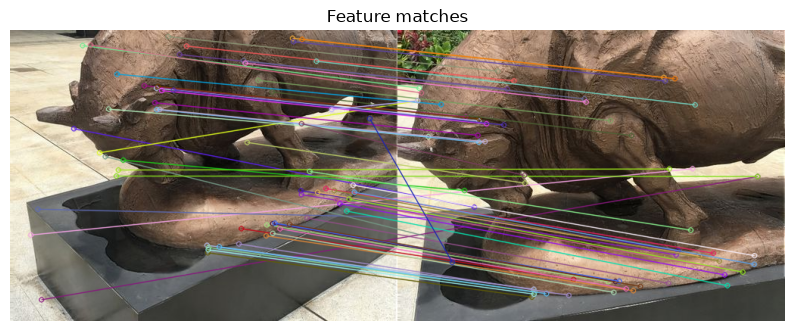}
\caption{Feature matches between the two views.}
\label{fig:l29-2}
\end{figure}

SIFT finds $504$ keypoints in view 1 and $654$ in view 2; Lowe's ratio test keeps $64$ good matches, shown in Figure~\ref{fig:l29-2}.

\subsection*{Step 2: robust essential matrix and pose recovery}

Feed the matched points through the same robust pipeline built in Chapters~\ref{ch:lesson23} and \ref{ch:lesson27} to recover the essential matrix and the cameras' relative pose: $48$ of $64$ matches pass as RANSAC inliers. The recovered rotation matches the true rotation closely (e.g.\ $0.9245$ vs.\ true $0.9208$ for the leading diagonal entry), and the recovered translation direction ($-0.9967, 0.0791, 0.0174$) is close to the true direction ($-0.999, 0.0125, 0.0427$). As before, \code{recoverPose} returns only a \textbf{unit-length} translation direction --- the baseline distance is fundamentally unrecoverable from image correspondences alone.

\subsection*{Step 3: triangulation}

Given two camera projection matrices $P_1,P_2$ (each $3\times4$) and a matched pixel pair $(x_1,x_2)$, find the 3D point $X$ satisfying both $x_1\propto P_1X$ and $x_2\propto P_2X$. Each view contributes 2 independent linear equations in $X$'s 4 homogeneous unknowns, by the same DLT/SVD recipe used for homographies (Chapter~\ref{ch:lesson24}) and the fundamental matrix (Chapter~\ref{ch:lesson28}):
\begin{equation}
A = \begin{bmatrix} u_1 P_1^{(3)} - P_1^{(1)} \\ v_1 P_1^{(3)} - P_1^{(2)} \\ u_2 P_2^{(3)} - P_2^{(1)} \\ v_2 P_2^{(3)} - P_2^{(2)} \end{bmatrix}, \qquad AX = 0,
\end{equation}
where $P^{(i)}$ denotes row $i$ of $P$; the null space of $A$ gives $X$. A from-scratch DLT triangulation matches \code{cv2.triangulatePoints} to $10^{-7}$.

\subsection*{Cheirality: a free sanity check}

Decomposing $E$ into $(R,t)$ actually yields four mathematically valid sign combinations; only one places the observed points in front of \emph{both} cameras, and \code{cv2.recoverPose} uses exactly this \textbf{cheirality constraint} to pick the right one internally. The epipolar constraint alone doesn't rule out a correspondence that triangulates \emph{behind} a camera, so a mismatched point can still pass RANSAC's inlier test; requiring positive depth in both views is a free extra filter --- it drops the reconstruction from $48$ to $47$ inliers here. Triangulating those 47 points with the \emph{true} camera poses gives a mean reprojection error of just $0.169$ px (max $1.680$ px).

\section{Putting it together: reconstruction using estimated pose}

The real test: triangulate using the camera matrix built from \code{recoverPose}'s \emph{estimated} $(R,t)$, not the dataset's true pose. Because $t$ was only recovered up to scale, so is the reconstruction --- every 3D point comes out a fixed factor smaller than reality. Multiplying by the true baseline length ($0.788$, a bit of cheating, just for evaluation) recovers the correct metric scene: mean 3D reconstruction error (vs.\ triangulation from the true poses) is just $0.058$, max $0.096$ --- a small fraction of the baseline.

With real (imperfect) correspondences, the entire pipeline --- essential matrix, pose, triangulation --- reconstructs the scene to within a small fraction of the camera baseline, using nothing but 2D pixel correspondences and known intrinsics, plus \emph{one} external number (the true baseline) to fix the scale ambiguity. In practice, that scale reference might come from a known object size in the scene, a second sensor (GPS, IMU, LiDAR), or a calibrated stereo rig (Chapter~\ref{ch:lesson22}) instead of two arbitrary independent cameras.

\subsection*{Visualizing the reconstruction}

\begin{figure}[h]
\centering
\includegraphics[width=0.6\textwidth]{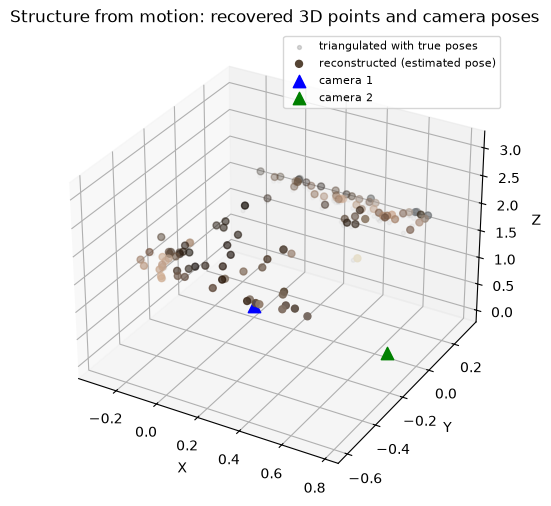}
\caption{The recovered 3D points (colored, estimated pose) alongside points triangulated with the true poses (gray), plus both camera centers.}
\label{fig:l29-3}
\end{figure}

Loosening the feature-matching thresholds pulls in more (slightly noisier) correspondences for a denser cloud: $112$ points, colored by their pixel color from view 1, shown in Figure~\ref{fig:l29-3}.

\subsection*{From sparse to dense, and from two views to many}

Even with loosened thresholds, feature matching only yields a \emph{sparse} cloud --- one 3D point per distinctive keypoint. With camera poses known, \textbf{multi-view stereo (MVS)} estimates a depth for \emph{every} pixel in \emph{every} view (a generalization of Chapter~\ref{ch:lesson22}'s stereo matching to arbitrarily-posed calibrated cameras), and those per-view depth maps are fused into the dense colored point cloud or mesh that toolkits like COLMAP (\citelink{schonberger2016colmap}{Sch\"onberger \& Frahm, 2016}) produce.

Real reconstructions rarely stop at two views. The standard recipe sequentially incorporates new images: match features against existing images, compute the new camera's pose relative to the existing 3D point cloud (Perspective-n-Point, \code{cv2.solvePnP}), then update the cloud by triangulation. As views accumulate, errors compound --- pose corrupts triangulated points, which corrupt the next recovered pose, and so on. \textbf{Bundle adjustment} fixes this by refining all camera poses and 3D points \emph{together} in one large nonlinear least-squares optimization that minimizes total reprojection error across every point in every view at once --- arguably the single most important step separating a simple two-view pipeline like this chapter's from a real SfM system.

\subsection*{Exercises}
\begin{enumerate}
  \item Add extra synthetic pixel noise on top of the already-real detections before triangulating. How much does the reconstruction error grow, and does it grow uniformly, or worse for points farther from the cameras (Chapter~\ref{ch:lesson22}'s disparity-depth relationship: distant points produce smaller, noisier parallax)?
  \item Sketch, in words, how you'd extend this notebook to a third view using \code{cv2.solvePnP}: which 3D points from the two-view reconstruction would you match into the new image, and how would \code{solvePnP}'s inputs and outputs map onto the pieces already in hand ($K$, matched 2D points, the existing 3D point cloud)?
  \item This pair has a deliberately generous baseline for reliable matching. How would you expect a much shorter baseline to affect the number of RANSAC inliers, and the accuracy of the recovered pose and reconstruction, compared to the wide-baseline pair used above?
\end{enumerate}

\vfill
\fullcode{lesson29}{lesson29_structure_from_motion}

%% file: chapters/lesson30.tex
\chapter{Projection --- The Central Idea}
\label{ch:lesson30}

This begins Part 3: deep learning for computer vision, which focuses on the modern approaches that use neural networks. Before writing a single neural network, though, it's worth pausing to reflect upon an idea that rarely gets much attention despite its fundamental importance. \textbf{Projection} is a simple idea: map points from one space into another, usually a simpler one, by a linear combination of their coordinates. Projection quietly ran through much of Parts I and II --- many of the techniques so far have secretly been projections. Projection also underlies everything from here on: neural networks turn out to be nothing more than \emph{learned}, \emph{composed} projections.

\section{Two projections you've already built}

\begin{figure}[h]
\centering
\includegraphics[width=0.75\textwidth]{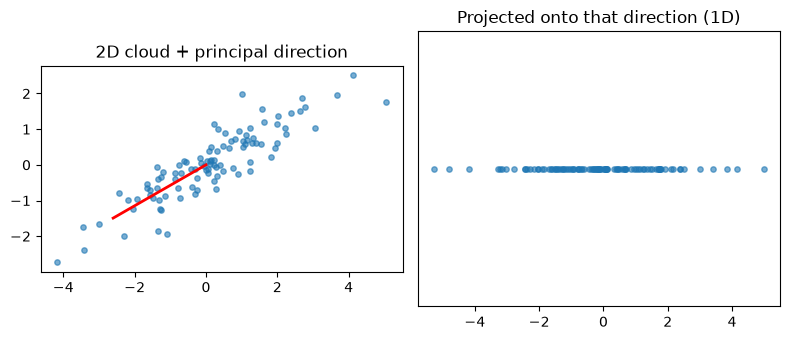}
\caption{A 2D cloud with its principal direction, and the same points projected onto that direction (1D).}
\label{fig:l30-1}
\end{figure}

\textbf{PCA} (Chapter~\ref{ch:lesson06}). Given a cloud of points, the eigenvectors of their covariance matrix gave directions to project onto. Projecting onto the top eigenvector, $y=v^\top x$, collapses each 2D point to a single number --- the coordinate along the direction of greatest spread, as Figure~\ref{fig:l30-1} shows.

\textbf{Camera projection} (Chapters~\ref{ch:lesson24}--\ref{ch:lesson25}). $P=K[R|t]$ maps a 3D world point to a 2D pixel. Every row of that matrix, before the final homogeneous divide, is itself a linear projection: a dot product between the point's coordinates and a fixed direction, plus an offset.

Both examples share the same mechanics: pick a direction $w$ (and maybe an offset $b$), then compute $y=w^\top x+b$. What differs is \emph{why} the direction was chosen: PCA picks $w$ to maximize the spread of the projected data (an unsupervised, purely geometric objective); a camera's rows are fixed by its physical geometry. Neither is chosen to solve a classification or recognition problem --- but nothing stops us from choosing $w$ for exactly that purpose.

\section{Matrix multiplication: many projections at once}

Everything above computed one number $y=w^\top x$ from one direction $w$. Stack $m$ different directions as the rows of a matrix $W$ (shape $m\times n$), and $y=Wx$ computes all $m$ projections at once --- one dot product per row. This is exactly what matrix multiplication \emph{is}: a bank of simultaneous projections. Several seemingly different linear operations from earlier in this course are secretly the same operation, just with a different, structured choice of $W$:
\begin{itemize}
  \item \textbf{Convolution} (Chapter~\ref{ch:lesson10}): each output value is a dot product between a small neighborhood and a kernel --- a projection. Sliding the kernel across the whole signal is a matrix multiply by a sparse matrix whose rows are shifted copies of the kernel (a Toeplitz matrix).
  \item \textbf{The Fourier transform} (Chapter~\ref{ch:lesson15}): each output value is a dot product between the \emph{whole} signal and one sinusoid at a fixed frequency --- a projection onto that frequency. The full transform is a matrix multiply by a fixed matrix of sines and cosines.
  \item \textbf{A homography or camera projection matrix} (Chapters~\ref{ch:lesson24}--\ref{ch:lesson25}): each output coordinate is a dot product between a homogeneous point and one row of $H$ or $P$.
\end{itemize}
Verified directly: a hand-built Toeplitz matrix reproduces \code{np.convolve} to within $5.6\times10^{-17}$, and a hand-built matrix of sinusoids reproduces \code{np.fft.fft} to within $2.9\times10^{-15}$ --- both effectively exact, floating-point roundoff only.

A fully-connected neural network layer is nothing more than this same $y=Wx+b$, with $W$'s rows \emph{learned} to solve a task instead of fixed by hand-derived structure (a kernel shape, a bank of sinusoids, a camera's geometry).

\section{A projection for classification}

Suppose instead of ``maximize spread,'' the goal is ``separate two classes.'' The same formula $y=w^\top x+b$ still applies --- project each point onto a direction $w$, and classify by the \emph{sign} of the resulting score. This is, in its entirety, a single artificial neuron with no activation function: the linear core that every neural network layer is built from.

\begin{figure}[h]
\centering
\includegraphics[width=0.9\textwidth]{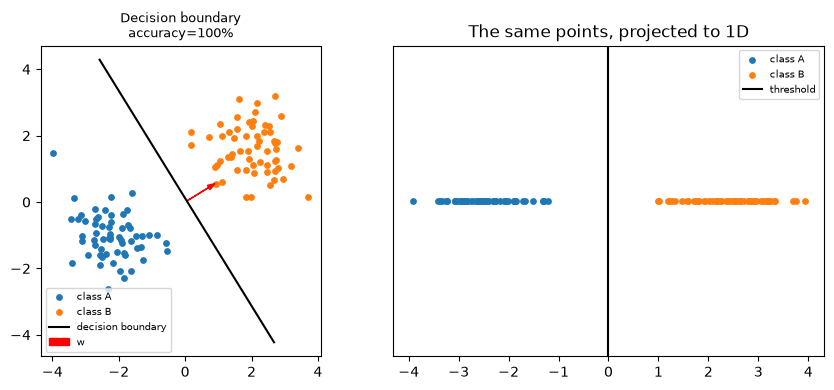}
\caption{Left: the decision boundary in the original 2D space, perpendicular to $w$. Right: the same points projected down onto $w$, with classification reduced to thresholding a single number at zero. The 2D picture is more intuitive, but the 1D picture is what actually generalizes --- in 100 dimensions there's no picture to draw of the ``boundary,'' but ``project to a single number, then threshold'' still works exactly the same way.}
\label{fig:l30-2}
\end{figure}

A principled hand-picked direction --- pointing from one class's mean toward the other's, with the threshold at the midpoint --- reaches $100.0\%$ classification accuracy on two well-separated synthetic classes (Figure~\ref{fig:l30-2}).

\section{What a single projection cannot do}

\begin{figure}[h]
\centering
\includegraphics[width=0.45\textwidth]{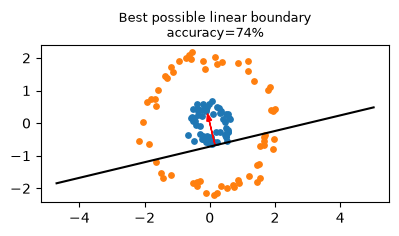}
\caption{The best possible linear boundary between a disk and a surrounding ring --- no rotation or shift of a single straight line can separate them, so the best any linear projection can manage is a mediocre compromise.}
\label{fig:l30-3}
\end{figure}

A single linear projection can only produce a \emph{straight-line} decision boundary (a hyperplane, in higher dimensions). Some datasets have no straight-line separator at all --- the classic example is one class surrounding the other. Searching 2000 random directions and 61 thresholds each for the best possible linear separator between a ring and a surrounded disk finds a best achievable accuracy of just $74.2\%$, as Figure~\ref{fig:l30-3} shows.

This is the wall every purely linear method runs into, and it's exactly what motivates the next two chapters: first, \emph{learning} $w$ and $b$ automatically instead of hand-picking them, and then, more importantly, \emph{composing several projections with nonlinearities in between}, which, given enough capacity, can separate regions of essentially any shape.

\subsection*{Exercises}
\begin{enumerate}
  \item In the PCA recap, project the cloud onto the \emph{second} (smaller) eigenvector instead of the principal one. How does the spread of the resulting 1D projection compare to projecting onto the principal direction, and why does that match what the eigenvalue itself tells you (Chapter~\ref{ch:lesson06})?
  \item Modify the two classes' means and spread so they overlap more. At what point does the mean-difference direction $w$ stop achieving high accuracy, and can you find a \emph{better} $w$ than the mean-difference one by hand for that harder case?
  \item For the ring-and-disk dataset, instead of a linear projection, try classifying by a simple nonlinear function of the points: the distance from the origin, $r=\sqrt{x_1^2+x_2^2}$, thresholded at some value. What accuracy does this one nonlinear feature achieve, and what does that suggest about \emph{what kind} of projection would actually solve this problem?
\end{enumerate}

\vfill
\fullcode{lesson30}{lesson30_projection}

%% file: chapters/lesson31.tex
\chapter{Neural Network Fundamentals}
\label{ch:lesson31}

Chapter~\ref{ch:lesson30} hand-picked a projection direction $w$ (the difference of class means) to separate two classes. That worked, but it required a human to notice a good heuristic. This chapter replaces the human with \textbf{gradient descent}: an automatic procedure that \emph{learns} $w$ and $b$ directly from data by repeatedly nudging them to reduce a loss. The recipe --- forward pass, loss, backward pass, update --- is the entire training loop behind every neural network in this course, no matter how large.

\section{A new library: PyTorch and tensors}

From here on, most chapters lean on \textbf{PyTorch} rather than OpenCV, because training neural networks needs two things OpenCV was never designed for: automatic differentiation (computing gradients of arbitrarily complex functions) and GPU acceleration for large matrix multiplications. A \textbf{tensor} is simply the general term for a grid of numbers of any dimensionality --- every array used so far in this course has really been a tensor all along; PyTorch's \code{torch.Tensor} is NumPy's \code{ndarray} with two extras: it can track a \code{.grad} for automatic differentiation, and it can live on a GPU.

Two PyTorch conventions worth flagging up front: it stores color images \textbf{channel-first}, as \code{(channel, height, width)} rather than NumPy/OpenCV's \code{(height, width, channel)} from Chapter~\ref{ch:lesson01}; and its layers almost always expect a \textbf{batch dimension} first, even for a single example.

\section{A binary classification problem}

\begin{figure}[h]
\centering
\includegraphics[width=0.6\textwidth]{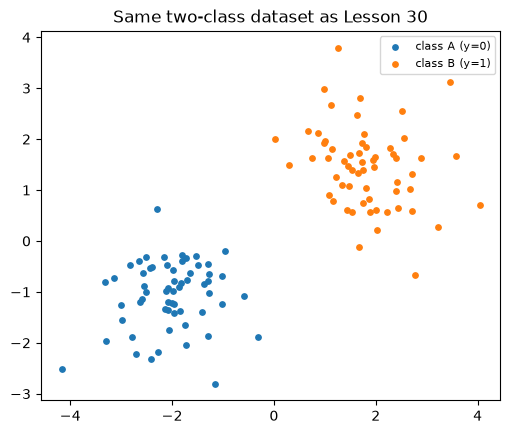}
\caption{The same two-class dataset from Chapter~\ref{ch:lesson30}.}
\label{fig:l31-1}
\end{figure}

We reuse the same two-class dataset from Chapter~\ref{ch:lesson30} (Figure~\ref{fig:l31-1}), so the learned direction below can be compared directly against the hand-picked one.

\section{From a raw score to a probability}

Recall the projection from Chapter~\ref{ch:lesson30}: a single neuron computes a raw score $z=w^\top x+b$, then classifies by that score's sign. The goal of training is to find a $w$ and $b$ that make $z$ come out positive for class B and negative for class A --- but instead of hand-picking them, gradient descent should find them automatically.

\begin{figure}[h]
\centering
\includegraphics[width=0.55\textwidth]{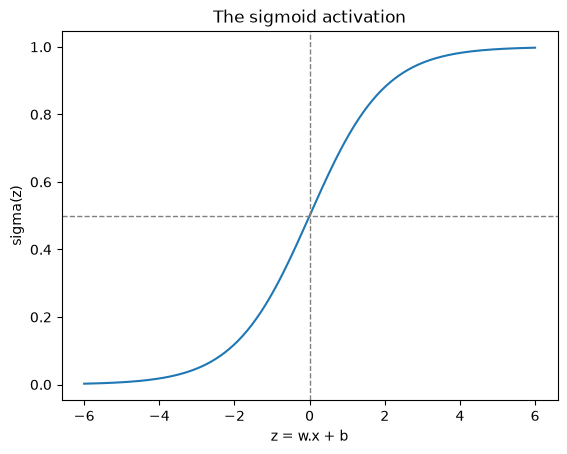}
\caption{The sigmoid activation.}
\label{fig:l31-2}
\end{figure}

The trouble is that classification accuracy (right or wrong) is flat almost everywhere and jumps discontinuously at the decision boundary --- it has no useful gradient to follow. The fix: squash the raw score through the \textbf{sigmoid} function, which maps any real number to a smooth value between 0 and 1, read as ``how confident is the model that this point is class B?''
\begin{equation}
\sigma(z) = \frac{1}{1+e^{-z}}.
\end{equation}
A large positive $z$ maps close to 1, a large negative $z$ close to 0, and $z=0$ maps to exactly 0.5, right on the fence (Figure~\ref{fig:l31-2}).

\section{Binary cross-entropy: turning probability into a trainable loss}

\begin{figure}[h]
\centering
\includegraphics[width=0.6\textwidth]{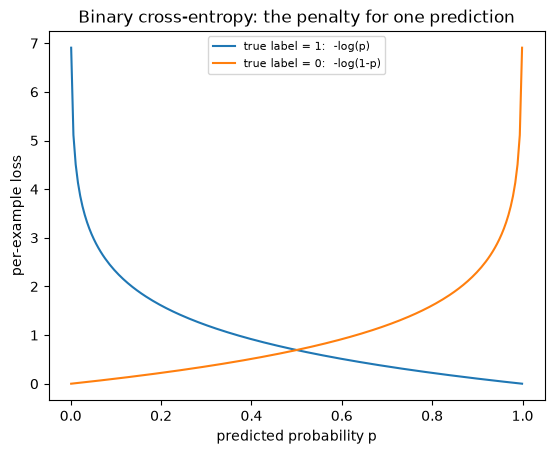}
\caption{Binary cross-entropy: the penalty for one prediction, as a function of the predicted probability.}
\label{fig:l31-3}
\end{figure}

A probability alone isn't yet something to optimize. Training needs a single number --- \textbf{loss} --- that's small when the model's probabilities are good and large when they're bad. \textbf{Binary cross-entropy} penalizes each prediction according to how confident it was, and whether that confidence was justified: if the true label is 1, the penalty is $-\log(p)$ (near zero when $p$ is close to 1, exploding as $p\to0$); if the true label is 0, the mirror image, $-\log(1-p)$ (Figure~\ref{fig:l31-3}).

Averaging that per-example penalty over the whole dataset gives the full loss:
\begin{equation}
L = -\frac{1}{n}\sum_i\big[y_i\log\sigma(z_i) + (1-y_i)\log(1-\sigma(z_i))\big].
\end{equation}
This is exactly the two curves above, selected by $y_i$ (only one term is ever nonzero, since $y_i$ is 0 or 1) and averaged over every point. Because both the sigmoid and the logarithm are smooth, $L$ has a well-defined gradient everywhere --- the loss decreases smoothly as predictions get closer to being correct, instead of only changing at the decision boundary.

\section{Backpropagation: the chain rule, applied}

``Backprop'' is just the chain rule from Chapters~\ref{ch:lesson12}--\ref{ch:lesson13}, applied to a composed function: $L$ depends on $\sigma(z)$, which depends on $z=w^\top x+b$, which depends on $w$ and $b$. Working backward through that chain (skipping the algebra) gives a remarkably clean result:
\begin{equation}
\frac{\partial L}{\partial z_i} = \sigma(z_i) - y_i, \qquad \frac{\partial L}{\partial w} = \frac{1}{n}X^\top(\sigma(z)-y), \qquad \frac{\partial L}{\partial b} = \frac{1}{n}\sum_i(\sigma(z_i)-y_i).
\end{equation}
In words: the gradient is just the \emph{prediction error}, averaged (weighted by $x$ for $w$'s gradient). Wildly wrong predictions push the weights hard; correct ones barely move them at all.

\subsection*{Sanity checks: numerical gradient and PyTorch autograd}

Before trusting any hand-derived gradient, it's standard practice to check it two independent ways. First, numerically: perturb each parameter by a tiny amount and see how much the loss actually changes, compared to the analytical formula --- the finite-difference and analytic gradients agree to a max discrepancy of $1.01\times10^{-9}$. Second, via PyTorch's automatic differentiation (\code{.backward()}, using its built-in \code{binary\_cross\_entropy\_with\_logits}) --- agreeing with the hand-derived gradient to $5.55\times10^{-17}$. This kind of double-checking is standard practice whenever a gradient is derived by hand.

\section{Training: the full loop}

Initialize $w,b$ randomly (not with a clever heuristic this time), then repeat: forward pass, compute loss, backward pass, take a small step \emph{against} the gradient (gradient \emph{descent}, since the gradient points toward increasing loss).

\begin{lstlisting}
def train(X, y, n_epochs=500, lr=0.5, seed=0):
    w = rng.normal(size=X.shape[1]) * 0.1
    b = 0.0
    for _ in range(n_epochs):
        p, _ = forward(X, w, b)
        grad_w, grad_b = backward(X, p, y)
        w -= lr * grad_w
        b -= lr * grad_b
    return w, b
\end{lstlisting}

\begin{figure}[h]
\centering
\includegraphics[width=0.5\textwidth]{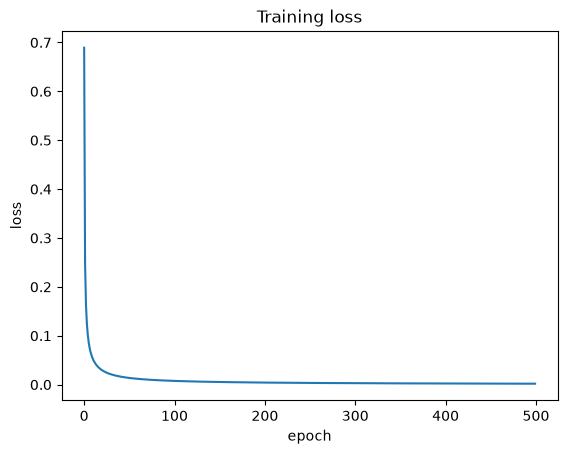}
\caption{Training loss over 500 epochs of gradient descent.}
\label{fig:l31-4}
\end{figure}

Training converges to $w=[2.954, 2.089]$, $b=-0.115$, with final accuracy $100.0\%$, shown in Figure~\ref{fig:l31-4}.

\subsection*{Compare to Chapter~\ref{ch:lesson30}'s hand-picked direction}

Normalizing both directions to unit length: the hand-picked direction is $[0.838, 0.546]$, the learned direction is $[0.816, 0.577]$ --- an angle of just $2.2^\circ$ apart. Gradient descent, starting from nothing but random noise, rediscovers essentially the same direction a human picked by reasoning about class means --- reassuring, but also a preview of the limits of a single neuron: it can only ever \emph{rediscover} what a single linear projection is capable of.

\section{Training on the unsolvable problem}

Chapter~\ref{ch:lesson30} showed that \emph{no} linear projection separates a ring from the disk it surrounds --- the best exhaustive search over directions and thresholds found was $74\%$ accuracy. What happens when gradient descent, rather than brute-force search, tries to solve the same problem? Trained for 2000 epochs: the learned weight vector shrinks to near-zero ($\|w\|=0.0824$), and accuracy lands at $50.0\%$ --- chance level, \emph{worse} than the brute-force search's $74\%$.

Gradient descent actually does worse here than brute-force search --- it converges to a near-zero weight vector instead of finding the small, lopsided arc that let a brute-force search eke out $74\%$. The dataset is (approximately) symmetric around the origin, so the \emph{average} gradient pull from all the training points nearly cancels out in every direction, and the optimizer settles near $w\approx0$ rather than hunting for an asymmetric corner-case solution. Different failure mode, same underlying truth: a single linear projection cannot solve this problem, no matter how it's found. Chapter~\ref{ch:lesson32} fixes this --- not by using a smarter optimizer, but by giving the model more than one projection to work with.

\subsection*{Exercises}
\begin{enumerate}
  \item Increase \code{lr} in \code{train} well past a reasonable value (e.g.\ \code{lr=20}) on the two-blob dataset. What happens to the loss curve, and how does this relate to the step size overshooting the loss surface's curvature?
  \item Retrain on the two-blob dataset with several different random seeds. Does the learned direction always end up close to the hand-picked one from Chapter~\ref{ch:lesson30}, or does it vary a lot? What does that suggest about how many good solutions exist for a well-separated dataset?
  \item Make the outer ring asymmetric instead of a full circle (e.g.\ only half the ring). Retrain the single neuron. Does this genuinely broken symmetry noticeably change the final accuracy compared to the $\sim50\%$ chance-level result on the full ring?
\end{enumerate}

\vfill
\fullcode{lesson31}{lesson31_neural_network_fundamentals}

%% file: chapters/lesson32.tex
\chapter{Multi-Layer Perceptrons}
\label{ch:lesson32}

\begin{figure}[h]
\centering
\includegraphics[width=0.55\textwidth]{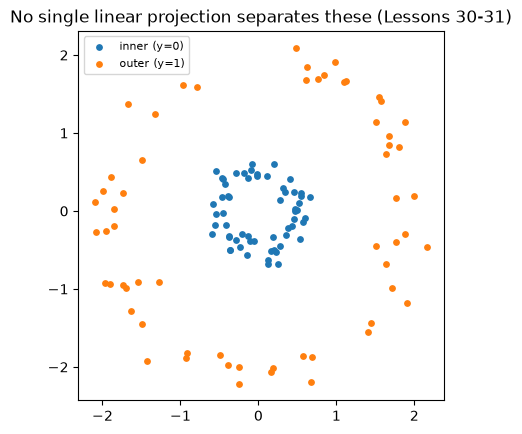}
\caption{The ring-and-disk dataset --- no single linear projection separates these (Chapters~\ref{ch:lesson30}--\ref{ch:lesson31}).}
\label{fig:l32-1}
\end{figure}

Chapter~\ref{ch:lesson31}'s single neuron, trained however carefully, could not beat chance on the ring-inside-a-disk dataset --- a single linear projection just isn't expressive enough (Figure~\ref{fig:l32-1}). This chapter adds one more layer: instead of \emph{one} projection followed by a threshold, use \emph{two} projections with a nonlinearity in between. That's it. That's the entire idea behind every deep network in this course --- stack more of these.

\section{Two projections, one nonlinearity}

A \textbf{multi-layer perceptron (MLP)} chains a \emph{hidden} projection into a new space, a nonlinear \textbf{activation} applied elementwise, and then a final linear projection (exactly Chapter~\ref{ch:lesson31}'s neuron) on top of the \emph{transformed} coordinates:
\begin{equation}
h = \text{ReLU}(W_1x+b_1), \qquad z=w_2^\top h+b_2, \qquad p=\sigma(z).
\end{equation}
$\text{ReLU}(u)=\max(0,u)$ is the simplest common activation: linear everywhere except a single kink at zero. That one kink, applied independently to every unit of $h$, is enough --- the hidden layer can bend, fold, and stretch the input space so a \emph{linear} boundary in the transformed space corresponds to a highly nonlinear boundary back in the original coordinates.

\subsection*{Why the nonlinearity is essential}

Without ReLU --- if the hidden layer were left as plain $h=W_1x+b_1$ --- the two layers would collapse into a single linear projection:
\begin{equation}
z = w_2^\top(W_1x+b_1)+b_2 = \underbrace{(w_2^\top W_1)}_{w'^\top}x + \underbrace{(w_2^\top b_1+b_2)}_{b'}.
\end{equation}
That's exactly Chapter~\ref{ch:lesson31}'s single neuron again, just with the two layers' weights multiplied together into one $w'$ and $b'$. This is the flip side of Chapter~\ref{ch:lesson30}'s observation that matrix multiplication is projection, stacked: composing two matrix multiplications is \emph{still} just one matrix multiplication, so stacking purely linear layers buys nothing, no matter how many are chained. ReLU's kink is what breaks that collapse --- because it isn't a linear function, $w_2^\top\text{ReLU}(W_1x+b_1)+b_2$ can no longer be rewritten as any single $w'^\top x+b'$. Verified directly: passing the same random data through two \emph{linear} layers matches the single combined layer $W_1W_2$, $b_1W_2+b_2$ to within $1.78\times10^{-15}$.

\section{Backprop through two layers}

The chain rule extends cleanly: propagate the error signal $\partial L/\partial z$ from Chapter~\ref{ch:lesson31} backward through the output projection to get $\partial L/\partial h$, then backward through the ReLU and the hidden projection to get $\partial L/\partial W_1,\partial L/\partial b_1$. Every step is either ``multiply by a weight matrix's transpose'' or ``multiply elementwise by an activation's derivative'' --- this is \emph{all} backpropagation ever is, no matter how many layers are stacked. A from-scratch implementation of forward and backward passes through a 2-layer MLP matches PyTorch's autograd gradients to within $3\times10^{-17}$ on every parameter.

\section{Training}

\begin{lstlisting}
def forward(X, W1, b1, W2, b2):
    a1 = relu(X @ W1 + b1)
    z2 = (a1 @ W2 + b2).ravel()
    return sigmoid(z2), (X, a1)
\end{lstlisting}

\begin{figure}[h]
\centering
\includegraphics[width=0.5\textwidth]{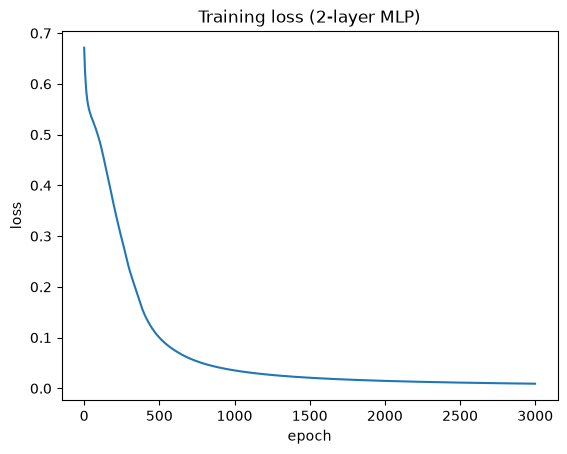}
\caption{Training loss (2-layer MLP) over 3000 epochs.}
\label{fig:l32-2}
\end{figure}

With $H=4$ hidden units, trained for 3000 epochs: final accuracy $100.0\%$ (Chapter~\ref{ch:lesson31}'s single neuron managed only $50\%$ on this dataset), shown in Figure~\ref{fig:l32-2}.

\section{The hidden layer's projections, made visible}

\begin{figure}[h]
\centering
\includegraphics[width=0.5\textwidth]{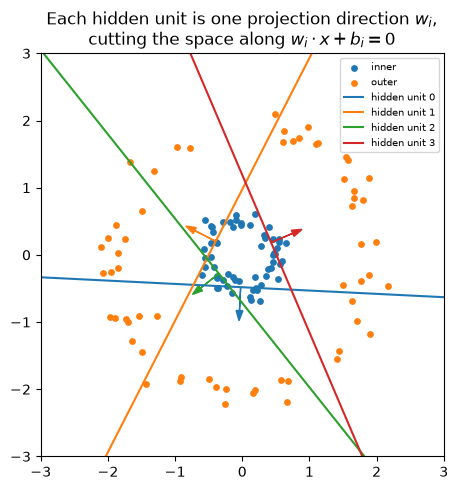}
\caption{Each hidden unit is one projection direction $w_i$, cutting the space along $w_i\cdot x+b_i=0$.}
\label{fig:l32-3}
\end{figure}

Each hidden unit is itself a Chapter~\ref{ch:lesson30}-style projection: a direction $w_i$ (one column of $W_1$) and an offset $b_i$ (one entry of $b_1$), together defining a cut $w_i\cdot x+b_i=0$ that ReLU zeroes out on one side of. With only 2 input dimensions, all $H$ of these can be drawn directly over the original data, as Figure~\ref{fig:l32-3} shows.

On its own, each cut is just a straight line --- no single one of them separates the ring from the disk. But the output layer combines all $H$ of their (ReLU'd) results together, and a handful of straight cuts arranged around the ring, combined with the right weights, can approximate a closed curve well enough to enclose it.

\subsection*{What the hidden layer actually did}

\begin{figure}[h]
\centering
\includegraphics[width=0.9\textwidth]{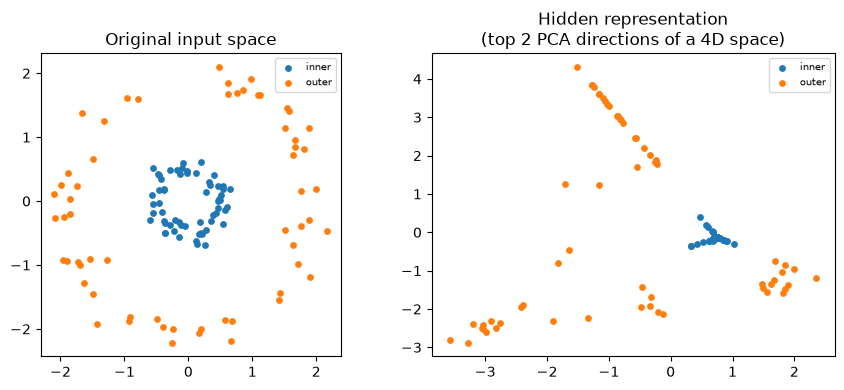}
\caption{Original input space vs.\ the hidden representation (top 2 PCA directions of a 4D space).}
\label{fig:l32-4}
\end{figure}

The output layer is \emph{just} a linear projection (Chapter~\ref{ch:lesson31}'s neuron) --- but it acts on $h$, the hidden layer's output, not on the original $x$. If the hidden layer did its job, the \emph{transformed} points should already be much easier to separate with a straight line. Since $h$ lives in $\mathbb{R}^4$ here, PCA (Chapter~\ref{ch:lesson06}) visualizes it in 2D, shown alongside the original space in Figure~\ref{fig:l32-4}.

Even this lossy 2D snapshot of the full 4D hidden space shows the two classes pulled apart into something close to linearly separable --- a big improvement over the $74\%$ ceiling that was the best \emph{any} line could do in the original space (Chapter~\ref{ch:lesson30}). The actual output neuron works in the full 4D hidden space, where it achieves 100\% accuracy exactly. This is the entire mechanism of deep learning in miniature: each layer \emph{reshapes the space} so that the next layer's job gets easier, until the final layer's job is trivial --- a single linear projection.

\section{Model capacity: how many hidden units are enough?}

A model's \textbf{capacity} is how rich a family of functions it can represent. Here that's set directly by $H$, the hidden layer's width: each hidden unit contributes one straight cut, and the output layer combines these units. Too few units, and the model simply cannot represent the boundary needed by the dataset --- no amount of training can find a solution that doesn't exist. Sweeping $H$ and retraining from scratch at each width, across 5 random initializations each:

\begin{center}
\renewcommand{\arraystretch}{1.3}
\begin{tabular}{rrrr}
\hline
\textbf{$H$} & \textbf{mean acc} & \textbf{min} & \textbf{max} \\
\hline
1  & 72.0\%  & 70.8\% & 73.3\% \\
2  & 89.5\%  & 67.5\% & 95.8\% \\
3  & 89.5\%  & 72.5\% & 100.0\% \\
4  & 100.0\% & 100.0\% & 100.0\% \\
6  & 100.0\% & 100.0\% & 100.0\% \\
8  & 100.0\% & 100.0\% & 100.0\% \\
16 & 100.0\% & 100.0\% & 100.0\% \\
\hline
\end{tabular}
\end{center}

\begin{figure}[h]
\centering
\includegraphics[width=0.55\textwidth]{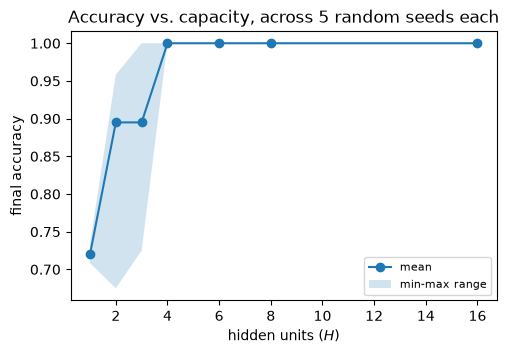}
\caption{Accuracy vs.\ capacity, across 5 random seeds at each width.}
\label{fig:l32-5}
\end{figure}

$H=1$ (Figure~\ref{fig:l32-5}) caps out well below 100\% no matter how long it trains or which seed it starts from --- that's not an optimization failure, it's a representational ceiling: one hidden unit is just a single cut, so it inherits Chapter~\ref{ch:lesson30}'s $\sim74\%$ linear limit almost exactly. $H=2$ is better, but still capacity-limited --- none of the five seeds reach 100\%. $H=3$ is the actual knife's edge: the capacity to represent a perfect solution is there (one seed reaches exactly 100\%), but whether training finds it depends on \emph{which} seed it starts from (a preview of Chapter~\ref{ch:lesson33}'s optimization landscape problems).

By $H=4$, every seed reaches 100\% reliably: there's enough capacity that the network stops being the bottleneck, and training reliably finds a solution within that capacity. This is the general shape of a capacity curve --- a hard floor set by what the model \emph{can} represent, with optimization difficulty riding on top of it near the floor's edge. This same curve is revisited in Chapter~\ref{ch:lesson35} from the opposite direction: there, capacity is held generously high and \emph{data} is scarce, which turns ``enough capacity to represent the right answer'' into ``also enough capacity to memorize the wrong one.''

Width isn't the only knob. \textbf{Depth} --- stacking more layers instead of making a layer wider --- is the other axis, and it isn't just ``width in disguise'': a deep, narrow network can represent some functions far more compactly than a shallow, wide one needs to, but it introduces its own failure mode instead. Chapter~\ref{ch:lesson37} picks this up directly: stacking enough layers can make gradients vanish before they ever reach the earliest ones, a problem not encountered by this chapter's shallow 2-layer network.

\subsection*{Exercises}
\begin{enumerate}
  \item The capacity sweep above used ReLU. Rerun it with \code{tanh} instead. Does the same $H=1$ ceiling and $H=3$-to-$H=4$ transition appear, or does a smooth activation change where the floor sits?
  \item Replace \code{relu}/\code{relu\_deriv} with \code{tanh}/its derivative ($1-\tanh^2$) throughout, and retrain. Does it still solve the problem? Compare the resulting loss curve's shape to the ReLU version's.
  \item Remove the nonlinearity entirely (replace \code{relu(z1)} with just \code{z1}). Confirm the network can no longer beat Chapter~\ref{ch:lesson31}'s $\sim50\%$ ceiling --- consistent with the collapse-to-a-single-projection argument above, now demonstrated on actual training dynamics instead of just static algebra.
\end{enumerate}

\vfill
\fullcode{lesson32}{lesson32_multilayer_perceptrons}

%% file: chapters/lesson33.tex
\chapter{Optimization}
\label{ch:lesson33}

Chapters~\ref{ch:lesson31}--\ref{ch:lesson32} used plain gradient descent: take the gradient, take a fixed-size step against it, $p\leftarrow p-\eta\nabla f(p)$, where $p$ is whatever's being learned, $\eta$ is the \textbf{learning rate}, and $\nabla f(p)$ is the loss gradient at the current parameters. That's enough to prove the \emph{idea} works, but it's rarely how real networks are trained. This chapter covers four practical upgrades --- \textbf{momentum}, \textbf{adaptive step sizes (Adam)}, \textbf{weight initialization schemes}, and \textbf{learning rate schedules} --- each fixing a specific, concrete failure mode of plain gradient descent.

In real training, the gradient is almost never computed over the \emph{entire} training set at once --- it wouldn't fit into memory. Instead, it's estimated from a small, randomly sampled \textbf{mini-batch} (say, 32 or 256 examples) at every step. The resulting noisy, sample-to-sample-varying estimate is what gives \textbf{stochastic gradient descent (SGD)} its name. This chapter mostly uses the exact, noise-free gradient of a fixed function (calling it ``SGD'' for convenience); the last section simulates mini-batch noise directly to show why it matters.

\section{A loss landscape where plain gradient descent struggles}

\begin{figure}[h]
\centering
\includegraphics[width=0.55\textwidth]{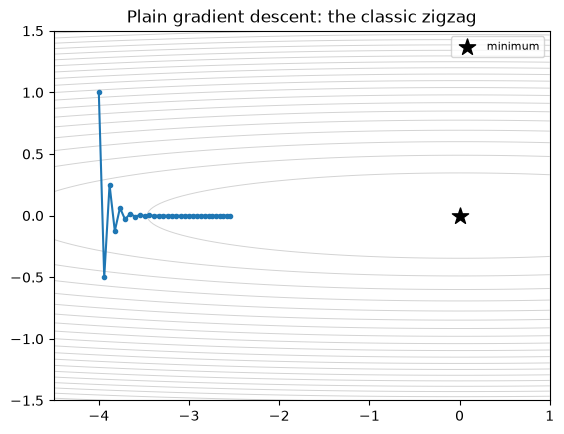}
\caption{Plain gradient descent: the classic zigzag, 30 steps on a narrow bowl.}
\label{fig:l33-1}
\end{figure}

$f(x,y)=0.05x^2+5y^2$ is a bowl that's steep in $y$ and shallow in $x$. A step size large enough to make progress along $x$ overshoots along $y$, causing the classic \emph{zigzag}: 30 steps of plain gradient descent leave a final loss of $0.3230$ (Figure~\ref{fig:l33-1}).

\section{Momentum: remember where you were heading}

Instead of stepping purely along the current gradient, accumulate a running \textbf{velocity} --- a weighted average of past gradients --- and step along that instead:
\begin{equation}
v \leftarrow \beta v - \eta\nabla f(p), \qquad p \leftarrow p + v.
\end{equation}
Momentum's real benefit is acceleration, not damping. Along a consistent direction (the shallow $x$ axis here), every gradient points the same way, so the running velocity keeps growing step after step --- the effective step size along that axis becomes much larger than $\eta$ alone would give, exactly like a ball picking up speed rolling downhill. Along an oscillating direction (the steep $y$ axis, flipping sign every step), consecutive gradients partially cancel in that same running average --- but only \emph{partially}: a single momentum coefficient $\beta$ is shared across every direction, so that cancellation isn't tuned to any one direction's curvature, and the steep axis can still swing substantially before it decays. The net effect, over a full run, is usually still a win: with $\beta=0.9$, 30 steps land at loss $0.2785$ --- better than plain SGD, but by acceleration plus a rough, untuned brake, not a clean fix for oscillation. A from-scratch implementation matches \code{torch.optim.SGD(..., momentum=0.9)} to $2.22\times10^{-16}$.

\section{Adam: per-parameter adaptive step sizes}

\textbf{Adam} (\citelink{kingmaba2014}{Kingma \& Ba, 2015}) tracks both a momentum-like running mean of the gradient ($m$) \emph{and} a running mean of the squared gradient ($v$), then divides the step by $\sqrt{v}$ --- automatically shrinking the step size for parameters with consistently large gradients (like the steep $y$ direction here) and boosting it for parameters with small ones:
\begin{equation}
m \leftarrow \beta_1 m + (1-\beta_1)g, \qquad v \leftarrow \beta_2 v + (1-\beta_2)g^2, \qquad p \leftarrow p - \eta\frac{\hat m}{\sqrt{\hat v}+\epsilon},
\end{equation}
where $\hat m,\hat v$ are bias-corrected versions of $m,v$ (mattering mainly in the first few steps). With $\eta=0.3$ (twice plain SGD's rate), 30 steps reach loss $0.1244$ --- the lowest of the three optimizers, matching \code{torch.optim.Adam} to $1.11\times10^{-16}$.

\begin{figure}[h]
\centering
\includegraphics[width=0.6\textwidth]{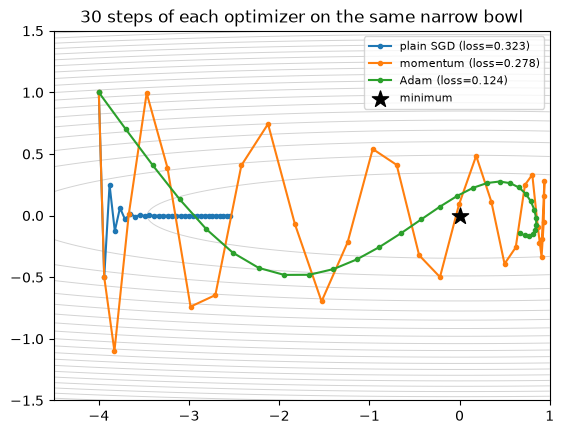}
\caption{30 steps of plain SGD, momentum, and Adam on the same narrow bowl.}
\label{fig:l33-2}
\end{figure}

Figure~\ref{fig:l33-2} makes the difference visible: plain SGD is stuck oscillating across the narrow valley, barely progressing along the shallow direction. Momentum's accumulated velocity clearly accelerates progress along that shallow direction, and it ends with a lower loss than plain SGD --- but notice its path actually swings \emph{wider} in $y$ than plain SGD's own, not narrower, since a single shared momentum coefficient can't damp each axis to the degree that axis specifically needs. Adam, which independently rescales each direction, is what actually tames the steep axis directly, while still racing along the shallow one.

\section{Weight initialization: why zero doesn't work}

It's tempting to initialize all weights to zero. Although this seems like the most ``neutral'' starting point, it's actually catastrophic for any layer with more than one unit: every hidden unit computes the exact same function of the input, gets the exact same gradient, and gets updated by the exact same amount, forever. This \textbf{symmetry} never breaks on its own: training a 4-hidden-unit MLP on the ring-and-disk dataset from zero initialization for 3000 epochs reaches only $50.0\%$ accuracy, with all four hidden units still identical (all zero); randomly initialized weights reach $100.0\%$. Small \emph{random} initialization breaks the symmetry: every unit starts out computing something slightly different, so gradient descent can push them apart and let them specialize. This is why every framework's default layer initialization uses small random values, never zeros.

\section{Weight initialization schemes: how large should ``small random'' be?}

Zero-init fails completely; ``small random'' fixes it. But \emph{how} small? Pick a fixed standard deviation and it works for one layer width, then quietly fails at another. Tracking the variance of activations after passing random data through a deep stack of \code{tanh}-activated linear layers, using a fixed weight std of $0.05$ at every width:

\begin{center}
\renewcommand{\arraystretch}{1.3}
\begin{tabular}{rr}
\hline
\textbf{width} & \textbf{variance after 20 layers} \\
\hline
16  & $7.94\times10^{-28}$ \\
64  & $1.02\times10^{-16}$ \\
256 & $3.12\times10^{-5}$ \\
\hline
\end{tabular}
\end{center}

A fixed std of $0.05$ vanishes to essentially zero after 20 layers --- and the amount it vanishes depends heavily on layer width. Each layer's pre-activation variance scales as $\text{fan\_in}\times\text{std}^2$, where \textbf{fan-in} is the layer's number of inputs: at width 16, that's $16\times0.05^2\approx0.04$, so each layer keeps only about 4\% of the previous layer's variance, and 20 compounding layers erase the signal almost completely. At width 256, the \emph{same} fixed std keeps $256\times0.05^2\approx0.64$ of the variance per layer --- still shrinking, but far more slowly. A fixed std's effect depends entirely on fan-in, so the same number that badly vanishes a narrow layer would, at a wide enough layer, push $\text{fan\_in}\times\text{std}^2$ past 1 and start \emph{exploding} instead.

\textbf{Xavier initialization} (\citelink{glorotbengio2010}{Glorot \& Bengio, 2010}), also known as \emph{Glorot initialization}, fixes this by scaling the std with the layer's fan-in: $\text{std}=\sqrt{1/\text{fan\_in}}$, which keeps $\text{fan\_in}\times\text{std}^2$ pinned at exactly 1 regardless of width. Repeating the experiment with Xavier-scaled weights keeps variance in the same ballpark at every width ($0.026$, $0.023$, $0.027$ at widths 16, 64, 256), instead of vanishing catastrophically --- and \code{nn.init.xavier\_normal\_}'s empirical weight std ($0.0627$) matches the formula's prediction ($0.0625$) closely.

But Xavier's derivation assumes a symmetric activation like \code{tanh}. \textbf{Kaiming initialization} (\citelink{he2015kaiming}{He et al., 2015}), also known as \emph{He initialization}, is the same idea adjusted for \code{ReLU}, which zeros out roughly half its inputs and so needs twice the variance to compensate: $\text{std}=\sqrt{2/\text{fan\_in}}$. Using Xavier's \code{tanh}-derived scale on a \code{ReLU} network still vanishes: at width 256, Xavier scale gives a variance of $3.06\times10^{-7}$ after 20 layers, while Kaiming scale keeps it at $1.324$ --- and \code{nn.init.kaiming\_normal\_}'s empirical std ($0.0887$) again matches the formula ($0.0884$).

Rule of thumb: use Kaiming init for \code{ReLU}-family networks (the default for \code{nn.Conv2d}/\code{nn.Linear} is actually already a variant of this), and Xavier for \code{tanh}/\code{sigmoid}. Both are strictly better than picking a fixed std and hoping. They're derived, not tuned, from a simple requirement: keep activation variance roughly constant from layer to layer, so a network can be made arbitrarily deep without silently losing its signal before training even starts.

\section{Learning rate schedules}

\begin{figure}[h]
\centering
\includegraphics[width=0.6\textwidth]{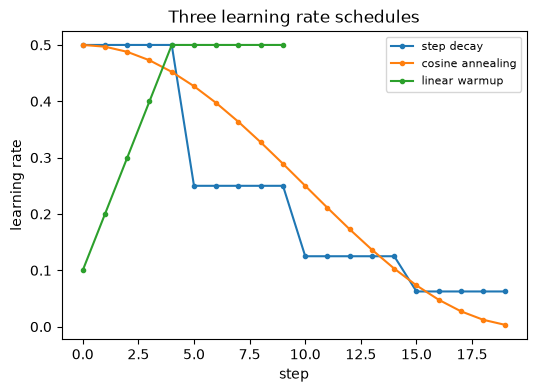}
\caption{Three learning rate schedules: step decay, cosine annealing, and linear warmup.}
\label{fig:l33-3}
\end{figure}

A single fixed learning rate for the whole training run is tempting, but it's often the wrong choice: a rate large enough for fast early progress is usually too large to settle precisely once training nears a minimum. \textbf{Learning rate schedules} change the rate over time. Three common ones, each validated against \code{torch.optim.lr\_scheduler} to within numerical roundoff (Figure~\ref{fig:l33-3}):
\begin{itemize}
  \item \textbf{Step decay}: multiply the rate by a fixed factor every $N$ steps.
  \item \textbf{Cosine annealing}: smoothly decay along a cosine curve from the initial rate down to (near) zero --- the name is borrowed directly from \textbf{simulated annealing}, the classical optimization technique of starting ``hot'' (free to explore, even accepting worse moves) and gradually ``cooling'' toward a fixed solution.
  \item \textbf{Linear warmup}: \emph{ramp up} from zero over the first few steps, before applying the main schedule --- used because early updates, before a model's gradient statistics have stabilized, can be unreliable enough that a large rate from step one causes damage a few steps of gradual ramp-up would have avoided.
\end{itemize}

\subsection*{Why decay actually helps: noisy gradients}

\begin{figure}[h]
\centering
\includegraphics[width=0.6\textwidth]{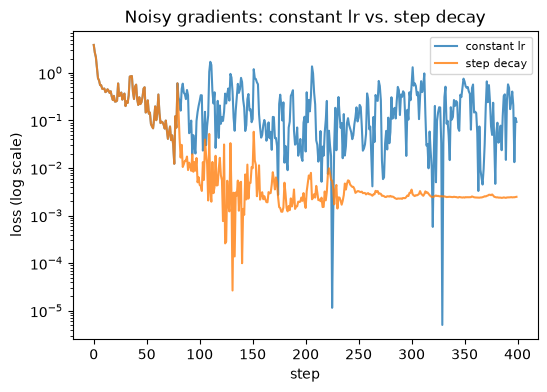}
\caption{Noisy gradients: constant learning rate vs.\ step decay, loss on a log scale.}
\label{fig:l33-4}
\end{figure}

On the exact, noise-free narrow bowl from the top of this chapter, a well-chosen constant learning rate converges just fine --- there's nothing to decay away from. Real training almost never sees the exact gradient, though: each step uses a mini-batch estimate, which is noisy. Simulating that by adding random noise to the gradient every step, and comparing a constant rate against step decay: constant learning rate leaves a mean loss over the final 50 steps of $0.2142$, while step decay reaches $0.0025$ --- nearly two orders of magnitude lower (Figure~\ref{fig:l33-4}).

With noisy gradients, a constant learning rate large enough to make fast early progress never actually settles --- it keeps bouncing around the minimum by an amount proportional to the learning rate itself, forever. Step decay makes the same fast early progress, then shrinks the rate as training continues, tightening that bounce and landing far closer to the true minimum. This is the real justification for learning rate schedules: not that a fixed rate is ``wrong'' on some idealized noise-free problem, but that it can't simultaneously be large (for speed) and small (for precision) when every gradient is a noisy estimate, which is the normal situation for real mini-batch training.

\subsection*{Exercises}
\begin{enumerate}
  \item Increase \code{momentum} from 0.9 to 0.99 on the narrow-bowl problem. Does it converge faster, or does it start to overshoot the minimum and oscillate on the \emph{shallow} axis now?
  \item Try initializing the MLP's weights identically but nonzero instead of all-zero. Does symmetry still fail to break?
  \item Adam's learning rate (0.3) is much larger than plain SGD's (0.15), yet Adam remains stable while a plain SGD run at that rate would diverge on the steep axis. Try it and confirm. What does that suggest about why Adam is often described as being more forgiving of the learning-rate choice?
  \item Rerun the Xavier-vs-Kaiming comparison at width 16 and width 1024. Does the gap between the two get bigger or smaller as width grows, matching the fan-in scaling argument used to derive both formulas?
  \item In the noisy-gradient learning-rate demo, try \code{noise\_std=0.0} (no noise at all). Does step decay still help, hurt, or make no difference relative to a constant rate?
\end{enumerate}

\vfill
\fullcode{lesson33}{lesson33_optimization}

%% file: chapters/lesson34.tex
\chapter{Convolutional Neural Networks (CNNs)}
\label{ch:lesson34}

The layers in Chapters~\ref{ch:lesson31}--\ref{ch:lesson33} treat every input as a flat vector, and that creates two distinct problems for images. First, each pixel gets its own independent weight, with no notion that neighboring pixels are \emph{near} each other --- the entire spatial structure images have is thrown away. Second, because every position gets its own separate weights, a fully-connected layer has no way to reuse a filter it's already learned: it must independently re-learn what an edge looks like at every single position, all over again. \textbf{Convolution} --- already built manually in Chapter~\ref{ch:lesson10} --- fixes both problems at once by baking in an \textbf{inductive bias}: restrict each output to a small local patch, and share that patch's weights across every position.

\section{A conv layer is a projection, restricted and shared}

Recall Chapter~\ref{ch:lesson30}'s central idea: $y=w^\top x+b$, a projection. A convolutional layer computes \emph{exactly this}, but for $x$, it only looks at a small local patch (say $3\times3$) instead of the whole image --- and, crucially, it reuses the \emph{same} $w$ at every patch location. Two consequences:
\begin{itemize}
  \item \textbf{Vastly fewer parameters.} A $3\times3$ filter over a single input channel has 9 weights, no matter how big the image is --- a fully-connected layer over a $16\times16$ image needs 256 weights \emph{per output unit}.
  \item \textbf{Translation equivariance.} Because the same filter slides everywhere, a feature learned at one position is automatically detected at every other position too, for free. \emph{Equivariant} means the output shifts along with the input, rather than staying fixed --- shift the image by 2 pixels and the feature map's detections shift by exactly 2 pixels too. That's a distinct, weaker property than \emph{invariance} (an unchanged output despite a shifted input), which pooling adds a bit of, later in this chapter.
\end{itemize}

\subsection*{A conv layer kernel is 3D}

A ``$3\times3$ filter'' in a CNN is really $3\times3\times C_{in}$, not just $3\times3$. Chapter~\ref{ch:lesson10}'s kernels were flat 2D grids, because they slid over a single grayscale channel. A conv layer's filter has to reach across \emph{every} input channel at once --- a $3\times3$ filter on a 3-channel RGB image has $3\times3\times3=27$ weights, one full depth-slice per input channel, summed together into a single output value at each position. Stack $C_{out}$ of these $3\times3\times C_{in}$ filters side by side (one per output channel) and the full weight tensor is 4D: \code{(C\_out, C\_in, 3, 3)}.

\section{Conv2d, forward and backward, from scratch}

This is the same sliding-window computation as Chapter~\ref{ch:lesson10}'s convolution, just with the kernel now a set of \emph{learnable} weights instead of a fixed, hand-designed one --- which means it needs a backward pass too.

\begin{lstlisting}
def conv2d_forward(img, kernel, bias):
    N, C, H, W = img.shape
    OC, IC, KH, KW = kernel.shape
    OH, OW = H - KH + 1, W - KW + 1
    out = np.zeros((N, OC, OH, OW))
    for n in range(N):
        for oc in range(OC):
            for i in range(OH):
                for j in range(OW):
                    patch = img[n, :, i:i+KH, j:j+KW]
                    out[n, oc, i, j] = np.sum(patch * kernel[oc]) + bias[oc]
    return out
\end{lstlisting}

A from-scratch forward pass matches \code{F.conv2d} to $8.88\times10^{-16}$; the from-scratch backward pass matches PyTorch autograd's gradients on the input, kernel, and bias to $8.88\times10^{-16}$, $7.99\times10^{-15}$, and $1.78\times10^{-15}$ respectively.

\subsection*{Filters you already know: Chapter~\ref{ch:lesson12}'s edge detectors}

\begin{figure}[h]
\centering
\includegraphics[width=0.5\textwidth]{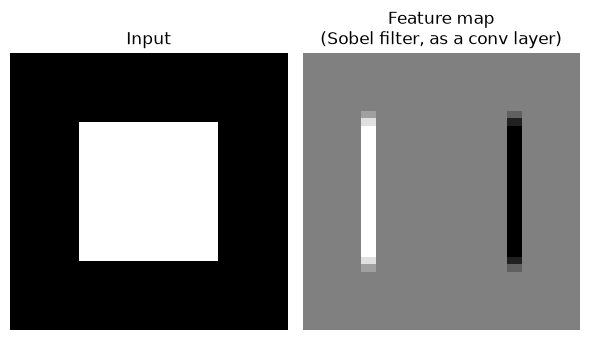}
\caption{Sobel's edge kernel, applied as a conv layer, on a simple square.}
\label{fig:l34-1}
\end{figure}

The output of a conv layer before training is nonsense; the point is that gradient descent will \emph{find} useful filters. To see what a useful filter's output already looks like, apply a filter designed by hand back in Chapter~\ref{ch:lesson12}: Sobel's edge kernel. As it turns out, trained CNNs' first-layer filters often converge to something visually similar to this --- oriented edge and blob detectors --- regardless of what the network was trained to do (Figure~\ref{fig:l34-1}).

\section{Pooling: downsampling with a purpose}

\textbf{Max pooling} slides a window over the feature map and keeps only the maximum value in each window, shrinking the spatial size (like the pyramids of Chapter~\ref{ch:lesson11}) while adding a small amount of local translation invariance --- a feature detected at slightly different positions within one pooling window still produces the same output. A from-scratch $2\times2$ max pool matches \code{F.max\_pool2d} exactly (max diff $0.00\times10^{0}$).

\section{Putting them together: convolution and pooling, interleaved}

A real CNN is neither all convolution nor all pooling --- it's the two stacked in alternation: convolution, pooling, convolution, pooling, and so on. Each pooling step roughly \emph{halves} the spatial size ($H$ and $W$), and the rule of thumb is to roughly \emph{double} the number of channels in the conv layer that follows. That's not a coincidence: pooling throws away spatial resolution, so doubling the channel count right after keeps the total amount of information ($C\times H\times W$) roughly constant from stage to stage, even as the representation trades ``a little detail at many positions'' for ``more detail at fewer positions.''

\section{Why translation equivariance actually matters: a generalization test}

\begin{figure}[h]
\centering
\includegraphics[width=0.9\textwidth]{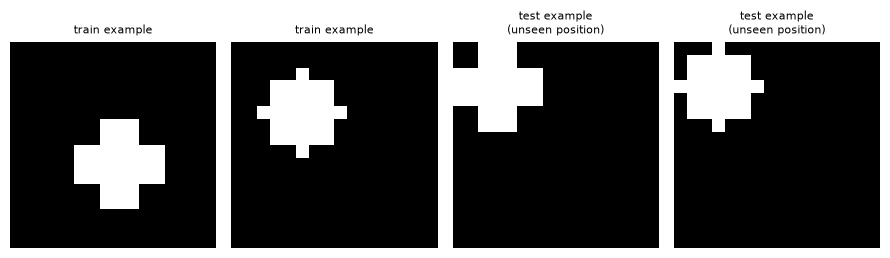}
\caption{Train examples (centered positions) vs.\ test examples (unseen, corner positions).}
\label{fig:l34-2}
\end{figure}

Build a small classification task: a $16\times16$ image contains either a plus sign or a circle, at a \emph{random position}. Train on shapes placed only near the center; test on shapes placed only near the corners --- positions the model never saw during training (Figure~\ref{fig:l34-2}). A network that has genuinely learned ``what a plus looks like,'' rather than ``which pixels tend to be on for a plus at these particular training positions,'' should have no trouble with this.

\begin{lstlisting}
class CNNClassifier(nn.Module):
    def __init__(self):
        super().__init__()
        self.conv = nn.Sequential(
            nn.Conv2d(1, 8, 5, padding=2), nn.ReLU(),
            nn.MaxPool2d(2),
            nn.Conv2d(8, 16, 5, padding=2), nn.ReLU(),
            nn.AdaptiveMaxPool2d(1),  # global max pool: collapses ALL spatial position info
        )
        self.fc = nn.Linear(16, 1)
\end{lstlisting}

Training a flatten-based MLP and this CNN on the same data:

\begin{center}
\renewcommand{\arraystretch}{1.3}
\begin{tabular}{lrr}
\hline
\textbf{model} & \textbf{train acc} & \textbf{test acc (unseen positions)} \\
\hline
MLP & 100.0\% & 73.3\% \\
CNN & 100.0\% & 100.0\% \\
\hline
\end{tabular}
\end{center}

Both models fit the training data perfectly. On positions neither has ever seen, the flatten-based MLP does little better than a coin flip --- it memorized \emph{which pixels} tend to be on for each class at the training positions, and that knowledge doesn't transfer. The CNN, whose global max pooling forces the final decision to depend only on \emph{what filters fired somewhere}, not \emph{where}, generalizes to the new positions with no loss in accuracy at all.

\subsection*{Exercises}
\begin{enumerate}
  \item Replace \code{nn.AdaptiveMaxPool2d(1)} in \code{CNNClassifier} with \code{nn.Flatten()} directly on the conv features (removing the global pooling, so spatial position is preserved all the way to the final linear layer). Retrain and re-evaluate on the unseen-position test set. Does the CNN's generalization advantage survive?
  \item Increase the gap between train and test position ranges, right at the image border where shapes get clipped. Does the CNN's accuracy hold up, or does it degrade --- and if it degrades, is that a translation-invariance failure or something else entirely (think about what happens to a shape's \emph{appearance}, not just its position, right at an edge)?
  \item \code{conv2d\_forward} above pads nothing, so the output shrinks with every layer. Modify it to support zero-padding and confirm the output size matches the input size, validated against \code{F.conv2d(..., padding=1)}.
\end{enumerate}

\vfill
\fullcode{lesson34}{lesson34_convolutional_neural_networks}

%% file: chapters/lesson35.tex
\chapter{Training a CNN}
\label{ch:lesson35}

Chapter~\ref{ch:lesson34}'s CNN was trained and evaluated on data drawn from the same, fairly generous distribution. Real training has a much sharper failure mode lurking: with too little data and too much model capacity, a network can perfectly memorize its training set while learning nothing that generalizes. Chapter~\ref{ch:lesson32} showed that more capacity means a model \emph{can} represent more; this chapter shows the flip side --- that same surplus capacity, with too little data to constrain it, is exactly what lets a network memorize instead of generalize. This chapter makes that failure concrete, then fixes it three different ways: \textbf{data augmentation} (Chapter~\ref{ch:lesson08}'s transforms, repurposed), \textbf{regularization} (weight decay), and \textbf{dropout}.

\section{A deliberately hard, small dataset}

\begin{figure}[h]
\centering
\includegraphics[width=0.85\textwidth]{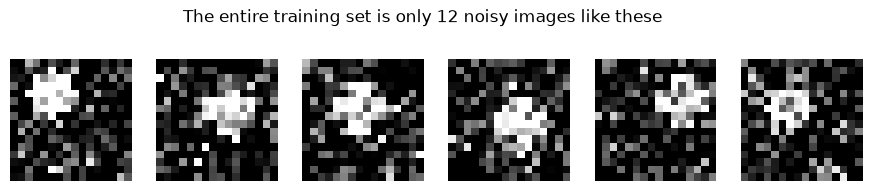}
\caption{The entire training set is only 12 noisy images like these.}
\label{fig:l35-1}
\end{figure}

The same plus-vs-circle task as Chapter~\ref{ch:lesson34}, but now with only \textbf{12 training images} and pixel noise added to every image (Figure~\ref{fig:l35-1}), while the validation set stays large (150 images) so its accuracy is a reliable estimate.

\section{Watching it overfit}

\begin{figure}[h]
\centering
\includegraphics[width=0.6\textwidth]{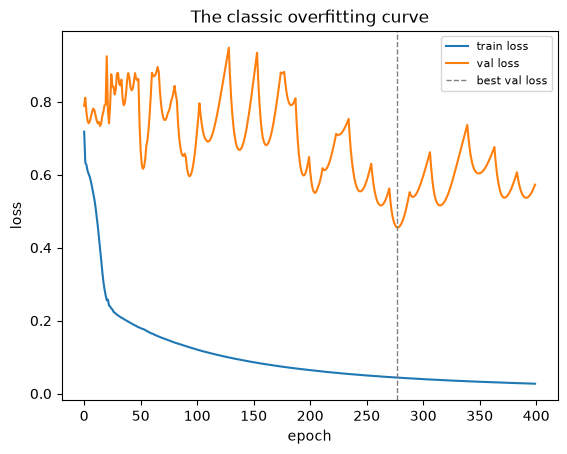}
\caption{The classic overfitting curve: training loss falls monotonically while validation loss bottoms out and rises again.}
\label{fig:l35-2}
\end{figure}

Train a reasonably large CNN (Chapter~\ref{ch:lesson34}'s architecture) on just these 12 images, and track both training loss and \emph{validation} loss (on the held-out 150 images) at every epoch:

\begin{lstlisting}
for _ in range(epochs):
    loss = F.binary_cross_entropy_with_logits(model(Xtr), ytr)
    loss.backward()
    opt.step()
    val_losses.append(F.binary_cross_entropy_with_logits(model(Xval), yval).item())
\end{lstlisting}

Final train accuracy reaches $100.0\%$, but final validation accuracy is only $72.7\%$; validation loss bottoms out at $0.456$ at epoch 277 (of 400), then climbs back up to $0.573$ by the end (Figure~\ref{fig:l35-2}).

Training loss marches steadily to zero --- the network perfectly memorizes all 12 images, noise included. Validation loss, meanwhile, bottoms out part-way through training and then climbs back up: past that point, every further epoch makes the model \emph{more} confidently wrong about data it hasn't seen. Final validation accuracy lands well short of the training set's perfect score, despite the training set being fit exactly.

\section{Fix 1: data augmentation}

\begin{figure}[h]
\centering
\includegraphics[width=0.85\textwidth]{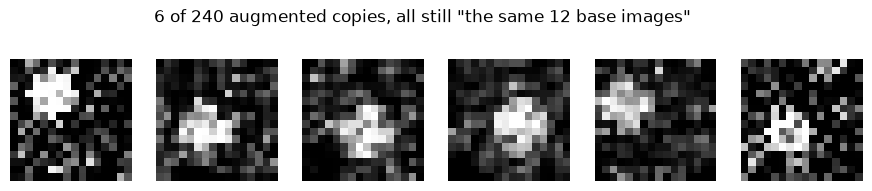}
\caption{6 of 240 augmented copies, all still ``the same 12 base images.''}
\label{fig:l35-3}
\end{figure}

If there isn't enough real data, manufacture more from what's there. Apply random transformations from Chapter~\ref{ch:lesson08} (flips, small rotations) to each training image --- the label doesn't change, but the pixels do, so the network sees a much wider variety of ``what a plus/circle can look like'' instead of memorizing 12 exact images (Figure~\ref{fig:l35-3}). With 20 augmented copies of each of the 12 base images: train accuracy stays at $100.0\%$, but validation accuracy jumps from $72.7\%$ to $95.3\%$.

\section{Fix 2: weight decay}

\textbf{Weight decay} adds a penalty proportional to the squared weight magnitudes directly into the loss (equivalently, it shrinks every weight slightly toward zero on every update). Large, highly-tuned weights are exactly what a network needs to memorize 12 specific noisy images; penalizing weight magnitude makes that memorization more costly relative to finding a simpler, smoother function --- without adding a single extra training example. With \code{weight\_decay=0.05}: train accuracy stays at $100.0\%$, validation accuracy rises to $84.0\%$.

\begin{center}
\renewcommand{\arraystretch}{1.3}
\begin{tabular}{lrr}
\hline
\textbf{approach} & \textbf{train acc} & \textbf{val acc} \\
\hline
no fix       & 100.0\% & 72.7\% \\
augmentation & 100.0\% & 95.3\% \\
weight decay & 100.0\% & 84.0\% \\
\hline
\end{tabular}
\end{center}

Both fixes recover a large chunk of the lost validation accuracy, from two different angles: augmentation attacks the problem by giving the model more (synthetic) data to be right about; weight decay attacks it by making the model less willing to contort itself around a small dataset in the first place. In practice, both are normally used together, along with other regularizers like \textbf{dropout}, covered next, and \textbf{batch normalization} (Chapter~\ref{ch:lesson37}), which incidentally also acts as a mild regularizer.

\section{Fix 3: dropout}

\textbf{Dropout} (\citelink{srivastava2014}{Srivastava et al., 2014}) randomly zeroes out a fraction $p$ of a layer's activations on every training step, forcing the surviving units to not rely on any one specific other unit always being present. To keep the layer's output at the same overall scale whether or not dropout is active, the surviving activations are rescaled by $1/(1-p)$ --- this is ``inverted dropout,'' what every framework's \code{Dropout} layer actually implements. At evaluation time, dropout does nothing at all: the full, unmodified layer runs, which is why \code{model.eval()} matters --- forgetting it would leave dropout randomly firing at test time. A from-scratch inverted-dropout implementation zeroes $30.0\%$ of activations against a target $p=0.3$ (PyTorch's \code{F.dropout} zeroes $30.3\%$ on the same seed), and eval-mode dropout is confirmed to be a true no-op.

Applying it to this chapter's overfitting problem needs \emph{some} redundancy to work with --- zeroing half of a 32-unit feature vector going straight into a 1-unit output leaves little room to help, so a wider hidden layer (32 $\to$ 64 $\to$ 1) is added, with dropout on the 64-unit layer. With only 12 training images, results are noisy from one random seed to the next, so mean validation accuracy over 8 seeds is compared rather than a single run: no dropout reaches $62.8\%\pm7.0\%$, dropout($p=0.5$) reaches $70.2\%\pm8.0\%$.

In a toy setting this tiny (just 12 training images), there isn't much redundancy for dropout to exploit yet. Although only a modest improvement is shown here, dropout's benefit is well established at the scale of real datasets and real networks (hundreds of redundant units, thousands of examples). In practice dropout is almost always combined with the other regularizers on this page, not used alone.

\section{Putting it together: real photos, real overfitting}

Every fix so far has been tested one at a time, on a synthetic 12-image toy problem chosen specifically to make the failure mode unmistakable. This closing section repeats the experiment on real \textbf{CIFAR-10} (\citelink{krizhevsky2009cifar}{Krizhevsky, 2009}) photographs: a small, deliberately data-scarce training set of cats and trucks (40 images per class), with augmentation, weight decay, and dropout combined together. Real photographs contain lighting, clutter, and genuine visual ambiguity between classes. Each configuration is averaged over 3 seeds, since 80 training images is still small enough for individual runs to vary:

\begin{center}
\renewcommand{\arraystretch}{1.3}
\begin{tabular}{lrr}
\hline
\textbf{approach} & \textbf{train acc} & \textbf{val acc} \\
\hline
no fix       & 100.0\% & 78.8\% \\
augmentation & 100.0\% & 80.6\% \\
weight decay & 94.6\%  & 75.9\% \\
dropout      & 100.0\% & 77.9\% \\
all combined & 100.0\% & 82.3\% \\
\hline
\end{tabular}
\end{center}

No single fix is a clear winner here. Weight decay alone actually lands \emph{below} the unregularized baseline --- with \code{weight\_decay=0.01}, training accuracy no longer even reaches 100\%, meaning the penalty is strong enough to fight the objective without finding a better solution in its place. That's a useful, humbling result: the \code{weight\_decay=0.05} value used earlier in this chapter was tuned for a much smaller network on 12 tiny synthetic images, and a regularization strength that works well in one setting doesn't automatically transfer to a different architecture or real photographic data.

Augmentation and dropout each recover a modest amount of validation accuracy individually. ``All combined'' beats every single fix and the unregularized baseline, echoing the same lesson from the synthetic dataset: these three regularizers compound rather than substitute for each other. The gain over the unregularized baseline is real but modest (a few percentage points here), which is itself the honest takeaway: regularization narrows the gap between training and validation performance, it doesn't manufacture data or model capacity that isn't there.

Closing that gap further, on real images, takes what later chapters introduce --- more training data, better architectures (Chapter~\ref{ch:lesson37}), and eventually \textbf{transfer learning} (Chapter~\ref{ch:lesson38}): starting from a network already trained on millions of images, instead of memorizing from scratch on a handful of them.

\subsection*{Exercises}
\begin{enumerate}
  \item Try \code{weight\_decay} values of $0.001$, $0.05$, and $1.0$. Is there a point where it starts to \emph{hurt} training accuracy along with (eventually) validation accuracy?
  \item Sweep \code{dropout\_p} over $[0.0, 0.2, 0.4, 0.6, 0.8]$, averaging over the same 8 seeds at each value. Is there a value that's clearly best, or does the mean stay within one standard deviation across most of the range given how little data there is?
  \item Increase the augmentation multiplier from 20 to 100 copies per base image. Does validation accuracy keep improving, or does it plateau --- and if it plateaus, what does that suggest about the fundamental limit of augmenting a dataset that only contains 12 \emph{underlying} examples to begin with?
\end{enumerate}

\vfill
\fullcode{lesson35}{lesson35_training_a_cnn}

%% file: chapters/lesson36.tex
\chapter{Image Classification in Practice}
\label{ch:lesson36}

Every classifier so far has been evaluated with a single number: overall accuracy. That number hides a lot. This chapter builds a 4-class classifier on a deliberately imbalanced subset of \textbf{CIFAR-10} (\citelink{krizhevsky2009cifar}{Krizhevsky, 2009}) --- real photographs, not synthetic shapes --- and shows why accuracy alone can be misleading, using the tools that reveal what's actually going wrong: the \textbf{confusion matrix}, and \textbf{per-class precision and recall}.

\section{A 4-class, imbalanced dataset}

\begin{figure}[h]
\centering
\includegraphics[width=0.7\textwidth]{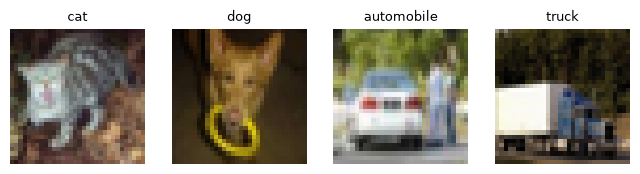}
\caption{One example from each of the four classes.}
\label{fig:l36-1}
\end{figure}

Four real photo categories from CIFAR-10: cat, dog, automobile, and truck (Figure~\ref{fig:l36-1}). The test set has an even 150 images per class, but the \emph{training} set is deliberately starved of trucks (90 examples, versus 400 each for the other three) --- a stand-in for the common real-world situation where some classes are just rarer to collect than others. Cat and dog are included deliberately as a visually similar pair, regardless of how much training data either gets.

\begin{lstlisting}
class CNN(nn.Module):
    def __init__(self):
        super().__init__()
        self.conv = nn.Sequential(
            nn.Conv2d(3, 16, 5, padding=2), nn.ReLU(), nn.MaxPool2d(2),
            nn.Conv2d(16, 32, 5, padding=2), nn.ReLU(), nn.MaxPool2d(2),
            nn.Conv2d(32, 64, 3, padding=1), nn.ReLU(),
            nn.AdaptiveMaxPool2d(1),
        )
        self.fc = nn.Linear(64, 4)
\end{lstlisting}

Trained for 150 epochs: overall test accuracy is $50.3\%$. Just over 50\% overall accuracy, on a 4-class task where chance is 25\%, sounds like a reasonable result for a small from-scratch CNN on real photos. It hides something important: the model is not equally good at all four classes.

\section{Confusion matrix}

Row $i$, column $j$ counts test examples of true class $i$ predicted as class $j$. A perfect classifier is diagonal; everything off the diagonal is a specific, nameable mistake.

\begin{center}
\renewcommand{\arraystretch}{1.3}
\begin{tabular}{lrrrr}
\hline
\textbf{true \textbackslash\ predicted} & \textbf{cat} & \textbf{dog} & \textbf{automobile} & \textbf{truck} \\
\hline
cat        & 60 & 68 & 20  & 2 \\
dog        & 47 & 95 & 7   & 1 \\
automobile & 9  & 5  & 134 & 2 \\
truck      & 21 & 13 & 103 & 13 \\
\hline
\end{tabular}
\end{center}

\begin{figure}[h]
\centering
\includegraphics[width=0.45\textwidth]{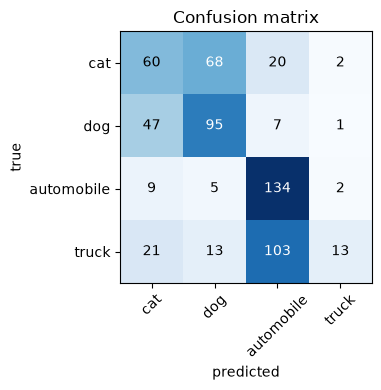}
\caption{The confusion matrix, visualized: rows are true class, columns are predicted class.}
\label{fig:l36-2}
\end{figure}

Figure~\ref{fig:l36-2} makes the pattern jump out immediately: truck's row is scattered everywhere except its own diagonal cell, while automobile's column absorbs a disproportionate share of every other row's mistakes.

\section{True positives, false positives, true negatives, false negatives}

Every cell of the confusion matrix above is a specific kind of correctness or mistake, but the standard vocabulary for talking about them is binary: pick one class and ask only ``is it this, or not?'' Collapsing the 4-class matrix down to ``truck vs.\ everything else'' gives exactly four outcomes:
\begin{itemize}
  \item \textbf{True positive (TP)}: actually a truck, predicted truck. A hit.
  \item \textbf{False negative (FN)}: actually a truck, predicted something else. A miss --- the model let it slip past.
  \item \textbf{False positive (FP)}: actually \emph{not} a truck, predicted truck anyway. A false alarm.
  \item \textbf{True negative (TN)}: actually not a truck, correctly predicted not-truck.
\end{itemize}
This is the same $2\times2$ table underlying every binary classifier's evaluation (a medical test's ``positive/negative'' result, a spam filter's ``spam/not spam'' decision) --- a multi-class confusion matrix is just this table computed once per class, with everything off that class's row/column collapsed into ``not this class.'' For truck: $\text{TP}=13$, $\text{FP}=5$, $\text{FN}=137$, $\text{TN}=445$ (total $600$, matching the test set size).

\section{Precision and recall}

Two numbers per class, computed directly from TP/FP/FN:
\begin{itemize}
  \item \textbf{Recall} $=\text{TP}/(\text{TP}+\text{FN})$ --- of everything that really \emph{was} class $i$, what fraction did the model catch? Low recall means the model misses that class often.
  \item \textbf{Precision} $=\text{TP}/(\text{TP}+\text{FP})$ --- of everything the model \emph{called} class $i$, what fraction actually was? Low precision means the model cries wolf on that class often.
\end{itemize}
(A less commonly needed but related pair, built from the other two quadrants: \textbf{specificity} $=\text{TN}/(\text{TN}+\text{FP})$, how well the model avoids false alarms on the negative class, and its complement the \textbf{false positive rate} $=\text{FP}/(\text{FP}+\text{TN})=1-\text{specificity}$.)

\begin{center}
\renewcommand{\arraystretch}{1.3}
\begin{tabular}{lrrr}
\hline
\textbf{class} & \textbf{precision} & \textbf{recall} & \textbf{support} \\
\hline
cat        & 0.44 & 0.40 & 150 \\
dog        & 0.52 & 0.63 & 150 \\
automobile & 0.51 & 0.89 & 150 \\
truck      & 0.72 & 0.09 & 150 \\
\hline
\end{tabular}
\end{center}

Truck --- the class starved to 90 training images, versus 400 for each other class --- has high precision but very low recall: when the model does say ``truck,'' it's usually right, but it fails to recognize the overwhelming majority of actual trucks, defaulting instead to whichever classes it saw plenty of during training (mostly ``automobile,'' the visually closest well-represented class). That is the standard signature of class imbalance, and it is completely invisible in the single overall-accuracy number from before. Cat and dog, by contrast, are both well-represented in training but get confused \emph{with each other} far more than with automobile or truck --- a different failure mode entirely, caused by genuine visual similarity between the two animals rather than by a lack of data. Automobile ends up with unusually high recall but only middling precision, for the same reason truck's recall suffered: it's absorbing the starved truck class's misclassifications, since ``wheeled vehicle'' is an easy fallback guess once the model can't tell trucks apart from automobiles.

The practical lesson: always inspect the confusion matrix and per-class metrics before trusting a single accuracy figure, especially on any dataset where classes aren't naturally balanced.

\subsection*{Exercises}
\begin{enumerate}
  \item Increase the truck training count from 90 to 400 (matching the other three classes) and rerun. Does truck recall recover? Does the cat/dog confusion also improve, or does it persist --- and why would more data not fix a problem caused by genuine visual similarity rather than data scarcity?
  \item Try weighting the loss by inverse class frequency (\code{F.cross\_entropy(logits, yt, weight=class\_weights)}, where \code{class\_weights[i] = 1 / count(class i)}) instead of collecting more truck images. Does it recover truck recall, and at what cost to the other classes' precision?
  \item Compute the \textbf{F1 score} (the harmonic mean of precision and recall, $2pr/(p+r)$) for each class. Why might F1 be a better single number to track per-class than accuracy, when accuracy is only meaningful in aggregate?
\end{enumerate}

\vfill
\fullcode{lesson36}{lesson36_image_classification_practice}

%% file: chapters/lesson37.tex
\chapter{Classic Architectures}
\label{ch:lesson37}

Chapters~\ref{ch:lesson34}--\ref{ch:lesson35} built and trained one small CNN. This chapter looks at how CNN \emph{architectures} evolved over roughly two decades, and works through the single biggest architectural idea in that history in detail: the \textbf{residual connection}, and the vanishing-gradient problem it was designed to fix.

\section{ImageNet and ILSVRC: the benchmark that drove this}

\textbf{ImageNet} (\citelink{deng2009}{Deng et al., 2009}) is a \emph{dataset}: over 14 million images, hand-labeled across more than 20,000 categories, built by crowdsourcing annotations onto WordNet's existing noun hierarchy. A big labeled dataset alone doesn't drive progress, though --- a shared, standardized \emph{competition} does. Starting in 2010, the \textbf{ILSVRC} (ImageNet Large Scale Visual Recognition Challenge, \citelink{russakovsky2015}{Russakovsky et al., 2015}) turned a 1000-category, $\sim$1.2-million-image subset of ImageNet into an annual benchmark: every group trained on the same data and was scored on the same held-out test set, making architectures directly, objectively comparable at a scale no one had managed before. The competition's winning top-5 error rate by year is a fair proxy for the field's progress across the decade this chapter covers:

\begin{center}
\renewcommand{\arraystretch}{1.3}
\begin{tabular}{lll}
\hline
\textbf{Year} & \textbf{Winner} & \textbf{Top-5 error} \\
\hline
2010 & (pre-deep-learning features)     & 28.2\% \\
2012 & AlexNet                          & 16.4\% \\
2014 & GoogLeNet (VGG close behind, at 7.3\%) & 6.7\% \\
2015 & ResNet                           & 3.57\% \\
2017 & SENet                            & 2.25\% \\
\hline
\end{tabular}
\end{center}

On this same task, a trained human achieves $\sim5\%$ error rate. ResNet was the first to beat this number --- sometimes referred to as ``super-human'' performance. By 2017, with the winning error under 2.5\%, ILSVRC's classification track was retired --- its job as a benchmark essentially done.

\section{A brief historical tour}

A handful of architectures were invented during this history, each solving a specific problem:
\begin{itemize}
  \item \textbf{LeNet-5} (\citelink{lecun1998}{LeCun et al., 1998}) --- the template Chapter~\ref{ch:lesson34} already implements: conv, pool, conv, pool, then fully-connected layers. Built for digit recognition on $32\times32$ images.
  \item \textbf{AlexNet} (\citelink{krizhevsky2012}{Krizhevsky, Sutskever \& Hinton, 2012}) --- the same template, just much bigger (5 conv layers, $\sim$60M parameters), trained on ImageNet using GPUs and ReLU instead of tanh/sigmoid. Its 2012 ILSVRC win is the watershed event that began the deep learning revolution in computer vision --- reviving mainstream interest in neural networks after a decade-long winter (akin to the ``AI winter'').
  \item \textbf{VGG} (\citelink{simonyanzisserman2014}{Simonyan \& Zisserman, 2015}) --- replaced AlexNet's large $11\times11$/$5\times5$ filters with a deep stack of small $3\times3$ convolutions. Two stacked $3\times3$ convs have the same \emph{receptive field} as one $5\times5$ conv (Chapter~\ref{ch:lesson11}'s pyramid idea again) but fewer parameters and an extra nonlinearity in between.
  \item \textbf{GoogLeNet / Inception} (\citelink{szegedy2015}{Szegedy et al., 2015}) --- went wider instead of just deeper: each ``Inception module'' runs several convolution sizes ($1\times1$, $3\times3$, $5\times5$) side by side on the same input and concatenates their outputs, so the network doesn't have to commit to one filter size per layer. $1\times1$ convolutions are used first to cheaply shrink the channel count, keeping the wider module affordable.
  \item \textbf{ResNet} (\citelink{he2015resnet}{He et al., 2016}) --- pushed depth from VGG's $\sim$19 layers to 50, 101, even 152, by solving the problem that had made naively stacking more layers \emph{worse}, not better.
  \item \textbf{SENet} (\citelink{hu2018senet}{Hu, Shen \& Sun, 2018}), the 2017 ILSVRC winner --- not a new backbone shape, but a refinement: its ``squeeze-and-excitation'' block lets each layer learn per-channel weights (which feature channels matter most for this input) rather than treating every channel equally.
\end{itemize}

The rest of this chapter focuses on ResNet's residual connection as the single most consequential idea in the table above.

\section{Building a ResNet block}

\subsection*{The vanishing-gradient problem}

Why does stacking more layers make plain networks \emph{worse}? Backprop's chain rule multiplies a gradient by every layer's local Jacobian on its way back to the input. If those per-layer factors are consistently smaller than 1 --- which happens easily with a saturating activation like sigmoid, whose derivative is at most 0.25 --- a deep enough stack multiplies the gradient by a very small number many times over. The gradient reaching early layers shrinks toward zero, and those layers stop learning at all.

Build a 30-layer, sigmoid-activated network of \code{Linear} layers and track the gradient magnitude at every depth, for two versions: a \textbf{plain} stack \code{x = sigmoid(layer(x))}, and a \textbf{residual} stack \code{x = x + 0.3 * sigmoid(layer(x))} where each layer only has to learn a small \emph{correction} added to its input, rather than replacing it outright.

\begin{lstlisting}
class ResidualDeepNet(nn.Module):
    def forward(self, x):
        for layer in self.layers:
            x = x + 0.3 * torch.sigmoid(layer(x))
        return x
\end{lstlisting}

\begin{figure}[h]
\centering
\includegraphics[width=0.6\textwidth]{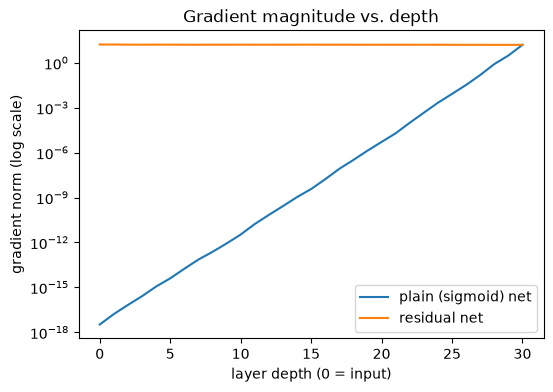}
\caption{Gradient magnitude vs.\ depth, plain sigmoid network vs.\ residual network, log scale.}
\label{fig:l37-1}
\end{figure}

For a 30-layer stack: the plain network's ratio of gradient norm at the input layer to the last layer is $2.07\times10^{-19}$; the residual network's ratio is $1.05\times10^0$ --- essentially flat (Figure~\ref{fig:l37-1}).

The plain network's gradient shrinks by roughly nineteen orders of magnitude between the last layer and the first --- for all practical purposes, the early layers receive no learning signal at all. The residual network's (ResNet's) gradient stays essentially flat across all 30 layers.

The reason is structural, not a matter of tuning: with a residual connection, $x_{l+1}=x_l+F(x_l)$, so $dx_{l+1}/dx_l=I+dF/dx_l$. Backprop multiplies these Jacobians together across layers, but every one of them contains an identity matrix $I$ as a direct term. That gives the gradient a path straight back to the input that never gets multiplied by a small sigmoid derivative --- an unobstructed shortcut, regardless of how deep the stack gets. The plain network has no such path: every layer's Jacobian is purely $dF/dx_l$, so there's nothing to stop repeated multiplication from driving the product toward zero.

\subsection*{Batch normalization}

ResNets use one more ingredient: \textbf{batch normalization} (\citelink{ioffeszegedy2015}{Ioffe \& Szegedy, 2015}). The idea is simple --- for each channel, subtract that channel's mean and divide by its standard deviation, computed \emph{across the current mini-batch} (over the batch, height, and width dimensions, separately per channel), so every channel's activations always have mean 0 and standard deviation 1 going into the next layer. Two learnable parameters, a per-channel scale $\gamma$ and shift $\beta$, are then applied on top, so the layer can still recover a different mean/scale if that's actually useful --- normalization to 0/1 is a \emph{starting point} the network can undo, not a hard constraint.

Deep networks without batch norm are prone to a related but distinct problem from vanishing gradients: as training updates early layers, the \emph{distribution} of activations feeding into later layers keeps shifting (sometimes called ``internal covariate shift''), so later layers are constantly chasing a moving target. Renormalizing at every layer keeps that distribution stable, which in practice lets much higher learning rates be used and makes deep networks noticeably easier to train --- one of the reasons ResNet could push to 50+ layers where earlier architectures struggled past $\sim$20.

A from-scratch per-channel normalize-then-scale-and-shift matches \code{nn.BatchNorm2d} to $4.77\times10^{-7}$: per-channel means and standard deviations that started at, e.g., $[-0.046, 0.165, -0.061, -0.165]$ and $[0.939, 1.037, 0.911, 0.977]$ come out at exactly $[0,0,0,0]$ and $[1,1,1,1]$ after normalization.

\subsection*{Putting it together: a convolutional residual block}

In an actual ResNet, $F$ is a pair of small convolutions (with batch normalization and ReLU in between), and the shortcut --- also called a \textbf{skip connection}, since it skips over $F$ entirely --- adds the \emph{input feature map} back onto their output, channel-for-channel and pixel-for-pixel:

\begin{lstlisting}
class ResidualBlock(nn.Module):
    def __init__(self, channels):
        super().__init__()
        self.conv1 = nn.Conv2d(channels, channels, 3, padding=1)
        self.bn1 = nn.BatchNorm2d(channels)
        self.conv2 = nn.Conv2d(channels, channels, 3, padding=1)
        self.bn2 = nn.BatchNorm2d(channels)

    def forward(self, x):
        out = torch.relu(self.bn1(self.conv1(x)))
        out = self.bn2(self.conv2(out))
        return torch.relu(x + out)  # the shortcut: add the block's input back in
\end{lstlisting}

Passing a $(4,16,20,20)$ tensor through the block produces an output of the identical shape $(4,16,20,20)$ --- a residual block preserves shape, so blocks can be stacked freely.

\subsection*{Exercises}
\begin{enumerate}
  \item In \code{PlainDeepNet}/\code{ResidualDeepNet} above, change the activation from \code{torch.sigmoid} to \code{torch.relu} (ReLU's derivative is 1 for any positive input, not capped at 0.25 like sigmoid's) and rerun the gradient-norm comparison. Does the plain network's vanishing problem get better, worse, or stay about the same?
  \item Change the residual net's scale factor from \code{0.3} to \code{1.0} (i.e.\ \code{x + torch.sigmoid(layer(x))} with no damping) and rerun. Does the gradient ratio change much? What does that suggest about \emph{why} the residual connection works --- is it the scale factor, or the \code{+ x} shortcut itself?
  \item \code{ResidualBlock} above requires the input and output to have the same number of channels, since they're added directly. Real ResNets sometimes need to change channel count between blocks (e.g.\ 16 $\to$ 32). Sketch (in words, or in code) what the shortcut path would need to do in that case for the addition to still make sense.
\end{enumerate}

\vfill
\fullcode{lesson37}{lesson37_classic_architectures}

%% file: chapters/lesson38.tex
\chapter{Transfer Learning}
\label{ch:lesson38}

Training a good CNN from scratch (Chapters~\ref{ch:lesson34}--\ref{ch:lesson35}) took hundreds of labeled examples even for a toy task. Real target tasks are often data-starved: a handful of labeled medical scans, a new product category with 20 photos. \textbf{Transfer learning} sidesteps this by reusing a network already trained on a \emph{different}, data-rich task, on the theory that early-layer features (edges, blobs, simple textures) are useful for almost any visual task, not just the one they were originally trained on. This chapter demonstrates it on real photos, using a CIFAR-10 subset (\citelink{krizhevsky2009cifar}{Krizhevsky, 2009}).

\section{Source task and target task}

\begin{figure}[h]
\centering
\includegraphics[width=0.9\textwidth]{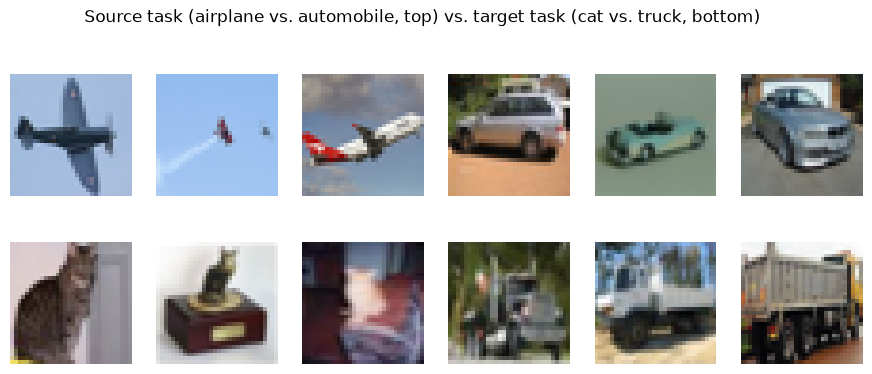}
\caption{Sample target-task images: cats (top row) and trucks (bottom row).}
\label{fig:l38-1}
\end{figure}

Set up two related but distinct tasks, both from CIFAR-10. The \textbf{source task} has plenty of data: distinguishing airplanes from automobiles, 300 training images each. The \textbf{target task} is the one we actually care about, and it's deliberately starved: distinguishing cats from trucks --- two classes the source task never saw at all --- with only \textbf{15 training images per class} (Figure~\ref{fig:l38-1}).

\section{Three strategies}

\begin{enumerate}
  \item \textbf{From scratch} --- train a fresh CNN on only the 30 target images. This is the baseline: no transfer at all.
  \item \textbf{Frozen backbone (feature extraction)} --- pretrain a CNN backbone on the source task, then freeze its weights entirely and train only a new linear classifier on top of the features it produces for target images.
  \item \textbf{Fine-tuning} --- start from the same pretrained backbone, but keep updating it on the target data too, using a \emph{much smaller} learning rate for the backbone than for the new classifier head (the backbone already encodes useful structure; large updates from just 30 examples would wreck it).
\end{enumerate}

\begin{lstlisting}
def train_target(backbone, freeze_backbone, seed, epochs=200, lr=0.01, backbone_lr=None):
    model = Classifier(backbone, freeze_backbone)
    if freeze_backbone:
        opt = torch.optim.Adam(model.fc.parameters(), lr=lr)
    elif backbone_lr is not None:
        opt = torch.optim.Adam([
            {'params': model.backbone.parameters(), 'lr': backbone_lr},
            {'params': model.fc.parameters(), 'lr': lr},
        ])
    else:
        opt = torch.optim.Adam(model.parameters(), lr=lr)
\end{lstlisting}

Pretraining a small CNN backbone once on the source task, then evaluating all three strategies on the target task, averaged over 8 random seeds:

\begin{center}
\renewcommand{\arraystretch}{1.3}
\begin{tabular}{lrr}
\hline
\textbf{strategy} & \textbf{mean test acc} & \textbf{std} \\
\hline
from scratch    & 63.6\% & 6.8\% \\
frozen backbone & 69.5\% & 1.0\% \\
fine-tuned      & 76.7\% & 1.2\% \\
\hline
\end{tabular}
\end{center}

Both transfer strategies clearly beat training from scratch, and they're also dramatically more \emph{consistent} (lower standard deviation). Even though the source task never saw a cat or a truck, it learned generic low-level features (edges, color blobs, textures) that transferred to the new task.

Note that fine-tuning used a learning rate for the backbone roughly $30\times$ smaller than the classifier head's. With just 30 target examples, a larger backbone update would overfit those 30 images, discarding everything useful it had learned from the 600-image source task. This is the same catastrophic-forgetting failure mode as Chapter~\ref{ch:lesson35}'s overfitting curve, just applied to a network that started out already knowing something.

\section{In practice: pretrained backbones in one line}

The toy backbone above is trained from scratch on a toy task for illustrative purposes, but real transfer learning leverages existing networks available for reuse. \code{torchvision} ships architectures already pretrained on the full ImageNet dataset (Chapter~\ref{ch:lesson37}) --- 1.2 million images across 1,000 classes. Loading such a network, ready to use as a frozen feature extractor, is a single line of code:

\begin{lstlisting}
resnet = torchvision.models.resnet18(weights=torchvision.models.ResNet18_Weights.IMAGENET1K_V1)
resnet.fc = nn.Identity()  # strip the original 1000-class ImageNet head, keep the 512-d features
resnet.eval()
for p in resnet.parameters():
    p.requires_grad = False  # frozen, feature-extraction style
\end{lstlisting}

Total ResNet-18 parameters: $11{,}689{,}512$, all pretrained on ImageNet, none of them touched here.

ResNet-18 was trained on $224\times224$ images normalized by ImageNet's per-channel mean and standard deviation --- not the plain $[0,1]$ pixel scaling used everywhere else in this chapter --- so the target images need to be resized and renormalized before going through it. Extracting a 512-dimensional feature vector for every target image once (the backbone is frozen, so this only needs to happen once, not on every training step), then training a small linear head on top --- exactly the frozen-backbone strategy above, just with a very different backbone --- reaches $98.1\%\pm0.2\%$ mean test accuracy over 8 seeds, versus $69.5\%$ for the toy frozen backbone above.

Frozen ImageNet features push accuracy well past either toy strategy above, using nothing but a new linear layer trained on the same 30 target images --- no backbone training at all.

\textbf{A caveat on this comparison:} some of that gap is an unfair advantage, not just richer pretraining. ImageNet's 1,000 classes already include photos of cats and several vehicle types, unlike the toy source task, which only saw airplanes and automobiles --- so this isn't a perfectly controlled comparison of ``big general pretraining'' versus ``tiny 2-class pretraining'' in isolation; part of ResNet-18's edge comes from having directly seen the target classes during pretraining.

Nevertheless, the qualitative point holds, and it's the actual reason ImageNet-pretrained backbones are the default starting point in practice: a network trained on 1.2 million images learns far richer, more broadly transferable features than one trained on a couple thousand images of two categories ever could --- which is exactly what Chapter~\ref{ch:lesson37}'s architectures, and the vastly larger datasets they were trained on, are built to produce.

\subsection*{Exercises}
\begin{enumerate}
  \item Try \code{backbone\_lr=0.01} (i.e.\ no learning-rate difference between backbone and head) in the fine-tuning call. Does fine-tuned accuracy get better or worse, and does that match the catastrophic-forgetting explanation above?
  \item Try freezing \emph{most} of the backbone but fine-tuning only its last conv layer. Where does that land relative to fully-frozen and fully-fine-tuned?
  \item Swap the target task's classes from cat/truck to dog/automobile --- a target pair that overlaps the source task's \emph{domain} (vehicles) even more closely than cat/truck did. Does frozen-backbone accuracy improve further, and does that match the intuition that transfer works best when source and target share more underlying visual structure?
  \item Try unfreezing just ResNet-18's last residual stage (\code{resnet.layer4}) and fine-tuning it on the 30 target images at a small learning rate, the same way the toy backbone was fine-tuned above. Does it beat the fully-frozen ResNet-18 result, or is 30 images too few to safely update even a small fraction of 11 million parameters?
\end{enumerate}

\vfill
\fullcode{lesson38}{lesson38_transfer_learning}

%% file: chapters/lesson39.tex
\chapter{Visualizing and Understanding CNNs}
\label{ch:lesson39}

A trained CNN is a black box because its millions of weights don't have obvious individual meanings. But \emph{where in the input image} a prediction comes from is answerable, and answering it is often what separates ``the model got the right answer'' from ``the model got the right answer for the right reason.'' This chapter builds two visualization tools from scratch --- \textbf{saliency maps} (\citelink{simonyan2013saliency}{Simonyan et al., 2014}) and \textbf{Grad-CAM} (\citelink{selvaraju2017}{Selvaraju et al., 2017}) --- on real photos from CIFAR-10 (\citelink{krizhevsky2009cifar}{Krizhevsky, 2009}), then uses them to catch a model that's cheating. Finally, \textbf{t-SNE} (\citelink{vandermaaten2008tsne}{Van der Maaten \& Hinton, 2008}) is used to visualize the learned feature vectors.

\section{Setup: a cat-vs-automobile CNN, with feature maps exposed}

A CIFAR-10 binary task: cat vs.\ automobile, 300 training images per class. The architecture is Chapter~\ref{ch:lesson34}'s CNN pattern, except \code{forward} now also returns the last convolutional layer's feature map (before global pooling), so both visualization methods have access to it. Trained for 300 epochs: test accuracy $86.5\%$.

\section{Saliency maps}

The idea of saliency maps (\citelink{simonyan2013saliency}{Simonyan et al., 2014}): take the gradient of the predicted class \emph{score} with respect to every input pixel. A pixel with a large-magnitude gradient is one where a small change would most change the prediction --- i.e., a pixel the network is ``looking at.''

\begin{lstlisting}
def saliency_map(model, img_hw3):
    x = torch.tensor(img_hw3).permute(2, 0, 1).unsqueeze(0)
    x.requires_grad_(True)
    score, feat = model(x)
    score.backward()
    return x.grad[0].abs().amax(dim=0).numpy(), feat  # max abs gradient across 3 color channels
\end{lstlisting}

\begin{figure}[h]
\centering
\includegraphics[width=0.55\textwidth]{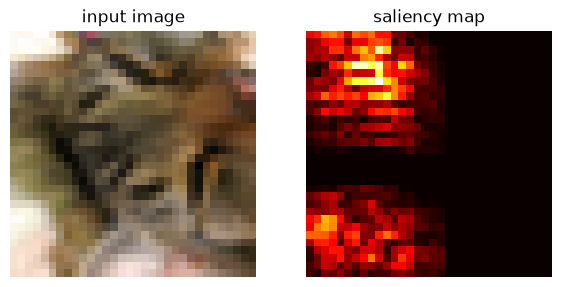}
\caption{An input image and its saliency map --- brightest where the gradient of the class score with respect to that pixel is largest.}
\label{fig:l39-1}
\end{figure}

\section{Grad-CAM}

\begin{figure}[h]
\centering
\includegraphics[width=0.85\textwidth]{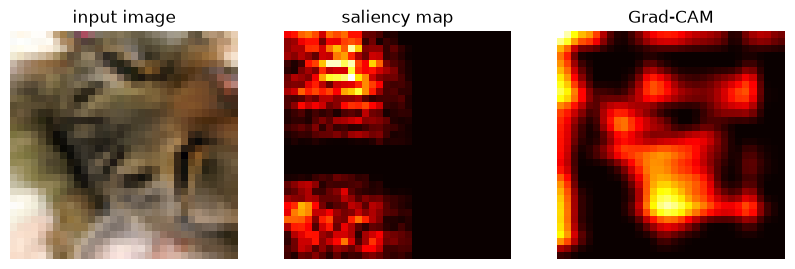}
\caption{Input image, saliency map, and Grad-CAM heatmap, side by side.}
\label{fig:l39-2}
\end{figure}

Raw saliency maps (Figure~\ref{fig:l39-1}) are pixel-level and tend to be noisy. \textbf{Grad-CAM} instead operates on the feature maps of a chosen convolutional layer, typically the last convolutional layer. Although these feature maps are lower-resolution than a pixel-level saliency map (downsampled by the pooling layers before them), they tend to carry more class-discriminative semantic information than individual raw-pixel gradients do. The method:
\begin{enumerate}
  \item Compute the gradient of the target class score with respect to each feature-map channel.
  \item Globally average each channel's gradients over its spatial dimensions to obtain one importance weight per channel.
  \item Take the weighted sum of the feature-map channels and apply ReLU, retaining the positive contributions to the target class.
\end{enumerate}
The resulting coarse heatmap has the spatial resolution of the selected convolutional layer; it can be upsampled to the input resolution for visualization (Figure~\ref{fig:l39-2}).

\begin{lstlisting}
def grad_cam(model, img_hw3, out_size=32):
    x = torch.tensor(img_hw3).permute(2, 0, 1).unsqueeze(0)
    x.requires_grad_(True)
    score, feat = model(x)
    feat.retain_grad()
    score.backward()
    weights = feat.grad[0].mean(dim=(1, 2))  # importance per channel
    cam = F.relu((weights[:, None, None] * feat[0]).sum(dim=0))
    return F.interpolate(cam[None, None], size=(out_size, out_size), mode='bilinear')[0, 0]
\end{lstlisting}

\section{Do these maps actually track what the model uses?}

To test whether these tools are useful, deliberately give the model a shortcut, and check whether they correctly catch it. Add a small, unmistakable $5\times5$ white marker to the top-left corner of every \textbf{automobile} training image only (never on cat images) --- a stand-in for a real-world confound, like a watermark, a lab-specific artifact, or a capture-device quirk that happens to correlate with one class. Train a second model, \code{model\_shortcut}, on this corrupted dataset, and compare it against the original \code{model\_original} above (never exposed to any marker):

\begin{center}
\renewcommand{\arraystretch}{1.3}
\begin{tabular}{lrr}
\hline
 & \textbf{clean test} & \textbf{marked test} \\
\hline
model\_original & 86.5\% & 83.5\% \\
model\_shortcut & 63.0\% & 98.0\% \\
\hline
\end{tabular}
\end{center}

\code{model\_original}'s accuracy barely moves whether the marker is present or not --- since it never learned to use it, it has nothing to lose when it's absent. \code{model\_shortcut} looks \emph{better} than \code{model\_original} when the marker is present, but it collapses when the marker is removed. It learned the shortcut ``white corner square = automobile.''

Now use Grad-CAM to check whether the visualization tool actually catches this. Since the marker's location is known exactly (rows 0--4, columns 0--4), checking becomes: what fraction of test images does each model's heatmap peak land inside that exact $5\times5$ region?

\begin{center}
\renewcommand{\arraystretch}{1.3}
\begin{tabular}{llrr}
\hline
\textbf{model} & \textbf{test images} & \textbf{saliency} & \textbf{Grad-CAM} \\
\hline
model\_original & marked & 0.0\%   & 0.0\% \\
model\_shortcut & marked & 96.0\%  & 100.0\% \\
model\_shortcut & clean  & 56.0\%  & 92.0\% \\
\hline
\end{tabular}
\end{center}

\code{model\_original} (never trained on the marker) rarely lands either method's peak in that exact corner. \code{model\_shortcut} is a different story: on marked images both tools are nearly perfect (saliency 96\%, Grad-CAM 100\%), but on \emph{clean} images --- where the marker was never added, yet \code{model\_shortcut} still relies on it internally --- Grad-CAM's peak still lands there 92\% of the time versus only 56\% for saliency. Grad-CAM pools gradients over an entire feature-map channel before localizing, which smooths out pixel-level noise and makes it the more trustworthy detector when the shortcut's literal trigger isn't visible in a given image; \code{model\_shortcut}'s internal machinery is anchored to that spot regardless of what's actually there, exactly the failure mode described at the top of this chapter.

The practical lesson: accuracy alone (Chapter~\ref{ch:lesson36}) can't distinguish whether the model has learned the real signal or a shortcut correlated with it in the training data. Visualization tools such as saliency and Grad-CAM can --- but only if you know to check, if you pick a tool sturdy enough to trust, and if you interpret the results with some skepticism about what else in the image might be drawing gradient attention.

\section{Feature-space visualization: t-SNE}

Saliency maps and Grad-CAM both answer a \emph{spatial} question about one image at a time: where is the network looking? \textbf{t-SNE} (\citelink{vandermaaten2008tsne}{van der Maaten \& Hinton, 2008}) asks a broader question across many images: does the network's internal representation separate the classes it was trained to recognize? It takes each image's 16-d feature vector --- the representation fed directly to the classifier --- and projects these vectors into 2D while preserving local neighborhoods. Unlike PCA (Chapter~\ref{ch:lesson06}), which uses a linear projection, t-SNE emphasizes local structure, making it useful for checking whether images from the same class cluster together in feature space without using their labels during the projection.

\begin{lstlisting}
def tsne(X, n_iter=500, perplexity=15.0, lr=100.0, seed=0):
    # binary-search a per-point Gaussian bandwidth to hit the target perplexity,
    # then fit a 2D embedding Y by gradient descent on KL(P || Q),
    # with Q built from a heavy-tailed Student-t kernel to avoid crowding
    ...
    return Y
\end{lstlisting}

\begin{figure}[h]
\centering
\includegraphics[width=0.9\textwidth]{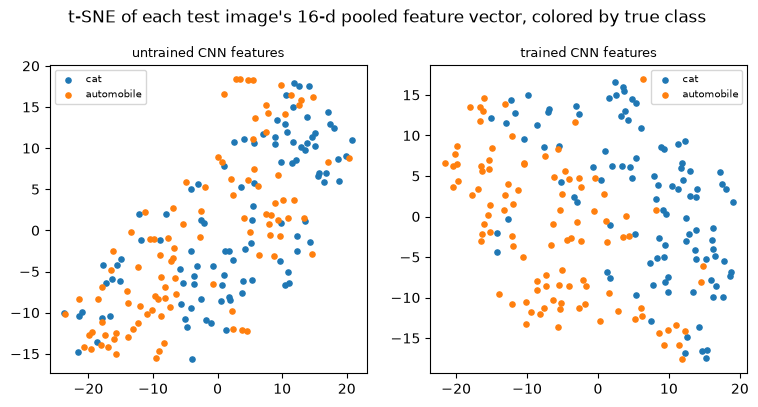}
\caption{t-SNE of each test image's 16-d pooled feature vector, colored by true class: untrained CNN features (left) vs.\ trained CNN features (right).}
\label{fig:l39-3}
\end{figure}

The untrained network's features scatter the two classes together with no visible structure (Figure~\ref{fig:l39-3}) --- unsurprising, since its convolutional weights are still random and have never seen a single labeled example. The trained network's features form two visibly separated clumps, cat and automobile, even though t-SNE itself was never told which point belonged to which class; it only saw the 16-d feature vectors and their pairwise distances. (Quantitatively, a simple 1-nearest-neighbor check in the 2D embedding --- does each point's nearest neighbor share its true label? --- reveals 78\% for the trained features versus about 62\% for the untrained ones.) This is evidence that training didn't just adjust the final linear layer --- it reshaped the whole feature space so that cats and automobiles genuinely live in different neighborhoods of it.

\subsection*{Exercises}
\begin{enumerate}
  \item Grad-CAM here uses the \emph{last} conv layer. Modify \code{grad\_cam} to instead use the intermediate feature map after \code{conv1} (before the pool and \code{conv2}). Does the resulting heatmap get sharper (closer to pixel-perfect, like the saliency map) or coarser, and why would an earlier layer behave that way?
  \item Shrink the marker from $5\times5$ to $2\times2$, or dim it from pure white ($1.0$) to a faint gray ($0.6$). Does \code{model\_shortcut} still learn to rely on it as strongly (check the clean-vs-marked accuracy gap), and does Grad-CAM still catch it as reliably?
  \item The saliency-map gradient in this chapter is taken with respect to the raw logit (\code{score}), not the sigmoid probability. Try computing it with respect to \code{torch.sigmoid(score)} instead --- does the resulting map look meaningfully different, and can you explain why using the chain rule?
  \item Color the trained-feature t-SNE plot by \emph{predicted} label instead of true label. Do the handful of points that land on the ``wrong side'' of the cluster boundary correspond to images the model actually misclassifies? Then repeat with \code{model\_shortcut}'s features on marked vs.\ clean images --- does removing the marker collapse the clean class separation the way the earlier accuracy numbers predicted it would?
\end{enumerate}

\vfill
\fullcode{lesson39}{lesson39_visualizing_cnns}

%% file: chapters/lesson40.tex
\chapter{Object Detection from Sliding Windows}
\label{ch:lesson40}

Every classifier so far has focused on images as a whole rather than image patches --- asking ``what \emph{is} this image?,'' assuming it already contains exactly one thing, centered and cropped. Object \textbf{detection} asks a harder question: given an image that contains zero, one, or several objects at unknown locations and scales, find the \emph{location} of each one.

The classic approach is the \textbf{sliding window}: train a classifier for a fixed-size image patch, then slide the classifier across the image to run it at every location (and scale). \citelink{rowleybalujakanade1996}{Rowley, Baluja, and Kanade (1996)} did exactly this with a small neural network as the window classifier. This idea was truly ahead of its time --- one of the first successful uses of a neural net for a real vision task, more than a decade before deep learning's resurgence (Chapter~\ref{ch:lesson37}). This chapter builds the approach end to end: a small window classifier, applied at every position, followed by a key step that any real system needs --- merging duplicate detections.

\section{Step 1: a synthetic ``face'' and a window classifier}

\begin{figure}[h]
\centering
\includegraphics[width=0.85\textwidth]{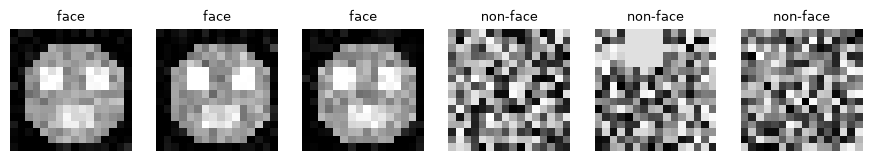}
\caption{Synthetic face examples vs.\ non-face clutter.}
\label{fig:l40-1}
\end{figure}

To demonstrate the mechanics, real data isn't needed --- only a class with consistent internal structure (a head outline, two eyes, a mouth, always in the same relative arrangement) versus clutter that lacks that structure (Figure~\ref{fig:l40-1}). Train a small CNN as a pure window classifier: given a fixed-size crop, is there a face filling it, yes or no?

\begin{lstlisting}
class WindowClassifier(nn.Module):
    def __init__(self):
        super().__init__()
        self.net = nn.Sequential(
            nn.Conv2d(1, 8, 3, padding=1), nn.ReLU(), nn.MaxPool2d(2),
            nn.Conv2d(8, 16, 3, padding=1), nn.ReLU(), nn.AdaptiveMaxPool2d(1),
        )
        self.fc = nn.Linear(16, 1)
\end{lstlisting}

Trained for 300 epochs: window classifier test accuracy $100.0\%$.

\section{Step 2: slide the window across a scene}

\begin{figure}[h]
\centering
\includegraphics[width=0.9\textwidth]{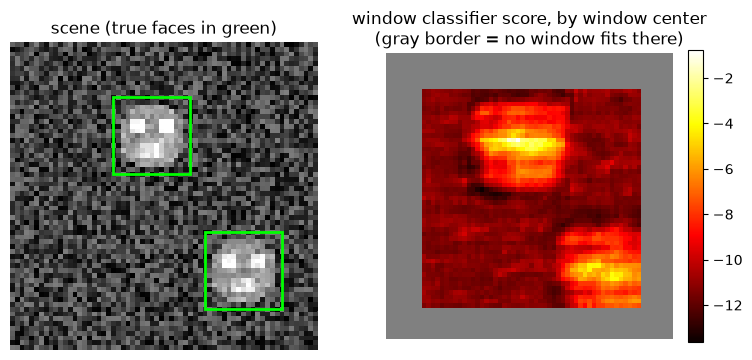}
\caption{The scene (true faces outlined in green) and the window classifier's score map, indexed by window center (gray border = no window fits there).}
\label{fig:l40-2}
\end{figure}

Build a larger scene containing two faces at unknown locations plus background clutter (true face centers at $(29.0,19.0)$ and $(48.0,47.0)$), then run the trained window classifier at every position on a dense grid (a \textbf{sliding window}). Each position gets a raw confidence score, which is recorded at that window's center. Since windows can't be centered within \code{win // 2} pixels of the border (a window close to the edge doesn't fit inside the image), the score is not computed within a border strip, shown in gray in Figure~\ref{fig:l40-2}.

\begin{lstlisting}
def sliding_window_scores(model, scene, win=16, stride=1):
    half = win // 2
    scores = np.full(scene.shape, np.nan, dtype=np.float32)
    for y0 in range(0, scene.shape[0] - win + 1, stride):
        for x0 in range(0, scene.shape[1] - win + 1, stride):
            patch = scene[y0:y0+win, x0:x0+win]
            scores[y0 + half, x0 + half] = model(torch.tensor(patch[None, None]).float()).item()
    return scores
\end{lstlisting}

\section{Step 3: threshold, then merge duplicates}

The score map peaks near the true faces, but thresholding produces a \emph{cluster} of detections around each face --- every window that overlaps a face heavily enough scores above threshold. This is precisely the duplicate-detection problem Chapter~\ref{ch:lesson12} first raised for both Canny and the Hough transform: many near-identical hypotheses need to collapse into one. The fix is \textbf{non-maximum suppression (NMS)}: repeatedly keep the highest-scoring remaining detection and discard every other detection that overlaps it by more than an IoU (intersection-over-union) threshold.

\begin{lstlisting}
def nms(detections, iou_thresh=0.3):
    dets = sorted(detections, key=lambda d: -d[4])
    keep = []
    while dets:
        best = dets.pop(0)
        keep.append(best)
        dets = [d for d in dets if iou(best, d) < iou_thresh]
    return keep
\end{lstlisting}

\begin{figure}[h]
\centering
\includegraphics[width=0.5\textwidth]{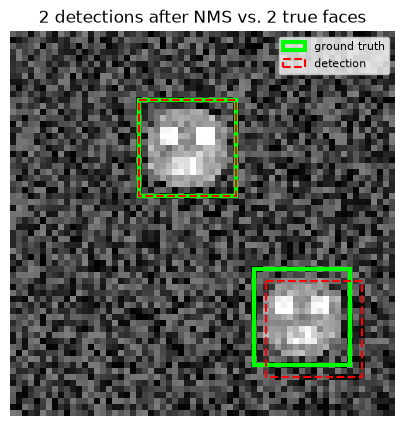}
\caption{2 detections after NMS (red, dashed) vs.\ 2 true faces (green).}
\label{fig:l40-3}
\end{figure}

Using a threshold set at the 97.5th percentile of all window scores ($-5.14$): 60 raw detections lie above threshold; NMS collapses these to just 2, landing at $(cx{=}29,cy{=}19)$ and $(cx{=}50,cy{=}49)$ --- one exactly on a true center, the other within a couple of pixels (Figure~\ref{fig:l40-3}).

With this threshold and \code{iou\_thresh=0.3}, NMS collapses the raw detections down to two boxes that closely match the two true face locations. As the exercises below explore, results on other scenes and seeds yield false positives and false negatives, depending on the threshold and other parameter choices.

\section{Handling scale: the image pyramid}

Everything above assumes faces are always exactly $16\times16$, but real faces appear at unknown scales. The solution is Chapter~\ref{ch:lesson11}'s Gaussian pyramid (\code{cv2.pyrDown}): build a stack of the image at multiple resolutions, and run the \emph{same fixed-size} window classifier over every level. A face that's too big for the window at full resolution will fit the window at some coarser pyramid level, since shrinking the image is equivalent to enlarging the effective window size relative to image content.

\section{Viola-Jones}

The Rowley-Baluja-Kanade window classifier (what this chapter just built, in miniature) was computationally heavy at the time. In 1996 (pre-GPUs) evaluating a small neural network at every position and scale of every pyramid level was slow. \citelink{violajones2001}{Viola and Jones (2001)} modified the same sliding-window idea to run in real time by replacing the neural network with a \textbf{cascade} of extremely cheap Haar-like features (the same local sum/difference idea as Chapter~\ref{ch:lesson16}'s Haar wavelet, applied to 2D rectangular regions): a sequence of stages, each a simple threshold on a rectangular-region intensity difference, ordered so that the vast majority of non-face windows get rejected by the \emph{first} stage or two, and only the rare promising windows pay for the full cascade of classifiers. As a result, the cascade's average cost per window is tiny --- which is why Viola-Jones is the algorithm that ended up running live on 2000s-era digital cameras.

\begin{figure}[h]
\centering
\includegraphics[width=0.75\textwidth]{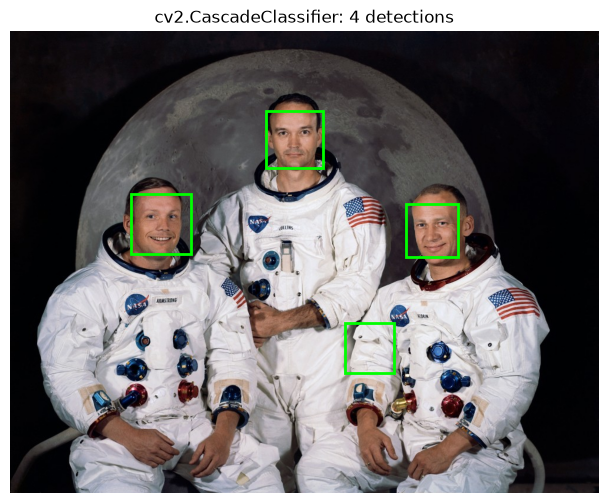}
\caption{\code{cv2.CascadeClassifier} on a real photo: 4 detections. \attrib{Image source: NASA (public domain), via Wikimedia Commons.}}
\label{fig:l40-4}
\end{figure}

OpenCV ships the Viola-Jones cascade as \code{cv2.CascadeClassifier}, pretrained on real faces. Just load an XML file of learned cascade stages and run it. On a real photo of the Apollo 11 crew (Figure~\ref{fig:l40-4}): 4 detections.

All three astronauts' faces (Armstrong, Collins, Aldrin) are found correctly, plus a small false positive on the fabric texture on Aldrin's sleeve. \code{minNeighbors} (how many overlapping detections a region must have) and \code{scaleFactor} (how coarsely the image pyramid steps between scales) trade off false positives against missed faces, playing a role similar to this chapter's NMS threshold.

\subsection*{Exercises}
\begin{enumerate}
  \item Change \code{iou\_thresh} in \code{nms} from $0.3$ to $0.7$. Rerun detection on the scene. Does NMS now under-merge (report more than 2 boxes) or over-merge (miss a face)? Explain why in terms of how much overlap real duplicate detections around the same face actually have.
  \item The threshold here is set from the 97.5th percentile of \emph{this scene's own} score distribution --- a form of cheating, since a real detector doesn't get to see the test scene's scores before deciding. Instead, compute a threshold from the training set's known face/non-face scores only, and check whether it still successfully detects both faces in the scene.
  \item Increase \code{n\_faces} in \code{make\_scene} to 4 and shrink \code{size} to 48 so faces are packed closer together. Does NMS still separate them correctly, or does IoU-based suppression start merging genuinely distinct nearby faces into one detection? At what spacing does it break down?
  \item Lower \code{minNeighbors} on the real Viola-Jones cascade from $5$ to $2$ and rerun. Does the fabric false positive get worse, or do new false positives appear elsewhere? Then try raising \code{minNeighbors} to $8$ --- does it clean up the false positive, and does it cost any true detections?
\end{enumerate}

\vfill
\fullcode{lesson40}{lesson40_object_detection_sliding_windows}

%% file: chapters/lesson41.tex
\chapter{Object Detection as Regression}
\label{ch:lesson41}

Chapter~\ref{ch:lesson40}'s sliding-window detector classified thousands of fixed-size windows and merged the results with non-maximum suppression (NMS). That works, but it is fundamentally a \emph{classification} approach bolted onto a search --- the network never predicts a box directly, only ``face or not, at this exact window.'' Modern detectors instead treat localization as a \textbf{regression} problem: given an image, directly predict bounding box coordinates. This chapter builds the simplest possible version of that idea, then surveys how real detectors (R-CNN, YOLO, SSD) scale it up.

\section{One object per image: predict a box directly}

\begin{figure}[h]
\centering
\includegraphics[width=0.85\textwidth]{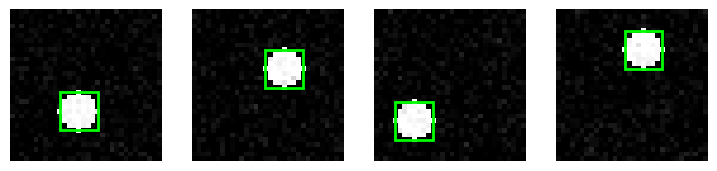}
\caption{Sample training scenes, each with a single object and its ground-truth box.}
\label{fig:l41-1}
\end{figure}

The simplest possible regression problem: one circular blob per image, at an unknown location and size (Figure~\ref{fig:l41-1}). Instead of a class label, the network's target is now four numbers --- $(c_x, c_y, w, h)$ indicating the bounding box, normalized to $[0,1]$ by image size.

\section{The model and the IoU metric}

The network is a CNN backbone (Chapter~\ref{ch:lesson34}'s pattern) followed by a 4-output regression head with a sigmoid, so every prediction lands in $[0,1]$ --- a valid normalized box coordinate. It's trained with plain MSE loss against the true box, and evaluated with \textbf{IoU} (Chapter~\ref{ch:lesson40}'s intersection-over-union), the metric that actually matters for detection: how much the predicted and true boxes overlap, not how close the four numbers are in isolation.

\begin{lstlisting}
class Detector(nn.Module):
    def __init__(self):
        super().__init__()
        self.conv = nn.Sequential(
            nn.Conv2d(1, 16, 5, padding=2), nn.ReLU(), nn.MaxPool2d(2),
            nn.Conv2d(16, 32, 5, padding=2), nn.ReLU(), nn.AdaptiveMaxPool2d(1),
        )
        self.fc = nn.Linear(32, 4)  # cx, cy, w, h, normalized

    def forward(self, x):
        return torch.sigmoid(self.fc(self.conv(x).flatten(1)))
\end{lstlisting}

\begin{figure}[h]
\centering
\includegraphics[width=0.85\textwidth]{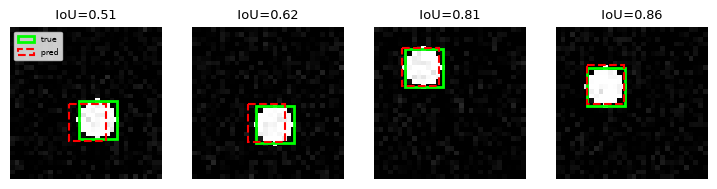}
\caption{Predicted boxes (red, dashed) vs.\ true boxes (green) on four test scenes, with per-example IoU.}
\label{fig:l41-2}
\end{figure}

Trained for 400 epochs with MSE loss: mean IoU on the test set is $0.758$, with $95.0\%$ of test boxes reaching $\text{IoU}>0.5$ (Figure~\ref{fig:l41-2}).

\section{Scaling this up: two families of real detectors}

This detector above only handles exactly one object per image, because a fixed-size output vector (4 numbers) can only describe one box. But real scenes have a variable, unknown number of objects. Two different families of detectors dominate the landscape:

\textbf{Two-stage (R-CNN family):} first generate a modest number of \emph{region proposals} --- candidate boxes likely to contain something, via a cheap, class-agnostic method (the original R-CNN used classical segmentation; \textbf{Faster R-CNN} (\citelink{ren2015}{Ren et al., 2015}) learns a small ``region proposal network'' instead) --- then run a classifier-plus-box-regressor (this chapter's whole architecture) on each proposal independently, exactly like running the sliding-window classifier from Chapter~\ref{ch:lesson40} but only at a handful of promising locations instead of every window. Accurate, but only as fast as (proposals) $\times$ (one forward pass) allows.

\textbf{Single-stage (YOLO, SSD):} skip proposals entirely. \textbf{YOLO} (You Only Look Once, \citelink{redmon2016}{Redmon et al., 2016}) divides the image into a coarse grid of cells, and has each grid cell directly predict (as this chapter's network does) a fixed number of boxes plus a class label plus a confidence score, all in one forward pass. \textbf{SSD} (Single Shot MultiBox Detector, \citelink{liu2016ssd}{Liu et al., 2016}) follows the same one-pass recipe but predicts boxes from \emph{several} feature-map resolutions at once (not just one final grid), so coarser layers naturally catch larger objects and finer layers catch smaller ones. To let a single cell describe objects of different aspect ratios, single-stage detectors use \textbf{anchor boxes}: several predefined box shapes (tall, wide, square) per cell, with the network predicting an \emph{offset} from each anchor rather than a box from scratch. Faster to run, historically somewhat less accurate than two-stage methods, though the gap has narrowed considerably.

Both families end with the same postprocessing step: non-maximum suppression (NMS, Chapter~\ref{ch:lesson40}) to merge the overlapping candidate boxes any real multi-object scene produces. (For simplicity, this chapter's toy detector omitted NMS by only predicting a single bounding box.)

\section{In practice: real detectors on real images}

\begin{figure}[h]
\centering
\includegraphics[width=0.75\textwidth]{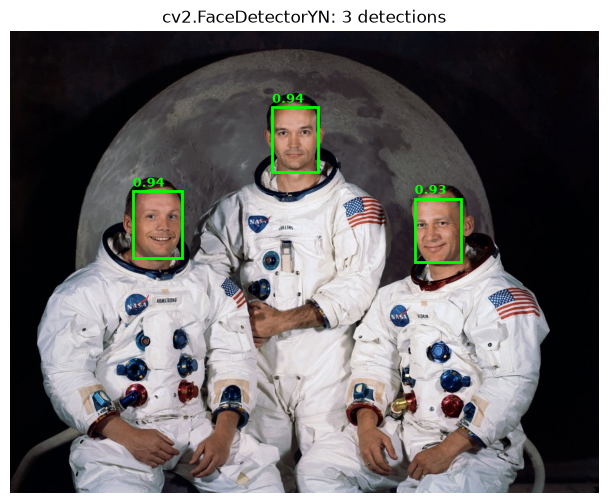}
\caption{\code{cv2.FaceDetectorYN}: 3 detections, each with a confidence score. \attrib{Image source: NASA (public domain), via Wikimedia Commons.}}
\label{fig:l41-3}
\end{figure}

OpenCV ships a single-stage detector, \code{cv2.FaceDetectorYN} (\textbf{YuNet}, \citelink{wu2023yunet}{Wu et al., 2023}), that follows the single-stage recipe above for faces specifically. Unlike Chapter~\ref{ch:lesson40}'s Viola-Jones cascade, it's a small single-shot CNN --- the same family as SSD above, just specialized to one class --- and it runs its own NMS internally before returning boxes. Run on the same Apollo 11 crew photo as Chapter~\ref{ch:lesson40} (Figure~\ref{fig:l41-3}), YuNet finds all three astronauts correctly, with no false positive --- an improvement over Viola-Jones's spurious detection on fabric texture. This is the payoff of a learned single-shot CNN over a cascade of hand-designed Haar-like features: richer features, trained end-to-end on real face/non-face data rather than assembled stage by stage.

But YuNet only answers ``face or not'' --- one class. For general, multi-class detection, Faster R-CNN pretrained on \textbf{COCO} (\citelink{lin2014coco}{Lin et al., 2014}, about 330k real photos labeled across 80 object categories) is a few lines away via \code{torchvision}. This is the same ``pretrained model in one line'' pattern as Chapter~\ref{ch:lesson38}'s ResNet-18. It natively handles a variable, unknown number of objects per image --- the actual payoff of the region-proposal architecture described above.

\begin{lstlisting}
weights = torchvision.models.detection.FasterRCNN_ResNet50_FPN_Weights.COCO_V1
real_detector = torchvision.models.detection.fasterrcnn_resnet50_fpn(weights=weights)
real_detector.eval()  # frozen, no training at all -- Faster R-CNN exactly as released
\end{lstlisting}

\begin{figure}[h]
\centering
\includegraphics[width=0.7\textwidth]{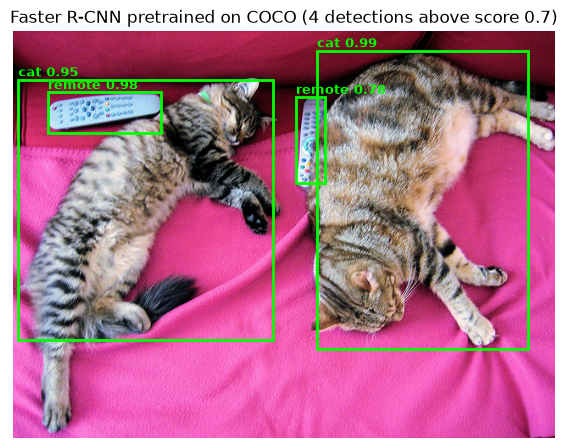}
\caption{Faster R-CNN pretrained on COCO: 4 detections above score $0.7$. \attrib{Image source: COCO dataset (val2017, image 000000039769).}}
\label{fig:l41-4}
\end{figure}

Four confident, correct detections survive the $\text{score\_thresh}=0.7$ cutoff (Figure~\ref{fig:l41-4}) --- two cats and two remote controls, each above $0.78$ --- despite this network using only pretrained weights. Lowering the threshold reveals what the cutoff was hiding: several overlapping, lower-confidence boxes (a ``couch'' and a ``bed'' guess both covering most of the image, around score $0.54$).

\subsection*{Exercises}
\begin{enumerate}
  \item Change \code{obj\_size} in \code{make\_scene} from a fixed $8$ to a random value (e.g.\ \code{rng.integers(4, 12)}) so objects vary in size, and retrain. Does mean IoU hold up, get worse, or barely change --- and why would variable object scale be harder for a single fixed-size regression head than variable position?
  \item The loss function here is plain MSE on $(c_x,c_y,w,h)$, but the metric that matters is IoU. Replace the loss with \code{1 - iou\_batch(pred, target).mean()} (directly optimizing IoU) and compare final mean test IoU to the MSE-trained version. Real detectors (e.g.\ Faster R-CNN, YOLO variants) do exactly this with generalized IoU losses --- can you see why MSE loss and IoU metric might disagree on which of two similar predictions is ``better''?
  \item This chapter's detector has no anchor boxes and only ever handles one object. Sketch (in words) how you would modify the architecture's \emph{output} to handle up to 3 objects per image using a $3\times3$ grid where each grid cell predicts one box and a ``there's an object here'' confidence --- the core idea behind YOLO's grid.
  \item Lower \code{score\_thresh} on the real Faster R-CNN to $0.3$ and rerun. Count how many boxes appear versus at $0.7$; which of the new ones are genuinely additional objects and which are duplicates/near-misses on the cats or remotes that NMS would normally merge away?
\end{enumerate}

\vfill
\fullcode{lesson41}{lesson41_object_detection_regression}

%% file: chapters/lesson42.tex
\chapter{Semantic Segmentation}
\label{ch:lesson42}

Classification labels a whole image. Detection (Chapters~\ref{ch:lesson40}--\ref{ch:lesson41}) labels a handful of boxes. \textbf{Semantic segmentation} goes one step further: label \emph{every pixel} with a class/category. The original deep-learning method is the \textbf{fully convolutional network (FCN)} (\citelink{long2015fcn}{Long, Shelhamer \& Darrell, 2015}), which replaces a classifier's fully-connected layers with convolutions, then upsamples the result back to the input resolution. The output is a dense, per-pixel prediction instead of a single label for the whole image. This chapter builds that idea from scratch, checks it against a real pretrained FCN, and then adds one specific enhancement to it --- skip connections.

\section{A 3-class pixel labeling task}

\begin{figure}[h]
\centering
\includegraphics[width=0.9\textwidth]{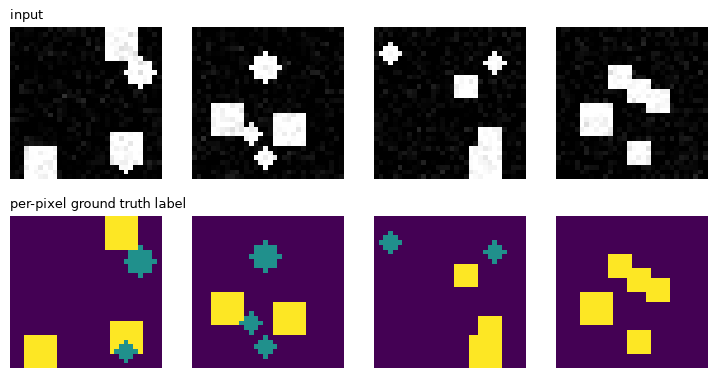}
\caption{Input images (top row) and their per-pixel ground-truth labels (bottom row).}
\label{fig:l42-1}
\end{figure}

Create an image with several small circles and squares scattered on a background. The target is a full per-pixel class map: 0 = background, 1 = circle, 2 = square. Class pixel fractions in the training set are heavily imbalanced: background $0.866$, circle $0.046$, square $0.088$ (Figure~\ref{fig:l42-1}).

\section{A minimal FCN-style network}

A segmentation network outputs a class-probability vector for every pixel, so its output is the same spatial size as the input. The \textbf{encoder} is an ordinary CNN, downsampling twice via max pooling (Chapter~\ref{ch:lesson34}) to build up semantically rich features with a wide receptive field (Chapter~\ref{ch:lesson11}'s pyramid). The \textbf{decoder} then upsamples that coarse bottleneck straight back to full resolution. That's the entire FCN recipe: just downsample then upsample.

\begin{lstlisting}
class FCNTiny(nn.Module):
    def forward(self, x):
        f1 = self.enc1(x)                # (B,16,H,W)
        f2 = self.enc2(self.pool(f1))    # (B,32,H/2,W/2)
        f3 = self.enc3(self.pool(f2))    # (B,64,H/4,W/4)  -- the bottleneck
        d2 = self.dec2(self.up(f3))      # upsample only -- no fusion with f2
        d1 = self.dec1(self.up(d2))      # upsample only -- no fusion with f1
        return self.out(d1)
\end{lstlisting}

Trained for 200 epochs with cross-entropy loss: pixel accuracy $98.1\%$, per-class IoU of $0.984$ (background), $0.736$ (circle), $0.891$ (square), giving a mean IoU of $0.870$.

\section{In practice: FCN on a real photo}

\code{torchvision} ships \textbf{FCN} (\citelink{long2015fcn}{Long et al., 2015}), pretrained on COCO images (labeled with the 21 Pascal VOC categories --- 20 object classes plus background). This is architecturally the same recipe as \code{FCNTiny} above: encoder, then upsampling.

\begin{figure}[h]
\centering
\includegraphics[width=0.95\textwidth]{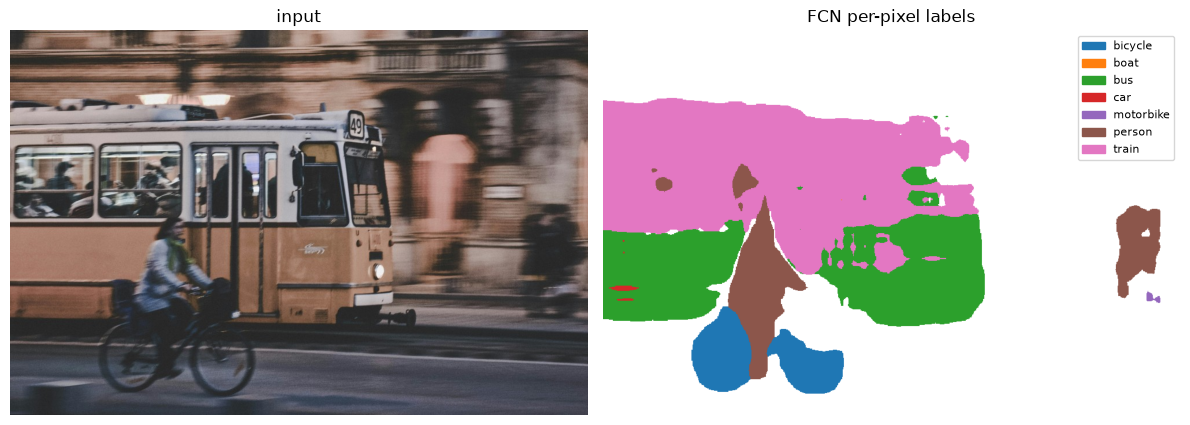}
\caption{Real FCN per-pixel labels on a street photo. \attrib{Photo by Felix Hanspach on Unsplash.}}
\label{fig:l42-2}
\end{figure}

The cyclist is cleanly labeled \code{person}, with the bicycle's wheels correctly labeled \code{bicycle} just below --- both real, correctly-placed detections (Figure~\ref{fig:l42-2}). The tram is where it gets interesting: the model splits it between \code{train} (the upper, windowed section) and \code{bus} (the lower body) rather than picking one label for the whole vehicle. That's not really a mistake so much as an honest reflection of VOC's limited vocabulary --- a tram is a genuinely ambiguous case between ``train'' and ``bus'' for a model that was never given a ``tram'' class to choose from, and the two guesses roughly track a real visual seam in the vehicle (windows vs.\ body) rather than being random. A second, smaller \code{person} blob on the right edge catches the blurred pedestrian in the background.

\section{Adding skip connections: U-Net}

\code{FCNTiny}'s decoder only sees the pooled, upsampled bottleneck features. After two rounds of pooling, that bottleneck has just a quarter of the input's spatial resolution, and small or closely-packed shapes can blur together or lose their boundaries entirely at that resolution. \textbf{U-Net} (\citelink{ronneberger2015}{Ronneberger, Fischer \& Brox, 2015}) fixes this with \textbf{skip connections}: at each decoder stage, it concatenates the encoder's feature map from the \emph{matching resolution}, preserving the fine-grained details that would otherwise be lost during pooling. This is similar in spirit to Chapter~\ref{ch:lesson37}'s residual connections, but the purpose and mechanism differ: U-Net concatenates features across the encoder-decoder boundary, rather than adding a residual update within a single network stack.

\begin{lstlisting}
class UNetTiny(nn.Module):
    def forward(self, x):
        f1 = self.enc1(x)
        f2 = self.enc2(self.pool(f1))
        f3 = self.enc3(self.pool(f2))
        d2 = self.dec2(torch.cat([self.up(f3), f2], dim=1))    # skip from f2
        d1 = self.dec1(torch.cat([self.up(d2), f1], dim=1))    # skip from f1
        return self.out(d1)
\end{lstlisting}

Trained identically for 200 epochs:

\begin{center}
\renewcommand{\arraystretch}{1.3}
\begin{tabular}{lrr}
\hline
 & \textbf{pixel acc} & \textbf{mean IoU} \\
\hline
FCN (no skip)  & 98.1\% & 0.870 \\
U-Net (skip)   & 99.5\% & 0.943 \\
\hline
\end{tabular}
\end{center}

\begin{figure}[h]
\centering
\includegraphics[width=0.9\textwidth]{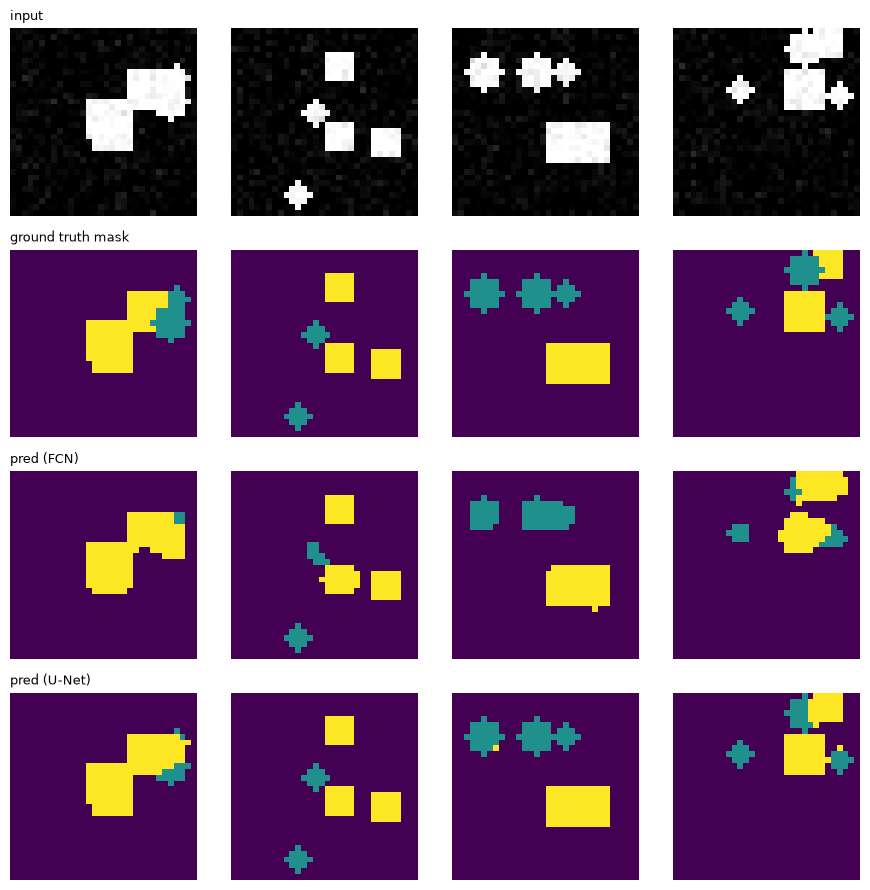}
\caption{Input, ground truth, FCN prediction, and U-Net prediction, for four test scenes.}
\label{fig:l42-3}
\end{figure}

For small, tightly packed shapes, skip connections provide a substantial mean-IoU improvement over a plain FCN, visible directly in Figure~\ref{fig:l42-3}. An FCN cannot recover detail that was lost in the low-resolution bottleneck, because interpolation can only work on existing information. Skip connections fix this problem by giving the decoder a second, non-bottlenecked path to the encoder's high-resolution features, so precise boundaries never have to survive the bottleneck in the first place.

This helps explain why U-Net remains popular for tasks requiring precise boundaries, especially medical imaging. But skip connections are not universal: popular networks like plain FCN and DeepLabV3 have none, and even DeepLabV3+ adds only a single one, not a full U-Net-style connection at every resolution. So while skip connections are a useful tool, they are not a requirement --- their value depends on how much fine spatial detail the task demands.

\subsection*{Exercises}
\begin{enumerate}
  \item Increase \code{n\_shapes} from 5 to 10, making the scene more crowded. Does the \code{FCNTiny}-vs-\code{UNetTiny} gap in mean IoU get larger or smaller? What does that suggest about when skip connections matter most?
  \item This chapter's loss is plain per-pixel cross-entropy. Try \code{F.cross\_entropy(logits, Mt, weight=inverse\_class\_freq)} (Chapter~\ref{ch:lesson36}'s imbalance fix, applied here) to see whether it changes the circle/square IoU balance.
  \item \code{nn.Upsample(mode='nearest')} was used for simplicity. Try \code{mode='bilinear', align\_corners=False} instead (Chapter~\ref{ch:lesson09}'s bilinear interpolation, now inside a network) and compare mean IoU for both \code{FCNTiny} and \code{UNetTiny}. Does the smoother upsampling help more or less than the skip connection does?
  \item Print the confidence (softmax probability of the predicted class) at a few pixels along the seam where the real FCN result switches from \code{train} to \code{bus} on the tram. Is the model confidently split, or genuinely uncertain right at that boundary?
\end{enumerate}

\vfill
\fullcode{lesson42}{lesson42_semantic_segmentation}

%% file: chapters/lesson43.tex
\chapter{Instance Segmentation}
\label{ch:lesson43}

Semantic segmentation (Chapter~\ref{ch:lesson42}) labels each pixel with a class, but it has no notion of \emph{how many} instances of each class there are --- two touching instances are just labeled as one connected blob of pixels. \textbf{Instance segmentation} answers the harder question: which pixels belong to \emph{this specific} object, as opposed to some other object of the same class. This chapter augments Chapter~\ref{ch:lesson42}'s segmentation network directly to solve it, with an idea that traces straight back to Chapter~\ref{ch:lesson12}'s Hough transform: instead of only classifying pixels, have the network also predict, for every foreground pixel, a vote for where its object's center is --- then cluster the votes to recover individual instances.

\section{Overlapping instances}

\begin{figure}[h]
\centering
\includegraphics[width=0.9\textwidth]{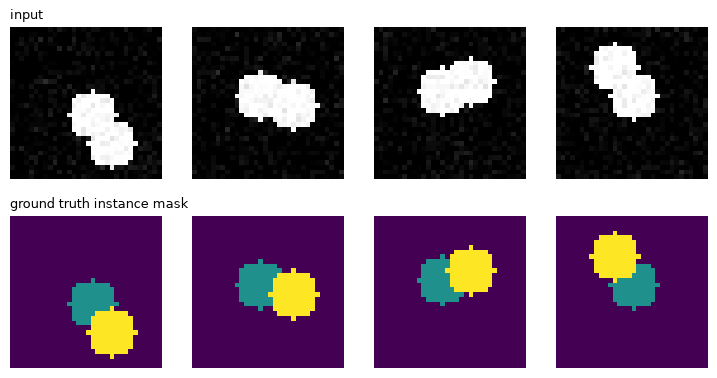}
\caption{Input images (top row) and their ground-truth instance masks (bottom row).}
\label{fig:l43-1}
\end{figure}

Two circles per scene, deliberately placed close enough to touch or overlap. Ground truth includes both a semantic mask (0=background, 1=circle) and an \emph{instance} mask (0=background, 1=first circle, 2=second circle), as shown in Figure~\ref{fig:l43-1}.

\section{Augmenting the network with an instance head}

Take Chapter~\ref{ch:lesson42}'s exact U-Net architecture and give it a second output head alongside the semantic head: for every foreground pixel, predict a $(dx,dy)$ offset pointing toward its own instance's center. The semantic head still does the same job as Chapter~\ref{ch:lesson42} (background vs.\ circle); the new offset head is what turns plain segmentation into \emph{instance} segmentation.

\begin{lstlisting}
class InstanceNet(nn.Module):
    """U-Net, with a second head: per-pixel offset-to-center regression"""
    def forward(self, x):
        f1 = self.enc1(x)
        f2 = self.enc2(self.pool(f1))
        d1 = self.dec1(torch.cat([self.up(f2), f1], dim=1))
        return self.sem_head(d1), self.offset_head(d1)
\end{lstlisting}

Trained for 300 epochs with a weighted sum of cross-entropy (semantic head) and MSE (offset head, masked to foreground pixels only): semantic pixel accuracy $100.0\%$.

\section{Recovering instances: vote for the center, then cluster (Hough, again)}

The offset head above was trained to predict, for every foreground pixel, an offset pointing toward \emph{its own instance's} center --- supervised directly from the ground-truth instance masks. Each foreground pixel casts one vote (its predicted center location) into an accumulator, exactly like Chapter~\ref{ch:lesson12}'s Hough line/circle voting. Because every pixel belonging to the same circle votes for approximately the same point, the accumulator forms one tight cluster of votes per instance, however tangled the pixels themselves are. Finding instances becomes: find the vote clusters, then assign each pixel to its nearest cluster.

\begin{lstlisting}
# greedily take the highest-voted peak, suppress its neighborhood, repeat
# (the same non-max-suppression idea as Chapter 40's detection boxes)
for _ in range(6):
    idx = np.unravel_index(votes_work.argmax(), votes_work.shape)
    if votes_work[idx] < vote_thresh:
        break
    peaks.append((idx[1], idx[0]))
    votes_work[y0:y1, x0:x1] = 0  # suppress the neighborhood around this peak
\end{lstlisting}

\begin{figure}[h]
\centering
\includegraphics[width=0.9\textwidth]{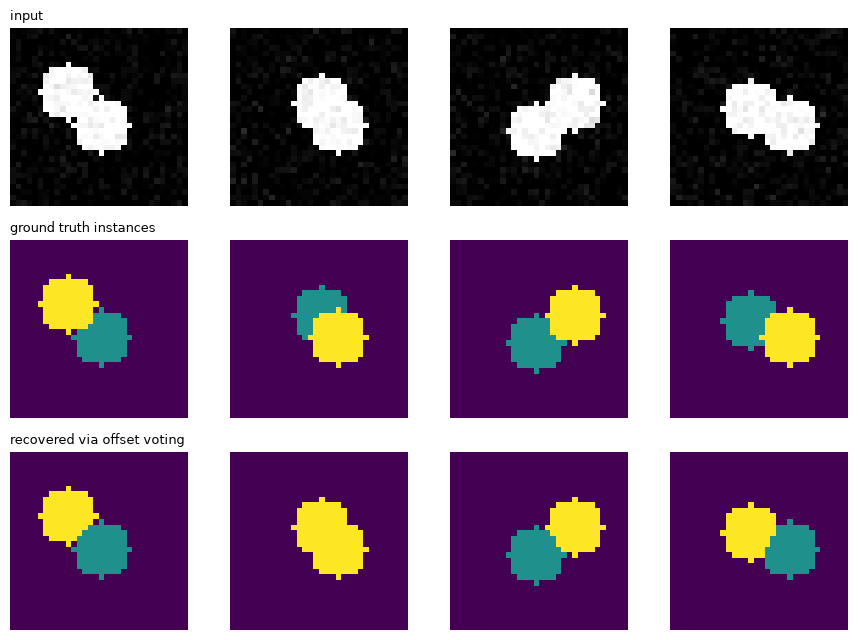}
\caption{Input, ground-truth instances, and instances recovered via offset voting, for four test scenes.}
\label{fig:l43-2}
\end{figure}

Offset-voting recovers the correct instance count in $44/45$ test scenes ($97.8\%$), shown in Figure~\ref{fig:l43-2}.

\section{Mask R-CNN: Faster R-CNN plus a mask branch}

The offset-voting approach above is one family of instance segmentation methods (related to techniques sometimes called ``instance embedding'' or center-voting). The other dominant approach, \textbf{Mask R-CNN} (\citelink{he2017maskrcnn}{He et al., 2017}), does not start from scratch --- it is \textbf{Faster R-CNN} (Chapter~\ref{ch:lesson41}) with one new piece bolted on. The region proposal network, the per-box classifier, and the per-box coordinate refinement are all unchanged. Mask R-CNN just adds a third, parallel branch on every detected box: alongside the existing class label and refined box coordinates, a small fully-convolutional head predicts a binary mask \emph{within that box}. Detection already solves ``how many objects, and roughly where'' via region proposals and NMS; the mask branch adds ``and here's this one's exact silhouette'' --- exactly the piece built and tested in isolation below.

One more change was needed to make that mask branch actually work well. Faster R-CNN extracts a fixed-size feature grid for each proposed box by snapping (quantizing) the box's coordinates to the nearest feature-map cell --- a step called RoIPool. That rounding barely affects classifying a box or nudging its coordinates by a few pixels, but it visibly misaligns a per-pixel mask against the object it is supposed to outline. Mask R-CNN replaces RoIPool with \textbf{RoIAlign}, which reads features via bilinear interpolation instead of rounding to the nearest cell, so the extracted features --- and therefore the predicted mask --- stay aligned with the actual box coordinates. This is a small change to the detector, but it is the specific fix that makes pixel-accurate masks possible.

Putting Chapters~\ref{ch:lesson36}, \ref{ch:lesson40}, \ref{ch:lesson41}, and this chapter together: \textbf{classification} (Chapter~\ref{ch:lesson36}) gives one label for the whole image, \textbf{semantic segmentation} (Chapter~\ref{ch:lesson42}) gives one label per pixel with no notion of separate objects of the same class, and \textbf{instance segmentation} (this chapter) adds a distinct identity per object --- ``background,'' ``circle \#1,'' ``circle \#2,'' not just ``background,'' ``circle.'' A fourth term, \textbf{panoptic segmentation}, unifies the last two: every pixel gets both a semantic class and, for countable ``thing'' classes like circles or cars, an instance ID --- while uncountable ``stuff'' classes like sky or road are labeled semantically only, since instance identity doesn't make sense for them.

\section{A mask head}

Detection (Chapter~\ref{ch:lesson41}) and per-instance masking are two separate, already-solved pieces at this point --- the only genuinely new ingredient Mask R-CNN adds is the \textbf{mask head}: a small network that looks \emph{only} inside one proposed box and predicts a binary mask for the single object that box was proposed for, ignoring anything else visible in the crop.

Build that piece in isolation. Take the ground-truth box around each instance's true center (assume localization is already solved, so only the new mask-head idea is under test), crop a fixed $14\times14$ window there, and train a tiny fully-convolutional head to predict which pixels in that crop belong to \emph{this} instance --- not to any neighboring circle that happens to also be visible in the same crop. This produces $510$ training crops and $90$ test crops, each $14\times14$.

\begin{lstlisting}
class MaskHead(nn.Module):
    def __init__(self):
        super().__init__()
        self.net = nn.Sequential(
            nn.Conv2d(1, 16, 3, padding=1), nn.ReLU(),
            nn.Conv2d(16, 16, 3, padding=1), nn.ReLU(),
            nn.Conv2d(16, 1, 1),  # per-pixel mask logit, full crop resolution throughout
        )
\end{lstlisting}

\begin{figure}[h]
\centering
\includegraphics[width=0.9\textwidth]{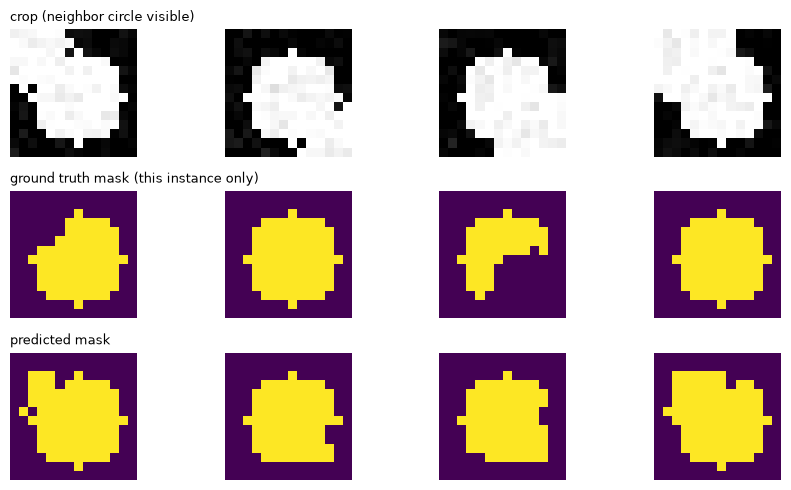}
\caption{Crops (with a neighboring circle visible), ground-truth per-instance masks, and predicted masks.}
\label{fig:l43-3}
\end{figure}

Trained for 300 epochs with binary cross-entropy: mask-head IoU (per-instance) $0.844$, versus a naive ``everything foreground in the crop'' baseline IoU of $0.687$ (Figure~\ref{fig:l43-3}).

Because this dataset's circles are generated close enough to always touch or overlap, \emph{every single test crop} contains visible pixels from the neighboring circle --- there is no ``easy'' case here to inflate the numbers. The per-instance mask head still reaches a respectably high IoU against the true single-instance mask, clearly ahead of the naive baseline that just marks every foreground pixel in the crop as belonging to ``the'' object (which is systematically wrong whenever a neighbor intrudes --- the same problem that motivated giving the network an instance-center head in the first place, just replayed at crop scale). The mask head doesn't need to solve ``where are the circles'' --- the box already answered that --- it only has to learn ``given that this box was proposed for one specific circle, which pixels are that circle's, and not its neighbor's.'' That's a strictly easier, more local question, which is exactly why Mask R-CNN's mask branch can be small and fast: a good box already does most of the work.

This also clarifies why Mask R-CNN and this chapter's offset-voting approach rarely fail the same way. Offset-voting can mis-cluster votes when two instances' centers are close enough that their accumulator peaks blur together. Mask R-CNN instead depends on the upstream detector proposing a correct box for every instance in the first place --- miss a box, and that instance's mask is never even attempted, no matter how good the mask head is.

\section{In practice: Mask R-CNN on a real photo}

The mask head above proves the idea works once localization is solved for free (ground-truth box centers). Here is the same idea, fully assembled and pretrained: a real \textbf{Mask R-CNN} (\code{torchvision.models.detection.maskrcnn\_resnet50\_fpn}, trained on COCO) run end to end --- detection and mask head both included, nothing given away --- on a photo with exactly the kind of touching, same-class instances this chapter has been building toward: the same Apollo 11 crew photo from Chapters~\ref{ch:lesson40}--\ref{ch:lesson41}.

\begin{figure}[h]
\centering
\includegraphics[width=0.95\textwidth]{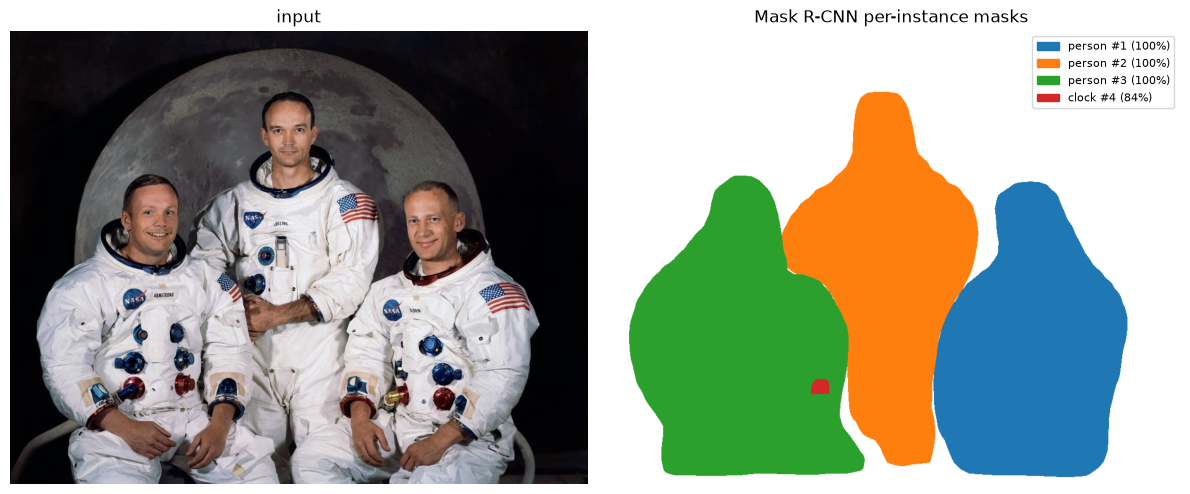}
\caption{Real Mask R-CNN per-instance masks. \attrib{Image source: NASA (public domain), via Wikimedia Commons.}}
\label{fig:l43-4}
\end{figure}

Above $80\%$ confidence: 4 instances --- three \code{person} detections (each $\geq99.8\%$) and one spurious \code{clock} detection ($83.6\%$), shown in Figure~\ref{fig:l43-4}.

The three astronauts are shoulder-to-shoulder, touching along multiple arm and torso boundaries, and belong to the exact same COCO class (\code{person}) --- precisely the case where a semantic mask would merge them into one undifferentiated blob. Mask R-CNN separates them cleanly into three distinct instances anyway, because its detector proposes three separate boxes first (one per astronaut) and the mask head only looks inside a single box at a time, exactly like the mask head above. The fourth, spurious detection is a small \code{clock} mask on the left astronaut's wrist --- the round, mirrored cuff-mounted checklist on Armstrong's arm, misread as a clock face.

\subsection*{Exercises}
\begin{enumerate}
  \item Reduce \code{dist} in \code{make\_scene} from the range $(6,9)$ to $(2,5)$, making the circles overlap much more heavily (centers closer together). Does the offset-voting method's instance-count accuracy hold up, or does it start failing too --- and if so, what does the failure look like (merged instances, split instances, or something else)?
  \item The clustering here uses \code{peak\_dist=4} for non-max suppression on the vote accumulator. Try \code{peak\_dist=8}. Does accuracy improve, get worse, or become sensitive to exactly how close two true circle centers happen to be in a given scene?
  \item This chapter only ever has 2 circles per scene. Extend \code{make\_scene} to place a random number of circles (1 to 4). Does the offset-voting approach's instance count accuracy hold up as the true number of instances grows?
  \item The mask head above was trained and evaluated using the \emph{ground-truth} box center, sidestepping detection entirely. Perturb each crop's center by a few random pixels before cropping (simulating an imperfect detector) and rerun. How much does mask IoU degrade as the box gets less accurately centered?
\end{enumerate}

\vfill
\fullcode{lesson43}{lesson43_instance_segmentation}

%% file: chapters/lesson44.tex
\chapter{Precision and Parallel Training}
\label{ch:lesson44}

Every network in Chapters~\ref{ch:lesson31}--\ref{ch:lesson43} trained on a small, CPU-friendly toy problem in default 32-bit floating point (fp32), on a single process. Real training runs rarely look like that: models are trained in reduced precision to go faster and use less memory, and split across many GPUs because no single device holds a modern model plus its training data. This chapter closes out Part 3 with the engineering underneath \emph{how} training actually happens at scale, building each mechanism in miniature and checking it against the real thing: \textbf{floating-point precision} and why reduced precision needs help to work at all, \textbf{mixed-precision training}'s specific fix for that, and the two ways of splitting work across devices, \textbf{data parallelism} and \textbf{model parallelism}.

\section{Floating-point precision: fp32 vs.\ fp16}

A 32-bit float (fp32) and a 16-bit float (fp16) both use the same idea --- sign, exponent, mantissa --- but fp16 spends far fewer bits on the exponent, so it simply cannot represent very small or very large magnitudes. \code{torch.finfo} reports the smallest positive \emph{normal} value each format can hold: about $1.2\times10^{-38}$ for fp32, but only $6.1\times10^{-5}$ for fp16 (fp16 subnormals stretch a little further, down to about $6\times10^{-8}$, at the cost of losing precision entirely). Below that floor, a value doesn't round to something small --- it rounds to exactly zero.

\begin{center}
\renewcommand{\arraystretch}{1.3}
\begin{tabular}{rrl}
\hline
\textbf{fp32 value} & \textbf{cast to fp16} & \textbf{underflowed to 0?} \\
\hline
$1.0\times10^{-2}$ & $1.0\times10^{-2}$ & False \\
$1.0\times10^{-4}$ & $1.0\times10^{-4}$ & False \\
$1.0\times10^{-6}$ & $1.0\times10^{-6}$ & False \\
$1.0\times10^{-7}$ & $1.2\times10^{-7}$ & False \\
$1.0\times10^{-8}$ & $0.0$ & \textbf{True} \\
$1.0\times10^{-9}$ & $0.0$ & \textbf{True} \\
\hline
\end{tabular}
\end{center}

Values down to $10^{-6}$ survive the cast intact; $10^{-7}$ survives but loses precision (rounds to $1.2\times10^{-7}$); $10^{-8}$ and smaller vanish to exactly $0.0$. That's not a rounding error in the usual sense --- it's total information loss, and it matters enormously for training because gradients routinely take on exactly these magnitudes, especially in deep networks. Chapters~\ref{ch:lesson33} and \ref{ch:lesson37} built and studied the vanishing-gradient problem in fp32, where early-layer gradients shrink but stay nonzero. In fp16, ``shrink'' can mean ``disappear.''

A 12-layer sigmoid network (deliberately deep and derivative-shrinking, the same vanishing-gradient setup Chapter~\ref{ch:lesson37} diagnosed) produces first-layer gradients with mean magnitude $2.65\times10^{-12}$, max $2.33\times10^{-11}$ --- comfortably nonzero in fp32 ($0.0\%$ exactly zero), but $100.0\%$ exactly zero the instant they're cast to fp16. Train this network directly in fp16 and the first layer would never update at all; the optimizer would faithfully apply a learning rate to a gradient of exactly zero, forever. This is the actual problem mixed-precision training has to solve, not a hypothetical one.

\section{Mixed-precision training: the recipe}

Training entirely in fp16 hits the wall demonstrated above; training entirely in fp32 gives up the speed and memory savings that were the whole point. \textbf{Mixed-precision training} (\citelink{micikevicius2017}{Micikevicius et al., 2018}) keeps both: the forward and backward pass run in fp16 for speed, but two safeguards protect against exactly the failure shown above:
\begin{enumerate}
  \item \textbf{fp32 master weights}: a full-precision copy of every weight is kept off to the side. Updates apply to the fp32 copy (where small updates don't get rounded away), and an fp16 copy is re-derived from it before each forward pass.
  \item \textbf{Loss scaling}: multiply the loss by a large constant \emph{before} calling \code{.backward()}. Since gradients are linear in the loss, every gradient scales up by the same factor --- pushing values that would have underflowed back into fp16's representable range. The update step divides the gradient back down by that same factor before it's applied, so the math is otherwise unaffected.
\end{enumerate}

\begin{center}
\renewcommand{\arraystretch}{1.3}
\begin{tabular}{rr}
\hline
\textbf{loss scale} & \textbf{fraction of fp16 grads still zero} \\
\hline
1           & 100.0\% \\
65{,}536    & 35.9\% \\
1{,}048{,}576   & 7.8\% \\
16{,}777{,}216  & 1.6\% \\
268{,}435{,}456 & 0.0\% \\
\hline
\end{tabular}
\end{center}

With no scaling ($\text{scale}=1$), every first-layer gradient underflows, exactly as before. Scaling by $65{,}536$ ($2^{16}$, a common real-world default) already rescues most of them, but over a third are still small enough to vanish. Only at $2^{28}$ does the last gradient clear fp16's floor. This is precisely why real implementations (PyTorch's \code{torch.cuda.amp.GradScaler}) use \emph{dynamic} loss scaling rather than a fixed constant: start with some scale, and increase it further whenever a step completes without producing an \code{inf}/\code{NaN} (a sign the scale could safely be larger), or back off whenever one appears (a sign the scale overflowed something). One more detail worth naming: \textbf{bfloat16} (bf16), an increasingly common alternative to fp16, sidesteps loss scaling entirely --- \code{torch.finfo} shows it has the \emph{same} exponent range as fp32 ($\text{smallest\_normal}$ around $1.2\times10^{-38}$ for both), so it can't underflow the way fp16 does. What it gives up instead is precision ($\text{eps}$ around $0.008$ for bf16 versus $0.001$ for fp16) --- a different trade-off along the same speed-versus-representable-range axis, not a free lunch.

\section{Data parallelism: correctness proven in miniature}

Precision controls how much work one device does per step; parallelism controls how many devices share that work. \textbf{Data parallelism} (the standard recipe behind large-batch training, e.g.\ \citelink{goyal2017}{Goyal et al., 2017}) splits a batch into shards, runs an identical copy of the model on each shard on its own device, and averages the resulting gradients before the update step. On one CPU there's no speedup to show, but the \emph{correctness} of that recipe --- that averaging per-shard gradients gives the same answer as computing the gradient on the whole batch at once --- is exactly checkable, because MSE loss is itself an average, and the gradient of an average of per-shard averages is the average of the per-shard gradients.

\begin{lstlisting}
# "data parallel, 4 shards": an identical model copy per shard, each computing its own
# gradient independently; only the gradients get communicated and averaged, never the data
per_shard_grads = []
for s in range(n_shards):
    shard_model = copy.deepcopy(model)
    F.mse_loss(shard_model(xs), ys).backward()
    per_shard_grads.append([p.grad.clone() for p in shard_model.parameters()])
averaged_grad = [sum(shard[i] for shard in per_shard_grads) / n_shards for i in range(len(params))]
\end{lstlisting}

Every parameter's averaged-shard gradient matches the single-device gradient to within $1.49\times10^{-8}$ --- floating-point roundoff, not a real discrepancy. This is the whole correctness argument for data parallelism in one demonstration: splitting the batch and averaging afterward is mathematically the same computation as never splitting it at all. What this toy version doesn't --- can't --- show is the actual cost real data-parallel training pays: the \textbf{all-reduce}, the communication step that sums/averages gradients across every device before any of them can proceed to the update. At small scale that communication is nearly free; at hundreds of GPUs it can dominate the step time, which is why large-scale training spends real engineering effort on overlapping communication with computation rather than doing them one after another.

\section{Model parallelism: shipping activations across a boundary}

Data parallelism copies the whole model onto every device and splits the \emph{data}. \textbf{Model parallelism} (\citelink{shoeybi2019}{Shoeybi et al., 2019}) does the opposite: split the \emph{model} itself across devices, because it's too large to fit on one. In the simplest form (pipeline parallelism), each device holds a contiguous block of layers; the forward pass computes one block's output, hands that activation across the device boundary, and the next device picks up from there. The backward pass runs the same handoff in reverse: each device needs only the \emph{gradient of the loss with respect to the activation it sent}, not anything about what happens on the other side of the boundary.

\begin{lstlisting}
# "two devices": stage1 on device A, stage2 on device B, with the activation
# manually shipped across the boundary in each direction
activation_out = split_model.stage1(X)                        # entirely on "device A"
activation_in = activation_out.detach().requires_grad_(True)  # crosses the boundary
prediction = split_model.stage2(activation_in)                # entirely on "device B"
F.mse_loss(prediction, y).backward()                          # backprop stops at the boundary
activation_out.backward(activation_in.grad)                   # resumes backprop into stage1
\end{lstlisting}

Every gradient matches exactly --- $0.0$ difference, not just roundoff, because the split-and-reattach mechanism (\code{detach()} at the boundary going forward, then manually resuming \code{.backward()} with the shipped-back gradient) is mathematically identical to autograd tracing straight through an unsplit model; it just makes the ``device boundary'' explicit as a place activations get detached and gradients get reattached by hand instead of automatically. Real pipeline parallelism adds one more concern this two-device, one-batch toy sidesteps entirely: while device B works on the tail of the pipeline for one batch, device A sits idle unless a \emph{second} batch is already flowing in behind it --- the ``pipeline bubble'' that real implementations fight with careful scheduling (splitting each batch into smaller \emph{microbatches} so every stage stays busy). Data parallelism's bottleneck is communication (the all-reduce above); model parallelism's is idle time.

\section{Two more practical tricks}

Two more techniques round out the practical picture without needing their own full demonstration, because both are natural extensions of mechanisms already built above:
\begin{itemize}
  \item \textbf{Gradient accumulation} simulates a larger batch than fits in memory by running several small forward/backward passes without an optimizer step in between, summing their gradients, and only then updating --- mathematically the same averaging idea as the data-parallel shards above, just spread across \emph{time} on one device instead of across \emph{devices} at once.
  \item \textbf{Gradient checkpointing} trades compute for memory: instead of storing every intermediate activation for the backward pass (what autograd normally does), it stores only a handful of checkpoints and \emph{recomputes} the activations in between during the backward pass. More forward-pass compute, much less memory --- useful when a model's activations, not its weights, are what doesn't fit.
\end{itemize}

\subsection*{Exercises}
\begin{enumerate}
  \item In the fp16 underflow demo, change \code{DeepSigmoidNet}'s \code{depth} from 12 to 4. Does the first-layer gradient still underflow completely in fp16? Relate the answer back to Chapter~\ref{ch:lesson37}'s discussion of \emph{why} depth and sigmoid activations together cause vanishing gradients.
  \item The loss-scaling table stops at $\text{scale}=2^{28}$. Keep increasing the scale (try $2^{32}$, $2^{40}$) --- at what point does the \emph{scaled} gradient itself start to risk overflowing fp16's max of $65{,}504$ instead of underflowing? What does that imply about why dynamic loss scaling needs to search for a scale, rather than always using the largest one possible?
  \item Change \code{n\_shards} in the data-parallelism demo from 4 to 3, with a batch size that doesn't divide evenly (try 65). Does the averaged-shard gradient still match the single-device gradient exactly, or does the mismatch reveal something about how real data-parallel training handles uneven shard sizes?
  \item Add a third stage to \code{TwoStageNet} (three \code{nn.Sequential} blocks instead of two) and extend the manual forward/backward handoff to three devices. Does the gradient still match a whole-model backward pass exactly?
\end{enumerate}

\vfill
\fullcode{lesson44}{lesson44_precision_and_parallel_training}

%% file: chapters/references.tex
\chapter*{References}
\addcontentsline{toc}{chapter}{References}

Entries marked \landmark are highly cited, landmark papers.

\bibentry{otsu1979}{\landmark Otsu, N.}{ (1979). A threshold selection method from gray-level histograms. \textit{IEEE Transactions on Systems, Man, and Cybernetics}, 9(1), 62--66.}

\bibentry{canny1986}{\landmark Canny, J.}{ (1986). A computational approach to edge detection. \textit{IEEE Transactions on Pattern Analysis and Machine Intelligence}, 8(6), 679--698.}

\bibentry{hough1962}{Hough, P.~V.~C.}{ (1962). Method and means for recognizing complex patterns. U.S. Patent 3,069,654.}

\bibentry{dudahart1972}{Duda, R.~O., \& Hart, P.~E.}{ (1972). Use of the Hough transformation to detect lines and curves in pictures. \textit{Communications of the ACM}, 15(1), 11--15.}

\bibentry{marrhildreth1980}{Marr, D., \& Hildreth, E.}{ (1980). Theory of edge detection. \textit{Proceedings of the Royal Society of London B}, 207(1167), 187--217.}

\bibentry{daubechies1988}{Daubechies, I.}{ (1988). Orthonormal bases of compactly supported wavelets. \textit{Communications on Pure and Applied Mathematics}, 41(7), 909--996.}

\bibentry{hubelwiesel1959}{Hubel, D.~H., \& Wiesel, T.~N.}{ (1959). Receptive fields of single neurones in the cat's striate cortex. \textit{Journal of Physiology}, 148(3), 574--591.}

\bibentry{marcelja1980}{Marcelja, S.}{ (1980). Mathematical description of the responses of simple cortical cells. \textit{Journal of the Optical Society of America}, 70(11), 1297--1300.}

\bibentry{daugman1985}{Daugman, J.~G.}{ (1985). Uncertainty relation for resolution in space, spatial frequency, and orientation optimized by two-dimensional visual cortical filters. \textit{Journal of the Optical Society of America A}, 2(7), 1160--1169.}

\bibentry{arthur2007}{\landmark Arthur, D., \& Vassilvitskii, S.}{ (2007). k-means++: The advantages of careful seeding. \textit{Proceedings of the ACM-SIAM Symposium on Discrete Algorithms (SODA)}.}

\bibentry{dempster1977}{\landmark Dempster, A.~P., Laird, N.~M., \& Rubin, D.~B.}{ (1977). Maximum likelihood from incomplete data via the EM algorithm. \textit{Journal of the Royal Statistical Society, Series B}, 39(1), 1--38.}

\bibentry{ester1996}{\landmark Ester, M., Kriegel, H.-P., Sander, J., \& Xu, X.}{ (1996). A density-based algorithm for discovering clusters in large spatial databases with noise. \textit{Proceedings of KDD}.}

\bibentry{harrisstephens1988}{Harris, C., \& Stephens, M.}{ (1988). A combined corner and edge detector. \textit{Proceedings of the Alvey Vision Conference}.}

\bibentry{shitomasi1994}{Shi, J., \& Tomasi, C.}{ (1994). Good features to track. \textit{Proceedings of IEEE CVPR}.}

\bibentry{lowe2004}{\landmark Lowe, D.~G.}{ (2004). Distinctive image features from scale-invariant keypoints. \textit{International Journal of Computer Vision}, 60(2), 91--110.}

\bibentry{lucaskanade1981}{Lucas, B.~D., \& Kanade, T.}{ (1981). An iterative image registration technique with an application to stereo vision. \textit{Proceedings of IJCAI}.}

\bibentry{hornschunck1981}{Horn, B.~K.~P., \& Schunck, B.~G.}{ (1981). Determining optical flow. \textit{Artificial Intelligence}, 17(1--3), 185--203.}

\bibentry{farneback2003}{Farneb\"ack, G.}{ (2003). Two-frame motion estimation based on polynomial expansion. \textit{Proceedings of the Scandinavian Conference on Image Analysis (SCIA)}.}

\bibentry{fischlerbolles1981}{\landmark Fischler, M.~A., \& Bolles, R.~C.}{ (1981). Random sample consensus: A paradigm for model fitting with applications to image analysis and automated cartography. \textit{Communications of the ACM}, 24(6), 381--395.}

\bibentry{bayer1976}{Bayer, B.~E.}{ (1976). Color imaging array. U.S. Patent 3,971,065.}

\bibentry{zhang2000}{Zhang, Z.}{ (2000). A flexible new technique for camera calibration. \textit{IEEE Transactions on Pattern Analysis and Machine Intelligence}, 22(11), 1330--1334.}

\bibentry{hartley1997}{Hartley, R.}{ (1997). In defense of the eight-point algorithm. \textit{IEEE Transactions on Pattern Analysis and Machine Intelligence}, 19(6), 580--593.}

\bibentry{yao2020}{Yao, Y., Luo, Z., Li, S., Zhang, J., Ren, Y., Zhou, L., Fang, T., \& Quan, L.}{ (2020). BlendedMVS: A large-scale dataset for generalized multi-view stereo networks. \textit{Proceedings of IEEE CVPR}.}

\bibentry{kingmaba2014}{\landmark Kingma, D.~P., \& Ba, J.}{ (2015). Adam: A method for stochastic optimization. \textit{Proceedings of ICLR}.}

\bibentry{glorotbengio2010}{Glorot, X., \& Bengio, Y.}{ (2010). Understanding the difficulty of training deep feedforward neural networks. \textit{Proceedings of AISTATS}.}

\bibentry{he2015kaiming}{He, K., Zhang, X., Ren, S., \& Sun, J.}{ (2015). Delving deep into rectifiers: Surpassing human-level performance on ImageNet classification. \textit{Proceedings of IEEE ICCV}.}

\bibentry{srivastava2014}{\landmark Srivastava, N., Hinton, G., Krizhevsky, A., Sutskever, I., \& Salakhutdinov, R.}{ (2014). Dropout: A simple way to prevent neural networks from overfitting. \textit{Journal of Machine Learning Research}, 15(56), 1929--1958.}

\bibentry{krizhevsky2009cifar}{Krizhevsky, A.}{ (2009). Learning multiple layers of features from tiny images. \textit{Technical Report, University of Toronto}.}

\bibentry{deng2009}{\landmark Deng, J., Dong, W., Socher, R., Li, L.-J., Li, K., \& Fei-Fei, L.}{ (2009). ImageNet: A large-scale hierarchical image database. \textit{Proceedings of IEEE CVPR}.}

\bibentry{russakovsky2015}{\landmark Russakovsky, O., Deng, J., Su, H., Krause, J., Satheesh, S., Ma, S., Huang, Z., Karpathy, A., Khosla, A., Bernstein, M., Berg, A.~C., \& Fei-Fei, L.}{ (2015). ImageNet large scale visual recognition challenge. \textit{International Journal of Computer Vision}, 115(3), 211--252.}

\bibentry{lecun1998}{\landmark LeCun, Y., Bottou, L., Bengio, Y., \& Haffner, P.}{ (1998). Gradient-based learning applied to document recognition. \textit{Proceedings of the IEEE}, 86(11), 2278--2324.}

\bibentry{krizhevsky2012}{\landmark Krizhevsky, A., Sutskever, I., \& Hinton, G.~E.}{ (2012). ImageNet classification with deep convolutional neural networks. \textit{Proceedings of NeurIPS}.}

\bibentry{simonyanzisserman2014}{\landmark Simonyan, K., \& Zisserman, A.}{ (2015). Very deep convolutional networks for large-scale image recognition. \textit{Proceedings of ICLR}.}

\bibentry{szegedy2015}{\landmark Szegedy, C., Liu, W., Jia, Y., Sermanet, P., Reed, S., Anguelov, D., Erhan, D., Vanhoucke, V., \& Rabinovich, A.}{ (2015). Going deeper with convolutions. \textit{Proceedings of IEEE CVPR}.}

\bibentry{he2015resnet}{\landmark He, K., Zhang, X., Ren, S., \& Sun, J.}{ (2016). Deep residual learning for image recognition. \textit{Proceedings of IEEE CVPR}.}

\bibentry{hu2018senet}{\landmark Hu, J., Shen, L., \& Sun, G.}{ (2018). Squeeze-and-excitation networks. \textit{Proceedings of IEEE CVPR}.}

\bibentry{ioffeszegedy2015}{\landmark Ioffe, S., \& Szegedy, C.}{ (2015). Batch normalization: Accelerating deep network training by reducing internal covariate shift. \textit{Proceedings of ICML}.}

\bibentry{simonyan2013saliency}{Simonyan, K., Vedaldi, A., \& Zisserman, A.}{ (2014). Deep inside convolutional networks: Visualising image classification models and saliency maps. \textit{ICLR Workshop}.}

\bibentry{selvaraju2017}{\landmark Selvaraju, R.~R., Cogswell, M., Das, A., Vedantam, R., Parikh, D., \& Batra, D.}{ (2017). Grad-CAM: Visual explanations from deep networks via gradient-based localization. \textit{Proceedings of IEEE ICCV}.}

\bibentry{vandermaaten2008tsne}{\landmark van der Maaten, L., \& Hinton, G.}{ (2008). Visualizing data using t-SNE. \textit{Journal of Machine Learning Research}, 9, 2579--2605.}

\bibentry{rowleybalujakanade1996}{Rowley, H.~A., Baluja, S., \& Kanade, T.}{ (1996). Neural network-based face detection. \textit{Proceedings of IEEE CVPR}.}

\bibentry{violajones2001}{\landmark Viola, P., \& Jones, M.}{ (2001). Rapid object detection using a boosted cascade of simple features. \textit{Proceedings of IEEE CVPR}.}

\bibentry{ren2015}{\landmark Ren, S., He, K., Girshick, R., \& Sun, J.}{ (2015). Faster R-CNN: Towards real-time object detection with region proposal networks. \textit{Proceedings of NeurIPS}.}

\bibentry{redmon2016}{\landmark Redmon, J., Divvala, S., Girshick, R., \& Farhadi, A.}{ (2016). You only look once: Unified, real-time object detection. \textit{Proceedings of IEEE CVPR}.}

\bibentry{liu2016ssd}{\landmark Liu, W., Anguelov, D., Erhan, D., Szegedy, C., Reed, S., Fu, C.-Y., \& Berg, A.~C.}{ (2016). SSD: Single shot multibox detector. \textit{Proceedings of ECCV}.}

\bibentry{wu2023yunet}{Wu, W., Peng, H., \& Yu, S.}{ (2023). YuNet: A tiny millisecond-level face detector. \textit{Machine Intelligence Research}, 20(5), 656--665.}

\bibentry{lin2014coco}{\landmark Lin, T.-Y., Maire, M., Belongie, S., Hays, J., Perona, P., Ramanan, D., Doll\'ar, P., \& Zitnick, C.~L.}{ (2014). Microsoft COCO: Common objects in context. \textit{Proceedings of ECCV}.}

\bibentry{long2015fcn}{\landmark Long, J., Shelhamer, E., \& Darrell, T.}{ (2015). Fully convolutional networks for semantic segmentation. \textit{Proceedings of IEEE CVPR}.}

\bibentry{ronneberger2015}{\landmark Ronneberger, O., Fischer, P., \& Brox, T.}{ (2015). U-Net: Convolutional networks for biomedical image segmentation. \textit{Proceedings of MICCAI}.}

\bibentry{he2017maskrcnn}{\landmark He, K., Gkioxari, G., Doll\'ar, P., \& Girshick, R.}{ (2017). Mask R-CNN. \textit{Proceedings of IEEE ICCV}.}

\bibentry{micikevicius2017}{Micikevicius, P., Narang, S., Alben, J., Diamos, G., Elsen, E., Garc\'ia, D., Ginsburg, B., Houston, M., Kuchaiev, O., Venkatesh, G., \& Wu, H.}{ (2018). Mixed precision training. \textit{Proceedings of ICLR}.}

\bibentry{goyal2017}{Goyal, P., Doll\'ar, P., Girshick, R., Noordhuis, P., Wesolowski, L., Kyrola, A., Tulloch, A., Jia, Y., \& He, K.}{ (2017). Accurate, large minibatch SGD: Training ImageNet in 1 hour. \textit{arXiv:1706.02677}.}

\bibentry{shoeybi2019}{Shoeybi, M., Patwary, M., Puri, R., LeGresley, P., Casper, J., \& Catanzaro, B.}{ (2019). Megatron-LM: Training multi-billion parameter language models using model parallelism. \textit{arXiv:1909.08053}.}

\bibentry{schonberger2016colmap}{Sch\"onberger, J.~L., \& Frahm, J.-M.}{ (2016). Structure-from-motion revisited. \textit{Proceedings of IEEE CVPR}.}